%% file: Oxford_Thesis.tex
\documentclass[a4paper]{OxfordTeXThesis} %CAMELGRAPH: Part II theses use one-sided binding.
\backgroundsetup{contents={}} 

\correctionstrue

\PassOptionsToPackage{numbers, compress}{natbib}
\usepackage[style=alttree, nonumberlist, toc, acronym, sort=def]{glossaries-extra}
\makeglossaries % QS: If not using Overleaf, to increase speed, this can read "\makeglossaries" instead, and "\printglossary" below. In other compilers, glossaries package requires two compiles to work, unless "[automake]" is an option in the package.
\loadglsentries{text/frontmatter/acronyms_and_abbreviations}
\loadglsentries{text/frontmatter/notation}
\glsaddall

\title{Towards Scaling Deep Neural \\ Networks with Predictive Coding: Theory and Practice}
\author{Francesco Innocenti}
\college{School of Engineering and Informatics}

\degree{Doctor of Philosophy}
\degreedate{13 October 2025}
\supervisors{Christopher L. Buckley and Anil Seth}

\DeclareSIUnit\bar{bar} % QS: from the siunitx package
\DeclareSIUnit\mrad{mrad}
\DeclareSIUnit{\wtpercent}{wt\%}

\hypersetup{
  colorlinks,
  citecolor=orange,
  linkcolor=blue,
  urlcolor=blue
}

\usepackage[T1]{fontenc}    % use 8-bit T1 fonts
\usepackage{url}            % simple URL typesetting
\usepackage{booktabs}       % professional-quality tables
\usepackage{amsfonts}       % blackboard math symbols
\usepackage{nicefrac}       % compact symbols for 1/2, etc.
\usepackage{microtype}      % microtypography

\usepackage{wrapfig}
\usepackage{float}

\usepackage[dvipsnames]{xcolor}

\usepackage{minitoc}

\usepackage{amsmath}
\usepackage{amsthm}
\usepackage{amsfonts}
\usepackage{amssymb}
\usepackage{mathtools}

\newcommand{\matr}[1]{\mathbf{#1}}
\DeclareMathOperator*{\argmin}{arg\,min}
\DeclareMathOperator{\vect}{vec}
\DeclareMathOperator{\diag}{diag}
\DeclareMathOperator{\offdiag}{offdiag}

\theoremstyle{plain}
\newtheorem{theorem}{Theorem}

\theoremstyle{definition}
\newtheorem{definition}[theorem]{Definition}

\theoremstyle{remark}

\theoremstyle{definition}
\newtheorem{desideratum}{Desideratum}

\newenvironment{mythm}[1]{%
  \renewcommand\thetheorem{#1}
  \theorem
}{\endtheorem}

\usepackage{algorithm}
\usepackage{algpseudocode}
\usepackage[most]{tcolorbox}
\usepackage{listings}

\definecolor{codecellbg}{RGB}{245,245,245}
\definecolor{codecellborder}{RGB}{231,231,231}
\definecolor{codered}{RGB}{171,49,42}
\definecolor{codegreen1}{RGB}{80,126,127}
\definecolor{codegreen2}{RGB}{55,126,33}
\definecolor{codeblue}{RGB}{0,0,245}
\definecolor{codeorange}{RGB}{255,128,0}
\definecolor{codepurple}{RGB}{156,47,246}

\lstdefinestyle{pythonStyle}{
    language=Python,
    basicstyle=\ttfamily,
    breaklines=true,
    showstringspaces=false,
    commentstyle=\color{codegreen1},
    keywordstyle=\color{codegreen2},
    stringstyle=\color{codered},
    numberstyle=\tiny\color{gray},
    escapeinside={(*@}{@*)},
    numbers=left,
    numbersep=5pt,
    morekeywords={self},
    breakatwhitespace=false,
    breaklines=true,
    keepspaces=true,
    showspaces=false,
    showstringspaces=false,
    showtabs=false,
    tabsize=4,
    postbreak=\mbox{\textcolor{red}{$\hookrightarrow$}\space},
}
\newtcblisting{jupyterCell}{
    colback=codecellbg,
    colframe=codecellborder,
    boxrule=0.5pt,
    arc=0mm,
    leftrule=3pt,
    rightrule=0.5pt,
    toprule=0.5pt,
    bottomrule=0.5pt,
    left=5mm,
    right=2mm,
    top=0mm,
    bottom=0mm,
    listing only,
    listing options={
        style=pythonStyle,
    },
}

\newtcbox{\codebox}[1][]{
    on line,
    colback=codecellbg,
    colframe=codecellborder,
    boxrule=0.5pt,
    arc=0pt,
    outer arc=0pt,
    left=2pt,
    right=2pt,
    top=2pt,
    bottom=2pt,
    boxsep=0pt,
    fontupper=\ttfamily,
    #1
}
\newcommand{\code}[1]{\codebox{#1}}

\tikzstyle{mynode}=[thick,draw=black,circle,minimum size=1]
\begin{document}

%CAMELGRAPH: The Part II word count includes the whole document from the table of contents to the references, not including the Project Management Forms. 

%%%%% CHOOSE YOUR LINE SPACING HERE

% This is the official option.  Use it for your submission copy and library copy:
\setlength{\textbaselineskip}{22pt plus2pt} %CAMELGRAPH: The Part II requires double line-spaced with at least 11pt text, so...
% \doublespacing % QS: this can be added in if you'd like less tinkering.

% This is closer spacing (about 1.5-spaced) that you might prefer for your personal copies:
%\setlength{\textbaselineskip}{18pt plus2pt minus1pt}

% You can set the spacing here for the roman-numbered pages (acknowledgements, table of contents, etc.)
\setlength{\frontmatterbaselineskip}{22pt plus2pt} %CAMELGRAPH: The Part II requires double line-spaced with at least 11pt text. 

% Leave this line alone; it gets things started for the real document.
\setlength{\baselineskip}{\textbaselineskip}

% \usepackage{parskip} % RJS: Comment this out to remove the need to use \newline function

%%%%% CHOOSE YOUR SECTION NUMBERING DEPTH HERE
% You have two choices.  First, how far down are sections numbered?  (Below that, they're named but don't get numbers.)  Second, what level of section appears in the table of contents?  These don't have to match: you can have numbered sections that don't show up in the ToC, or unnumbered sections that do.  Throughout, 0 = chapter; 1 = section; 2 = subsection; 3 = subsubsection, 4 = paragraph...

% The level that gets a number:
\setcounter{secnumdepth}{2}
% The level that shows up in the ToC:
\setcounter{tocdepth}{2}
% QS: the level that shows up in the miniToCs:
\setcounter{minitocdepth}{2}

%%%%% ABSTRACT SEPARATE
% This is used to create the separate, one-page abstract that you are required to hand into the Exam Schools (QS: only true for some degrees).  You can comment it out to generate a PDF for printing or whatnot.
% \begin{abstractseparate}
% \input{text/abstract} % Create an abstract.tex file in the 'text' folder for your abstract.
% \end{abstractseparate}

% JEM: Pages are roman numbered from here, though page numbers are invisible until ToC.  This is inkeeping with most typesetting conventions.

\begin{romanpages} %CAMELGRAPH: The Part II requires every page to be numbered sequentially, so this will be deactivated. 
% CAMELGRAPH: The above means that the OxfordTeXThesis.cls must also be edited to have the titlepage, absract, and acknowledgements into the 'fancy' pagestyle, not empty. 

%\def\crest{{\includegraphics[width=0.1\textwidth]{figures/sample/laser.png}}} % QS: Put your own image in here if you don't want the standard belted crest. 

%TC:ignore % QS: This tells TeXCount to not count words from here until the next endignore. 

% Title page is created here
\maketitle

\newpage
This thesis is the result of my own work and includes nothing which is the outcome of work done in collaboration except as declared in the introduction and specified in the text. It is not substantially the same as any work that has already been submitted, or is being concurrently submitted, for any degree, diploma or other qualification at the University of Sussex or any other University or similar institution except as declared in the introduction and specified in the text.

%%%%% DEDICATION -- If you'd like one, un-comment the following.
\begin{dedication}
To my parents,\\
who gave me everything\\
\end{dedication}

%%%%% ABSTRACT -- Nothing to do here except comment out if you don't want it.
%CAMELGRAPH: The Part II word count does inclue the abstract, so add in an endignore and ignore around the abstract. 
\begin{abstract}
    \input{text/frontmatter/abstract}
\end{abstract}
\addcontentsline{toc}{chapter}{Abstract}

%%%%% ACKNOWLEDGEMENTS -- Nothing to do here except comment out if you don't want it.
\begin{acknowledgements}
    \input{text/frontmatter/acknowledgements}
\end{acknowledgements}
\addcontentsline{toc}{chapter}{Acknowledgements}

%%%%% MINI TABLES
% This lays the groundwork for per-chapter, mini tables of contents.  Comment the following line
% (and remove \minitoc from the chapter files) if you don't want this.  Un-comment either of the
% next two lines if you want a per-chapter list of figures or tables.
\dominitoc % include a mini table of contents
%\dominilof  % include a mini list of figures
%\dominilot  % include a mini list of tables

% This aligns the bottom of the text of each page.  It generally makes things look better.
\flushbottom

% This is where the whole-document ToC appears:
\tableofcontents

%TC:endignore % QS: This tells TeXCount to start the word count again after the Table of Contents. Add extra sets of ignore and endignore to include and exclude portions from here on.
%CAMELGRAPH: The Part II word count excludes the title page, acknowledgements, and table of contents.

% \listoffigures
\mtcaddchapter  % uncommented - Francesco
\mtcaddchapter
% \mtcaddchapter is needed when adding a non-chapter (but chapter-like) entity to avoid confusing minitoc

% Uncomment to generate a list of tables:
% \listoftables
% \mtcaddchapter

%%%%% LIST OF ABBREVIATIONS
% This example includes a list of abbreviations.  Look at text/abbreviations.tex to see how that file is formatted.  The template can handle any kind of list though, so this might be a good place for a glossary, etc.
%\include{text/The Frontmatter/abbreviations} %QS: this is for JEM's version.

\renewcommand{\glossarypreamble}{\glsfindwidesttoplevelname[\currentglossary]}% This finds the widest 'glossary' entry and scales automatically scales the 
\printglossary[title=Abbreviations, toctitle=Abbreviations, type=acronym]
\printglossary[title=Notation, toctitle=Notation, type=main]

% \renewcommand{\glossarypreamble}{\glsfindwidesttoplevelname[\currentglossary]}
% \printnoidxglossary[title=Glossary, toctitle=Glossary, type=main] % QS: Include this line if you would like to include a glossary. 

% The Roman pages, like the Roman Empire, must come to its inevitable close.
\end{romanpages}%CAMELGRAPH: The Part II requires every page to be numbered sequentially, so this will be deactivated. 

%%%%% CHAPTERS
% Add or remove any chapters you'd like here, by file name (excluding '.tex'):

\include{text/mainmatter/ch1-introduction}
\mtcaddchapter
\include{text/mainmatter/ch2-pcns}
\mtcaddchapter
\include{text/mainmatter/ch3-trust-region}
\include{text/mainmatter/ch4-saddles}
\include{text/mainmatter/ch5-mupc}
\include{text/mainmatter/ch6-jpc}
\include{text/mainmatter/ch7-conclusions}

% moved references before appendices - Francesco 
\renewcommand{\bibname}{References}
\addcontentsline{toc}{chapter}{References}
\bibliography{references}

%%%%% APPENDICES 

% Starts lettered appendices, adds a heading in table of contents, and adds a page that just says "Appendices" to signal the end of your main text.

% Add or remove any appendices you'd like here:
\startappendices
\include{text/appendices/ch3}
\include{text/appendices/ch4}
\include{text/appendices/ch5}
\include{text/appendices/ch6}

\clearpage

%%%%% REFERENCES
%TC:ignore % QS: Often the references are not counted in a word count. 
%CAMELGRAPH: The Part II word count ignores references.

% JEM: Quote for the top of references (just like a chapter quote if you're using them).  
% \begin{savequote}[8cm]
% The first kind of intellectual and artistic personality belongs to the hedgehogs, the second to the foxes \dots
%   \qauthor{--- Sir Isaiah Berlin \cite{berlin_hedgehog_2013}}
% \end{savequote}

% \setlength{\baselineskip}{0pt} % JEM: Single-space References

% \renewbibmacro*{urldate}{
% (retrieved \printfield{urlday}/\printfield{urlmonth}/\printfield{urlyear})
% } % QS: Automates URL date text for citations, with references necessarily being in 'YYYY-MM-DD' format. 

% {\renewcommand*\MakeUppercase[1]{#1}%
% \printbibliography[heading=bibintoc,title={\bibtitle}]}
%\bibliography{references}

%TC:endignore
\end{document}

%% file: text/frontmatter/acronyms_and_abbreviations.tex
% QS: For use with the glossaries package, as the 'new' method for doing abbreviations and a glossary in the OxfordTeXThesis class. For ease, I have combined acronyms and abbreviations, as that will serve most people. To change this, separate this document into acronyms.tex and abbreviations.tex, insert the [abbreviation] option into glossaries-extra, load the new file, and add abbreviations under \newabbreviation. 

% alphabetical order
\newacronym{1MLP}{1MLP}{Scalar \glsxtrshort{MLP} with a single hidden unit}
\newacronym{AI}{AI}{Artificial intelligence}
\newacronym{API}{API}{Application Programming Interface}
\newacronym{BP}{BP}{Backpropagation}
\newacronym{BPC}{BPC}{Bidirectional predictive coding}
\newacronym{DLN}{DLN}{Deep linear network}           
\newacronym{DNN}{DNN}{Deep neural network}
\newacronym{EBM}{EBM}{Energy-based model}
\newacronym{EP}{EP}{Equilibrium propagation}
\newacronym{FC}{FC}{Fully connected network layer}
\newacronym{GD}{GD}{Gradient descent}
\newacronym{GPU}{GPU}{Graphics processing unit}
\newacronym{HPC}{HPC}{Hybrid predictive coding}
\newacronym{LPM}{LPM}{Leading principal minor}
\newacronym{ML}{ML}{Machine learning}
\newacronym{MLP}{MLP}{Multi-layer perceptron; fully connected neural network}
\newacronym{MP}{MP}{Marčhenko-Pastur distribution}
\newacronym{MSE}{MSE}{Mean squared error}
\newacronym{NTK}{NTK}{Neural tangent kernel}
\newacronym{ODE}{ODE}{Ordinary differential equation}
\newacronym{PC}{PC}{Predictive coding}
\newacronym{PCN}{PCN}{Predictive coding network}
\newacronym{PD}{PD}{Positive definite}
\newacronym{ReLU}{ReLU}{Rectified Linear Unit}
\newacronym{ResNet}{ResNet}{Residual network}
\newacronym{RMT}{RMT}{Random matrix theory}
\newacronym{SEM}{SEM}{Standard error of the mean}
\newacronym{SGD}{SGD}{Stochastic \glsxtrshort{GD}}
\newacronym{SP}{SP}{Standard parameterisation}
\newacronym{Tanh}{Tanh}{Hyperbolic tangent}
\newacronym{TP}{TP}{Target propagation}
\newacronym{TR}{TR}{Trust region}
\newacronym{TRN}{TRN}{Trust region Newton}
\newacronym{μP}{μP}{Maximal Update Parameterisation}

% general
% \newacronym{PC}{PC}{Predictive coding}
% \newacronym{PCN}{PCN}{Predictive coding network}
% \newacronym{BP}{BP}{Backpropagation}
% \newacronym{TP}{TP}{Target propagation}
% \newacronym{EP}{EP}{Equilibrium propagation}
% \newacronym{AI}{AI}{Artificial intelligence}
% \newacronym{GPU}{GPU}{Graphics processing unit}
% \newacronym{EBM}{EBM}{Energy-based model}
% \newacronym{DNN}{DNN}{Deep neural network}
% \newacronym{MLP}{MLP}{Multi-layer perceptron; fully connected neural network}
% \newacronym{ReLU}{ReLU}{Rectified Linear Unit}
% \newacronym{Tanh}{Tanh}{Hyperbolic tangent}
% \newacronym{GD}{GD}{Gradient descent}
% \newacronym{SGD}{SGD}{Stochastic \glsxtrshort{GD}}
% \newacronym{MSE}{MSE}{Mean squared error}
% \newacronym{DLN}{DLN}{Deep linear network}
% \newacronym{ML}{ML}{Machine learning}

% trust region paper
% \newacronym{1MLP}{1MLP}{Scalar \glsxtrshort{MLP} with a single hidden unit}
% \newacronym{TR}{TR}{Trust region}
% \newacronym{TRN}{TRN}{Trust region Newton}

% mupc paper
% \newacronym{ResNet}{ResNet}{Residual network}
% \newacronym{NTK}{NTK}{Neural tangent kernel}
% \newacronym{SP}{SP}{Standard parameterisation}
% \newacronym{PD}{PD}{Positive definite}
% \newacronym{LPM}{LPM}{Leading principal minor}
% \newacronym{RMT}{RMT}{Random matrix theory}
% \newacronym{MP}{MP}{Marčhenko-Pastur distribution}

% jpc paper
% \newacronym{ODE}{ODE}{Ordinary differential equation}

%% file: text/frontmatter/notation.tex
\label{notation}

% general
\newglossaryentry{scalars}{
    name={\ensuremath{u, U}},
    description={Scalars}
}
\newglossaryentry{vector}{
    name={\ensuremath{\mathbf{v}}},
    description={Vector, assumed to be column-oriented \ensuremath{\mathbf{v} \in \mathbb{R}^{n \times 1}}}
}
\newglossaryentry{matrix}{
    name={\ensuremath{\mathbf{A}}},
    description={Matrix}
}
\newglossaryentry{identity}{
    name={\ensuremath{\mathbf{I}_n}},
    description={\ensuremath{n \times n} identity matrix}
}
\newglossaryentry{zero}{
    name={\ensuremath{\mathbf{0}_n}},
    description={\ensuremath{n}-zero vector or \ensuremath{n \times n} null matrix, depending on context}
}
\newglossaryentry{gaussian}{
    name={\ensuremath{\mathcal{N}(\mu, \sigma^2)}},
    description={Normal (Gaussian) distribution with mean \ensuremath{\mu} and variance \ensuremath{\sigma^2}}
}
\newglossaryentry{l2-norm}{
    name={\ensuremath{||\cdot||}},
    description={Euclidean \ensuremath{\ell^2} norm}
}
\newglossaryentry{vec-operator}{
    name={\ensuremath{\textnormal{vect}(\cdot)}},
    description={row-wise vector operator}
}
\newglossaryentry{kronecker}{
    name={\ensuremath{\otimes}},
    description={Kronecker product between two matrices}
}
\newglossaryentry{first-deriv}{
    name={\ensuremath{\nabla_{\mathbf{x}} f}},
    description={Gradient of some function \ensuremath{f} with respect to \ensuremath{\mathbf{x}}}
}
\newglossaryentry{second-deriv}{
    name={\ensuremath{\nabla^2_\mathbf{x} f}},
    description={Hessian of \ensuremath{f} with respect to \ensuremath{\mathbf{x}}}
}
\newglossaryentry{gradient}{
    name={\ensuremath{\mathbf{g}_f(\mathbf{x})}},
    description={Abbreviation for the gradient of \ensuremath{f} with respect to \ensuremath{\mathbf{x}}}
}
\newglossaryentry{hessian}{
    name={\ensuremath{\mathbf{H}_f(\mathbf{x})}},
    description={Abbreviation for the Hessian of \ensuremath{f} with respect to \ensuremath{\mathbf{x}}}
}
\newglossaryentry{critical-point}{
    name={\ensuremath{\mathbf{x}^*}},
    description={Critical point of some function \ensuremath{f} where \ensuremath{\nabla_{\mathbf{x}} f = 0}}
}
% \newglossaryentry{eigens}{
%     name={\ensuremath{\lambda(\cdot)}},
%     description={Matrix eigenvalues, e.g. \ensuremath{\lambda_{\text{max}}(\cdot)} as the maximum}
% }

% specific to PC or neural nets
\newglossaryentry{pc-activities}{
    name={\ensuremath{\mathbf{z}}},
    description={Set of activities, states or latent variables of a \glsxtrshort{PCN}}
}
\newglossaryentry{params}{
    name={\ensuremath{\boldsymbol{\theta}}},
    description={Set of weights or parameters of a neural network}
}
\newglossaryentry{pc-energy}{
    name={\ensuremath{\mathcal{F}(\boldsymbol{\theta}, \mathbf{z})}},
    description={\glsxtrshort{PC} energy function (e.g. Eq. \ref{pcns:eq:pc-energy})}
}
\newglossaryentry{theory-equilib-pc-energy}{
    name={\ensuremath{\mathcal{F}^*}},
    description={Abbreviation for the \glsxtrshort{PC} energy at a solution or equilibrium of the network activities \ensuremath{\mathcal{F}(\boldsymbol{\theta}, \mathbf{z}^*)} (e.g. Eq. \ref{ch4:eq:dln-equilib-energy})}
}
\newglossaryentry{batch-size}{
    name={\ensuremath{B}},
    description={Batch or dataset size}
}
\newglossaryentry{width}{
    name={\ensuremath{N}},
    description={Width of a neural network}
}
\newglossaryentry{layers}{
    name={\ensuremath{L}},
    description={Number of layers or weight matrices of a neural network}
}
\newglossaryentry{hidden-layers}{
    name={\ensuremath{H}},
    description={Number of hidden layers of a neural network (\ensuremath{H=L-1})}
}
\newglossaryentry{act-fn}{
    name={\ensuremath{\phi_\ell(\cdot)}},
    description={Activation function of a neural network layer (e.g. \glsxtrshort{ReLU})}
}

%% file: text/frontmatter/abstract.tex
% RJS: It is important to note that this template it built upon another, but includes and shows examples of packages that I found useful when working on my thesis.

Backpropagation (BP) is the standard algorithm for training the deep neural networks that power modern artificial intelligence including large language models. However, BP is energy inefficient and unlikely to be implemented by the brain. This thesis studies an alternative, potentially more efficient brain-inspired algorithm called predictive coding (PC). Unlike BP, PC networks (PCNs) perform inference by iterative equilibration of neuron activities before learning or weight updates. Recent work has suggested that this iterative inference procedure provides a range of benefits over BP, such as faster training. However, these advantages have not been consistently observed, the inference and learning dynamics of PCNs are still poorly understood, and deep PCNs remain practically untrainable. Here, we make significant progress towards scaling PCNs by taking a theoretical approach grounded in optimisation theory. First, we show that the learning dynamics of PC can be understood as an approximate trust-region method using second-order information, despite explicitly using only first-order local updates. Second, going beyond this approximation, we show that PC can in principle make use of arbitrarily higher-order information, such that for fully connected networks the effective landscape on which PC learns is far more benign and robust to vanishing gradients than the (mean squared error) loss landscape. Third, motivated by a study of the inference dynamics of PCNs, we propose a new parameterisation called ``$\mu$PC'', which for the first time allows stable training of 100+ layer networks with little tuning and competitive performance on simple classification tasks. We also introduce an open-source Python library for training PCNs in JAX. Overall, this thesis significantly advances our fundamental understanding of the inference and learning dynamics of PCNs, while highlighting the need for future research to focus on hardware co-design and more expressive architectures if PC is to compete with BP at scale.

%% file: text/frontmatter/acknowledgements.tex
This PhD would not have been possible without the guidance, collaboration, and support of many people. First and foremost is my main supervisor, Christopher L. Buckley, who pushed me to be a better researcher while giving me the freedom to pursue my own questions. Second, I am grateful to El Mehdi Achour, with whom I collaborated on two major works in this thesis (Chapters \ref{ch:saddles} \& \ref{ch:mupc}). In one of life's rare serendipities, El Mehdi and I met on a beach at a conference in Hawaii by introduction of a friend of his (who happened to sit next to me on the plane). What started as a fruitful collaboration developed into a friendship that I hope to maintain in the future.

I am also thankful to Paul Kinghorn, with whom I had many discussions about predictive coding in my first year that were the seed of much of the work in this thesis. Another person I am indebted to is Ryan Singh, who helped significantly with the theory presented in Chapter \ref{ch:trust-region} and who always had useful insights to share. Presenting this work together at my first conference is an experience that I will never forget.

In addition, I would like to thank my second supervisor, Anil Seth, for general advice on writing and the trajectory of my PhD; Dhruva V. Raman for early discussions about the work presented in Chapter \ref{ch:trust-region}; and Sussex Neuroscience for funding and support. I am also indebted to my undergraduate research supervisors, Ashok Jansari and Devin B. Terhune, for helping me become a better scientist and find a passion for research.

Other people who provided support during my PhD include (in alphabetical order) Fatima Arshad, Lionel Barnett, Poppy Collis, Benjamin Evans, Hannah Gong, Oluwaseyi Oladipupo Jesusanmi, Kasia Mojescik, Joshua Reyniers, Ivor Simpson, Ruth Staras, Miguel De Llanza Varona, and Will Yun-Farmbrough.

Last, but definitely not least, none of this would have been possible without the love and support of my parents, Carolina and Alessandro, who always believed in me and gave me everything they had and more. I dedicate this achievement to you.

%% file: text/mainmatter/ch1-introduction.tex
\chapter{Introduction}
\label{ch:intro}
\minitoc

\section{Thesis Overview}
This thesis explores an alternative approach to training deep neural networks (DNNs), the underlying models of modern artificial intelligence (AI) \cite{lecun2015deep}. The current standard for neural network training is the so-called ``backpropagation of error'' algorithm \cite{rumelhart1986learning} (BP). At its core, BP is an efficient method for computing derivatives of complex functions, enabled by specialised hardware such as graphics processing units (GPUs) and software libraries such as PyTorch \cite{paszke2019pytorch} and JAX \cite{bradbury2018jax}. 

However, BP has several inherent limitations. For example, BP requires storing the forward computational graph of the model, making it memory and energy inefficient \cite{faiz2023llmcarbon, thompson2021deep, strubell2020energy}. BP is also a sequential algorithm that cannot be parallelised across model layers \cite{jaderberg2017decoupled}. These limitations arise from the inherently \textit{non-local} nature of BP, in that the update of any given weight depends on information from all downstream layers in the network. For these and other reasons, BP is also widely regarded as ``biologically implausible'' or unlikely to be implemented in the brain \cite{crick1989recent, lillicrap2020backpropagation}. 

The alternative algorithm that we study in this thesis is called \textit{predictive coding} (PC) \cite{van2024predictive, salvatori2023brain, millidge2022predictivereview, millidge2021predictive}. PC belongs to a broad and diverse class of brain-inspired or biologically plausible learning algorithms, including equilibrium propagation \cite{scellier2017equilibrium, zucchet2022beyond}, target propagation \cite{meulemans2020theoretical}, and forward learning \cite{hinton2022forward}, among others \cite{dellaferrera2022error, payeur2021burst, ororbia2019biologically, lillicrap2016random}. While different in many aspects, these algorithms all share a key feature that distinguishes them from BP: local, ``Hebbian-like'' weight updates that rely solely on interactions between neighbouring neurons.

At a high level, PC is based on the basic idea that the brain's \textit{modus operandi} is to minimise the errors of its predictions with respect to a generative model of the environment. This idea has a long history in computational neuroscience. Originally proposed as a theory of retinal function \cite{srinivasan1982predictive}, PC later developed into a more general principle for information processing in the brain \cite{mumford1992computational, rao1999predictive, friston2003learning, friston2005theory, friston2008hierarchical}. 

In more recent years, starting with the seminal tutorials of \cite{buckley2017free, bogacz2017tutorial}, PC has been explored as a learning algorithm that could provide a biologically plausible alternative to BP. DNNs trained with PC have shown comparable performance to BP on simple machine learning tasks including classification, generation, and memory association \cite{salvatori2023brain, millidge2022predictivereview, millidge2021predictive}. Moreover, PC has been suggested to provide a range of benefits over BP \cite{song2022inferring}, including faster learning convergence and increased performance in more biologically realistic tasks such as online and continual learning. PC networks (PCNs) also support arbitrary computational graphs \citep{salvatori2022learning, byiringiro2022robust}, can perform hybrid and causal inference \citep{salvatori2023causal, tscshantz2023hybrid}, and can be extended to deal with temporal tasks \citep{millidge2024predictive}.

However, the main challenge—which we attempt to tackle in this thesis—has been to scale PC and other local learning algorithms to very deep (10+ layer) networks on large-scale datasets such as ImageNet \cite{deng2009imagenet} (let alone large language models trained on trillions of tokens). It is not unlikely that local algorithms could be practically scaled (i.e. with competitive compute and memory resources) only on alternative, non-digital hardware such as analog or neuromorphic chips. We will return to this point in the conclusion (\S\ref{ch:conclusions}). Nevertheless, this thesis will show that we can still make significant progress on this goal by studying PC on standard GPUs.

The way we attempt to meet the challenge of scaling PC is through a combination of theory and experiment. Following the nascent field of deep learning theory \cite{liu2019deep, he2020recent, roberts2022principles, suh2024survey, petersen2024mathematical, ziyin2025parameter}, we will take an optimisation-theoretic approach, with \textit{deep linear networks} (DLNs) as our main theoretical model. Indeed, many of the contributions of this thesis are found in adapting optimisation-theoretic analyses of DLNs to PC. This model will not only provide the most explanatory and predictive theory of the inference and learning dynamics of practical PCNs (Chapters~\ref{ch:saddles}-\ref{ch:mupc}), but also allow us, for the first time, to scale PC to 100+ layer networks with little tuning and competitive performance on simple tasks (Chapter~\ref{ch:mupc}). Other contributions, covered in more detail below (\S\ref{ch1:contributions}), include a novel interpretation of PC as a trust-region optimiser (Chapter~\ref{ch:trust-region}) and an open-source Python package for training PCNs in JAX (Chapter~\ref{ch:jpc}).

\subsection{Structure}
The thesis is structured as follows. The rest of this chapter presents a detailed breakdown of the contributions of this PhD. \textbf{Chapter~\ref{ch:pcns}} reviews PCNs as a foundation for the subsequent chapters. With the exception of the conclusion and appendices, the remaining chapters correspond to different research papers. \textbf{Chapter~\ref{ch:trust-region}} presents an approximate theory of PC as a second-order trust-region method. \textbf{Chapter~\ref{ch:saddles}} goes significantly beyond this theory and provides a characterisation of the learning landscape and dynamics of PCNs with surprising and insightful findings. Following from that, \textbf{Chapter~\ref{ch:mupc}} performs a similar analysis of the inference landscape and dynamics of PCNs and introduces ``$\mu$PC'', a new parameterisation of PCNs that allows stable training of 100+ layer networks. \textbf{Chapter~\ref{ch:jpc}} presents JPC, an open-source Python library developed to train a variety of PCNs that was used for many of the experiments in this thesis. Each of these chapters is associated with a comprehensive appendix, typically including relevant literature reviews, technical derivations, experimental details and supplementary figures. Finally, \textbf{Chapter~\ref{ch:conclusions}} concludes by discussing the main implications and limitations of this thesis, along with some speculations.

\section{Statement of Contributions}
\label{ch1:contributions}
This thesis makes the following main contributions, each associated with a chapter and paper (see Table~\ref{contributions-table} for a summary):
\begin{itemize}
    \item \textbf{Chapter~\ref{ch:trust-region} \cite{innocenti2023understanding}.} We show that the learning dynamics of PC can be understood as an implicit approximate second-order trust-region method, despite explicitly using only first-order (gradient) information. This theory (i) makes fewer assumptions than previous works, (ii) sheds new insights into the workings of PC, and (iii) suggests some novel neuroscience interpretations. This work was presented in \cite{innocenti2023understanding}, which won a Best Paper Award at the ICML 2023 Workshop on Localized Learning. The ICML talk is available \href{https://icml.cc/virtual/2023/workshop/21484}{here}. 
    %See also \href{https://francesco-innocenti.github.io/posts/2023/08/10/PC-as-a-2nd-Order-Method/}{this post} for a high-level summary. 
    \item \textbf{Chapter~\ref{ch:saddles} \cite{innocenti2025only}.} Going significantly beyond the above work, we develop a much more precise theory of the learning dynamics of PCNs by characterising the geometry of the effective landscape on which PC learns. For fully connected (non-residual) networks, we show that PC learns on a rescaled mean squared error loss that, under certain conditions, is much easier to navigate than the original loss. Among other things, our theory (i) corrects a previous mistake in the literature, (ii) provides a unifying explanation of seemingly contradictory findings, and (iii) makes new predictions which we verify. This work was accepted at NeurIPS 2024 \cite{innocenti2025only} and later republished \href{https://iopscience.iop.org/article/10.1088/1742-5468/ade2eb}{here} in the \textit{Journal of Statistical Mechanics: Theory and Experiment} as part of a Special Issue on Machine Learning 2025. 
    % A brief review of this work can be found \href{https://francesco-innocenti.github.io/posts/2024/10/01/The-Energy-Landscape-of-Predictive-Coding-Networks/}{here}.
    \item \textbf{Chapter~\ref{ch:mupc} \cite{innocenti2025mu}.} We develop a similar theory of the inference landscape and dynamics of PCNs, showing (i) that the landscape becomes increasingly ill-conditioned with model size (width and particularly depth) as well as training time, and (ii) that the forward pass of standard PCNs tends to vanish/explode with depth. Motivated by these findings, we introduce $\mu$PC, a new parameterisation of PCNs that for the first time allows stable training of 100+ layer networks with little tuning and competitive performance on simple classification tasks. To the best of my knowledge, \textit{no networks of such depths had been trained before with a local or brain-inspired learning algorithm}. This work lays a foundation for future attempts to scale PC and has been accepted at NeurIPS 2025. 
    % For a concise summary, see \href{https://francesco-innocenti.github.io/posts/2025/05/20/Scaling-Predictive-Coding-to-100+-Layer-Networks/}{this post}.
    \item \textbf{Chapter~\ref{ch:jpc} \cite{innocenti2024jpc}.} We introduce JPC \cite{innocenti2024jpc}, a Python library for training a variety of PCNs with JAX. JPC is available at \sloppy{\url{https://github.com/thebuckleylab/jpc}} including many examples and detailed documentation.
\end{itemize}
While the author of this thesis was the main contributor to all of the above works, for reference each of these chapters includes a final section on specific author contributions. We also note a contribution made during this PhD that does not form part of the thesis: ``A Simple Generalisation of the Implicit Dynamics of In-Context Learning'' as a paper to appear at the NeurIPS 2025 workshop on \textit{What Can('t) Transformers Do?}.

Overall, this thesis significantly advances our understanding of how inference and learning, and their interaction, unfold in PCNs, with clear practical implications for scaling PC and other energy-based learning algorithms (as discussed in detail in \S \ref{ch:conclusions}). Any future attempts to further scale or better understand PCNs would benefit from this work.
\begin{table}[h!]
  \centering
  \caption{\textbf{Summary of contributions.}}
  \label{contributions-table}
  \begin{tabular}{@{} p{1.5cm} p{5cm} p{6.5cm} @{}}
    \toprule
    \textbf{Chapter} & \textbf{Paper} & \textbf{Main results} \\
    \midrule
    \ref{ch:trust-region} & Understanding Predictive Coding as a Second-Order Trust-Region Method \cite{innocenti2023understanding} & The learning dynamics of PC can be interpreted as an approximate second-order trust-region method, despite explicitly using only first-order, local updates. \\
    \addlinespace
    \ref{ch:saddles} & Only Strict Saddles in the Energy Landscape of Predictive Coding Networks? \cite{innocenti2025only} & At the equilibrium of the inference dynamics, PCNs effectively learn on a rescaled mean squared error loss, and many highly degenerate saddle points of the loss become benign in the equilibrated energy. Under certain conditions, this makes feedforward (non-residual) networks easier to train with PC than BP. \\
    \addlinespace
    \ref{ch:mupc} & $\mu$PC: Scaling Predictive Coding to 100+ Layer Networks \cite{innocenti2025mu} & A reparameterisation of PCNs which we call ``$\mu$PC'' allows stable training of 100+ layer residual networks with little tuning and competitive performance on simple classification tasks, while also enabling zero-shot transfer of both the weight and activity learning rates across model widths and depths. \\
    \addlinespace
    \ref{ch:jpc} & JPC: Flexible Inference for Predictive Coding Networks in JAX \cite{innocenti2024jpc} & \textsc{JPC} is a simple, fast and flexible \textbf{J}AX library that allows training of neural networks with many different \textbf{P}redictive \textbf{C}oding schemes. \\
    \bottomrule
  \end{tabular}
\end{table}
% \begin{mccorrection}
% Corrections can be inserted as so. 
% \end{mccorrection}

% Shorter corrections can be inserted \mccorrect{like so}\todo{Comments can also be done \textit{like this}.}. 

%% file: text/mainmatter/ch2-pcns.tex
\chapter{Predictive Coding Networks (PCNs)}
\label{ch:pcns}

In this chapter, we review predictive coding networks (PCNs) as a foundation for the following chapters. Note, however, that we aim to make each chapter self-contained and so key equations will be re-presented.

\paragraph{PCN energy.} Training a deep neural network (DNN) with PC means modelling the activity of each layer (and neuron) as a random variable rather than some deterministic function as is assumed for BP. A hierarchical Gaussian model with identity covariances is the most common form of generative model used in practice. While other types of generative model have been explored \cite{salvatori2022learning, pinchetti2022predictive}, this is what we will focus on to keep the theory close to practice. For a multi-layer perceptron or fully connected network (with no biases), the activity of a layer $\mathbf{z}_\ell \in \mathbb{R}^{N_{\ell}}$ can then be modelled as $\mathbf{z}_\ell \sim \mathcal{N}(\phi_\ell(\matr{W}_\ell \mathbf{z}_{\ell-1}), \matr{I}_\ell)$, where $\matr{W}_\ell \in \mathbb{R}^{N_\ell \times N_{\ell-1}}$ is some learnable weight matrix and $\phi_\ell(\cdot)$ is an element-wise activation function such as ReLU. Under Dirac-delta or point-mass posterior distributions, we can derive an energy function, often referred to as the variational free energy, which reduces to a simple sum of squared prediction errors across $L$ network layers \cite{buckley2017free}
\begin{equation}
    \mathcal{F} = \frac{1}{B}\sum_{i=1}^{B} \sum_{\ell=1}^L ||\mathbf{z}_{\ell, i} - \phi_\ell(\matr{W}_\ell \mathbf{z}_{\ell-1, i})||^2/2,
    \label{pcns:eq:pc-energy}
\end{equation}
where $B$ is the batch size or number of data points fitted at any point during training. For simplicity, we will often drop the data index $i$. Eq.~\ref{pcns:eq:pc-energy} is not the most general form of PC energy that can be written, since one can also assume different layer-to-layer functions (other than fully connected), multiple transformations per layer, and non-identity covariances. However, this thesis will focus on this formulation (and slight variations thereof), again to remain faithful to typical PCNs trained in practice. Note also that Eq.~\ref{pcns:eq:pc-energy} can be rewritten to define an energy for every neuron, which will inevitably lead to local gradients with respect to both the activities and the weights. We will use $\boldsymbol{\theta} \coloneq \{\vect(\matr{W}_\ell)\}_{\ell=1}^L \in \mathbb{R}^p$ to represent all the weights, with $p$ as the total number of parameters, and $\mathbf{z} \coloneq \{\mathbf{z}_\ell \}_{\ell=1}^H \in \mathbb{R}^{NH}$ to denote all the activities free to vary, with $H=L-1$ as the number of hidden layers. We will also use subscripts to index either layers or time steps depending on the context.

For theoretical purposes, we will often (though not always) study deep linear networks (DLNs)\footnote{Specifically, the analyses of Chapters~\ref{ch:saddles} \& \ref{ch:mupc} will rely on DLNs, while Chapter~\ref{ch:trust-region} will consider arbitrary PCNs.}, assuming that the activation function is the identity $\phi_\ell = \matr{I}$ at every layer $\ell$. There are two main reasons for this choice. First, linearity makes the mathematical analysis more tractable in many respects. Second, DLNs have proved to be a useful model of non-linear networks as first famously shown by \cite{saxe2013exact}. As we will see in Chapters~\ref{ch:saddles} \& \ref{ch:mupc}, while capable of learning only linear representations, DLNs have non-convex loss landscapes and non-linear learning dynamics similar to their non-linear counterparts.

\paragraph{PCN training.} To train a PCN, the observations of the generative model need to be clamped to some target data, $\mathbf{z}_L \coloneq \mathbf{y} \in \mathbb{R}^{N_L}$. This could be a label for classification or an image for generation, and these two settings are typically referred to as \textit{discriminative} and \textit{generative} PC, respectively. In supervised (vs unsupervised) learning, the first layer is also fixed to some input, $\mathbf{z}_0 \coloneq \mathbf{x} \in \mathbb{R}^{N_0}$. The experiments in this thesis will focus on the (supervised) discriminative setting, but the theory will often generalise to any setting. Note that different papers use different notation and terminology depending on the setting of interest.

Once the network output and (optionally) input are clamped to some data, the energy (Eq.~\ref{pcns:eq:pc-energy}) is minimised in a bi-level, expectation-maximisation fashion \cite{dempster1977maximum}, as we explain in detail below.

\paragraph{Inference.} In the first phase, given some weights $\textcolor{blue!50}{\boldsymbol{\theta}_t}$, we minimise the energy with respect to the \textcolor{red!50}{activities} of the network:
% Define node styles locally for each neural network
\tikzset{
    freenode/.style={
        circle,
        draw=red!50,    % Border color
        fill=red!10,  % Fill color
        thick
    },
    fixednode/.style={
        circle,
        draw=black,    % Border color
        fill=white,    % Fill color for nodes at the bottom layer
        thick
    }
}
\pgfmathsetseed{3}  % 1, 2

\vspace{0.2cm}
\hspace{0.8cm}
\begin{minipage}{0.005\textwidth}
    \begin{tikzpicture}[x=0.75cm,y=0.6cm]
      \foreach \N [count=\lay,remember={\N as \Nprev (initially 0);}]
                   in {2,3,3,2}{ % loop over layers
        \foreach \i [evaluate={
          \x=\N/2-\i; 
          \y=-1.25*\lay; 
          \prev=int(\lay-1);
          \opacity=0.6+0.4*rand}]
                     in {1,...,\N}{ % loop over nodes
          % Store the opacity value globally
          \pgfmathsetmacro{\temp}{\opacity}
          \expandafter\xdef\csname opacity\lay\i\endcsname{\temp}
          
          \ifnum \lay=1
              \ifnum \i=2 % Check if it's the first or second node of the top layer (\lay = 1)
                  \node[fixednode] (N\lay-\i) at (\x,\y) {};
                  \node[anchor=east] at (N\lay-\i.west) {\hspace{1mm}$\mathbf{x}$};
              \else
                  \node[fixednode] (N\lay-\i) at (\x,\y) {};
              \fi
          \else
              \ifnum \lay=4
                  \ifnum \i=2 % Check if it's the first or second node of the bottom layer (\lay = 4)
                      \node[fixednode] (N\lay-\i) at (\x,\y) {};
                      \node[anchor=east] at (N\lay-\i.west) {\hspace{1mm}$\mathbf{y}$};
                  \else
                      \node[fixednode] (N\lay-\i) at (\x,\y) {};
                  \fi
              \else
                  \node[freenode, opacity=\opacity] (N\lay-\i) at (\x,\y) {};
              \fi
          \fi
          % Connect nodes to previous layer
          \ifnum\Nprev>0
            \foreach \j in {1,...,\Nprev}
              \draw[->] (N\prev-\j) -- (N\lay-\i);
          \fi
        }
      }
    \end{tikzpicture}
\end{minipage}
\begin{minipage}{0.85\textwidth}
    \begin{equation}
        \begin{aligned}
            \textit{Infer:  }
            \textcolor{red!50}{\mathbf{z}^*} = \argmin_{\textstyle\mathbf{z}} \mathcal{F}(\textcolor{blue!50}{\boldsymbol{\theta}_t}, \mathbf{z}).
        \end{aligned}
    \label{pcns:eq:pc-infer}
    \end{equation}
\end{minipage}
\vspace{0.2cm} \\
This process is called ``inference'' and can be intuitively thought as the network trying to find an equilibrium of its state that best accounts for all the data. This minimisation process can be performed in many different ways, using different state initialisations and algorithms, in continuous or discrete time. Typically, the activities are initialised with a forward pass, and (discrete-time) gradient descent (GD) is used such that $\mathbf{z}_{i+1} = \mathbf{z}_i - \beta \nabla_{\mathbf{z}} \mathcal{F}(\boldsymbol{\theta}_t, \mathbf{z}_i)$ with some step size $\beta$. The goal is often to reach convergence as implied by Eq.~\ref{pcns:eq:pc-infer} (though see \cite{salvatori2022incremental} for an exception), which is often determined by checking whether the activity gradients are close to zero $\nabla_{\mathbf{z}} \mathcal{F} \approx 0$\footnote{In Chapter~\ref{ch:mupc}, we will see that this is not a sufficient criterion to determine closeness to an inference solution.}. This iterative inference procedure (Eq.~\ref{pcns:eq:pc-infer}) is arguably the key aspect in which PC (and other energy-based algorithms) differs from BP, where inference is amortised and simply modelled by a feedforward pass.

\paragraph{Learning.} Once we have reached a fixed point of the network state $\textcolor{red!50}{\mathbf{z}^*}$, we minimise the energy evaluated at this equilibrium with respect to the \textcolor{blue!50}{weights}, by performing a single weight update:

\vspace{0.2cm}
\hspace{0.8cm}
\begin{minipage}{0.005\textwidth}
    \begin{tikzpicture}[x=0.75cm,y=0.6cm]
      \foreach \N [count=\lay,remember={\N as \Nprev (initially 0);}]
                   in {2,3,3,2}{ % loop over layers
        \foreach \i [evaluate={\x=\N/2-\i; \y=-1.25*\lay; \prev=int(\lay-1);}]
                     in {1,...,\N}{ % loop over nodes
                     % Retrieve stored opacity
                     \pgfmathsetmacro{\storedopacity}{\csname opacity\lay\i\endcsname}
                     
                     \ifnum \lay=1
                          \ifnum \i=2 % Check if it's the first or second node of the top layer (\lay = 1)
                              \node[anchor=east] at (N\lay-\i.west) {\hspace{1mm}$\mathbf{x}$};
                          \fi
                      \else
                          \ifnum \lay=4
                              \ifnum \i=2 % Check if it's the first or second node of the bottom layer (\lay = 4)
                                  \node[anchor=east] at (N\lay-\i.west) {\hspace{1mm}$\mathbf{y}$};
                              \fi
                          \fi
                      \fi
          %\node[fixednode] (N\lay-\i) at (\x,\y) {};
          % Use stored opacity for freenode styles
          \ifnum \lay=1
              \node[fixednode] (N\lay-\i) at (\x,\y) {};
          \else
              \ifnum \lay=4
                  \node[fixednode] (N\lay-\i) at (\x,\y) {};
              \else
                  \node[freenode, opacity=\storedopacity] (N\lay-\i) at (\x,\y) {};
              \fi
          \fi
          
          \ifnum\Nprev>0 % connect to previous layer
            \foreach \j in {1,...,\Nprev}{ % loop over nodes in previous layer
              \pgfmathsetmacro{\randomseed}{int(100*rand)}
              \pgfmathsetseed{\randomseed}
              \pgfmathsetmacro{\thickness}{0.6 + 0.3*rand}
              \draw[->, blue!50, line width=\thickness pt] (N\prev-\j) -- (N\lay-\i);
            }
          \fi
        }
      }
    \end{tikzpicture}
\end{minipage}
\begin{minipage}{0.85\textwidth}
    \begin{equation}
        \begin{aligned}
            \textit{Learn:  }
            \textcolor{blue!50}{\boldsymbol{\theta}_{t+1}} = \boldsymbol{\theta}_t - \eta \matr{P}_t \nabla_{\theta} \mathcal{F}(\boldsymbol{\theta}_t, \textcolor{red!50}{\mathbf{z}^*}),
        \end{aligned}
    \label{pcns:eq:pc-learn}
    \end{equation}
\end{minipage}
\vspace{0.2cm} \\
where $\nabla_{\theta} \mathcal{F}$ is the gradient of the energy with respect to the weights, $\matr{P}_t$ is some preconditioner matrix, and $\eta$ is a global learning rate. Note that standard GD is recovered by selecting an identity preconditioner $\matr{P}_t = \matr{I}$. This phase is called ``learning'' for obvious reasons and is in practice often performed using the Adam optimiser \cite{kingma2014adam}. Following a weight update, we restart the optimisation cycle with a new data batch (which we have not shown here for simplicity) and repeat this process, typically until we are satisfied with the test or generalisation performance on some held-out examples. See Algorithm~\ref{pcn:pseudoalgo} for some pseudo code. The way this bi-level optimisation is performed reflects the intuition that the neural (activity) dynamics (Eq.~\ref{pcns:eq:pc-infer}) operate at a faster timescale than the synaptic (weight) dynamics (Eq.~\ref{pcns:eq:pc-learn}). As alluded to above, in contrast to BP, both the activity and weight gradients of the energy are local, requiring information only about neighbouring neurons.
\begin{algorithm}[t]
\caption{Training a Neural Network with Predictive Coding}
\label{pcn:pseudoalgo}
    \begin{algorithmic}
        \State \textbf{Input:} Initial weights $\boldsymbol{\theta}_0$, dataset $\{(\mathbf{x}_i, \mathbf{y}_i)\}^{B}_{i=1}$
        \State \textbf{Hyperparameters:} Learning steps $T$, inference steps $N$, inference step size $\beta$, learning step size $\eta$
        \For{$t = 0, \dots, T - 1$}
            \State Initialise activities $\mathbf{z}_0$ with data sample $(\mathbf{x}_i, \mathbf{y}_i)$
            \For{$i = 0, \dots, N - 1$}
                \State $\mathbf{z}_{i+1} \gets \mathbf{z}_i - \beta \nabla_{\mathbf{z}} \mathcal{F}(\boldsymbol{\theta}_t, \mathbf{z}_i)$ \Comment{Inference (Eq.~\ref{pcns:eq:pc-infer})}
            \EndFor
            \State $\boldsymbol{\theta}_{t+1} \gets \boldsymbol{\theta}_t - \eta \nabla_{\boldsymbol{\theta}} \mathcal{F}(\boldsymbol{\theta}_t, \mathbf{z}_{N-1})$ \Comment{Learning (Eq.~\ref{pcns:eq:pc-learn})}
        \EndFor
    \end{algorithmic}
\end{algorithm}

It is not an understatement to say that this thesis focuses on understanding (and improving) these coupled optimisation problems (Eqs.~\ref{pcns:eq:pc-infer} \& \ref{pcns:eq:pc-learn}) when the energy parameterises standard DNNs. In particular, Chapters~\ref{ch:trust-region} \& \ref{ch:saddles} are about learning, while Chapter~\ref{ch:mupc} focuses on inference. It is important to note that previous attempts to understand PC relied mainly on a functional analysis of the energy \cite{millidge2022theoretical, alonso2022theoretical}, ignoring the rich structure of DNNs. As we will see in Chapters~\ref{ch:saddles} \& \ref{ch:mupc}, this structure is crucial for \textit{explaining, predicting and controlling} both the inference and learning dynamics of PCNs.

\paragraph{PCN testing.} PCNs can be tested in many different ways depending on the setting and task of interest. In any supervised setting (classification or generation), we can get a prediction for a given input with a forward pass in the same way as for BP. In addition, because PCNs implement a generative model, we can in principle clamp any part of the network and let it \textit{infer} or ``fill in'' the activities of all the nodes or layers left free to vary \cite{salvatori2022learning}. This can be done to complete masked images in memory association tasks, to infer a label given an image (and so allowing a single network to perform both generation and classification), or to infer some latent representation in an unsupervised setting \cite{van2024predictive, salvatori2023brain, millidge2022predictivereview, millidge2021predictive}. 

%% file: text/mainmatter/ch3-trust-region.tex
\chapter{Predictive Coding as Trust-region Optimisation}
\label{ch:trust-region}
\minitoc

\section{Abstract}
Predictive coding (PC) is a brain-inspired local learning algorithm that has recently been suggested to provide advantages over backpropagation (BP) in biologically relevant tasks. While theoretical work has mainly focused on the conditions under which PC can approximate or equal BP, how standard PC differs from BP is less well understood. Here, we develop a theory of PC as an approximate adaptive trust-region (TR) method that uses second-order information. We show that the weight gradient of PC can be interpreted as shifting the BP loss gradient towards a TR direction computed by the PC inference dynamics. Our theory suggests that PC should escape saddle points faster than BP, a prediction which we prove in a shallow linear model and support with experiments on deep networks. This work lays a theoretical framework for understanding other suggested benefits of PC.

\section{Introduction}
In recent years, there has been considerable effort in trying to find conditions under which predictive coding (PC) can reduce to backpropagation (BP). This work started with \citep{whittington2017approximation} showing that PC can approximate the gradients computed by BP on fully connected networks (or multi-layer perceptrons, MLPs) when the influence of the prior (input) is upweighted relative to the observations (output). \citep{millidge2022predictive} generalised this result to arbitrary computational graphs including convolutional and recurrent neural networks. A variation of PC, in which the weights are updated at precisely timed inference steps, was later shown to be equivalent to BP on MLPs \citep{song2020can}, a result further generalised by \citep{salvatori2021predictive} and \citep{rosenbaum2022relationship}. Finally, \citep{millidge2022backpropagation} provided a unification of these and other approximation results under certain equilibrium properties of energy-based models (EBMs).

On the other hand, the ways in which standard PC (without any modifications) differs from BP are much less understood. \citep{song2022inferring} proposed that PC, and EBMs more generally, implement a fundamentally different principle of credit assignment called ``prospective configuration''. According to this principle, neurons first change their activity to better predict the target and then update their weights to consolidate that activity pattern. This is in contrast to BP, where weights take precedence over activities. Based on a wide range of empirical results, \citep{song2022inferring} suggested that PC can provide a range of benefits over BP, including faster learning convergence and improved performance in more biologically realistic settings such as online and continual learning. 

Partly motivated by this conceptual principle, recent work has started to develop theories of standard PC. For example, \citep{millidge2022theoretical} showed (i) that in the linear case the PC inference equilibrium can be interpreted as an average of BP's forward pass values and the local targets computed by target propagation (TP) \citep{meulemans2020theoretical}, and (ii) that any critical point of the PC energy function is also a critical point of the BP loss. In the online setting (of data batches of size one), \citep{alonso2022theoretical} showed that PC approximates implicit gradient descent under specific rescalings of the layer activities and parameter learning rates. While I was writing the paper on which this chapter is based, \citep{alonso2023understanding} further showed that when that approximation holds, PC is sensitive to Hessian information for small learning rates. Despite these results, the fundamental relationship between standard PC and BP still remains to be fully elucidated.

Adding to this body of work, here we show that PC can be usefully understood as a form of an \textit{approximate adaptive trust-region} (TR) \textit{algorithm that exploits second-order information}. In particular, we show that the inference phase of PC can be thought of as solving a TR problem on the BP loss using a trust region defined by the Fisher information of the generative model (see \S \ref{ch3:pc-trust}). The PC weight gradient can then be interpreted as shifting the loss gradient computed by BP towards the TR inference solution. Our theory suggests that PC should escape saddles faster than BP, a well-known property of TR methods \citep{conn2000trust, dauphin2014identifying, yuan2015recent, levy2016power, murray2019revisiting}. We confirm this prediction in a toy model (\S \ref{ch3:toy}) and provide supporting experiments on deep networks (\S \ref{ch3:experiments}).

The rest of the chapter is structured as follows. After some relevant background on PC and TR methods (\S \ref{ch3:prelim}), we build some intuition for the differences between PC and BP by studying a toy model (\S \ref{ch3:toy}). Section~\ref{ch3:pc-trust} then presents our theoretical analysis of PC as a TR method, followed by some experiments consistent with the theory (\S \ref{ch3:experiments}). We conclude with the implications and limitations of this work (\S \ref{ch3:discussion}). Derivations, experiment details and supplementary figures are deferred to Appendix \ref{ch3:appendix}.

\section{Preliminaries}
\label{ch3:prelim}
For brevity, below we will use $\mathbf{g}_f(\mathbf{x})$ and $\matr{H}_f(\mathbf{x})$ to denote the gradient and Hessian, respectively, of some objective $f$ with respect to $\mathbf{x}$. We will omit their subscript and/or argument when clear from context.

\subsection{Predictive coding (PC)}
\label{ch3:pc}
We briefly recall relevant concepts and equations that were presented in detail in Chapter~\ref{ch:pcns}. PC networks (PCNs) are defined by an energy function $\mathcal{F}(\boldsymbol{\theta}, \mathbf{z})$ that depends on both the weights $\boldsymbol{\theta}$ and the activities $\mathbf{z}$ of the model. Note that below we will also sometimes refer to the weights as $\mathbf{w}$. Depending on the setting, different parts of the network are clamped to some data during training. Our theory will apply to arbitrary settings, but the experiments in \S \ref{ch3:experiments} will focus on the so-called ``discriminative'' setting, with images as inputs and labels as targets. To train a PCN, we minimise the energy in two separate phases, first with respect to the activities (inference) and then with respect to the weights (learning):
\begin{equation}
    \textit{Infer:  }
        \mathbf{z}^* = \argmin_{\mathbf{z}} \mathcal{F}(\boldsymbol{\theta}, \mathbf{z}),
    \label{ch3:eq:pc-infer}
\end{equation}
\begin{equation}
    \textit{Learn:  }
        \Delta\boldsymbol{\theta} \propto - \nabla_{\theta} \mathcal{F}(\boldsymbol{\theta}, \mathbf{z}^*).
    \label{ch3:eq:pc-learn}
\end{equation}
Note, importantly, that the aim is to update the weights at an equilibrium of the activities $\mathbf{z}^*$ (see \cite{salvatori2022incremental} for an exception). This optimisation cycle is repeated for multiple data batches until we are satisfied with the generalisation performance on some held-out samples.

\subsection{Trust region (TR) methods}
\label{ch3:tr-methods}
TR methods are often introduced as alternatives to ``line-search'' algorithms. Whereas line-search techniques such as gradient descent (GD) determine first a direction and then a step size (or learning rate), TR methods do the opposite. Namely, they begin by selecting a step (or region, known as the ``trust region'') and then optimise for the optimal direction within that region. More formally, given an objective $f(\boldsymbol{\theta}_t)$ we aim to minimise, a general TR problem \citep{conn2000trust, dauphin2014identifying, yuan2015recent} can be formulated as follows:
\begin{equation}
    \Delta \boldsymbol{\theta} = \argmin_{\Delta \boldsymbol{\theta}} \tilde{f}(\boldsymbol{\theta}_t) \quad \text{s.t.} \quad \Delta \boldsymbol{\theta}^T \matr{A} \Delta \boldsymbol{\theta} \leq p,
    \label{ch3:eq:tr-problem}
\end{equation}
where $\tilde{f}(\boldsymbol{\theta}_t)$ indicates different Taylor approximations of the objective, and $\matr{A}$ is some positive-definite matrix defining the norm or geometry of the trust region bounded by some radius $p$. Specific TR algorithms can be derived by (i) different approximations $\tilde{f}(\boldsymbol{\theta}_t)$, (ii) different geometries induced by $\matr{A}$, and by (iii) whether $\matr{A}$ depends on the current state of the parameters $\boldsymbol{\theta}_t$ and is therefore in some sense ``adaptive''. 

Line-search methods can be seen as special cases of TR problems \citep{conn2000trust}. For example, GD can be derived as a TR problem (Eq.~\ref{ch3:eq:tr-problem}) by assuming a linear approximation of the objective $\tilde{f}(\boldsymbol{\theta}_t) = f(\boldsymbol{\theta}_t) + \mathbf{g}^T \Delta \boldsymbol{\theta}$ and an Euclidean geometry (or $\ell^2$ penalty) given by $\matr{A} = \matr{I}$. Solving for the optimal parameter change gives the GD update $\Delta \boldsymbol{\theta}^* = - \alpha \mathbf{g}$, where the global learning rate is related to the trust region size $\alpha = \sqrt{p}/||\mathbf{g}||$. Note that this formulation also makes explicit that ``vanilla'' GD is a non-adaptive algorithm (unless some learning rate schedule with $\alpha_t$ is employed). Similarly, a damped or trust-region Newton (TRN) method can be obtained by using a quadratic approximation $\tilde{f}(\boldsymbol{\theta}_t) = f(\boldsymbol{\theta}_t) + \mathbf{g}^T \Delta \boldsymbol{\theta} + \Delta \boldsymbol{\theta}^T \matr{H} \Delta \boldsymbol{\theta}$, leading to the update $\Delta \boldsymbol{\theta}^* = - (\matr{H} + \frac{1}{\alpha} \matr{I})^{-1} \mathbf{g}$.

\section{A Toy Model} 
\label{ch3:toy}
\begin{figure}[t]
    \vskip 0.2in
    \begin{center}
        \centerline{\includegraphics[width=0.9\textwidth]{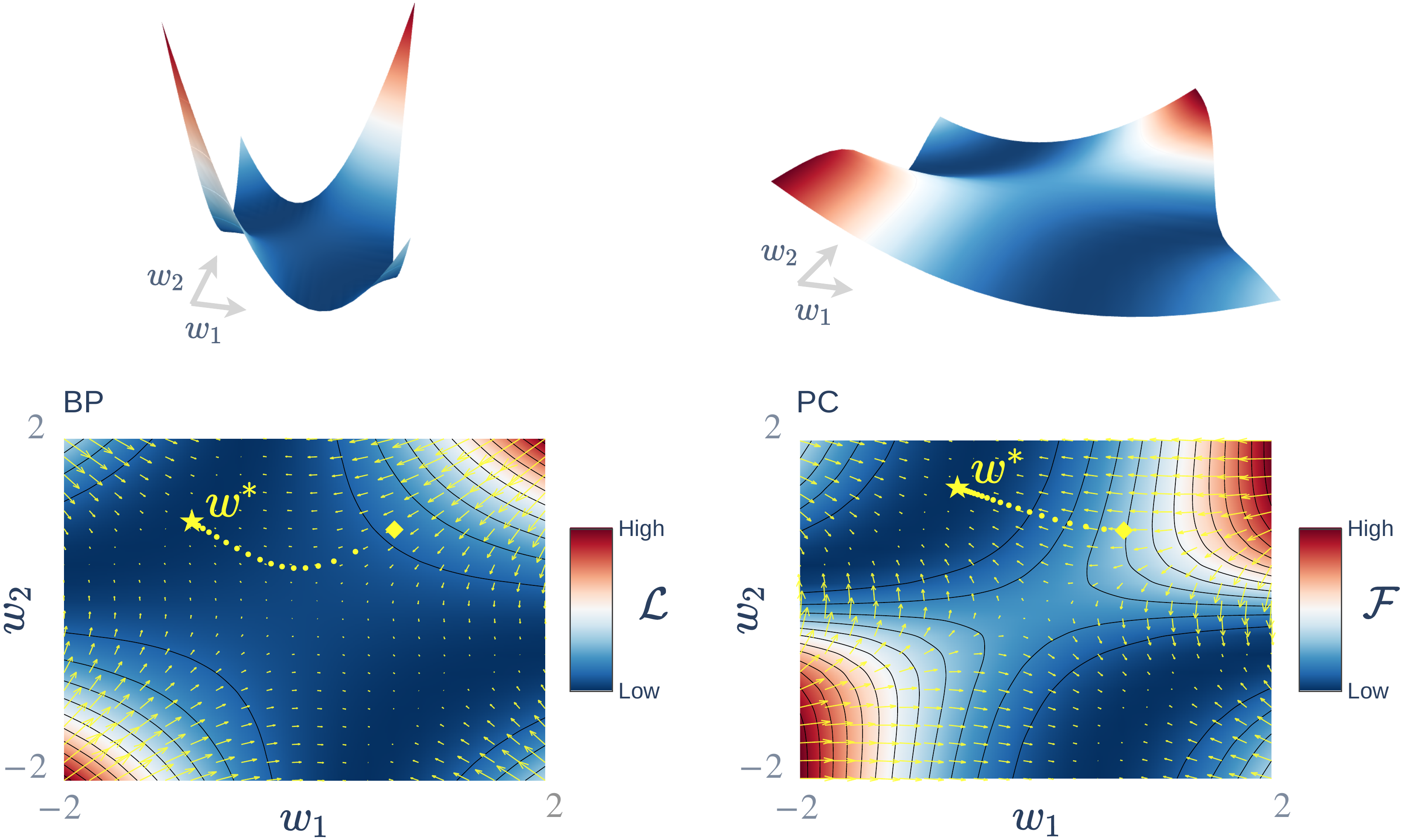}} %0.8
        \caption{\textbf{Landscape geometry and gradient descent dynamics of BP vs PC on a toy network.} Training loss and energy landscapes of an example 1MLP trained with BP (\textit{left}) and PC (\textit{right}), plotted both as surfaces (\textit{top}) and contours with superimposed gradient fields (\textit{bottom}). Surfaces are plotted at the same scale for comparison, and vector fields are standardised for visualisation (see \S \ref{ch3:toy-exp} for more details). The energy landscape of PC is plotted at the (approximate) inference equilibrium $\mathcal{F}|_{\nabla_z\mathcal{F} \approx 0}$ (see also Figure~\ref{ch3:fig:energy-land-infer-dynamics} for a visualisation of the landscape inference dynamics). Note that this is essentially the same plot as the left column of Figure~\ref{ch4:fig:toy-examples} in the next chapter.}
        \label{ch3:fig:toy-net}
    \end{center}
    \vskip -0.2in
\end{figure}
In this section, we study an MLP with a single linear hidden unit (1MLP) $f(x) = w_2w_1x$ as a toy model, allowing us to compare BP and PC exactly\footnote{In next chapter, we will see that with some extra effort we can perform this exact comparison for arbitrary linear networks.}. Figure~\ref{ch3:fig:toy-net} shows an example of the landscape geometry and GD dynamics of the 1MLP weights trained by BP and PC (for details, see \S \ref{ch3:toy-exp}). For BP, the landscape is simply the loss landscape, while the effective landscape on which PC learns is the energy landscape \textit{at the equilibrium of the states} or the inference equilibrium (Eq.~\ref{ch3:eq:pc-infer}).

Even in this simple setting, we can observe marked qualitative and quantitative differences between the two algorithms. In particular, PC seems to evade the saddle at the origin, taking a more direct path to the closest manifold of solutions. This is reflected in the geometry of the equilibrated energy landscape, which shows both a flatter ``trap'' direction leading to the saddle and a more negatively curved ``escape'' direction leading to a valley of solutions. For this toy model, it is straightforward to prove that, using (stochastic) GD (SGD), PC will escape this saddle faster than BP (Theorem~\ref{ch3:thm:saddle-escape}).

\begin{wrapfigure}{r}{0.5\textwidth}
    \vskip 0.1in
    \begin{minipage}{\linewidth}
        \centering
        \includegraphics[width=0.9\columnwidth]{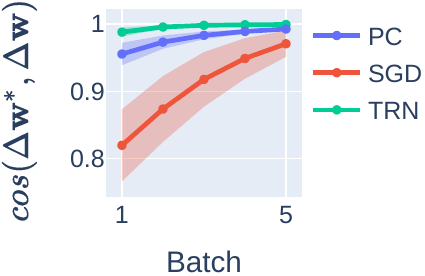} %0.7
        \caption{\textbf{The PC weight update direction is significantly closer to optimal than BP on 1MLPs.} For the first $5$ training batches, we plot the mean cosine similarity between the optimal weight direction $\Delta \mathbf{w}^*$ and the update $\Delta \mathbf{w}$ computed by (i) PC, $- \nabla_{\mathbf{w}} \mathcal{F}|_{\nabla_z \mathcal{F} \approx 0}$; (ii) BP with SGD, $- \nabla_{\mathbf{w}} \mathcal{L}$; and (iii) a trust-region Newton (TRN) method, $- (\matr{H} + \lambda \matr{I})^{-1}\nabla_{\mathbf{w}} \mathcal{L}$ with $\lambda = 2$. Shaded regions indicate the standard error of the mean (SEM) across 10 random weight initialisations.}
        \label{ch3:fig:cos-sims}        
    \end{minipage}
    \vskip -0.2in
\end{wrapfigure}

More generally, the gradient field of the equilibrated energy appears to be better aligned with the solutions than that of the loss. Indeed, Figure~\ref{ch3:fig:cos-sims} shows that on average the PC update points much closer and more reliably than BP to the optimal direction (i.e. towards the closest solution).

We also observe that the GD dynamics of PC seem to slow down near a minimum. In the 1MLP case, one can prove that this is because the manifold of minima of the equilibrated energy is \textit{flatter} than that of the loss (Theorem~\ref{ch3:thm:flat-min}). One implication is that during training PC will be more robust to weight perturbations near a minimum (see Figure~\ref{ch3:fig:weight-perturb}), which could be important in more biological, online settings.

To summarise, in this toy example we have shown that PC inference (Eq.~\ref{ch3:eq:pc-infer}) effectively reshapes the geometry of the weight landscape such that GD (i) escapes the origin saddle faster and (ii) takes longer to converge close to a minimum while being more robust to perturbations. Next, we develop a theory that helps to explain these findings. However, a much more precise and insightful explanation, as well as generalisation, of these observations will be presented in the next chapter.

\section{PC as an Approximate Second-order TR Method} 
\label{ch3:pc-trust}
Here we show that the inference phase of PC (Eq.~\ref{ch3:eq:pc-infer}) can be interpreted as solving a TR problem (Eq.~\ref{ch3:eq:tr-problem}) on the BP loss in activity space, while the learning phase (Eq.~\ref{ch3:eq:pc-learn}) essentially uses the TR solution to shift the GD direction of the weight update. To make this connection, we perform a second-order Taylor expansion of an arbitrary PC energy (e.g. see Eq.~\ref{pcns:eq:pc-energy}) centred around the feedforward pass values $\hat{\mathbf{z}}$ (see \S \ref{ch3:derivations} for a full derivation):
\begin{align}
    \mathcal{F}(\mathbf{z}) = \mathcal{L}(\hat{\mathbf{z}}) &+ \mathbf{g}_{\mathcal{L}}(\hat{\mathbf{z}})^T\Delta \mathbf{z} \nonumber \\
    &+ \frac{1}{2} \Delta \mathbf{z}^T \mathcal{I}(\hat{\mathbf{z}})\Delta \mathbf{z} + \mathcal{O}(\Delta \mathbf{z}^3)
    \label{ch3:eq:2-order-taylor-energy}
\end{align}
where $\Delta \mathbf{z} = (\mathbf{z} - \hat{\mathbf{z}})$, $\mathbf{g}_{\mathcal{L}}(\hat{\mathbf{z}})$ is the gradient of the loss with respect to the activities, and $\mathcal{I}(\hat{\mathbf{z}})$ is the Fisher information of the target given by the generative model $p(\mathbf{y} | \mathbf{z})$. This approximation allows us to characterise how (to second order) the PC energy diverges from the BP loss during inference. Indeed, a forward pass is in practice the most common method used to initialise the activities of PCNs for inference. We observe that Eq.~\ref{ch3:eq:2-order-taylor-energy} defines a TR problem (Eq.~\ref{ch3:eq:tr-problem}) in \textit{activity space} with a linear approximation of the loss plus an \textit{adaptive, second-order} geometry given by $\matr{A}=\mathcal{I}(\hat{\mathbf{z}})$. To second order, the solution to this TR problem (Eq.~\ref{ch3:eq:2-order-taylor-energy}) is given by
\begin{equation}
    \mathbf{z}^* \approx \hat{\mathbf{z}} - \mathcal{I}(\hat{\mathbf{z}})^{-1}g_{\mathcal{L}}(\hat{\mathbf{z}}).
    \label{ch3:eq:infer-sol}
\end{equation}
How does this TR solution found by the inference dynamics impact the weight gradient of PC and so its learning dynamics? Recall that in PC the weights are typically updated after the activities have converged (\S \ref{ch3:pc}). We therefore calculate the weight gradient of the energy evaluated at the approximate inference solution we just derived (see \S \ref{ch3:derivations}):
\begin{equation}
    \underbrace{\vphantom{\frac{\partial \mathcal{F}}{\partial \boldsymbol{\theta}}\biggr|_{\mathbf{z}^*}}\frac{\partial \mathcal{F}(\mathbf{z}^*)}{\partial \boldsymbol{\theta}}}_{\text{PC direction}} 
    \approx 
    \underbrace{\vphantom{\frac{\partial \mathcal{F}}{\partial \boldsymbol{\theta}}\biggr|_{\mathbf{z}^*}}\frac{\partial \hat{\mathbf{z}}}{\partial \boldsymbol{\theta}}\mathcal{I}(\hat{\mathbf{z}})^{-1}\mathbf{g}_{\mathcal{L}}(\hat{\mathbf{z}})}_{\text{TR direction}} 
    + 
    \underbrace{\vphantom{\frac{\partial \mathcal{F}}{\partial \boldsymbol{\theta}}\biggr|_{\mathbf{z}^*}}\mathbf{g}_{\mathcal{L}}(\boldsymbol{\theta})}_{\text{BP direction}},
    \label{ch3:eq:equilib-weight-grad-approx}
\end{equation}
where $\mathbf{g}_{\mathcal{L}}(\boldsymbol{\theta})$ is the loss gradient with respect to the weights, and $\partial \hat{\mathbf{z}} / \partial \boldsymbol{\theta}$ is a change of coordinates from activity to weight space. Thus, we see that the weight gradient on the equilibrated energy (Eq.~\ref{ch3:eq:equilib-weight-grad-approx}) effectively shifts the GD direction of the loss gradient in the direction of the TR inference solution (Eq.~\ref{ch3:eq:infer-sol}) mapped back into weight space. When $\mathcal{I}(\hat{\mathbf{z}})$ provides useful information, we can then intuitively think of the equilibrated energy landscape $\mathcal{F}(\boldsymbol{\theta}, \mathbf{z}^*)$ as a more ``trustworthy'' landscape---a landscape which should be easier to gradient descend---than the loss landscape.

We can gain some insight into the PC learning dynamics of Eq.~\ref{ch3:eq:equilib-weight-grad-approx} by considering the contribution of the Fisher information $\mathcal{I}(\hat{\mathbf{z}})$. For example, in directions of high Fisher information or model curvature (corresponding to directions of high latent variance), the PC weight gradient will be biased towards the TR solution. Interestingly, TR methods are known to be better at escaping saddles \citep{conn2000trust, dauphin2014identifying, yuan2015recent, levy2016power, murray2019revisiting}, which is exactly what we observe for the 1MLP model (\S \ref{ch3:toy}). We also find that the weight direction taken by PC is much closer to that of a TRN method than BP with GD (see Figure~\ref{ch3:fig:cos-sims}). In areas of low Fisher information, on the other hand, PC will tend to look more (but not exactly) like standard GD, since the curvature will not be zero (unless we are at a critical point where the gradient also vanishes). This is what we seem to observe in the 1MLP case near a minimum, where the model curvature does not seem to provide useful information and slows down convergence. Our theory, then, can be said to qualitatively recapitulate the landscape geometry and GD dynamics of PC in the 1MLP case (\S \ref{ch3:toy}).

\section{Experiments} 
\label{ch3:experiments}
This section reports some experiments consistent with the hypothesis, proved for 1MLPs (Theorem~\ref{ch3:thm:saddle-escape}) and suggested by our theoretical analysis of PC as a TR method (\S \ref{ch3:pc-trust}), that PC escapes saddles faster than BP when using (S)GD. 

\subsection{Deep chains} 
\label{ch3:deep-chains}
As a first step, we compared the loss dynamics of BP and PC on neural networks of unit width or ``deep chains'' $f(x) = w_L \phi_L(\dots \phi_1(w_1 x))$ trained on toy regression tasks (see \S\ref{ch3:chain-exps} for details). These simple networks are the ideal minimal case to test the hypothesis that PC escapes saddles faster than BP since the unit width keeps standard weight initialisations close to the saddle at the origin \cite[see ][]{orvieto2022vanishing}, and the degeneracy or flatness of this saddle grows with the number of hidden layers \cite{kawaguchi2016deep}. We will revisit these points in more detail in the next chapter. Since (S)GD is known to stall near saddles \citep{dauphin2014identifying, du2017gradient, jin2021nonconvex}, and many saddles grow flatter with the number of network layers \cite{achour2021loss}, we should expect the training dynamics of BP to slow down with depth, while PC should converge more quickly if it indeed avoids saddles faster.
\begin{figure}[t]
    \vskip 0.2in
    \begin{center}
        \centerline{\includegraphics[width=0.9\textwidth]{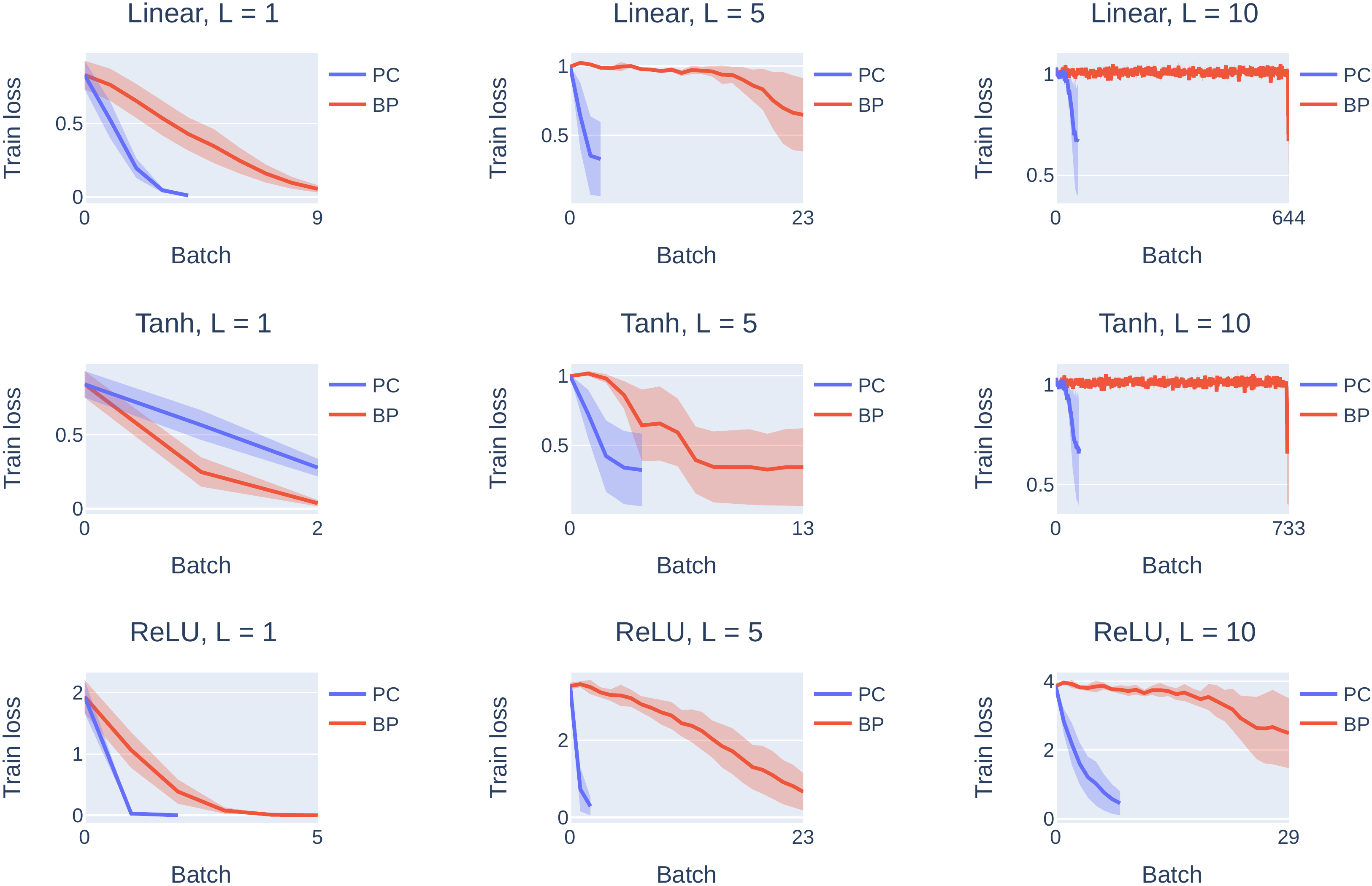}}  %0.8
        \caption{\textbf{PC can train deeper chains significantly faster than BP.} Mean training loss of 1D networks (deep chains) trained with BP and PC (see \ref{ch3:chain-exps} for details). Rows and columns indicate different activation functions (Linear, Tanh and ReLU) and number of hidden layers $H = \{1, 5, 10\}$, respectively. Each network type was optimised for learning rate, and training was terminated when the loss stopped decreasing. Shaded regions represent the SEM across 3 different initialisations.}
        \label{ch3:fig:chains}
    \end{center}
    \vskip -0.2in
\end{figure}

Following previous work \citep{alonso2022theoretical, song2022inferring}, for each experiment we performed a learning rate grid search to ensure that any differences in results were not due to inherently different optimal learning rates between PC and BP (see \S \ref{ch3:chain-exps}). Below, we plot the loss dynamics during training rather than testing because we are interested in the optimisation, as opposed to generalisation, dynamics. Nevertheless, the results do not significantly differ, and the test losses are reported in Figure~\ref{ch3:fig:chain-test-losses}.

Confirming our main prediction, we find that, with SGD PC can train deeper chains significantly faster than BP (Figure~\ref{ch3:fig:chains}). Note that training was terminated whenever the loss stopped decreasing. For linear and Tanh activations, we observe that BP’s convergence significantly slows down with more layers. We also see the emergence of phase transitions at increased depth, a phenomenon observed in the loss dynamics of deep linear networks \citep{saxe2013exact, jacot2021saddle}. Finally, we note that both BP and PC were unable to train very deep chains ($H = 15$), possibly due to vanishing/exploding gradients. We will revisit this point in Chapter~\ref{ch:mupc}.

\subsection{Deep and wide networks} 
\label{ch3:dnns}
Next, we compared PC and BP on wide, as well as deep, fully connected networks $f(\mathbf{x}) = \matr{W}_L \phi_L(\dots \phi_1(\matr{W}_1 \mathbf{x}))$. Wide networks introduce many more saddles due to, for example, the permutation symmetries between hidden units \citep{bishop2006pattern, brea2019weight, simsek2021geometry}. In particular, the network output is invariant to swapping any two neurons in the same layer (or equivalently, their incoming and outgoing weights). Note, however, that wide networks have many other symmetries and associated saddles \cite{achour2021loss, ziyin2025parameter, martinelli2025flat}, to which we will return in the next chapter.

\begin{figure}[H]
    \vskip 0.2in
    \begin{center}
        \centerline{\includegraphics[width=\textwidth]{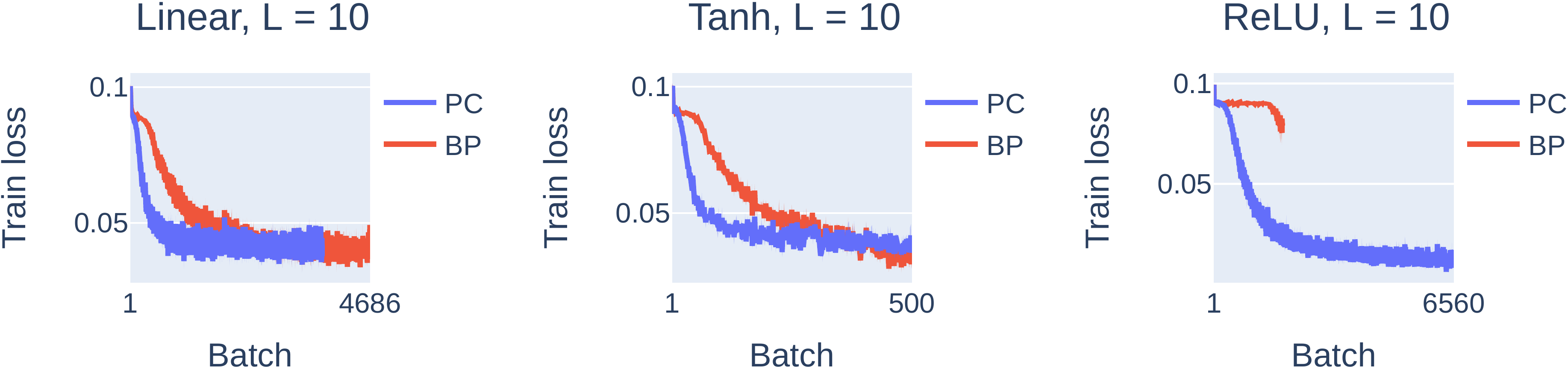}}
        \caption{\textbf{Faster convergence of PC in deep and wide networks trained on MNIST.} Mean training loss of deep ($H = 10$) and wide ($N = 500$) networks trained to classify MNIST for 3 random initialisation (see \ref{ch3:chain-exps} for details). As for Figure Figure~\ref{ch3:fig:chain-test-losses}, training was terminated whenever the loss stopped decreasing. SEMs are not visible.}
        \label{ch3:fig:dnns}
    \end{center}
    \vskip -0.2in
\end{figure}

We trained 10-layer networks of width $N_1 = \dots = N_{L-1} = 500$ to classify MNIST digits (see \S \ref{ch3:dnn-exps}) and found speed-ups for PC similar to those observed in deep chains for all the activation functions tested (Figure~\ref{ch3:fig:dnns}).

\section{Discussion} 
\label{ch3:discussion}
In summary, we showed that PC can be cast as an approximate adaptive trust-region method that exploits second-order information, despite explicitly using only first-order updates. 

\subsection{Implications}
Our theory suggested that PC should escape saddle points faster than BP with SGD, a prediction which we verified in a toy model and supported with experiments on deep networks. These results are consistent with previously reported speed-ups of PC over BP \citep{song2022inferring, alonso2022theoretical}. For example, \citep{song2022inferring} found that PC converged much faster than BP on a 15-layer, LeakyReLU network ($N = 64$) trained on Fashion-MNIST with Adam. In the online setting (of batch size 1), \citep{alonso2022theoretical} found similar speed-ups for relatively shallower ($L = 3$) and wider ($N = 1024$) ReLU networks trained to classify and reconstruct CIFAR-10. Our theory provides a potential explanation for these results in terms of faster saddle escape. The next chapter will formalise, as well as nuance, this prediction.

More generally, our results suggest that the second-order information used by PC contains information about the curvature of the loss landscape. Related, \citep{alonso2023understanding} showed that PC approximates TRN in the online learning setting. Note, however, that our theory is independent of batch size, and the empirical results suggest that PC exploits second-order information for large batches too. Nevertheless, the next chapter will expose the limitations of this theory, as we discuss below.

Although we did not explore this, our theory can also recover previous approximation results to BP and TP relying on the ratio of bottom-up vs top-down information \citep{whittington2017approximation, millidge2022theoretical}. In particular, manipulating this ratio can be seen as adjusting different axes of the trust region or, equivalently, per-parameter learning rates (see Figure~\ref{ch3:fig:energy-land-precision-ratio} for an illustration). Indeed, because of the duality between TR and line-search methods \citep{conn2000trust}, our theory admits an alternative interpretation of PC as an adaptive gradient method, conceptually similar to state-of-the-art deep learning optimisers such as Adam \citep{kingma2014adam}. Notably, adaptive methods have also been shown to escape saddle points faster than standard SGD \citep{staib2019escaping, orvieto2022vanishing}.

Recent work by \citep{pogodin2023synaptic} suggests that our theory could be potentially tested against biological data. The authors showed that under certain assumptions the geometry of weight updates can be inferred from the weight distributions, and suggested that an Euclidean geometry (as defined by standard GD) is inconsistent with the empirically observed log-normal distributions of synaptic weights. This is in line with our result that PC uses a non-Euclidean (natural) geometry, with the Fisher information as the metric. To distinguish between different non-Euclidean geometries, however, experimental data both before and after learning seems to be needed, since \citep{pogodin2023synaptic} showed that different geometries can lead to the same post-learning distribution depending on the pre-learning distribution.

Related, our study speaks to the question of whether the brain may approximate GD. It appears to be a widely accepted belief that the brain estimates gradients on some objective or loss function \citep{marblestone2016toward, richards2019deep, lillicrap2020backpropagation, hennig2021learning, richards2023study}. \citep{richards2023study} suggest that this claim could be experimentally tested by looking at how synaptic changes following learning on some task correlate with the true gradient of some loss for that task. Whether or not PC is a good model of learning in the brain, our results show that \textit{first-order, gradient updates on a sum of local objectives} (in this case the PC energy) \textit{can lead to second-order updates on a global objective}. This raises the possibility that the brain could use curvature information of the loss by still doing GD, but on a sum of local objectives. If so, synaptic changes may not correlate with the loss gradient and should also be compared with second-order updates.

Finally, our theory can be seen as an important step in providing a more solid theoretical footing to the conceptual principle of ``prospective configuration'' \citep{song2022inferring} and its associated empirical benefits. It could be interesting to extend this framework to explain, and perhaps uncover, other advantages and disadvantages of PC, such as robustness to small batch sizes and reduced weight interference. However, in the following chapter we will argue that any serious theory of the inference and learning dynamics of PCNs should take into account the rich architectural structure of neural networks.

\subsection{Limitations}
As alluded to above, one important limitation of our theory is that it is only valid to a second-order approximation (Eq.~\ref{ch3:eq:2-order-taylor-energy}). Indeed, in the next chapter we will show that PC not only in fact uses curvature information about the loss landscape but also arbitrarily \textit{higher-order} information. Another weakness of the theory is that, while applying to arbitrary energy functions, it does not take into account the structure or architecture of the network, which the next chapter will show to be crucial. In addition, while this work highlights the potential benefits of PC's inference scheme, its computational cost remains a major limitation, making it orders of magnitude more expensive than BP (at least on standard GPUs). Our results can be seen as explaining this high inference cost by revealing the implicit computation and inversion of a Fisher matrix. In this respect, we note that amortised PC schemes have been developed \citep{tscshantz2023hybrid}, and future work could investigate whether the benefits of iterative inference can be retained with amortisation.

\section*{Author contributions}
FI conceptualised the study, proved the toy model results, identified the connection with trust-region methods, ran all the experiments, and wrote the paper. RS helped with the development of the theory in \S \ref{ch3:pc-trust}. CLB contributed to conceptual discussions and supervised the project.

%% file: text/mainmatter/ch4-saddles.tex
\chapter{On the Geometry of the Energy Landscape of PCNs}
\label{ch:saddles}
\minitoc

\begin{quote}
    “Equations are just the boring part of mathematics. I attempt to see things in terms of geometry.”

    \hfill --- Stephen Hawking
\end{quote}

\section{Abstract}
Predictive coding (PC) is an energy-based learning algorithm that performs iterative inference over network activities before updating weights. Recent work suggests that PC can converge in significantly fewer learning steps than backpropagation thanks to its inference procedure. However, these advantages are not always observed, and the impact of PC inference on learning is not theoretically well understood. To address this gap, we study the geometry of the effective landscape on which PC learns: the weight landscape at the inference equilibrium of the network activities. For deep linear networks, we first show that the equilibrated PC energy is equal to a rescaled mean squared error loss with a weight-dependent rescaling. We then prove that many highly degenerate (non-strict) saddles of the loss including the origin become much easier to escape (strict) in the equilibrated energy. Experiments on both linear and non-linear networks strongly validate our theory and further suggest that all the saddles of the equilibrated energy are strict. Overall, this work shows that PC inference makes the loss landscape of feedforward networks more benign and robust to vanishing gradients, while also highlighting the fundamental challenge of scaling PC to very deep models.

\section{Introduction} 
As reviewed in Chapter~\ref{ch:pcns}, in contrast to backpropagation (BP), predictive coding (PC) performs iterative inference over network activities before weight updates. While this inference process incurs an additional computational cost, it has been suggested to provide many benefits, including faster learning convergence as we saw in the previous chapter \citep{song2022inferring, alonso2022theoretical, innocenti2023understanding}. However, these speed-ups are not consistently observed across datasets, models and optimisers \citep{alonso2022theoretical}, and the impact of PC inference on learning more generally is not theoretically well understood (see \S \ref{ch4:pc-theory} for a review of related work).

To address this gap, here we study the geometry of the effective landscape on which PC learns: \textit{the weight landscape at the inference equilibrium of the network activities} (defined in \S\ref{ch4:pc}). Our theory considers deep linear networks (DLNs), the standard model for theoretical studies of the loss landscape (see \S\ref{ch4:related-work}). Despite being able to learn only linear representations, DLNs have non-convex loss landscapes with non-linear learning dynamics that have proved to be a useful model for understanding non-linear networks \citep[e.g.][]{saxe2013exact}. In contrast to previous theories of PC \citep{alonso2022theoretical, alonso2023understanding, innocenti2023understanding}, we do not make any additional assumptions or approximations (again see \S\ref{ch4:related-work}), and perform exhaustive experiments to verify that our linear theory holds for non-linear networks. 

For DLNs, we first show that, at the inference equilibrium, the PC energy is equal to a rescaled mean squared error (MSE) loss with a non-trivial, weight-dependent rescaling (Theorem~\ref{ch4:thm1}). We then compare saddle points of the loss, which have been recently characterised \citep{kawaguchi2016deep, achour2021loss}, to those of the equilibrated energy. Such saddles, which are ubiquitous in the loss landscape of neural networks \citep{dauphin2014identifying, achour2021loss}, can be of two main types: ``strict'' (Def. \ref{ch4:def:strict-saddle}), with negative curvature; and ``non-strict'', where an escape direction is found in higher-order derivatives \citep{ge2015escaping, kawaguchi2016deep, achour2021loss}. Non-strict saddles are particularly problematic for first-order methods like (stochastic) gradient descent (SGD) since they are by definition at least second-order critical points. While SGD can be exponentially slowed in the vicinity of strict saddles \citep{du2017gradient}, it can effectively get stuck in non-strict ones \citep{sankar2018saddles, bottcher2024visualizing}. This is the phenomenon of vanishing gradients viewed from a landscape perspective \citep{orvieto2022vanishing, bengio1994learning}.

By contrast, here we prove that many non-strict saddles of the MSE loss, specifically saddles of rank zero, become strict in the equilibrated energy of any DLN (Theorems~\ref{ch4:thm2}-\ref{ch4:thm3}). These saddles include the origin, whose degeneracy (or flatness) in the loss grows linearly with the number of hidden layers. Our theoretical results are strongly validated by experiments on both linear and non-linear networks, and additional experiments suggest that other (higher-rank) non-strict saddles of the loss become strict in the equilibrated energy. Based on these results, we conjecture that all the saddles of the equilibrated energy are strict. Overall, this work suggests that PC inference makes the loss landscape of feedforward networks more benign and robust to vanishing gradients.

The rest of the chapter is structured as follows. After introducing the setup (\S \ref{ch4:prelim}), we present our theoretical results for DLNs (\S \ref{ch4:theory}), including some illustrative examples and thorough empirical verifications of each result. Section~\ref{ch4:experiments} then reports experiments on non-linear networks supporting our theory and more general conjecture. We conclude by discussing the implications and limitations of our work, as well as potential future directions (\S\ref{ch4:discussion}). Appendix \ref{ch4:appendix} includes a review of related work, derivations, experiment details and supplementary results. Code to reproduce all the experiments is available at \sloppy{\url{https://github.com/francesco-innocenti/pc-saddles}}.

\subsection{Summary of contributions} 
\begin{itemize}
    \item We derive an exact solution for the PC energy of DLNs at the inference equilibrium (Theorem~\ref{ch4:thm1}), which turns out to be a rescaled MSE loss with a weight-dependent rescaling. This corrects a previous mistake in the literature that the MSE loss is equal to the output energy \citep{millidge2022theoretical}, while enabling further studies of the PC energy landscape. We find an excellent match between our theory and experiment (Figure~\ref{ch4:fig:dln-equilib-energy}).
    \item Based on this result, we prove that, in contrast to the MSE, the origin of the equilibrated energy of DLNs is a strict saddle independent of network depth. We provide an explicit characterisation of the Hessian at the origin of the equilibrated energy (Theorem~\ref{ch4:thm2}), which is perfectly validated by experiments on linear networks (Figures~\ref{ch4:fig:hess-origin-toy}-\ref{ch4:fig:hess-origin-mnist} \& \ref{ch4:fig:hess-origin-toy-chain}).
    \item We further prove that other non-strict saddles of the MSE than the origin, specifically saddles of rank zero, become strict in the equilibrated energy of DLNs (Theorem~\ref{ch4:thm3}). We provide an empirical verification of one of these saddles as an example (Figures~\ref{ch4:fig:hess-other-saddle-toy}-\ref{ch4:fig:hess-other-saddle-mnist}).
    \item We empirically show that our linear theory holds for non-linear networks, including convolutional architectures, trained on standard image classification tasks. In particular, when initialised close to non-strict saddles of the MSE covered by Theorem~\ref{ch4:thm3}, we find that SGD on the equilibrated energy escapes much faster than on the loss given the same learning rate (Figures~\ref{ch4:fig:origin-escape} \& \ref{ch4:fig:other-saddle-escape}). In contrast to BP, PC exhibits no vanishing gradients (Figure~\ref{ch4:fig:origin-grad-norms}).
    \item We perform additional experiments, again on both linear and non-linear networks, showing that PC quickly escapes other (higher-rank) non-strict saddles of the MSE that we do not address theoretically (Figure~\ref{ch4:fig:matrix-completion}), supporting our conjecture that all the saddles of the equilibrated energy are strict.
\end{itemize}

\section{Preliminaries}
\label{ch4:prelim}
\paragraph{Notation.} We use the following shorthand $\matr{W}_{k:\ell} = \matr{W}_k \dots \matr{W}_\ell$ for $\ell, k \in 1,\dots, L$, denoting the total product of weight matrices as $\matr{W}_{L:1} = \matr{W}_L \dots \matr{W}_1$. For the identity matrix $\matr{I}_n$ of size $n \times n$ and the zero vector or null matrix $\mathbf{0}_n$, $n$ will be omitted when clear from context. $||\cdot||$ always denotes the $\ell_2$ norm, and $\otimes$ is the Kronecker product between two matrices. We will consider the gradient and Hessian of an objective $f$ only with respect to the network weights $\boldsymbol{\theta}$ and sometimes abbreviate them as $\matr{g}_f \coloneq \nabla_{\boldsymbol{\theta}} f$ and $\matr{H}_f \coloneq \nabla^2_{\boldsymbol{\theta}} f$, respectively, omitting the independent variable for simplicity. The largest and smallest eigenvalues of the Hessian are $\lambda_{\text{max}}(\matr{H}_f)$ and $\lambda_{\text{min}}(\matr{H}_f)$, with $\hat{\mathbf{v}}_{\text{max}}$ and $\hat{\mathbf{v}}_{\text{min}}$ as their associated eigenvectors. See \S\ref{ch4:notation} for more general notation.

\begin{definition}
\label{ch4:def:strict-saddle}
\textit{Strict saddle.} Following \citep{ge2015escaping} and later work, any critical point $\boldsymbol{\theta}^*$ of $f(\boldsymbol{\theta})$ where $\mathbf{g}_f (\boldsymbol{\theta}^*) = \mathbf{0}$ is defined as a strict saddle when the Hessian at that point has at least one positive $\lambda_{\text{max}}(\matr{H}_f(\boldsymbol{\theta}^*)) > 0$ and one negative eigenvalue $\lambda_{\text{min}}(\matr{H}_f(\boldsymbol{\theta}^*)) < 0$. Any other critical point with a positive semi-definite Hessian and at least one negative eigenvalue in a higher-order derivative is said to be a non-strict saddle.
\end{definition}

\subsection{Deep Linear Networks (DLNs)} 
\label{ch4:dln}
We consider DLNs with one or more hidden layers $H = L-1 \geq 1$ defining the linear mapping $\matr{W}_{L:1}: \mathbb{R}^{N_0} \rightarrow \mathbb{R}^{N_L}$ where $\matr{W}_\ell \in \mathbb{R}^{N_\ell \times N_{\ell-1}}$, with layer widths $\{N_\ell\}_{\ell=0}^L$. We ignore biases for simplicity. The standard MSE loss for DLNs can then be written as
\begin{equation}
    \mathcal{L} = \frac{1}{2B}\sum_{i=1}^{B}||\mathbf{y}_i - \matr{W}_{L:1} \mathbf{x}_i||^2
    \label{ch4:eq:mse-loss}
\end{equation}
for a dataset of $B$ examples $\{(\mathbf{x}_i, \mathbf{y}_i)\}_{i=1}^N$ where $\mathbf{x} \in \mathbb{R}^{N_0}$ and $\mathbf{y} \in \mathbb{R}^{N_L}$. The total number of weights is given by $p = \sum_{\ell=1}^L N_\ell N_{\ell-1}$, and we will denote the set of all network parameters as $\boldsymbol{\theta} \in \mathbb{R}^p$. For brevity, we will often refer to the MSE loss as simply the loss.

\subsection{Predictive coding (PC)}
\label{ch4:pc}
As reviewed in detail in Chapter~\ref{ch:pcns}, PC networks (PCNs) minimise an energy function $\mathcal{F}(\boldsymbol{\theta}, \mathbf{z})$ that depends on both the weights $\boldsymbol{\theta}$ and the activities $\mathbf{z}$ of the model. For DLNs, the PC energy reduces to
\begin{equation}
    \mathcal{F} = \frac{1}{2B}\sum_{i=1}^{B} \sum_{\ell=1}^L ||\mathbf{z}_{\ell, i} - \matr{W}_\ell \mathbf{z}_{\ell-1, i}||^2.
    \label{ch4:eq:pc-energy}
\end{equation}
To train a PCN, the last layer is clamped to some data $\mathbf{z}_{L, i} \coloneq \mathbf{y}_i$, which could be a label for classification or an image for generation. In a supervised task, the first layer is also fixed to some input, $\mathbf{z}_{0, i} \coloneq \mathbf{x}_i$. Our theory will apply to any supervised setting (discriminative or generative), but for simplicity the experiments in \S \ref{ch4:experiments} will focus on discriminative tasks. The energy (Eq.~\ref{ch4:eq:pc-energy}) is minimised first with respect to the activities (inference), and then with respect to the weights (learning):
\begin{equation}
    \textit{Infer:  }
        \mathbf{z}^* = \argmin_{\mathbf{z}} \mathcal{F}(\boldsymbol{\theta}, \mathbf{z}),
    \label{ch4:eq:pc-infer}
\end{equation}
\begin{equation}
    \textit{Learn:  }
        \Delta\boldsymbol{\theta} \propto - \nabla_{\theta} \mathcal{F}(\boldsymbol{\theta}, \mathbf{z}^*),
    \label{ch4:eq:pc-learn}
\end{equation}
where we omit the data index $i$ for simplicity. As highlighted in the previous chapter, the effective landscape on which PC learns is the energy at the inference equilibrium of the network activities $\mathcal{F}(\boldsymbol{\theta}, \mathbf{z}^*)$, which we will refer to as the equilibrated energy or sometimes simply the energy.
\begin{figure}[t]
    \begin{center}
        \centerline{\includegraphics[width=\textwidth]{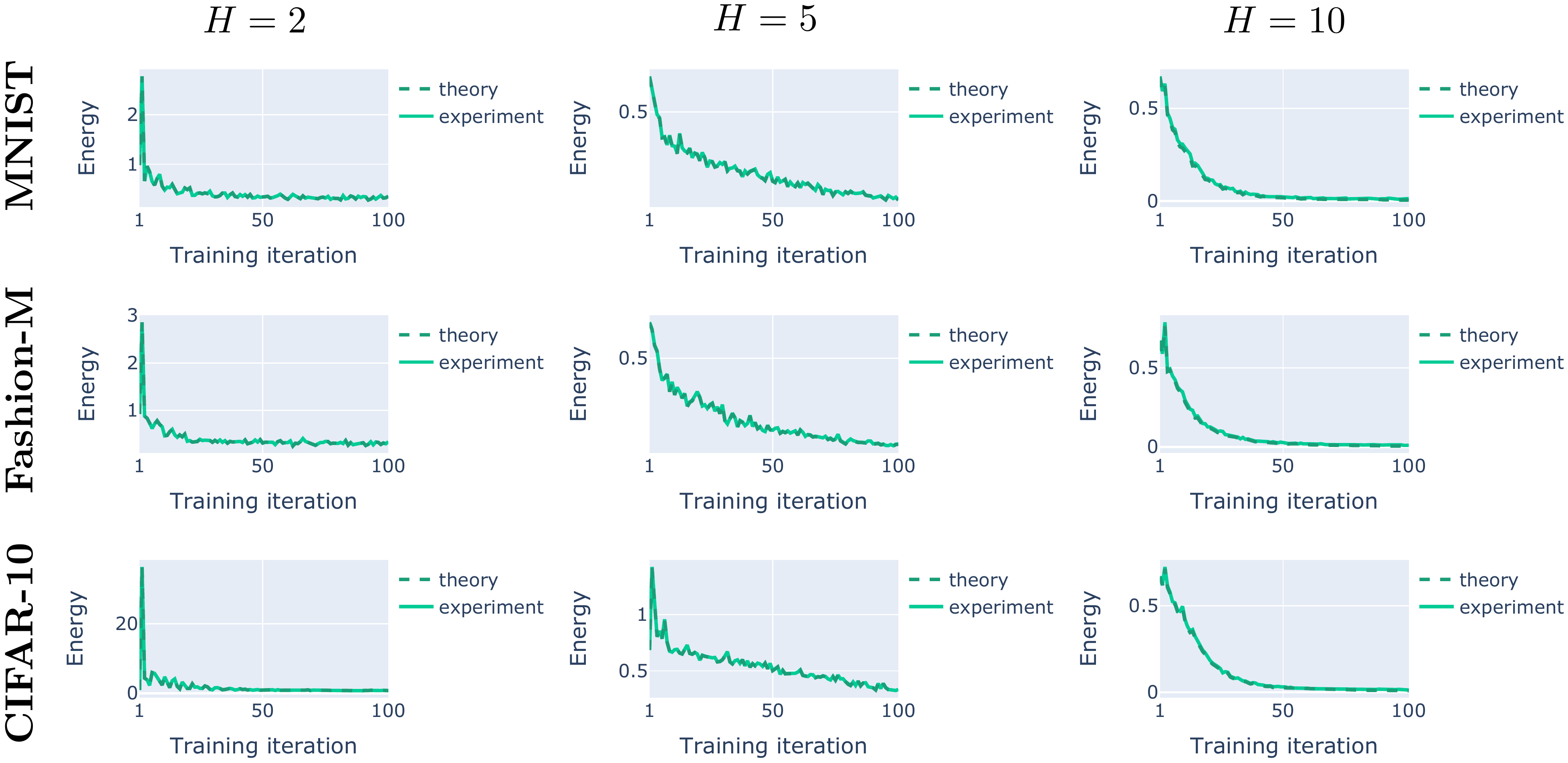}}
        \caption{\textbf{Empirical verification of the theoretical equilibrated energy of deep linear networks (Theorem~\ref{ch4:thm1}).} For different datasets, we plot the energy (Eq.~\ref{ch4:eq:pc-energy}) at the numerical inference equilibrium $\mathcal{F}|_{\nabla_{\mathbf{z}}\mathcal{F} \approx 0}$ for DLNs with different number of hidden layers $H \in \{2, 5, 10\}$ (see \S\ref{ch4:exp-details} for more details), observing an excellent match with the theoretical prediction (Eq.~\ref{ch4:eq:dln-equilib-energy}).}
        \label{ch4:fig:dln-equilib-energy}
    \end{center}
    \vskip -0.2in
\end{figure}

\section{Theoretical results} 
\label{ch4:theory}
\subsection{Equilibrated energy as rescaled MSE} 
As reviewed in \S \ref{ch4:pc}, the weights of a PCN are typically updated once the activities have converged to an equilibrium. The equilibrated energy $\mathcal{F}(\boldsymbol{\theta}, \mathbf{z}^*)$, which we will abbreviate as $\mathcal{F}^*(\boldsymbol{\theta})$, is therefore the \textit{effective weight landscape} navigated by PC and the object we are interested in studying. It turns out that we can derive a closed-form solution for the equilibrated energy of DLNs, which will form the basis of our subsequent results.
\begin{tcolorbox}[width=\linewidth, sharp corners=all, colback=white!95!black, colframe=white!95!black]
    \begin{mythm}{3.1}[Equilibrated energy of DLNs]\label{ch4:thm1}
        For any DLN parameterised by $\boldsymbol{\theta} \coloneq \vect(\matr{W}_1, \dots, \matr{W}_L)$ with input and output $(\mathbf{x}_i, \mathbf{y}_i)$, the PC energy (Eq.~\ref{ch4:eq:pc-energy}) at the exact inference equilibrium $\partial \mathcal{F}/\partial \mathbf{z} = \mathbf{0}$ is equal to the following rescaled MSE loss (see \S\ref{ch4:equilib-energy} for derivation)
            \begin{equation}
                \mathcal{F}^* = \frac{1}{2B} \sum_{i=1}^B (\mathbf{y}_i - \matr{W}_{L:1}\mathbf{x}_i)^T \matr{S}^{-1}(\mathbf{y}_i - \matr{W}_{L:1}\mathbf{x}_i)
                \label{ch4:eq:dln-equilib-energy}
            \end{equation}
        where the rescaling is $\matr{S}(\boldsymbol{\theta}) = \matr{I}_{N_L} + \sum_{\ell=2}^L (\matr{W}_{L:\ell})(\matr{W}_{L:\ell})^T$.
    \end{mythm}
\end{tcolorbox}
The proof relies on unfolding the hierarchical Gaussian model assumed by PC to work out an implicit generative model of the output, and the rescaling $\matr{S}(\boldsymbol{\theta})$ comes from the variance modelled by PC at each layer (see \S\ref{ch4:equilib-energy} for details). Figure~\ref{ch4:fig:dln-equilib-energy} shows an excellent empirical validation of the theory.

Intuitively, the PC inference process (Eq.~\ref{ch4:eq:pc-infer}) can then be thought of as reshaping the (MSE) loss landscape to take some layer-wise, weight-dependent variance into account. This immediately raises the question: how does the equilibrated energy landscape $\mathcal{F}^*(\boldsymbol{\theta})$ differ from the loss landscape $\mathcal{L}(\boldsymbol{\theta})$? Is the rescaling---and so the layer variance modelled by PC---useful for learning? Below we provide a partial positive answer to this question by comparing the geometry of saddle points of the two objectives.

\subsection{Analysis of the origin saddle} 
\label{ch4:origin-saddle}
\begin{figure}[h]
    \begin{center}
        \centerline{\includegraphics[width=0.8\textwidth]{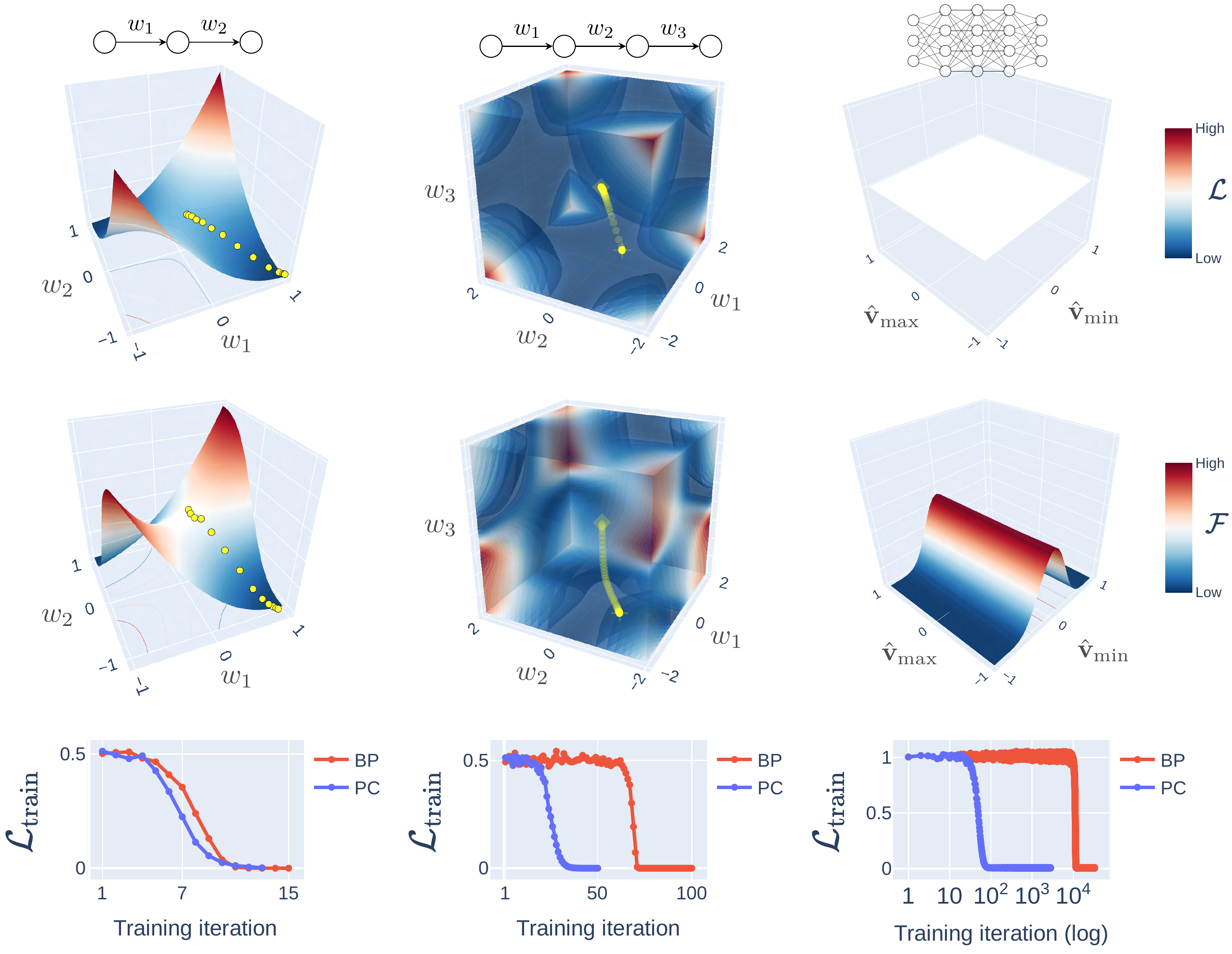}}
        \caption{\textbf{Toy examples illustrating the (Theorem~\ref{ch4:thm2}) result that the saddle at the origin of the equilibrated energy is strict independent of network depth.} We plot the MSE loss $\mathcal{L}(\boldsymbol{\theta})$ (\textit{top}) and equilibrated energy landscape $\mathcal{F}^*(\boldsymbol{\theta})$ (\textit{middle}) around the origin for 3 linear networks trained with SGD on a toy problem (see \S\ref{ch4:exp-details} for details). We also show the training losses for a representative run with initialisation close to the origin (\textit{bottom}). For the one-dimensional networks, we visualise the landscape around the origin as well as the SGD updates. For the wide network, we project the landscape onto the maximum and minimum eigenvectors of the Hessian, following \citep{bottcher2024visualizing}. Note that in this case the projection of the loss is flat because the Hessian at the origin is null for $H > 1$ (Eq.~\ref{ch4:eq:loss-hess-origin}).}
        \label{ch4:fig:toy-examples}
    \end{center}
    \vskip -0.2in
\end{figure}

Here we prove that, in contrast to the MSE loss, the origin of the equilibrated energy (Eq.~\ref{ch4:eq:dln-equilib-energy} where all the weights are zero $\boldsymbol{\theta} = \mathbf{0}$) is a strict saddle (Def. \ref{ch4:def:strict-saddle}) for DLNs of any depth. To do so, we derive an explicit expression for the Hessian at the origin of the equilibrated energy. For comparison, we first briefly recall the known results that, at the origin, the loss Hessian is indefinite for one-hidden-layer networks and null for any deeper network (see \S\ref{ch4:loss-hessian} for a re-derivation)
\begin{align}
    \matr{H}_{\mathcal{L}} (\boldsymbol{\theta} = \mathbf{0}) = 
    \begin{cases} 
    \begin{bmatrix} \mathbf{0} & - \widetilde{\matr{\Sigma}}_{\mathbf{xy}} \otimes \matr{I}_{N_1} \\ -\matr{I}_{N_1} \otimes \widetilde{\matr{\Sigma}}_{\mathbf{yx}} & \mathbf{0} \end{bmatrix}, & H = 1 \\ \\
    \mathbf{0}_p, & H > 1
    \end{cases},
    \label{ch4:eq:loss-hess-origin}
\end{align}
where following previous works $\widetilde{\matr{\Sigma}}_{\mathbf{xy}} \coloneq \frac{1}{B} \sum_i^B \mathbf{x}_i\mathbf{y}_i^T$ is the empirical input-output covariance. One-hidden-layer networks $H = 1$ are a special case where the origin saddle of the loss is strict (Def. \ref{ch4:def:strict-saddle}) and was studied in detail by \citep{saxe2013exact} (see left panel of Figure~\ref{ch4:fig:toy-examples} for an example). For deeper networks $H > 1$, the saddle is non-strict as first shown by \citep{kawaguchi2016deep}:
\begin{align}
    \begin{cases} 
    \lambda_{\text{min}}(\matr{H}_{\mathcal{L}}(\boldsymbol{\theta}=\mathbf{0})) < 0, & H = 1 \quad \text{[strict saddle]}  \\ \\
    \lambda_{\text{min}}(\matr{H}_{\mathcal{L}}(\boldsymbol{\theta}=\mathbf{0})) = 0, & H > 1 \quad \text{[non-strict saddle]}
    \end{cases}.
    \label{ch4:eq:loss-hess-origin-min-eigen}
\end{align}
More specifically, the origin saddle of the loss is of order $H$\footnote{The $n$th-order of a saddle simply indicates the ($n$th+1) derivative where the first negative (escape) direction is found. So, for example, a first-order (strict) saddle has a zero gradient and an indefinite Hessian, while a second-order (non-strict) saddle has a null Hessian but a third derivative with a negative direction.}, becoming increasingly degenerate (or flat) and harder to escape with depth, especially for first-order methods like SGD (see the middle and right panels of Figure~\ref{ch4:fig:toy-examples}). 
\begin{figure}[t]
    \begin{center}
        \centerline{\includegraphics[width=\textwidth]{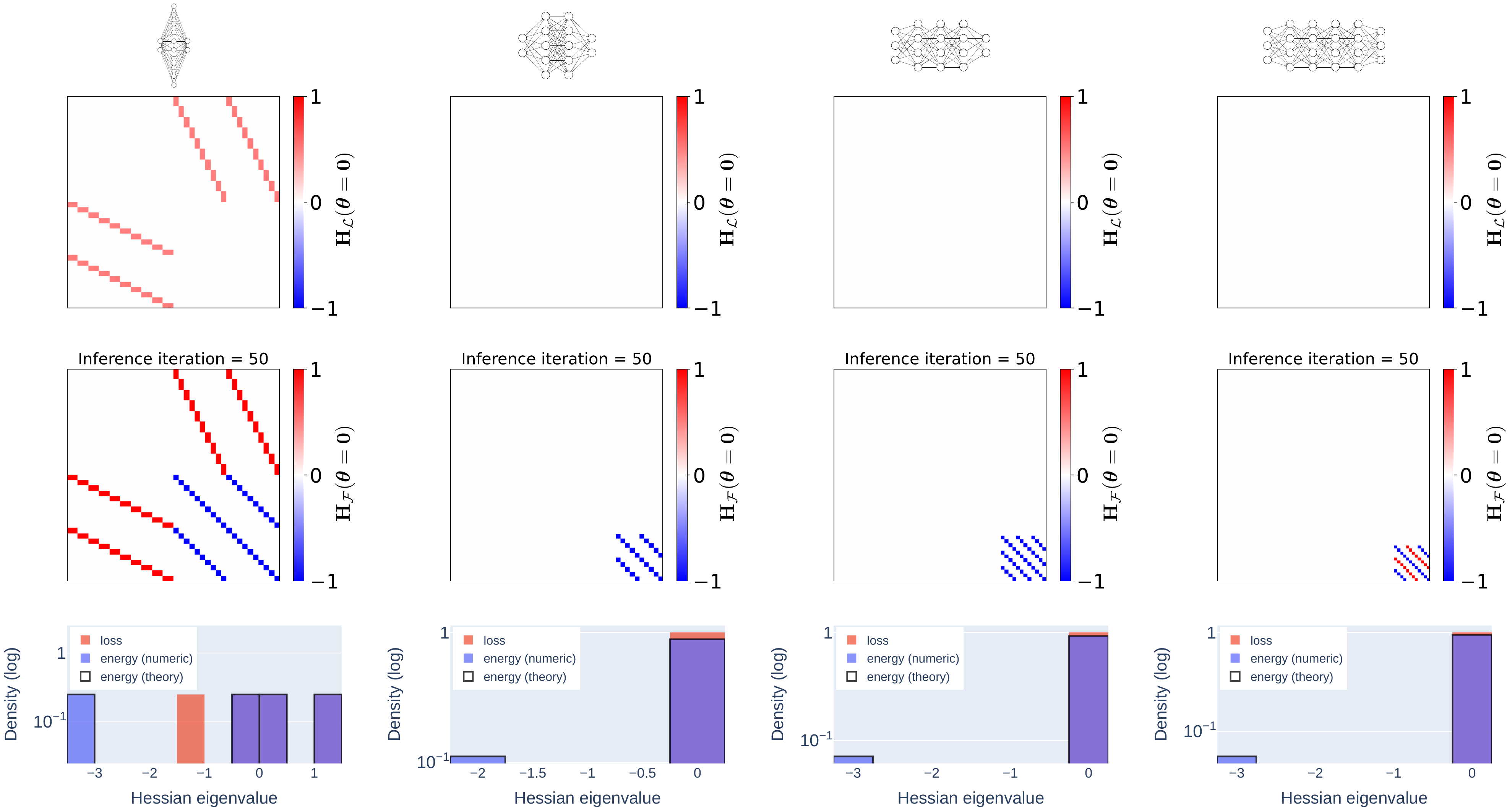}}
        \caption{\textbf{Empirical verification of the Hessian at the origin of the equilibrated energy for DLNs tested on toy data.} We show the Hessian and its eigenspectrum at the origin of the MSE loss (\textit{top}) and equilibrated energy (\textit{middle}) for DLNs with Gaussian target $\mathbf{y}=-\mathbf{x}$ where $\mathbf{x} \sim \mathcal{N}(1, 0.1)$ (see \S\ref{ch4:exp-details} for details). Note that purple bars show overlapping loss and energy Hessian eigendensity. In the right panel, we vary one of the output dimensions to be $y_2 = x_2$. We confirm the strictness of the origin saddle in the equilibrated energy and observe an excellent numerical validation of our theoretical Hessian (Eq.~\ref{ch4:eq:equilib-energy-hess-origin}). Figure~\ref{ch4:fig:hess-origin-toy-chain} shows the same results for one-dimensional networks, and Figure~\ref{ch4:fig:hess-origin-mnist} shows similar results for more realistic datasets.}
        \label{ch4:fig:hess-origin-toy}
    \end{center}
    \vskip -0.2in
\end{figure}

\begin{figure}[t]
    \begin{center}
        \centerline{\includegraphics[width=\textwidth]{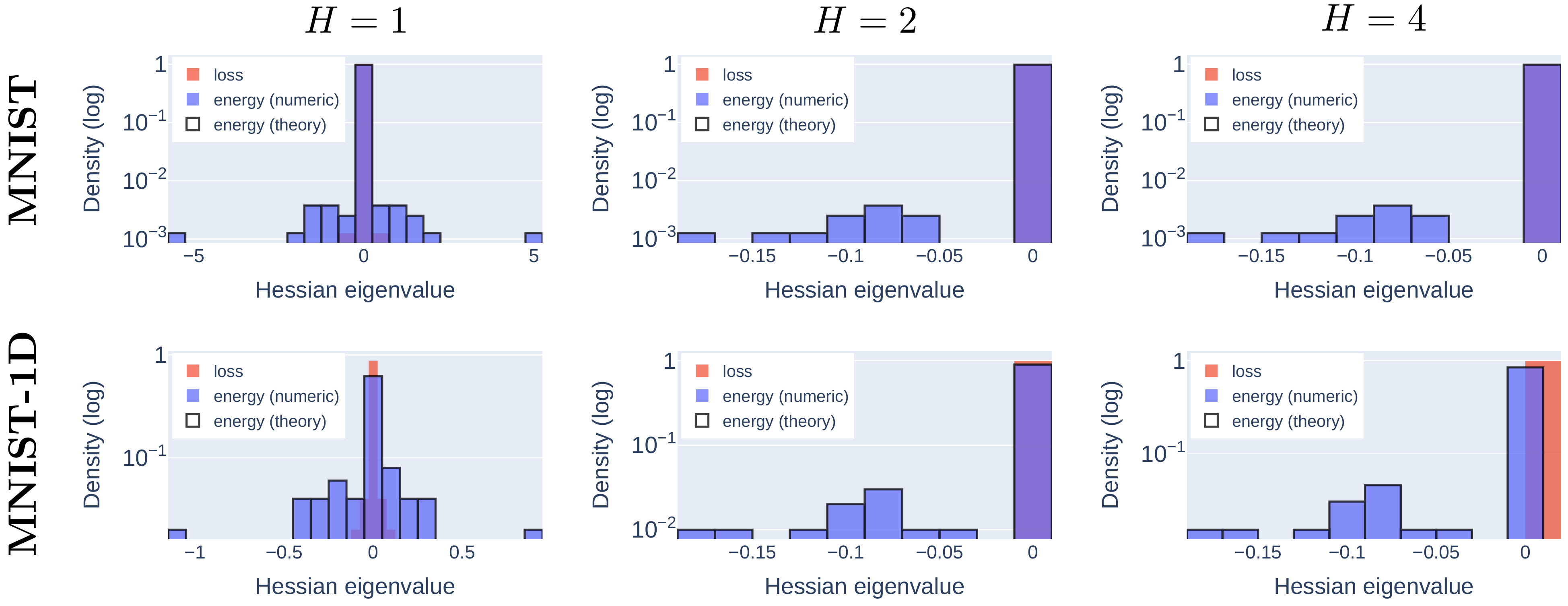}}
        \caption{\textbf{Empirical verification of the Hessian eigenspectrum at the origin of the equilibrated energy for DLNs tested on more realistic datasets.} This shows similar results to Figure~\ref{ch4:fig:hess-origin-toy} for the more realistic datasets MNIST and MNIST-1D \citep{greydanus2020scaling} (see \S\ref{ch4:exp-details} for details). We again find a perfect match between theory and experiment for DLNs with hidden layers $H \in \{1, 2, 4\}$, confirming the strictness of the origin saddle of the equilibrated energy.}
        \label{ch4:fig:hess-origin-mnist}
    \end{center}
    \vskip -0.2in
\end{figure}

By contrast, we now show that the origin saddle of the equilibrated energy is strict for DLNs of any number of hidden layers. Figure~\ref{ch4:fig:toy-examples} shows a few toy examples illustrating the result. In brief, we observe that, when initialised close to the origin saddle, SGD takes increasingly more steps to escape from the loss than the energy as a function of depth (for the same learning rate). Now we state the result more formally. The Hessian at the origin of the equilibrated energy turns out to be (see \S\ref{ch4:energy-hessian} for derivation) 
\begin{align}
    \matr{H}_{\mathcal{F}^*} (\boldsymbol{\theta} = \mathbf{0}) = 
    \begin{cases} 
    \begin{bmatrix} \mathbf{0} & - \widetilde{\matr{\Sigma}}_{\mathbf{xy}} \otimes \matr{I}_{N_1} \\ -\matr{I}_{N_1} \otimes \widetilde{\matr{\Sigma}}_{\mathbf{yx}} & -\widetilde{\matr{\Sigma}}_{\mathbf{yy}} \otimes I_{N_{L-1}} \end{bmatrix}, & H = 1 \\ \\
    \begin{bmatrix} \mathbf{0} & \dots & \mathbf{0} \\ \vdots & \ddots & \vdots \\ \mathbf{0} & \dots & -\widetilde{\matr{\Sigma}}_{\mathbf{yy}} \otimes I_{N_{L-1}} \end{bmatrix},  & H > 1
    \end{cases},
    \label{ch4:eq:equilib-energy-hess-origin}
\end{align}
where $\widetilde{\matr{\Sigma}}_{\mathbf{yy}} \coloneq \frac{1}{B} \sum_i^B \mathbf{y}_i\mathbf{y}_i^T$ is the empirical output covariance. We see that, in contrast to the loss Hessian (Eq.~\ref{ch4:eq:loss-hess-origin}), the energy Hessian has a non-zero last diagonal block given by $\partial^2 \mathcal{F}^*/\partial \matr{W}_L^2$, for any number of hidden layers $H$. It is then straightforward to show that the energy Hessian has always at least one negative eigenvalue, since the output covariance is positive definite.
\begin{tcolorbox}[width=\linewidth, sharp corners=all, colback=white!95!black, colframe=white!95!black]
    \begin{mythm}{3.2}[Strictness of the origin saddle of the equilibrated energy]\label{ch4:thm2}
        The Hessian at the origin of the equilibrated energy (Eq.~\ref{ch4:eq:dln-equilib-energy}) for any DLN has at least one negative eigenvalue (see \S\ref{ch4:energy-hessian} for proof)
            \begin{equation}
                \lambda_{\text{min}}(\matr{H}_{\mathcal{F}^*}(\boldsymbol{\theta}=\mathbf{0})) < 0, \quad \forall H \geq 1 \quad \text{[strict saddle, Def. \ref{ch4:def:strict-saddle}]}.
                \label{ch4:eq:equilib-energy-hess-origin-min-eigen}
            \end{equation}
    \end{mythm}
\end{tcolorbox}
Figures~\ref{ch4:fig:hess-origin-toy} \& \ref{ch4:fig:hess-origin-mnist} show a perfect match between the theoretical (Eq.~\ref{ch4:eq:equilib-energy-hess-origin}) and numerical Hessian at the origin of the equilibrated energy, which we computed for a range of DLNs on a random batch of toy as well as more realistic datasets.

Theorem~\ref{ch4:thm2} proves that the origin is a strict saddle of the equilibrated energy for DLNs of any depth. This is in stark contrast to the MSE loss where it is only true for one-hidden-layer networks $H = 1$ (Eq.~\ref{ch4:eq:loss-hess-origin-min-eigen}). The result predicts that, near the origin, (S)GD should escape the saddle faster on the equilibrated energy than on the loss given the same learning rate, and increasingly so as a function of depth. Figure~\ref{ch4:fig:toy-examples} confirms this prediction for some toy linear networks, and Figures~\ref{ch4:fig:origin-escape}-\ref{ch4:fig:matrix-completion} clearly show that it holds for non-linear networks as well.

\subsection{Analysis of other saddles} 
\label{ch4:other-saddles}
Is the origin a special case where the equilibrated energy has an easier-to-escape saddle than the loss? Or is this result pointing to something more general? Here we consider a specific type of non-strict saddle of the loss (of which the origin is one) and show that indeed they also become strict in the equilibrated energy. We address other saddle types experimentally in \S \ref{ch4:experiments} and leave their theoretical study for future work.

Specifically, we consider saddles of rank zero, which for the MSE can be identified as critical points where the product of weight matrices is zero $\matr{W}_{L:1} = \mathbf{0}$ \citep{achour2021loss}. For the equilibrated energy (Eq.~\ref{ch4:eq:dln-equilib-energy}), we consider the critical points $\boldsymbol{\theta}^*(\matr{W}_L = \mathbf{0}, \matr{W}_{L-1:1} = \mathbf{0})$, since the last weight matrix needs to be null in order for the energy gradient to be zero (see \S\ref{ch4:energy-hessian} for an explanation). It turns out that at these critical points there exists a direction of negative curvature.
\begin{tcolorbox}[width=\linewidth, sharp corners=all, colback=white!95!black, colframe=white!95!black]
    \begin{mythm}{3.3}[Strictness of zero-rank saddles of the equilibrated energy]\label{ch4:thm3}
        Consider the set of critical points of the equilibrated energy (Eq.~\ref{ch4:eq:dln-equilib-energy}) $\boldsymbol{\theta}^*(\matr{W}_L = \mathbf{0}, \matr{W}_{L-1:1} = \mathbf{0})$ where $\mathbf{g}_{\mathcal{F}^*}(\boldsymbol{\theta}^*) = \mathbf{0}$. The Hessian at these points has at least one negative eigenvalue  (see \S\ref{ch4:strict-zero-rank} for proof)
        \begin{equation}
            \exists \lambda(\matr{H}_{\mathcal{F}^*}(\boldsymbol{\theta}^*)) < 0  \quad \text{[strict saddles, Def. \ref{ch4:def:strict-saddle}]}.
            \label{ch4:eq:equilib-energy-hess-zero-rank-min-eigen}
        \end{equation}
    \end{mythm}
\end{tcolorbox}
Note that Theorem~\ref{ch4:thm2} can now be seen as a corollary of Theorem~\ref{ch4:thm3}, although for the origin we derived the full Hessian. This result also stands in contrast to the (MSE) loss, where many of the considered critical points (specifically when 3 or more weight matrices are zero) are non-strict saddles as proved by \citep{achour2021loss}. The prediction is again that, in the vicinity of any of these saddles, PC should escape faster than BP with (S)GD given the same learning rate. For space reasons, the subsequent experiments focus only on the origin as an example of a saddle covered by Theorem~\ref{ch4:thm3} (and Theorem~\ref{ch4:thm2}), but \S\ref{ch4:supp-figs} includes an empirical validation of another (zero-rank) strict saddle of the equilibrated energy (Figures~\ref{ch4:fig:hess-other-saddle-toy}-\ref{ch4:fig:hess-other-saddle-mnist} \& \ref{ch4:fig:other-saddle-escape}). Our released code also makes it relatively easy to test for other saddles.

\section{Experiments} 
\label{ch4:experiments}
Here we report experiments on linear and non-linear networks supporting our theoretical results as well as more general conjecture that all the saddles of the equilibrated energy are strict. In all the experiments, we trained networks with BP and PC using (S)GD with the same learning rate, since the goal was to test our theory of the saddle geometry of the equilibrated energy landscape. Code to reproduce all the results is available at \sloppy{\url{https://github.com/francesco-innocenti/pc-saddles}}.
\begin{figure}[t]
    \begin{center}
        \centerline{\includegraphics[width=0.9\textwidth]{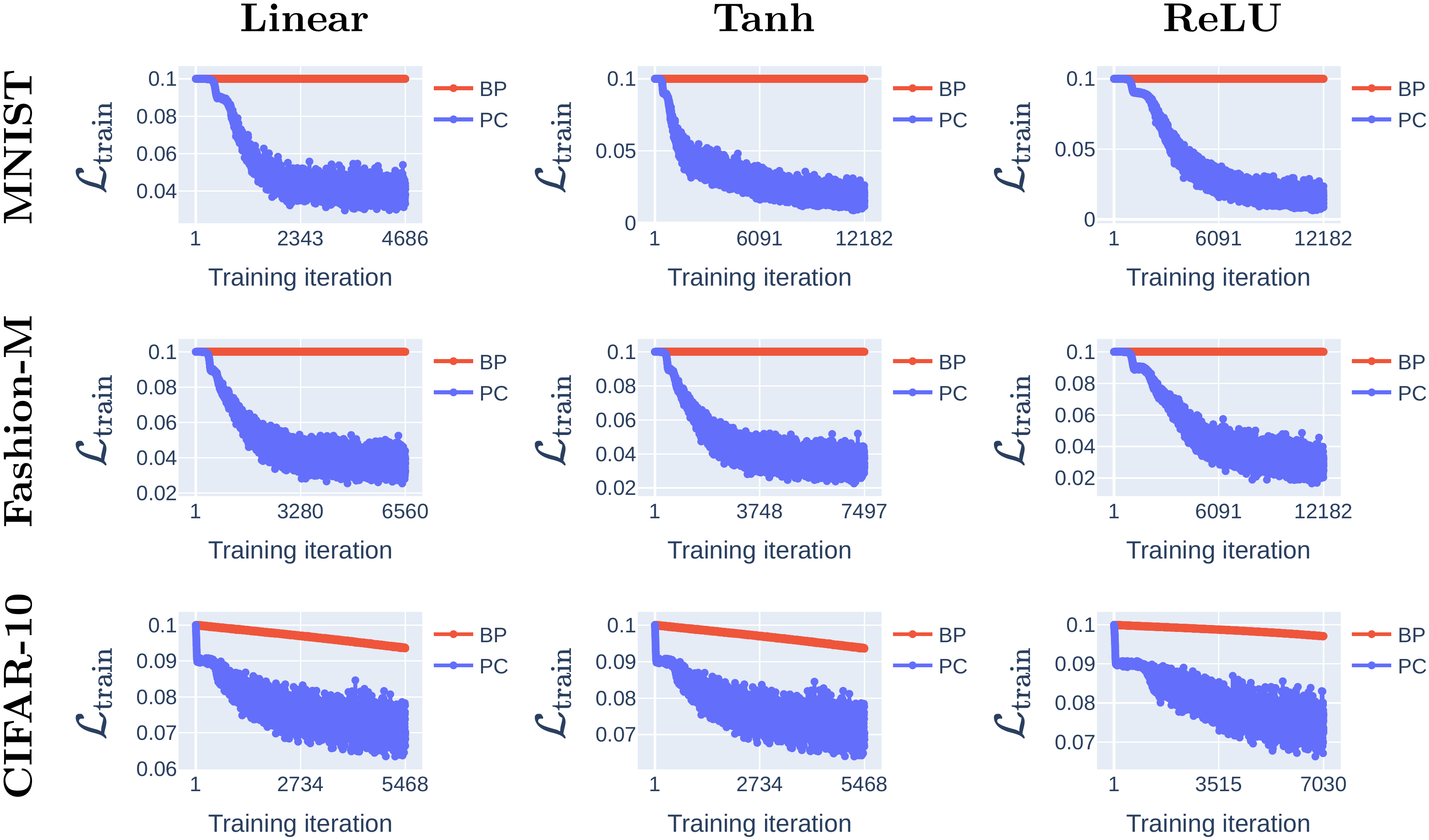}}
        \caption{\textbf{PC escapes the origin saddle much faster than BP with SGD on non-linear networks.} We plot the training (MSE) loss for a representative run of BP and PC for linear and non-linear networks trained on standard image classification tasks (see \S\ref{ch4:exp-details} for details). All networks were initialised close to the origin with scale $\sigma = 5e^{-3}$, and were trained with SGD and learning rate $\eta = 1e^{-3}$. The networks trained on MNIST and Fashion-MNIST had 5 fully connected layers, while those trained on CIFAR-10 had a convolutional architecture. See Figure~\ref{ch4:fig:origin-grad-norms} for the corresponding weight gradient norms during training. Results were consistent across different random seeds.}
        \label{ch4:fig:origin-escape}
    \end{center}
    \vskip -0.2in
\end{figure}

First, we compared the training (MSE) loss dynamics of linear and non-linear networks, including convolutional architectures, on standard image classification tasks with SGD initialised close to the origin (see \S\ref{ch4:exp-details} for details). For computational reasons, we did not run the BP-trained networks to convergence, underscoring the point that the origin saddle of the loss is highly degenerate for relatively deep networks and particularly hard to escape for first-order methods like SGD. In all cases, we observe that PC escapes the origin saddle substantially faster than BP (Figure~\ref{ch4:fig:origin-escape}), and Figure~\ref{ch4:fig:origin-grad-norms} shows that PC exhibits no vanishing gradients. We observe indistinguishable results when initialising close to another non-strict saddle of the loss covered by Theorem~\ref{ch4:thm3} (Figure~\ref{ch4:fig:other-saddle-escape}). These findings support our theoretical results beyond the linear case.

\begin{figure}[h]
    \begin{center}
        \centerline{\includegraphics[width=0.9\textwidth]{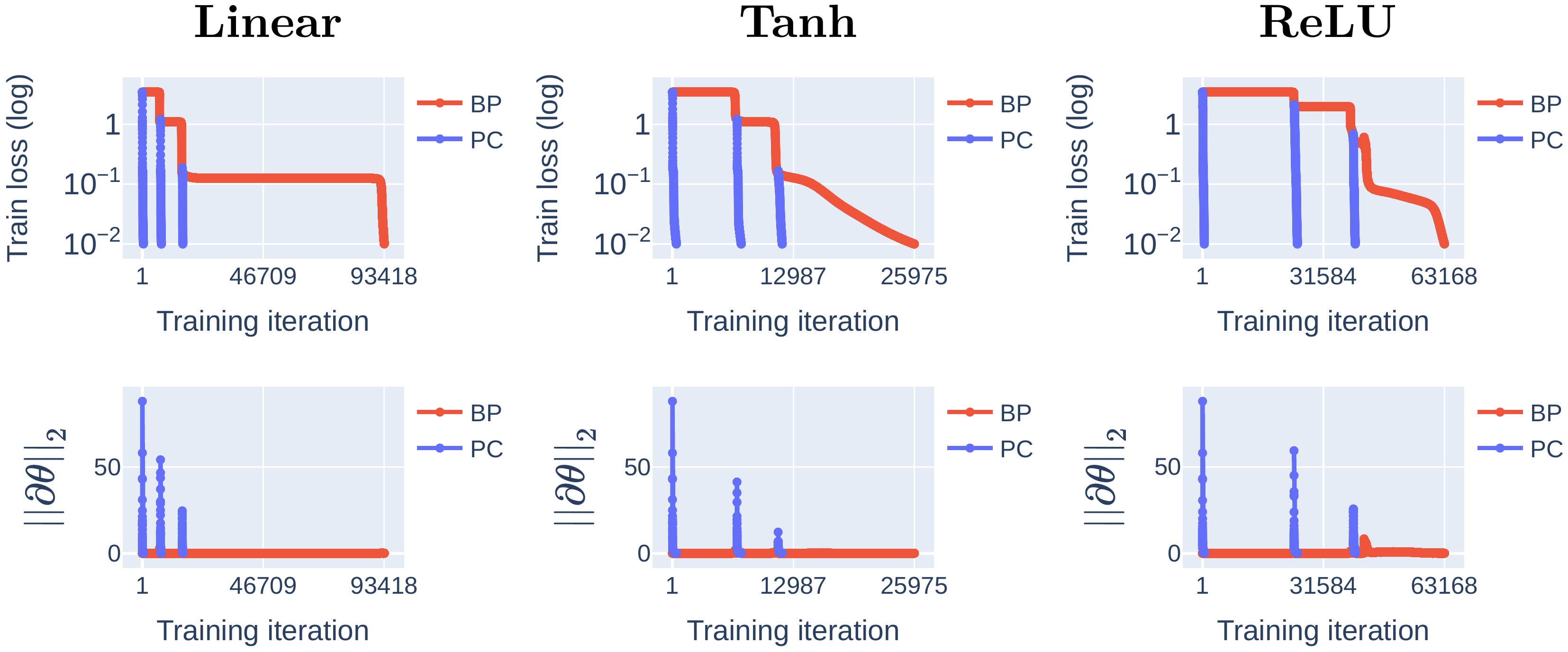}}
        \caption{\textbf{PC quickly escapes higher-rank saddles visited by BP with GD on a matrix completion task.} We plot the training loss (\textit{top}) and corresponding weight gradient norms of the loss (BP) and equilibrated energy (PC) (\textit{bottom}) for networks ($H = 3$, $N = 100$) trained with full-batch GD to fit a random rank-3 matrix, as studied in \citep{jacot2021saddle}. BP-trained networks were initialised near the origin with scale $\sigma = 5e^{-3}$, while PCNs were initialised at each saddle visited by BP (see \S\ref{ch4:exp-details} for details). Results were consistent across different random seeds.}
        \label{ch4:fig:matrix-completion}
    \end{center}
    \vskip -0.2in
\end{figure}

From Figure~\ref{ch4:fig:origin-escape}, we also observe a second plateau in the loss dynamics of PCNs, suggesting a saddle of higher rank (presumably rank 1). This is consistent with the saddle-to-saddle dynamics described for DLNs by \citep{jacot2021saddle}, where for small initialisation GD transitions through a sequence of saddles, each representing a solution of increasing rank. Motivated by this observation, we explicitly tested for higher-rank, non-strict saddles of the loss that we did not study theoretically by replicating one of the experiments of \citep[][cf. Figure~1]{jacot2021saddle}. In particular, we trained networks to fit a rank-3 matrix, which meant that starting near the origin GD visited 3 saddles (of successive rank 0, 1 and 2) before converging to a rank-3 solution as shown in Figure~\ref{ch4:fig:matrix-completion}. We find that, when initialised near any of the saddles visited by BP, PC escapes quickly and does not show vanishing gradients (Figure~\ref{ch4:fig:matrix-completion}), supporting the conjecture that all the saddles of the equilibrated energy are strict.

\section{Discussion} 
\label{ch4:discussion}
In summary, we took a first important step in characterising the effective landscape on which PC learns: the energy landscape at the inference equilibrium. For DLNs, we first showed that the equilibrated energy is equal to a rescaled MSE loss with a weight-dependent rescaling (Theorem~\ref{ch4:thm1}). This result corrects a previous mistake in the literature that the MSE loss is equal to the output energy \citep{millidge2022theoretical} and that the total energy (Eq.~\ref{ch4:eq:pc-energy}) can therefore  be decomposed into the loss and other layer energies (a relationship that only holds at the feedforward activity values). As we expand on below, Eq.~\ref{ch4:eq:dln-equilib-energy} also enables further studies of the PC learning landscape.

We then proved that many non-strict saddle points of the MSE loss, specifically zero-rank saddles, become strict in the equilibrated energy of any DLN (Theorems~\ref{ch4:thm2}-\ref{ch4:thm3}). These saddles include the origin, making PC effectively more robust to vanishing gradients (Figures~\ref{ch4:fig:matrix-completion} \& \ref{ch4:fig:origin-grad-norms}). We thoroughly validated our theory with experiments on both linear and non-linear architectures, and provided empirical support for the strictness of higher-rank saddles of the equilibrated energy. Based on these results, we conjecture that all the saddles of the equilibrated energy are strict. Overall, the PC inference process can therefore be interpreted as making the loss landscape of feedforward networks more benign and robust to vanishing gradients.

\subsection{Implications}
Our work goes significantly beyond existing theories of PC in terms of both explanatory and predictive power. The vast majority of previous works make non-standard assumptions or loose approximations that result in non-specific experimental predictions. For example, the interpretation of PC as implicit GD by \citep{alonso2022theoretical} holds only for small batch sizes and specific layerwise rescalings of the activities and parameter learning rates. (\citep{alonso2023understanding} generalised this result to remove the activity rescalings but not the learning rate ones.) By contrast, linearity is the only major assumption made by our theory, and we empirically verify that all the results hold for non-linear networks. Similarly, both \citep{alonso2023understanding} and \citep{innocenti2023understanding} (the latter of which was the subject of the previous chapter) make second-order approximations of the energy to argue that PC makes use of Hessian information. However, our results clearly show that PC can in principle leverage much \textit{higher-order} information, turning highly degenerate, $H$-order saddles into strict (first-order) ones.

Previous theories have also struggled to explain why faster learning convergence with PC is not always observed depending on the task, model, and optimiser \citep{alonso2022theoretical, song2022inferring}. Our landscape analysis, while incomplete (more on this below), acknowledges these factors and their interplay, helping to explain inconsistent findings and predict when speed-ups can and cannot be expected. All things being equal, PC should converge faster on deep and \textit{narrow} networks (though not too deep as we discuss below), since the distance between the origin saddle and standard initialisations scales with the network width \citep{orvieto2022vanishing}. This likely explains the speed-up reported by \citep{song2022inferring} on a narrow ($N = 64$) 15-layer fully connected network. 

However, in practice all things are not equal, and everything from not reaching an inference equilibrium to different losses, datasets, architectures and optimisers all interact to determine convergence of the learning dynamics. The latter two factors are particularly important. For example, residual networks (ResNets) \cite{he2016deep}, which are popularly known to help with vanishing gradients in deep networks, are locally convex around the origin in the linear case since they effectively shift the location of the origin saddle \cite{hardt2016identity}. In addition, as mentioned in the previous chapter, adaptive optimisers such as Adam \cite{kingma2014adam}---which remains one of the state-of-the-art algorithms---have been shown to escape saddles faster than standard SGD \cite{staib2019escaping, orvieto2022vanishing}. This raises the question of whether there are conditions under which minimising the equilibrated energy could be faster than the loss or lead to better performance, which we return to below.

Our work has also implications for theories of credit assignment in the brain. In particular, our results put the recent principle of prospective configuration \citep{song2022inferring} for energy-based learning on a more solid theoretical footing. While we clearly validate the intuition behind claim that PC inference facilitates learning, under standard conditions including deep and wide ResNets trained with adaptive optimisers, BP will likely converge as fast as, if not faster, than PC.

More broadly, our landscape theory closely relates to the work of \citep{stern2024physical}, who showed that learning in linear physical systems with equilibrium propagation \citep{scellier2017equilibrium, scellier2024energy} has beneficial effects on the activity (rather than weight) Hessian. Studying these connections, and more generally the benefits of inference for learning in energy-based systems, could be an interesting future direction.

\subsection{Limitations}
We conclude by addressing the main limitations of our work. First, the strictness of the energy saddles we studied holds, by derivation, only at the exact inference equilibrium (Theorem~\ref{ch4:thm1}, Eq.~\ref{ch4:eq:dln-equilib-energy}). It is important to note that these benefits are continuous, and one does not need to reach equilibrium to improve the degeneracy of the loss saddles (as also shown in the previous chapter in Figure~\ref{ch3:fig:energy-land-infer-dynamics}). In this sense, PC could be seen as a resource. However, as we will study in detail in the next chapter, PC inference seems to require increasingly more iterations to converge on deeper networks---which aligns with our landscape theory since the loss saddles become more degenerate with depth. Our results therefore highlight the important challenge of speeding up PC inference on very deep models if its claimed benefits for learning are to be realised on large-scale settings \citep{pinchetti2024benchmarking}, at least on standard hardware (GPUs). The next chapter will address and help overcome this challenge.

Even if this problem is solved, there seem to be two related questions that ultimately matter for the practical training of deep networks. First, are there conditions under which the equilibrated energy can be minimised faster than the loss in a more compute- or memory-efficient manner, with at least equal performance? As mentioned above, current architectures and optimisers such as skip connections \citep{he2016deep} and Adam \citep{kingma2014adam} help to deal with the origin saddle at an increased memory cost. Could this trade off with the compute cost of PC inference (again on GPUs)? The next chapter will help answer this question by studying the inference landscape and dynamics of PCNs.

Second, could there be scenarios where PC is slower or less efficient but at the benefit of significantly better performance? This is a hard question to answer since we are far from having a theory of generalisation in deep learning \citep{zhang2021understanding, jiang2019fantastic}. Given our origin saddle result (Theorem~\ref{ch4:thm2}), however, it is interesting to note that on problems such as matrix completion (Figure~\ref{ch4:fig:matrix-completion}) where a low-rank bias is useful, GD with small initialisation can converge to better-generalising solutions compared to standard initialisations \citep{jacot2021saddle}.

Finally, understanding the overall convergence behaviour of PC would also require characterising other types of critical point of the equilibrated energy, specifically its minima \citep{frieder2024bad}. Our work, and Eq.~\ref{ch4:eq:dln-equilib-energy} in particular, enables this. In \S \ref{ch4:energy-minima}, we present a preliminary investigation showing that, for linear chains, the global minima of the equilibrated energy are \textit{flatter} than those of the MSE loss, generalising a result in the previous chapter (Theorem~\ref{ch3:thm:flat-min}). This result potentially explains the common observation that PC convergence tends to slow down towards the end of training, but we leave its full implications for future work.

\section*{Author contributions}
FI conceptualised the study, proved Theorems~\ref{ch4:thm1} \& \ref{ch4:thm2}, ran all the experiments, and wrote the paper. EMA contributed to conceptual discussions and proved Theorem~\ref{ch4:thm3}. RS and CLB contributed to conceptual discussions.

%% file: text/mainmatter/ch5-mupc.tex
\chapter{$\mu$PC: Scaling Predictive Coding to 100+ Layer Networks}
\label{ch:mupc}
\minitoc

\section{Abstract}
The biological implausibility of backpropagation (BP) has motivated many alternative, brain-inspired algorithms that attempt to rely only on local information, such as predictive coding (PC) and equilibrium propagation. However, these algorithms have notoriously struggled to train very deep networks, preventing them from competing with BP in large-scale settings. Indeed, scaling PC networks (PCNs) has recently been posed as a challenge for the community \cite{pinchetti2024benchmarking}. Here, we show that 100+ layer PCNs can be trained reliably using a Depth-$\mu$P parameterisation \cite{yang2023tensorinfdepth, bordelon2023depthwise} which we call ``$\mu$PC''. By analysing the scaling behaviour of PCNs, we reveal several pathologies that make standard PCNs difficult to train at large depths. We then show that, despite addressing only some of these instabilities, $\mu$PC allows stable training of very deep (up to 128-layer) residual networks on simple classification tasks with competitive performance and little tuning compared to current benchmarks. Moreover, $\mu$PC enables zero-shot transfer of both weight and activity learning rates across widths and depths. Our results serve as a first step towards scaling PC to more complex architectures and have implications for other local algorithms. Code for $\mu$PC is made available as part of a JAX library for PCNs.\footnote{\url{https://github.com/thebuckleylab/jpc} \cite{innocenti2024jpc}.}
\begin{figure}[h!]
    %\vskip 0.2in
    \begin{center}
        \centerline{\includegraphics[width=\textwidth]{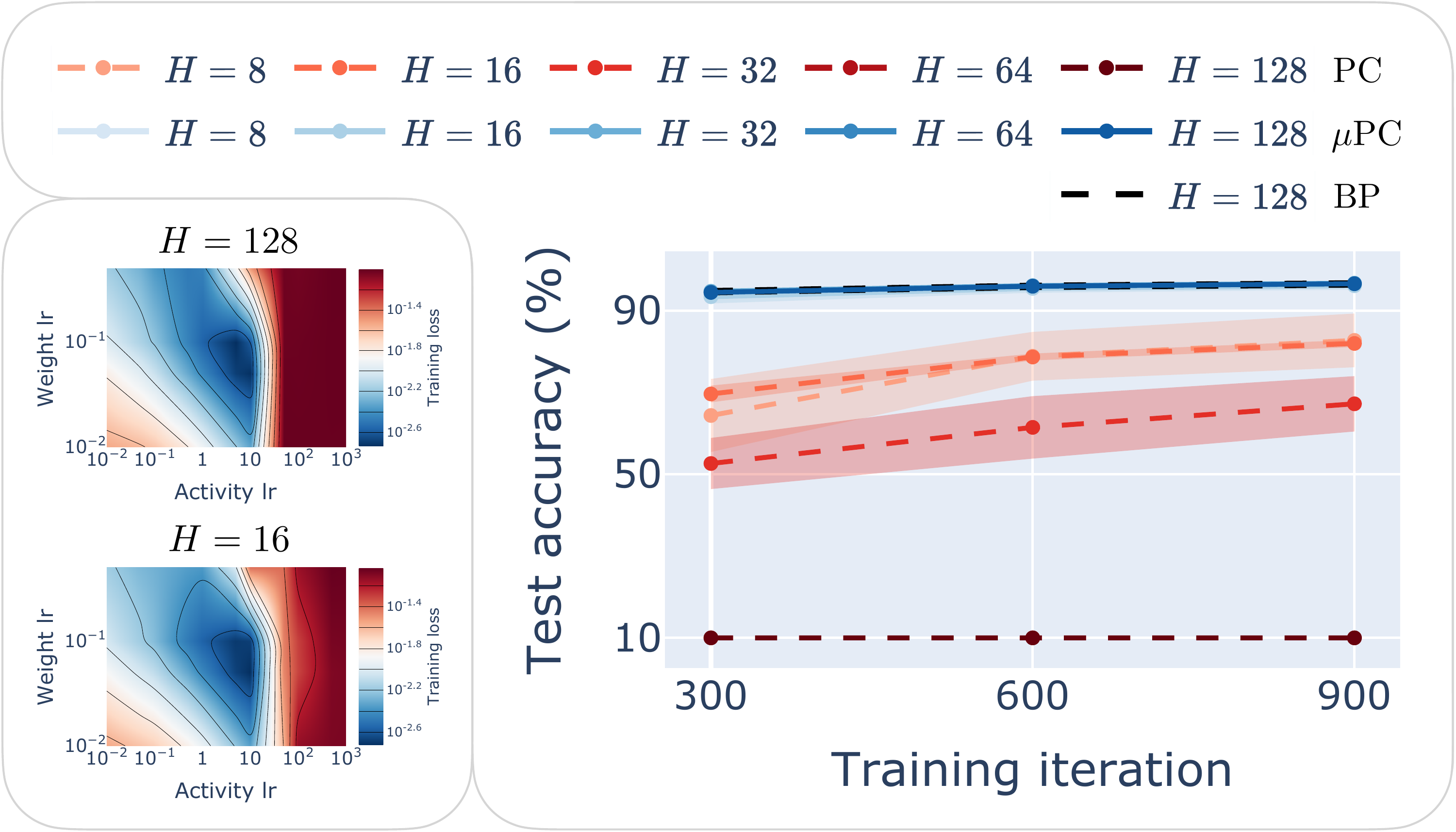}}
        \caption{\textbf{$\mu$PC enables stable training of 100+ layer ResNets with zero-shot learning rate transfer.} (\textit{Right}) Test accuracy of ReLU ResNets with depths $H = \{8, 16, 32, 64, 128 \}$ trained to classify MNIST for one epoch with standard PC, $\mu$PC and BP with Depth-$\mu$P (see \S\ref{ch5:exp-details} for details). Solid lines and shaded regions indicate the mean and $\pm 1$ standard deviation across 3 different random seeds. These results hold across other activation functions (see Fig.~\ref{ch5:fig:mupc-vs-pc-mnist-accs-all-act-fns}). See also Figs.~\ref{ch5:fig:mupc-mnist-5-epochs}-\ref{ch5:fig:cifar-20-epochs} for asymptotic results with 128-layer ReLU networks trained for multiple epochs on both MNIST, Fashion-MNIST and CIFAR10. (\textit{Left}) Example of zero-shot transfer of the weight and activity learning rates from 16- to 128-layer Tanh networks. See Figs.~\ref{ch5:fig:mupc-hyperparam-transfer-tanh} \& \ref{ch5:fig:mupc-hyperparam-transfer-linear}-\ref{ch5:fig:mupc-hyperparam-transfer-relu} for an explanation and the complete transfer results across widths as well as depths.}
        \label{ch5:fig:mupc-vs-pc-mnist-accs}
    \end{center}
    \vskip -0.25in
\end{figure}

\section{Introduction}
\label{ch5:intro}
In the previous chapter, we saw that the iterative inference procedure of PC (Eq. \ref{pcns:eq:pc-infer}) effectively allows the algorithm to learn on a reshaped loss landscape that is more benign and robust to vanishing gradients. All things being equal, this should make deep networks easier to train with PC than BP. However, in practice very deep (10+ layer) PCNs have highly unstable inference dynamics and become challenging to train \cite{pinchetti2024benchmarking}. More generally, local learning rules have notoriously struggled to train large and especially deep models on the scale of modern AI applications.\footnote{It is possible that these algorithms are more suited to alternative, non-digital hardware, but their scalability can still be investigated on standard GPUs. Indeed, the issues we expose with the standard parameterisation of PCNs can be argued to be hardware-independent (\S \ref{ch5:ill-cond-infer}).} 

For the first time, we show that very deep (100+ layer) networks can be trained reliably using a Depth-$\mu$P-inspired parameterisation \cite{yang2023tensorinfdepth, bordelon2023depthwise} of PC which we call ``$\mu$PC'' (Fig.~\ref{ch5:fig:mupc-vs-pc-mnist-accs}). To our knowledge, \textit{no networks of such depth have been trained before with a local algorithm}. Indeed, this has recently been posed as a challenge for the PC community \cite{pinchetti2024benchmarking}. We start by showing that the standard parameterisation of PC networks (PCNs) is inherently unscalable in that (i) the inference landscape becomes increasingly ill-conditioned with model size and training time, and (ii) the forward initialisation of the activities vanishes or explodes with the depth. We then show that, despite addressing only the second instability, $\mu$PC is capable of training up to 128-layer fully connected residual networks (ResNets) on standard classification tasks with competitive performance and little tuning compared to current benchmarks (Fig.~\ref{ch5:fig:mupc-vs-pc-mnist-accs}). Moreover, $\mu$PC enables zero-shot transfer of both the weight and activity learning rates across widths and depths (Fig.~\ref{ch5:fig:mupc-hyperparam-transfer-tanh}). We make code for $\mu$PC publicly available as part of a JAX library for PCNs at \sloppy{\url{https://github.com/thebuckleylab/jpc}} \cite{innocenti2024jpc}, which we introduce in the next chapter.

The rest of this chapter is structured as follows. Following a brief review of the maximal update parameterisation ($\mu$P) and PCNs (\S \ref{ch5:background}), Section~\ref{ch5:instability} exposes two distinct pathologies in standard PCNs which make training at large scale practically impossible. Motivated by these findings, we then suggest a minimal set of desiderata for a more scalable PCN parameterisation (\S \ref{ch5:desiderata}). Section~\ref{ch5:experiments} presents experiments with $\mu$PC, and Section~\ref{ch5:mupc-theory} studies a regime where $\mu$PC converges to BP. We conclude with the limitations of this work and promising directions for future research (\S \ref{ch5:discussion}). Appendix \ref{ch5:appendix} includes a review of related work and additional experiments, along with derivations, experimental details and supplementary figures.

\subsection{Summary of contributions}
\label{ch5:contributions}
\begin{itemize}
    \item We show that $\mu$PC, which reparameterises PCNs using Depth-$\mu$P \cite{yang2023tensorinfdepth, bordelon2023depthwise}, allows stable training of very deep (100+ layer) ResNets on simple classification tasks with competitive performance and little tuning compared to current benchmarks \cite{pinchetti2024benchmarking} (Figs.~\ref{ch5:fig:mupc-vs-pc-mnist-accs} \& \ref{ch5:fig:mupc-mnist-5-epochs}-\ref{ch5:fig:mupc-fashion-15-epochs}).
    \item $\mu$PC also empirically enables zero-shot transfer of both the weight and activity learning rates across widths and depths (Figs.~\ref{ch5:fig:mupc-hyperparam-transfer-tanh} \& \ref{ch5:fig:mupc-hyperparam-transfer-linear}-\ref{ch5:fig:mupc-hyperparam-transfer-relu}).
    \item We achieve these results by a theoretical and empirical analysis of the scaling behaviour of the inference landscape and dynamics of PCNs (\S \ref{ch5:instability}), revealing the following two pathologies:
    \begin{itemize}
        \item the inference landscape becomes increasingly ill-conditioned with model size (Fig.~\ref{ch5:fig:sp-cond-nums-init}) and training time (Fig.~\ref{ch5:fig:sp-train-cond-nums-GD-mnist}) (\S \ref{ch5:ill-cond-infer}); and
        \item the forward pass of standard PCNs vanishes or explodes with the depth (\S \ref{ch5:vanish-explode-ffwd}).
    \end{itemize}
    \item To address these instabilities, we propose a minimal set of desiderata that PCNs should aim to satisfy to be trainable at scale (\S \ref{ch5:desiderata}), revealing an apparent trade-off between the conditioning of the inference landscape and the stability of the forward pass (Fig.~\ref{ch5:fig:des-trade-off}). This analysis can be applied to other inference-based algorithms (\S \ref{ch5:other-algos}).
    \item To better understand $\mu$PC, we study a theoretical regime where the $\mu$PC energy converges to the mean squared error (MSE) loss and so PC effectively implements BP (Theorem~\ref{ch5:thm1}, Fig.~\ref{ch5:fig:mupc-loss-energy-ratios-init-1e-1}). However, we find that $\mu$PC can successfully train deep networks far from this regime.
\end{itemize}

\section{Background}
\label{ch5:background}

\subsection{The maximal update parameterisation ($\mu$P)} 
\label{ch5:mup}
The maximal update parameterisation was first introduced by \cite{yang2021tensor} to ensure that the order of the activation or feature updates at each layer remains stable with the width $N$. This was motivated by the lack of feature learning in the neural tangent kernel or ``lazy'' regime \cite{jacot2018neural}, where the activations remain practically unchanged during training \cite{chizat2019lazy, lee2019wide}. More formally, $\mu$P can be derived from the following 3 desiderata \cite{yang2021tensor}: (i) the layer preactivations are $\mathcal{O}_N(1)$ at initialisation, (ii) the network output is $\mathcal{O}_N(1)$ during training, and (iii) the layer features are also $\mathcal{O}_N(1)$ during training.\footnote{Throughout, we will use $\mathcal{O}_n(1)$ to mean $\Theta_n(1)$ such that the activations neither explode nor vanish with $n$.}

Satisfying these desiderata boils down to solving a system of equations for a set of scalars (commonly referred to as ``abcd'') parameterising the layer transformation, the (Gaussian) initialisation variance, and the learning rate \cite{yang2023tensoradaptive, pehlevan2023lecture}. Different optimisers and types of layer lead to different scalings. One version of $\mu$P (and the version we will be using here) initialises all the weights from a standard Gaussian and rescales each layer transformation by $1/\sqrt{N_{\ell-1}}$, with the exception of the output which is scaled by $1/N_{L-1}$. Remarkably, $\mu$P allows not only for more stable training dynamics but also for \textit{zero-shot hyperparameter transfer}: tuning a small model parameterised with $\mu$P guarantees that optimal hyperparameters such as the learning rate will transfer to a wider model \cite{yang2021tuning, noci2025super}.

More recently, $\mu$P has been extended to depth for ResNets (``Depth-$\mu$P'') \cite{yang2023tensorinfdepth, bordelon2023depthwise}, such that transfer is also conserved across depths $L$. This is done by mainly introducing a $1/\sqrt{L}$ scaling before each residual block. Extensions of standard $\mu$P for other algorithms have also been proposed \cite{ishikawa2023parameterization, ishikawa2024local, haas2024effective, dey2024sparse}.

\subsection{Predictive coding networks (PCNs)} 
\label{ch5:pcns}
We consider the following general parameterisation of the energy function of $L$-layered PCNs \cite{buckley2017free}:
\begin{equation}
    \mathcal{F} = \sum_{\ell=1}^{L} \frac{1}{2}||\mathbf{z}_\ell - a_\ell \matr{W}_\ell \phi_\ell(\mathbf{z}_{\ell-1}) - \tau_\ell \mathbf{z}_{\ell-1}||^2
    \label{ch5:eq:pc-energy}
\end{equation}
with weights $\matr{W}_\ell \in \mathbb{R}^{N_\ell \times N_{\ell-1}}$, activities $\mathbf{z}_\ell \in \mathbb{R}^{N_\ell}$ and activation function $\phi_\ell(\cdot)$. Dense weight matrices could be replaced by convolutions, all assumed to be initialised i.i.d. from a Gaussian $(\matr{W}_{\ell})_{ij} \sim \mathcal{N}(0, b_\ell)$ with variance scaled by $b_\ell$. We omit multiple data samples to simplify the notation, and ignore biases since they do not affect the main analysis, as explained in \S \ref{ch5:activity-grad-hess}. Compared to the general energy presented in \S\ref{ch:pcns} (Eq.~\ref{pcns:eq:pc-energy}), we also add scalings $a_\ell \in \mathbb{R}$ and optional skip or residual connections set by $\tau_\ell \in \{0, 1\}$. 

The energy of the last layer is defined as $\mathcal{F}_L = \frac{1}{2}||\mathbf{z}_L - a_L \matr{W}_L\phi_L(\mathbf{z}_{L-1})||^2$ for some target $\mathbf{z}_L \coloneqq \mathbf{y} \in \mathbb{R}^{N_L}$, while the energy of the first layer is $\mathcal{F}_1 = \frac{1}{2}||\mathbf{z}_1 - a_1 \matr{W}_1\mathbf{z}_0||^2$, with some optional input $\mathbf{z}_0 \coloneq \mathbf{x} \in \mathbb{R}^{N_0}$ for supervised (vs unsupervised) training.\footnote{Many of our theoretical results can be extended to the unsupervised case (see \S \ref{ch5:appendix}), but for ease of presentation we will focus on the supervised case.} We will refer to PC or SP as the ``standard parameterisation'' with unit premultipliers $a_\ell = 1$ for all $\ell$ and standard initialisations \cite{lecun2002efficient, glorot2010understanding, he2015delving} such as $b_\ell = 1/N_{\ell-1}$, and to $\mu$PC as that which uses (some of) the scalings of Depth-$\mu$P (\S \ref{ch5:mup}).\footnote{We distinguish between $\mu$PC and Depth-$\mu$P because the parameterisation impacts properties specific to the PC energy (Eq.~\ref{ch5:eq:pc-energy}) as we will see in \S \ref{ch5:desiderata}.} See Table~\ref{ch5:params-table} for a summary. 

We fix the width of all the hidden layers $N = N_1 = \dots = N_H$ where $H=L-1$ is the number of hidden layers. As the previous chapters, we use $\boldsymbol{\theta} \coloneq \{\vect(\matr{W}_\ell)\}_{\ell=1}^L \in \mathbb{R}^p$ to represent all the weights and $\mathbf{z} \coloneq \{\mathbf{z}_\ell \}_{\ell=1}^H \in \mathbb{R}^{NH}$ to denote all the activities free to vary. Note that, depending on the context, we will use both $H$ and $L$ to refer to the network depth.

As reviewed in Chapter~\ref{ch:pcns}, PCNs are trained by minimising the energy (Eq.~\ref{ch5:eq:pc-energy}) in two separate phases: first with respect to the activities (inference) and then with respect to the weights (learning),
\begin{equation}
    \textit{Infer:  }
        \mathbf{z}^* = \argmin_{\mathbf{z}} \mathcal{F}(\boldsymbol{\theta}, \mathbf{z})
    \label{ch5:eq:pc-infer}
\end{equation}
\begin{equation}
    \textit{Learn:  }
        \Delta\boldsymbol{\theta} \propto - \nabla_{\theta} \mathcal{F}(\boldsymbol{\theta}, \mathbf{z}^*).
    \label{ch5:eq:pc-learn}
\end{equation}
Inference acts on a single data point and is generally performed by gradient descent (GD), $\mathbf{z}_{t+1} = \mathbf{z}_t - \beta \nabla_{\mathbf{z}} \mathcal{F}$ with step size $\beta$. While the previous two chapters were concerned with the learning problem (Eq.~\ref{ch5:eq:pc-learn}), here we will mainly address the first optimisation problem (Eq.~\ref{ch5:eq:pc-infer}), namely the inference landscape and dynamics, but we discuss and numerically investigate the impact on the learning dynamics (Eq.~\ref{ch5:eq:pc-learn}) wherever relevant.

\section{Instability of the standard PCN parameterisation}
\label{ch5:instability}
In this section, we reveal through both theory and experiment that the standard parameterisation (SP) of PCNs suffers from two instabilities that make training and convergence of the PC inference dynamics (Eq.~\ref{ch5:eq:pc-infer}) at large scale practically impossible. First, the inference landscape of standard PCNs becomes increasingly ill-conditioned with model size and training time (\S \ref{ch5:ill-cond-infer}). Second, depending on the model, the feedforward pass either vanishes or explodes with the depth (\S \ref{ch5:vanish-explode-ffwd}). The second problem is shared with BP-trained networks, while the first instability is unique to PC and likely any other algorithm performing inference minimisation (\S \ref{ch5:other-algos}).

\subsection{Ill-conditioning of the inference landscape}
\label{ch5:ill-cond-infer}
\begin{figure}[t]
    %\vskip 0.2in
    \begin{center}
        \centerline{\includegraphics[width=\textwidth]{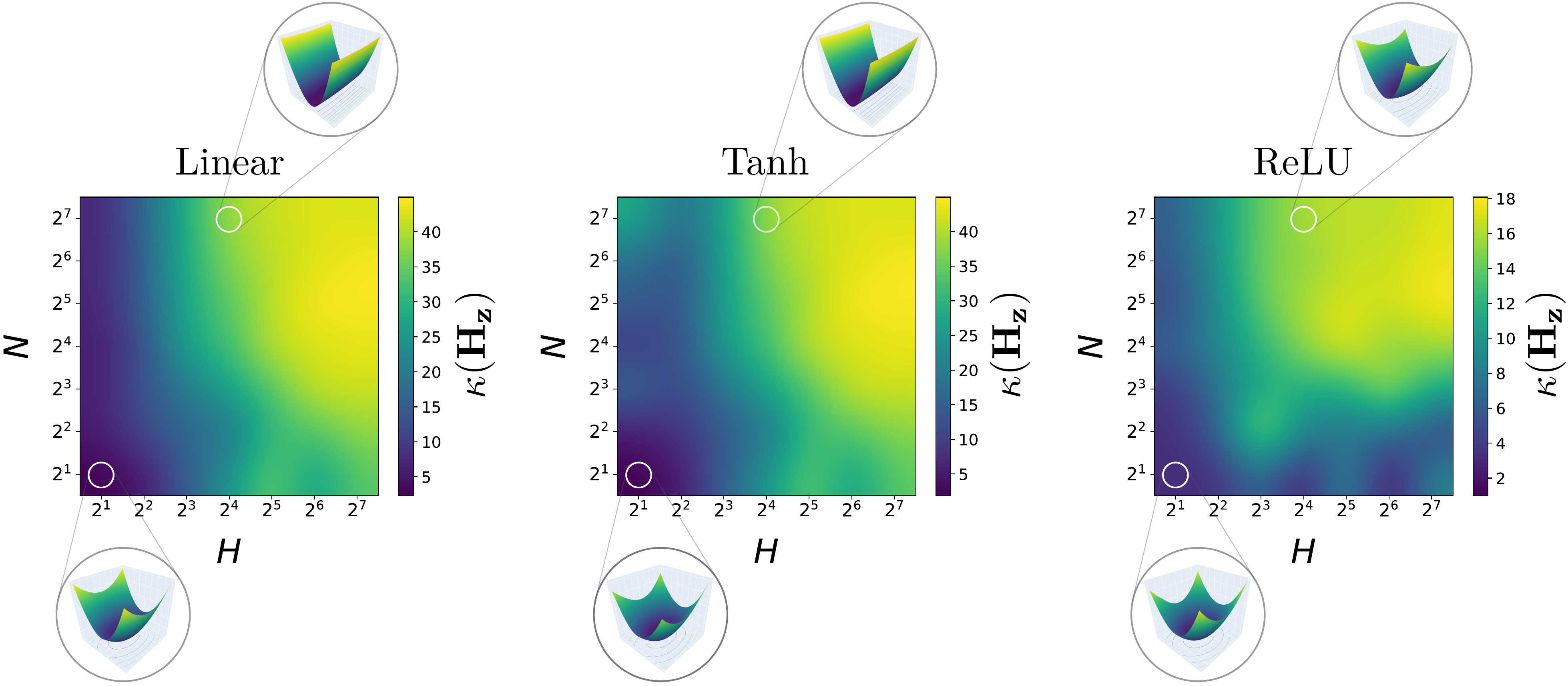}}
        \caption{\textbf{Wider and particulary deeper PCNs have a more ill-conditioned inference landscape.} We plot the condition number of the activity Hessian $\kappa(\matr{H}_{\mathbf{z}})$ (lower is better) of randomly initialised fully connected networks as function of the width $N$ and depth $H$ (see \S\ref{ch5:exp-details} for details). Insets show 2D projections of the landscape of selected networks around the linear solution (Eq.~\ref{ch5:eq:pc-infer-solution}) along the maximum and minimum eigenvectors of the Hessian $\mathcal{F}(\mathbf{z}^* + \alpha \hat{\mathbf{v}}_{\text{min}} + \beta \hat{\mathbf{v}}_{\text{max}})$. Note that the ill-conditioning is much more extreme for ResNets (see Fig.~\ref{ch5:fig:resnets-cond-nums-init}). Results were similar across different seeds.}
        \label{ch5:fig:sp-cond-nums-init}
    \end{center}
    \vskip -0.25in
\end{figure}
Here we show that the inference landscape of standard PCNs becomes increasingly ill-conditioned with network width, depth and training time. As reviewed in \S \ref{ch5:pcns}, the inference phase of PC (Eq.~\ref{ch5:eq:pc-infer}) is commonly performed by GD. For a deep linear network (DLN, Eq.~\ref{ch5:eq:pc-energy} with $\phi_\ell = \matr{I}$ for all $\ell$), one can solve for the activities in closed form as shown by \cite{ishikawa2024local},
\begin{equation}
  \nabla_{\mathbf{z}} \mathcal{F} = \matr{H}_{\mathbf{z}} \mathbf{z} - \mathbf{b}= 0 \quad \implies \quad \mathbf{z}^* = \matr{H}_{\mathbf{z}}^{-1}\mathbf{b}
  \label{ch5:eq:pc-infer-solution}
\end{equation}
where $(\matr{H}_{\mathbf{z}})_{\ell k} \coloneq \partial^2 \mathcal{F} / \partial \mathbf{z}_\ell \partial \mathbf{z}_k \in \mathbb{R}^{(NH) \times (NH)}$ is the Hessian of the energy with respect to the activities, and $\mathbf{b} \in \mathbb{R}^{NH}$ is a sparse vector depending only on the data and associated weights (see \S \ref{ch5:activity-grad-hess} for details). Eq.~\ref{ch5:eq:pc-infer-solution} shows that for a DLN, PC inference is a well-determined linear problem.\footnote{This contrasts with the weight landscape $\mathcal{F}(\boldsymbol{\theta})$, which grows nonlinear with the depth even for DLNs \cite{innocenti2025only}.}

For arbitrary DLNs, one can also prove that the inference landscape is strictly convex as the Hessian is positive definite\footnote{We note that this was claimed to be proved by \cite{millidge2022theoretical}; however, they only showed that the block diagonals of the Hessian are positive definite, ignoring the layer, off-diagonal interactions.}, $\matr{H}_{\mathbf{z}} \succ 0$ (Theorem~\ref{ch5:thmA1}; see \S \ref{ch5:pos-def} for proof). This makes intuitive sense since the energy (Eq.~\ref{ch5:eq:pc-energy}) is quadratic in $\mathbf{z}$. The result is empirically verified for DLNs in Figs.~\ref{ch5:fig:H_eigens_N_slice}-\ref{ch5:fig:max-min-eigens} and appears to generally hold for nonlinear networks (see Figs.~\ref{ch5:fig:max-min-eigens} \& \ref{ch5:fig:resnets-cond-nums-init}).

For such convex problems, the convergence rate of GD is known to be given by the condition number of the Hessian \cite{boyd2004convex, nesterov2013introductory}, $\kappa(\matr{H}_{\mathbf{z}}) = |\lambda_{\text{max}}| / |\lambda_{\text{min}}|$. Intuitively, the higher the condition number, the more elliptic the level sets of the energy $\mathcal{F}(\mathbf{z})$ become, and the more iterations GD will need to reach the solution (see Fig.~\ref{ch5:toy-ill-cond}), with the step size bounded by the highest curvature direction $\beta<2/\lambda_{\text{max}}$ (see Fig.~\ref{ch5:fig:sp-activity-inits} for an example). For non-convex problems, it can still be useful to have a notion of local conditioning \cite[e.g.][]{zhao2024theoretical}.
\begin{figure}[t]
    \vskip 0.2in
    \begin{center}
        \centerline{\includegraphics[width=\textwidth]{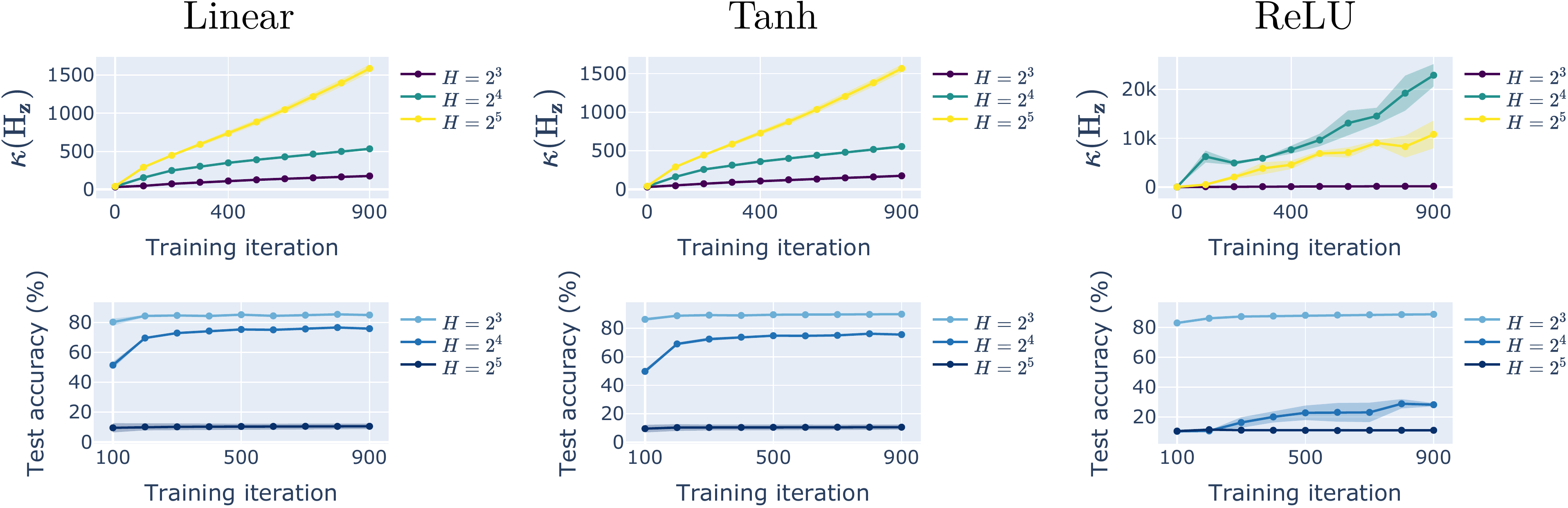}}
        \caption{\textbf{The inference landscape of PCNs grows increasingly ill-conditioned with training.} We plot the condition number of the activity Hessian (Eq.~\ref{ch5:eq:activity-hessian}) (\textit{top}) as well as test accuracies (\textit{bottom}) for fully connected networks of depths $H \in \{8, 16, 32\}$ during one epoch of training. All networks had width $N=128$ and were trained to classify MNIST (see \S\ref{ch5:exp-details} for more details). Similar results are observed for ResNets (Fig.~\ref{ch5:fig:sp-train-cond-nums-skips-mnist}) and Fashion-MNIST (Fig.~\ref{ch5:fig:sp-train-cond-nums-GD-fashion}). Solid lines and shaded regions indicate the mean and standard deviation over 3 random seeds.}
        \label{ch5:fig:sp-train-cond-nums-GD-mnist}
    \end{center}
    \vskip -0.25in
\end{figure}

What determines the condition number of $\matr{H}_{\mathbf{z}}$? Looking more closely at the structure of the Hessian
\begin{align}
    \frac{\partial^2 \mathcal{F}}{\partial \mathbf{z}_\ell \partial \mathbf{z}_k} =
    \begin{cases} 
    \matr{I} + a_{\ell+1}^2\matr{W}_{\ell+1}^T \matr{W}_{\ell+1}, & \ell = k \\
    -a_{k+1}\matr{W}_{k+1}, & \ell - k = 1 \\
    -a_{\ell+1}\matr{W}_{\ell+1}^T, & \ell - k = -1 \\
    \mathbf{0}, & \text{else}
    \end{cases},
    \label{ch5:eq:activity-hessian}
\end{align}
one realises that it depends on two main factors: (i) the \textit{network architecture}, including the width $N$, depth $L$ and connectivity; and (ii) the \textit{value of the weights} at any time during training $\boldsymbol{\theta}_t$. We first find that the inference landscape of standard PCNs becomes increasingly ill-conditioned with the width and particularly depth (Fig.~\ref{ch5:fig:sp-cond-nums-init}), and extremely so for ResNets (Fig.~\ref{ch5:fig:resnets-cond-nums-init}). See also \S \ref{ch5:rand-matrix} for a random matrix theory analysis of the scaling behaviour of the initialised Hessian eigenspectrum with $N$ and $L$. In addition, we observe that the ill-conditioning grows and spikes during training (Figs.~\ref{ch5:fig:sp-train-cond-nums-GD-mnist}, \ref{ch5:fig:sp-train-cond-nums-skips-mnist}, \ref{ch5:fig:sp-train-cond-nums-GD-fashion} \& \ref{ch5:fig:sp-train-cond-nums-skips-fashion}), and using an adaptive optimiser such as Adam \cite{kingma2014adam} does not seem to help (Figs.~\ref{ch5:fig:sp-train-cond-nums-adam-mnist} \& \ref{ch5:fig:sp-train-cond-nums-adam-fashion}). Together, these findings help to explain why the convergence of the GD inference dynamics (Eq.~\ref{ch5:eq:pc-infer}) can dramatically slow down on deeper models \cite{innocenti2024jpc, pinchetti2024benchmarking}, while also highlighting that small inference gradients---which are commonly used to determine convergence---do not necessarily imply closeness to a solution.

\subsection{Vanishing/exploding forward pass}
\label{ch5:vanish-explode-ffwd}
In the previous section (\S \ref{ch5:ill-cond-infer}), we saw that the growing ill-conditioning of the inference landscape with the model size and training time is one likely reason for the challenging training of PCNs at large scale. Another reason---and as we will see the key reason---is that the forward initialisation of the activities can vanish or explode with the depth. This is a classic finding in the neural network literature that has been surprisingly ignored for PCNs. For fully connected networks with standard initialisations \cite{lecun2002efficient, glorot2010understanding, he2015delving}, the forward pass vanishes with the depth, leading to vanishing gradients. This issue can be addressed with residual connections \cite{he2016deep} and various forms of activity normalisation \cite{ioffe2015batch, ba2016layer},\footnote{The development of adaptive optimisers such as Adam \cite{kingma2014adam} was of course also crucial to deal with vanishing gradients \cite{orvieto2022vanishing}, but here we are only interested in the statistics of the forward pass.} both of which remain key components of the modern transformer block \cite{vaswani2017attention}. 

However, while there have been attempts to train ResNets with PC \cite{pinchetti2024benchmarking}, they have been without activity normalisation. This is likely because any kind of normalisation of the activities seems at odds with convergence of the inference dynamics to a solution (Eq.~\ref{ch5:eq:pc-infer}). Without normalisation, however, the activations (and gradients) of vanilla ResNets explode with the depth (see Fig.~\ref{ch5:fig:fwd-pass-stability-depth-params}). A potential remedy would be to normalise only the forward pass, but here we will aim to take advantage of more principled approaches with stronger guarantees about the stability of the forward pass (\S \ref{ch5:desiderata}).

\section{Desiderata for stable PCN parameterisation}
\label{ch5:desiderata}
In \S \ref{ch5:instability}, we exposed two main pathologies in the scaling behaviour of standard PCNs: (i) the growing ill-conditioning of the inference landscape with model size and training time (\S \ref{ch5:ill-cond-infer}), and (ii) the instability of the forward pass with depth (\S \ref{ch5:vanish-explode-ffwd}). These instabilities motivate us to specify a minimal set of \textit{desiderata} that we would like a PCN to satisfy to be trainable at large scale.\footnote{We do not see these desiderata as strict (necessary or sufficient) conditions, since relatively small PCNs can be trained competitively without satisfying them, and other conditions might be needed for successful training.}
\begin{tcolorbox}[width=\linewidth, sharp corners=all, colback=white!95!black, colframe=white!95!black]
    \begin{desideratum} \label{ch5:des1}
        \textit{Stable forward pass at initialisation.} At initialisation, all the layer preactivations are stable independent of the network width and depth, $||\mathbf{z}_\ell|| \sim \mathcal{O}_{N, H}(1)$ for all $\ell$, where $\mathbf{z}_\ell = h_\ell(\dots h_1(\mathbf{x}))$ with $h_\ell(\cdot)$ as the map relating one layer to the next.
    \end{desideratum}
\end{tcolorbox}
To our knowledge, there are two approaches that provide strong theoretical guarantees about this desideratum: (i) orthogonal weight initialisation for both fully connected \cite{saxe2013exact, pennington2017resurrecting, pennington2018emergence} and convolutional networks \cite{xiao2018dynamical}, ensuring that $\matr{W}_\ell^T\matr{W}_\ell = \matr{I}$ at every layer $\ell$; and (ii) the recent Depth-$\mu$P parameterisation \cite{yang2023tensorinfdepth, bordelon2023depthwise} (see \S \ref{ch5:mup} for a review). For a replication of these results, see Fig.~\ref{ch5:fig:fwd-pass-stability-depth-params}. To apply Depth-$\mu$P to PC, we simply reparameterise the PC energy for ResNets (Eq.~\ref{ch5:eq:pc-energy} with $\tau_\ell = 1$ for $\ell = 2, \dots, H$ and $\tau_\ell = 0$ otherwise) with the layer scalings of Depth-$\mu$P (see Table~\ref{ch5:params-table}).\footnote{$\mu$P and Depth-$\mu$P also include an optimiser-dependent scaling of the learning rate. However, we found this scaling to be suboptimal for PC as discussed in \S \ref{ch5:discussion}.} We call this reparameterisation $\mu$PC.
\begin{table}[H]
  \caption{\textbf{Summary of parameterisations.} Standard PC has unit layer premultipliers and weights initialised from a Gaussian with variance scaled by the input width at every layer $N_{\ell-1}$. $\mu$PC uses a standard Gaussian initialisation and adds width- and depth-dependent scalings at every layer.}
  \label{ch5:params-table}
  \centering
  \begin{tabular}{lllll}
    \toprule
                & $a_1$ (input weights)   & $a_\ell$ (hidden weights)   &  $a_L$ (output weights)   &  $b_\ell$ (init. variance)   \\
    \midrule
    PC          & $1$                     & $1$                         & $1$                       &  $N_{\ell-1}^{-1}$                \\
    $\mu$PC     & $N_0^{-1/2}$         & $(N_{\ell-1}L)^{-1/2}$               & $N_{L-1}^{-1}$                  &  $1$                         \\
    \bottomrule
  \end{tabular}
\end{table}
\begin{figure}[t]
    \vskip 0.2in
    \begin{center}
        \centerline{\includegraphics[width=\textwidth]{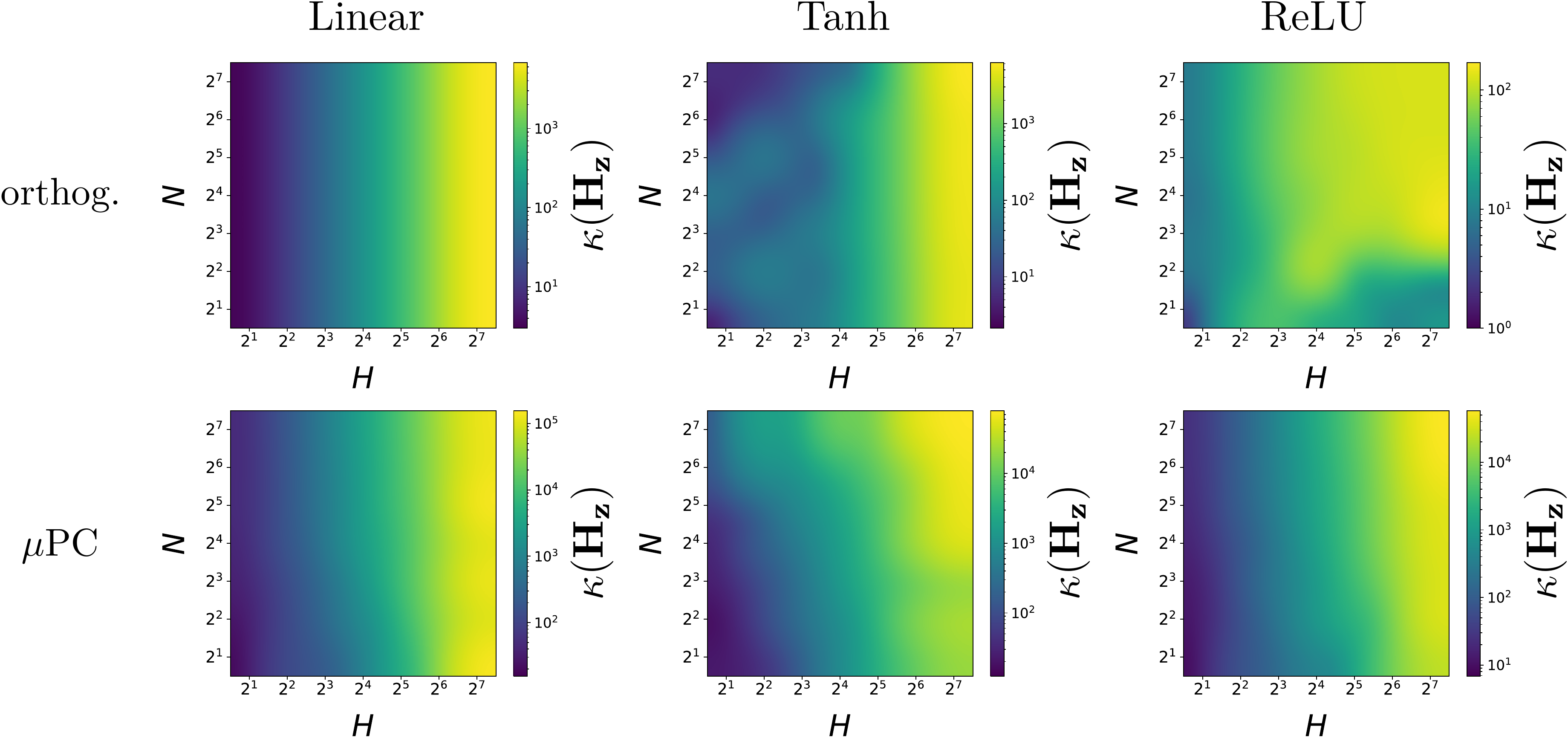}}
        \caption{\textbf{Parameterisations with stable forward passes induce highly ill-conditioned inference landscapes with depth.} We plot the conditioning of the activity Hessian of randomly initialised networks over width $N$ and depth $H$ for the $\mu$PC and orthogonal parameterisations. Networks with and without residual connections were used for these respective parameterisations. Note that ReLU networks with orthogonal initialisation cannot achieve stable forward passes (see Fig.~\ref{ch5:fig:fwd-pass-stability-depth-params}). Results were similar across different seeds.}
        \label{ch5:fig:des-trade-off}
    \end{center}
    \vskip -0.25in
\end{figure}

We would like Desideratum~\ref{ch5:des1} to hold throughout training as we state in the following desideratum.
\begin{tcolorbox}[width=\linewidth, sharp corners=all, colback=white!95!black, colframe=white!95!black]
    \begin{desideratum} \label{ch5:des2}
        \textit{Stable forward pass during training.} The forward pass is stable during training such that Desideratum~\ref{ch5:des1} is true for all training steps $t = 1, \dots, T$.
    \end{desideratum}
\end{tcolorbox}
Depth-$\mu$P ensures this desideratum for BP, but we do not know whether the same will apply to $\mu$PC. We return to this point in \S \ref{ch5:mupc-theory}. For the orthogonal parameterisation, the weights should remain orthogonal during training to satisfy Desideratum~\ref{ch5:des2}, which could be encouraged with some kind of regulariser. Next, we address the ill-conditioning of the inference landscape (\S \ref{ch5:ill-cond-infer}), again first at initialisation.
\begin{tcolorbox}[width=\linewidth, sharp corners=all, colback=white!95!black, colframe=white!95!black]
    \begin{desideratum} \label{ch5:des3}
        \textit{Stable conditioning of the inference landscape at initialisation.} The condition number of the activity Hessian (Eq.~\ref{ch5:eq:activity-hessian}) at initialisation stays constant with the network width and depth, $\kappa(\matr{H}_{\mathbf{z}}) \sim \mathcal{O}_{N, H}(1)$.
    \end{desideratum}
\end{tcolorbox}
Ideally, we would like the PC inference landscape to be perfectly conditioned, i.e. $\kappa(\matr{H}_{\mathbf{z}}) = 1$. However, this cannot be achieved without zeroing out the weights, $\matr{H}_{\mathbf{z}}(\boldsymbol{\theta} = \mathbf{0}) = \matr{I}$, since the Hessian is symmetric and so it can only have all unit eigenvalues if it is the identity. Starting with small weights $(\matr{W}_\ell)_{ij} \ll 1$ at the cost of slightly imperfect conditioning is not a solution, since the forward pass vanishes, thus violating Desideratum~\ref{ch5:des1}. See \S \ref{ch5:activity-decay} for another intervention that appears to come at the expense of performance.

What about the above parameterisations ensuring stable forward passes? Interestingly, both orthogonal initialisation and $\mu$PC induce highly ill-conditioned inference landscapes with the depth (Fig.~\ref{ch5:fig:des-trade-off}), similar to SP with skip connections (Fig.~\ref{ch5:fig:resnets-cond-nums-init}). This highlights a potential trade-off between the stability of the forward pass (technically, the conditioning of the input-output Jacobian) and the conditioning of the activity Hessian. Because PCNs with ill-conditioned inference landscapes can still be trained (e.g. see Fig.~\ref{ch5:fig:sp-train-cond-nums-GD-mnist}), we will choose to satisfy Desideratum~\ref{ch5:des1} at the expense of Desideratum~\ref{ch5:des3}, while seeking to prevent the condition number from exploding during training.
\begin{tcolorbox}[width=\linewidth, sharp corners=all, colback=white!95!black, colframe=white!95!black]
    \begin{desideratum} \label{ch5:des4}
        \textit{Stable conditioning of the inference landscape during training.} The condition number of the activity Hessian (Eq.~\ref{ch5:eq:activity-hessian}) is stable throughout training such that $\kappa(\matr{H}_{\mathbf{z}}(t)) \approx \kappa(\matr{H}_{\mathbf{z}}(t-1))$ for all training steps $t = 1, \dots, T$.
    \end{desideratum}
\end{tcolorbox}

\section{Experiments}
\label{ch5:experiments}
We performed experiments with parameterisations ensuring stable forward passes at initialisation (Desideratum~\ref{ch5:des1}), namely $\mu$PC and orthogonal, despite their inability to solve the ill-conditioning of the inference landscape with depth (Desideratum~\ref{ch5:des3}; Fig.~\ref{ch5:fig:des-trade-off}). Due to limited space, we report results only for $\mu$PC since orthogonal initialisation was not found to be as effective (see \S \ref{ch5:orthog-init}). We trained fully connected residual PCNs on simple image classification tasks (MNIST, Fashion-MNIST and CIFAR10). This simple setup was chosen because the main goal was to test whether $\mu$PC is capable of training deep PCNs---a task that has proved challenging with more complex datasets and architectures \cite{pinchetti2024benchmarking}. We note that all the networks used as many inference steps as hidden layers (see Figs.~\ref{ch5:fig:mupc-one-step-mnist} \& \ref{ch5:fig:mupc-one-step-fashion} for results with one step).

First, we trained ResNets of varying depth (up to 128 layers) to classify MNIST for a single epoch. Remarkably, we find that $\mu$PC allows stable training of networks of all depths across different activation functions (Figs.~\ref{ch5:fig:mupc-vs-pc-mnist-accs} \& \ref{ch5:fig:mupc-vs-pc-mnist-accs-all-act-fns}). These networks were tuned only for the weight and activity learning rates, with no other optimisation techniques such as momentum, weight decay, and nudging as used in previous studies \cite{pinchetti2024benchmarking}. Competitive performance ($\approx 98\%$) is achieved in 5 epochs (Fig.~\ref{ch5:fig:mupc-mnist-5-epochs}), $5\times$ faster than the current benchmark \cite{pinchetti2024benchmarking}. Similar results are observed on Fashion-MNIST, where competitive accuracy ($\approx 89\%$) is reached in fewer than 15 epochs (Fig.~\ref{ch5:fig:mupc-fashion-15-epochs}). On CIFAR10, performance is far from SOTA because of the fully connected (as opposed to convolutional) architectures used, but $\mu$PC remains trainable at large depth (Fig.~\ref{ch5:fig:cifar-20-epochs}).
\begin{figure}[t]
    \vskip 0.2in
    \begin{center}
        \centerline{\includegraphics[width=\textwidth]{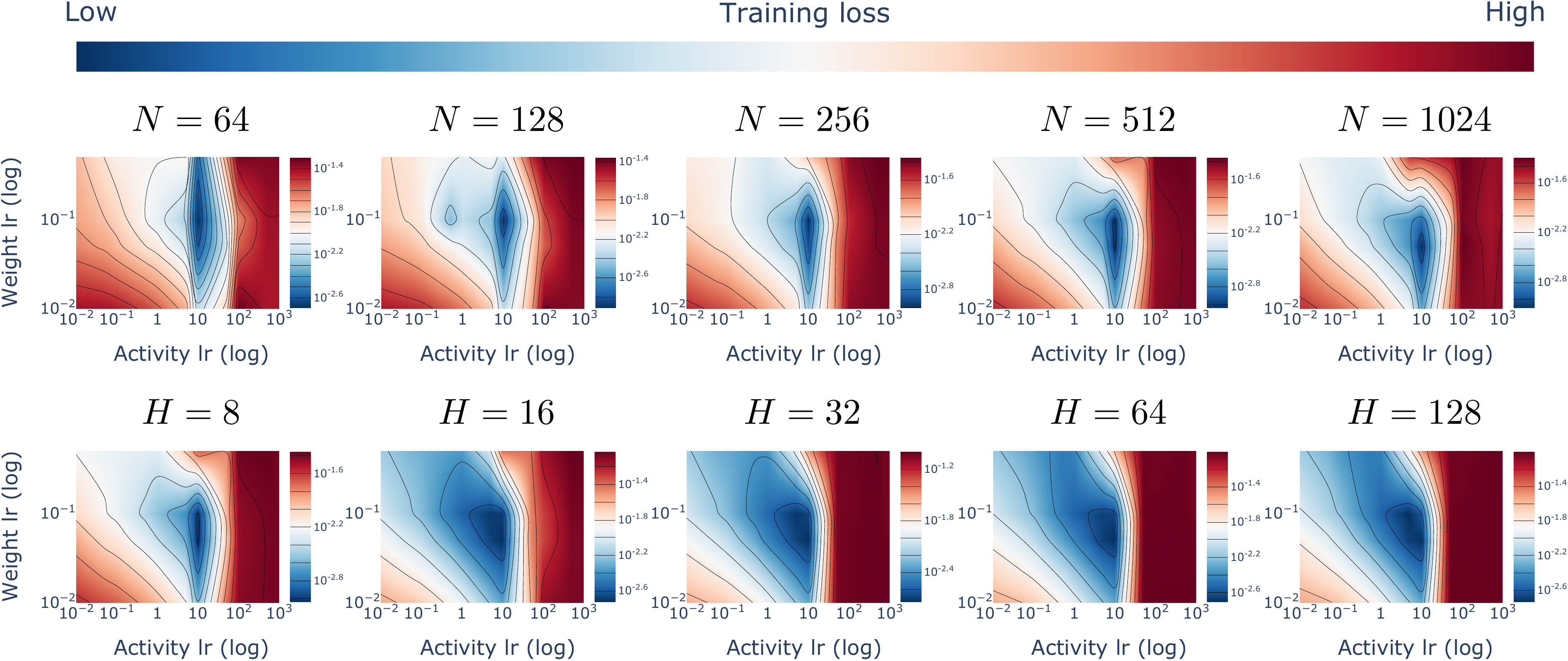}}
        \caption{\textbf{$\mu$PC enables zero-shot transfer of the weight and activity learning rates across widths $N$ and depths $H$.} Minimum training loss (log) achieved by ResNets of varying width and depth trained with $\mu$PC on MNIST across different weight and activity learning rates. All networks had Tanh as nonlinearity (see Figs.~\ref{ch5:fig:mupc-hyperparam-transfer-linear}-\ref{ch5:fig:mupc-hyperparam-transfer-relu} for other activation functions), those with varying width (first row) had 8 hidden layers, and those with varying the depth (second row) had 512 hidden units (see \S\ref{ch5:exp-details} for details). Each contour was averaged over 3 random seeds.}
        \label{ch5:fig:mupc-hyperparam-transfer-tanh}
    \end{center}
    \vskip -0.25in
\end{figure}

Strikingly, we also find that $\mu$PC enables zero-shot transfer of both the weight and activity learning rates across widths and depths (Figs.~\ref{ch5:fig:mupc-hyperparam-transfer-tanh} \& \ref{ch5:fig:mupc-hyperparam-transfer-linear}-\ref{ch5:fig:mupc-hyperparam-transfer-relu}), consistent with recent results with Depth-$\mu$P \cite{yang2023tensorinfdepth, bordelon2023depthwise}. This means that one can tune a small PCN and then transfer the optimal learning rates to wider and/or deeper PCNs---a process that is particularly costly for PC since it requires two separate learning rates. In fact, this is precisely how we obtained the Fashion-MNIST (Fig.~\ref{ch5:fig:mupc-fashion-15-epochs}) and (Fig.~\ref{ch5:fig:cifar-20-epochs}) results: by performing transfer from 8- to 128-layer networks, avoiding the expensive tuning at large scale.

\section{Is $\mu$PC BP?}
\label{ch5:mupc-theory}
\begin{figure}[h]
    \vskip 0.2in
    \begin{center}
        \centerline{\includegraphics[width=\textwidth]{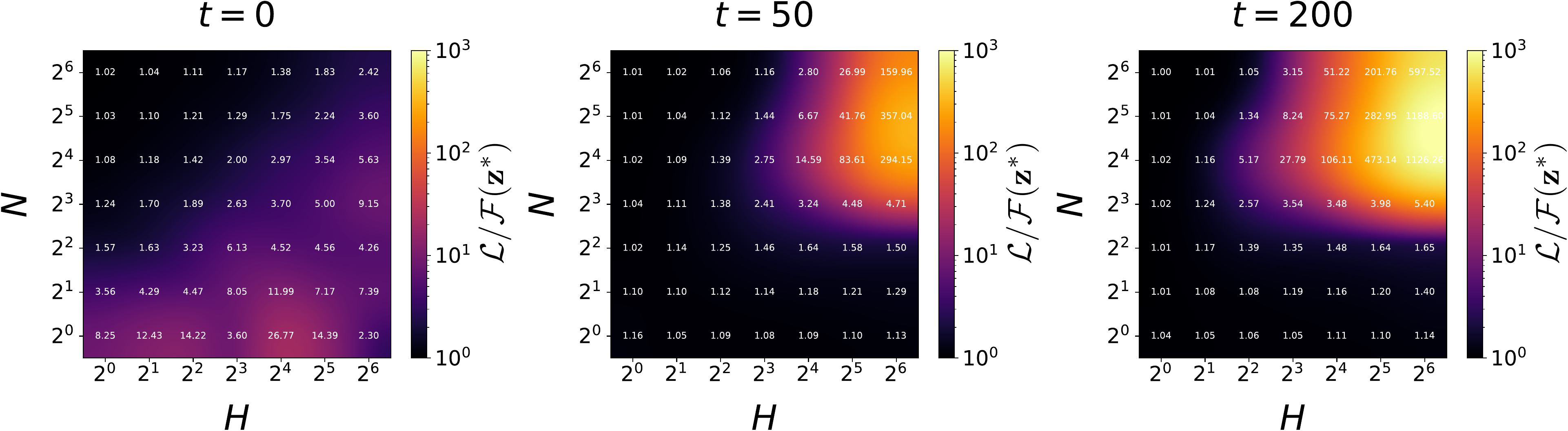}}
        \caption{\textbf{Convergence/Divergence of $\mu$PC to BP for linear ResNets.} To verify Theorem~\ref{ch5:thm1} (Eq.~\ref{ch5:eq:mupc-energy-convergence-loss}), we plot the ratio between the MSE loss and the equilibrated $\mu$PC energy of linear ResNets (Eq.~\ref{ch5:eq:resnet-equilib-energy}) at different training points $t$ as a function of the width $N$ and depth $H$ (see \S\ref{ch5:exp-details} for details). We observe that while at initialisation ($t=0$) the equilibrated energy converges to the loss as the the width grows relative to the depth (verifying Theorem \ref{ch5:thm1}), the correspondence breaks down with training at large depth and width. Results were similar across different runs.}
        \label{ch5:fig:mupc-loss-energy-ratios-init-1e-1}
    \end{center}
    \vskip -0.25in
\end{figure}

Why does $\mu$PC seem to work so well despite failing to solve the ill-conditioning of the inference landscape with depth (Fig.~\ref{ch5:fig:des-trade-off})? Depth-$\mu$P also satisfies other, BP-specific desiderata that PC might not require or benefit from. Here we show that while there is a practical regime where $\mu$PC approximates BP, it turns out to be brittle, and so BP cannot explain the success of $\mu$PC (at least on the tasks considered). In particular, it is possible to show that, when the width is much larger than the depth $N \gg L$, at initialisation the $\mu$PC energy at the inference equilibrium converges to the MSE loss. In this regime, PC computes the same gradients as BP and all the Depth-$\mu$P theory applies.
\begin{tcolorbox}[width=\linewidth, sharp corners=all, colback=white!95!black, colframe=white!95!black]
    \begin{mythm}{1}[Limit Convergence of $\mu$PC to BP.]\label{ch5:thm1}
        Let $\mathcal{F}_{\mu{\text{PC}}}(\boldsymbol{\theta}, \mathbf{z})$ be the PC energy of a randomly initialised linear ResNet (Eq.~\ref{ch5:eq:pc-energy} with $\tau_\ell = 1$ for $\ell = 2, \dots, H$ and $\tau_\ell = 0$ otherwise) parameterised with Depth-$\mu$P (Table~\ref{ch5:params-table}) and $\mathcal{L}_{\mu{\text{P}}}(\boldsymbol{\theta})$ its corresponding MSE loss. Then, as the aspect ratio of the network $r \coloneqq L/N$ vanishes, the equilibrated energy (Eq.~\ref{ch5:eq:resnet-equilib-energy}) converges to the loss (see \S \ref{ch5:limit-converge-proof} for proof)
            \begin{equation}
                r \rightarrow 0, \quad \mathcal{F}_{\mu{\text{PC}}}(\boldsymbol{\theta}, \mathbf{z}^*) = \mathcal{L}_{\mu{\text{P}}}(\boldsymbol{\theta}).
            \label{ch5:eq:mupc-energy-convergence-loss}
            \end{equation}
    \end{mythm}
\end{tcolorbox}
The result relies on the derivation in the previous chapter of the equilibrated energy as a rescaled MSE loss for DLNs \cite{innocenti2025only}. We simply generalise this to linear ResNets and show that the rescaling approaches the identity with $\mu$PC in the above limit. Fig.~\ref{ch5:fig:mupc-loss-energy-ratios-init-1e-1} shows that the result holds at initialisation ($t=0$), with the equilibrated energy converging to the loss when the width is around $32\times$ the depth. (Note that the deepest networks ($H = 128, N = 512$) we tested in the previous experiments (\S \ref{ch5:experiments}) had a much smaller aspect ratio, $r = 4$.) Nevertheless, we observe that the equilibrated energy starts to diverge from the loss with training at large width and depth (Fig.~\ref{ch5:fig:mupc-loss-energy-ratios-init-1e-1}). Note also that we do not know the inference solution for nonlinear networks. We therefore leave further theoretical study of $\mu$PC to future work. See also \S \ref{ch5:related-work} for a discussion of how Theorem~\ref{ch5:thm1} relates to previous correspondences between PC and BP. 

\section{Discussion}
\label{ch5:discussion}
In summary, we showed that it is possible to reliably train very deep (100+ layer) networks with a local learning algorithm. We achieved this via a Depth-$\mu$P-like reparameterisation of PCNs which we labelled $\mu$PC. We found that $\mu$PC is capable of training very deep networks with little tuning and competitive performance on simple classification tasks (Fig.~\ref{ch5:fig:mupc-vs-pc-mnist-accs}), while also enabling zero-shot transfer of weight and activity learning rates across widths and depths (Fig.~\ref{ch5:fig:mupc-hyperparam-transfer-tanh}). We make $\mu$PC available as part of \textsc{JPC} \cite{innocenti2024jpc}, a recent JAX library for PCNs (\sloppy{\url{https://github.com/thebuckleylab/jpc}}) presented in the next chapter.

\paragraph{$\mu$PC and inference ill-conditioning.} Despite its relative success, $\mu$PC did not solve the growing ill-conditioning of the inference landscape with the network depth (Desideratum~\ref{ch5:des3}; Fig.~\ref{ch5:fig:des-trade-off}). This can be explained by two additional findings. First, the forward pass of $\mu$PC seems to initialise the activities much closer to the analytical solution (Eq.~\ref{ch5:eq:pc-infer-solution}) for DLNs than standard PC (Fig.~\ref{ch5:fig:mupc-activity-inits}). Second, training $\mu$PC networks with a single inference step (as opposed to as many as hidden layers) led to performance degradation not only during training, but also with depth (Figs.~\ref{ch5:fig:mupc-one-step-mnist} \& \ref{ch5:fig:mupc-one-step-fashion}). Together, these results suggest that a stable forward pass, as ensured by $\mu$PC, is critical not only for performance but also for dealing with the ill-conditioning, by initialising the activities closer to a solution such that only a few (empirically determined) inference steps are needed. This is also consistent with the finding that while inference convergence is necessary for successful training of the SP, it does not appear sufficient for good generalisation (see \S\ref{ch5:infer-converge-sufficient}). It would be interesting to study $\mu$PC in more detail in linear networks given their analytical tractability.

Another recent study investigated the problem of training deep PCNs \cite{goemaere2025error}, showing an exponential decay in the activity gradients over depth. This result can be seen as a consequence of the ill-conditioning of the inference landscape with depth (Fig.~\ref{ch5:fig:sp-cond-nums-init}), since flat regions where the forward pass seems to initialise the activities (see \S \ref{ch5:activity-inits}) have small gradients, and depth drives ill-conditioning. \cite{goemaere2025error} proposed a reparameterisation of PCNs leveraging BP for faster inference convergence on GPUs, and it could be interesting to combine this approach with $\mu$PC, especially for more complex datasets and architectures where more inference steps might be necessary.

\paragraph{$\mu$PC and the other Desiderata.} Did $\mu$PC satisfy some other Desiderata (\S \ref{ch5:desiderata}) besides the stability of the forward pass at initialisation (Desideratum~\ref{ch5:des1})? When experimenting with $\mu$PC, we tried including the Depth-$\mu$P scalings only in the forward pass (i.e. removing them from the energy or even just the inference or weight gradients). However, this always led to non-trainable networks even at small depths, suggesting that the Depth-$\mu$P scalings are also beneficial for the PC inference and learning dynamics and that the resulting updates are likely to keep the forward pass stable during training (Desideratum~\ref{ch5:des2}). Deriving principled scalings specific to PC could help explain these findings or even lead to better scalings. Finally, $\mu$PC did not seem to prevent the ill-conditioning of the inference landscape from growing with training (see Figs.~\ref{ch5:fig:mupc-train-cond-nums-mnist} \& \ref{ch5:fig:mupc-train-cond-nums-fashion}), thus violating Desideratum~\ref{ch5:des4}.

\paragraph{Is $\mu$PC optimal?} $\mu$PC unlikely to be the optimal parameterisation for PCNs. This is because we adapted, rather than derived, principled (Depth-$\mu$P) scalings for BP, with only guarantees about the stability of the forward pass. Indeed, we did not rescale the learning rate of Adam (used in all our experiments) by $\sqrt{NL}$ as prescribed by Depth-$\mu$P \cite{yang2023tensorinfdepth}, since this scaling always led to non-trainable networks. We note that depth transfer has also been achieved without this scaling \cite{bordelon2023depthwise, noci2025super} and that the optimal depth scaling is still an active area of research \cite{dey2025don}. It would also be useful to better understand the relationship between $\mu$PC and the (width-only) $\mu$P parameterisation for PC proposed by \cite{ishikawa2024local} (see \S \ref{ch5:related-work} for a comparison). More generally, it would therefore be potentially impactful to derive principled scalings specific to PC. While an analysis far from inference equilibrium appears challenging, one could start with the order of the weight updates of the equilibrated energy of linear ResNets (Eq. \ref{ch5:eq:resnet-equilib-energy}).

\paragraph{Other future directions.} Given the recent successful application of Depth-$\mu$P to convolutional networks and transformers \cite{bordelon2023depthwise, noci2025super}, it would be interesting to investigate whether these more complex architectures can be successfully trained on large-scale datasets with $\mu$PC. In addition, our analysis of the inference landscape can be applied to any other algorithm performing some kind of inference minimisation (see \S \ref{ch5:other-algos} for a preliminary investigation of equilibrium propagation), and it could be interesting to see whether these algorithms could also benefit from $\mu$P.

\section*{Author contributions}
FI conceptualised the study, developed the theoretical results, ran all the experiments, and wrote the paper. EMA contributed to conceptual discussions and helped with theoretical derivations as well as code deep dives. CLB supervised the project.

%% file: text/mainmatter/ch6-jpc.tex
\chapter{JPC: Flexible Inference for PCNs in JAX}
\label{ch:jpc}
\minitoc

\section{Abstract}
We introduce \textsc{JPC}, a \textbf{J}AX library for training neural networks with \textbf{P}redictive \textbf{C}oding (PC). \textsc{JPC} provides a simple, fast and flexible interface to train a variety of PC networks (PCNs) including discriminative, generative and hybrid models. In addition to standard discrete optimisers, \textsc{JPC} offers ordinary differential equation solvers to integrate the continuous gradient flow inference dynamics of PCNs. \textsc{JPC} also provides a number of theoretical tools that can be used to study PCNs. We hope that \textsc{JPC} will facilitate future research of PC. The code is available at \sloppy{\url{www.github.com/thebuckleylab/jpc}}.

\section{Introduction}
As reviewed in previous chapters, in recent years predictive coding (PC) has been explored as a biologically plausible alternative to standard backpropagation \cite{van2024predictive, millidge2021predictive, millidge2022predictivereview, salvatori2023brain}. However, with a few recent notable exceptions \cite{legrand2024pyhgf, pinchetti2024benchmarking}, there has been a lack of unified open-source implementations of PC networks (PCNs) which would facilitate research and reproducibility\footnote{We also acknowledge earlier libraries such as \href{https://github.com/infer-actively/pypc}{\texttt{pypc}} and \href{https://github.com/RobertRosenbaum/Torch2PC}{\texttt{Torch2PC}} \cite{rosenbaum2022relationship}.}.

In this short chapter, we introduce ``\textsc{JPC}'', a \textbf{J}AX library for training neural networks with \textbf{PC}. \textsc{JPC} provides a simple, fast and flexible interface for training a variety of PCNs including discriminative, generative and hybrid models. Like JAX, \textsc{JPC} follows a fully functional programming paradigm that is close to the mathematics, and the core library is less than 1000 lines of code. This is in contrast to the recently introduced \textsc{PCX} \cite{pinchetti2024benchmarking}, another JAX-based PCN library that instead takes an object-oriented approach, leading to a less intuitive implementation. Unlike existing libraries, \textsc{JPC} also offers ordinary differential equation solvers (ODE) to integrate the continuous gradient flow inference dynamics of PCNs (Eq. \ref{pcns:eq:pc-infer}), in addition to standard discrete optimisers.\footnote{As discussed in \S \ref{ch6:conclusion}, subsequent work \cite{dherin2025learning} also investigated ODE solvers for standard neural network training.} \textsc{JPC} also provides some theoretical tools that can be used to study and potentially identify problems with PCNs.

In the rest of this chapter, we first present \textsc{JPC}'s core design (\S\ref{ch6:design}). For a review of PC, we refer the reader to Chapter~\ref{ch:pcns}. We then report some empirical results showing that a second-order ODE solver can achieve significantly faster runtimes than standard Euler integration of the gradient flow PC inference dynamics, with comparable performance on different datasets and networks (\S\ref{ch6:runtime}). We conclude with a brief discussion of the results and possible extensions of \textsc{JPC} (\S\ref{ch6:conclusion}).

\section{Design and Implementation} 
\label{ch6:design}
\textsc{JPC} provides both a simple high-level application programming interface (API) to train and test PCNs in a few lines of code (\S\ref{ch6:basic-api}) and more advanced functions offering greater flexibility as well as additional features (\S\ref{ch6:advanced-api}). It is built on top of three main JAX libraries:
\begin{itemize}
    \item \href{https://github.com/patrick-kidger/equinox}{\texttt{Equinox}} \cite{kidger2021equinox} to define neural networks with PyTorch-like syntax,
    \item \href{https://github.com/patrick-kidger/diffrax}{\texttt{Diffrax}} \cite{kidger2022neural} to leverage ODE solvers to integrate the gradient flow PC inference dynamics (Eq.~\ref{pcns:eq:pc-infer}), and
    \item \href{https://github.com/google-deepmind/optax}{\texttt{Optax}} \cite{bradbury2018jax} for parameter optimisation (Eq.~\ref{pcns:eq:pc-learn}).
\end{itemize}
Below we provide a sketch of \textsc{JPC} with pseudocode, referring the reader to the \href{https://thebuckleylab.github.io/jpc/}{documentation} and the \href{https://thebuckleylab.github.io/jpc/examples/discriminative_pc/}{example notebooks} for more details.

\subsection{Basic API}  
\label{ch6:basic-api}
The function \code{ jpc.make\_pc\_step } allows one to update the parameters of essentially any \href{https://github.com/patrick-kidger/equinox}{\texttt{Equinox}} network compatible with PC updates.

\begin{jupyterCell}
from (*@\textcolor{codeblue}{jpc}@*) import make_pc_step

result (*@\textcolor{codepurple}{=}@*) make_pc_step(
    model,     # equinox model with callable layers
    optim,     # optax optimiser
    opt_state, # optimiser state
    y,         # target
    x          # optional input
)

# updated model and optimiser
model (*@\textcolor{codepurple}{=}@*) result["model"]
opt_state (*@\textcolor{codepurple}{=}@*) result["opt_state"]
\end{jupyterCell}
As shown above, at a minimum \code{ jpc.make\_pc\_step } takes a model, an \href{https://github.com/google-deepmind/optax}{\texttt{Optax}} optimiser and its 
state, and some data. For a model to be compatible with PC updates, it needs to be split into callable layers (see the 
\href{https://thebuckleylab.github.io/jpc/examples/discriminative_pc/}{example notebooks}). Note also 
that an input is not needed for unsupervised training. In fact, 
\code{ jpc.make\_pc\_step } can be used for both classification and generation tasks by simply flipping the model's input and output, and for 
supervised as well as unsupervised training (again see the \href{https://thebuckleylab.github.io/jpc/examples/discriminative_pc/}{example notebooks}). 

Under the hood, \code{ jpc.make\_pc\_step }:
\begin{enumerate}
    \item integrates the gradient flow PC inference dynamics (Eq.~\ref{pcns:eq:pc-infer}) using a \href{https://github.com/patrick-kidger/diffrax}{\texttt{Diffrax}} ODE solver (a second-order explicit Runge–Kutta method called ``Heun'' by default), and
    \item updates the parameters at the converged value of the activities (Eq.~\ref{pcns:eq:pc-learn}) with a given \href{https://github.com/google-deepmind/optax}{\texttt{Optax}} optimiser.
\end{enumerate}
Default parameters such as the ODE solver and a step size controller can all be overridden. One has also the option of recording a variety of metrics including the energies and activities at each inference step. 

Importantly, \code{ jpc.make\_pc\_step } is designed to use JAX's native ``just-in-time'' (jit) compilation for optimised performance, and the user only needs to embed this function in a data loop to train a neural network. We also provide convenience, already-jitted functions for testing specific PC models, such as \code{ jpc.test\_discriminative\_pc } and \code{ jpc.test\_generative\_pc }.

A similar API is provided for hybrid PC (HPC) models \cite[see][]{tscshantz2023hybrid} with \code{ make\_hpc\_step }:
\begin{jupyterCell}
from (*@\textcolor{codeblue}{jpc}@*) import make_hpc_step

result (*@\textcolor{codepurple}{=}@*) make_hpc_step(
    generator,     # generative model
    amortiser,     # model for inference amortisation
    optims,        # optimisers, one for each network
    opt_states,    # optimisers' state
    y,                 
    x 
)

\end{jupyterCell}
where now one has to pass an additional model (and associated optimiser objects) for amortising the inference of the generative model. Again, there is an option to change the default ODE solver parameters and record different metrics, and the convenience function \code{ jpc.test\_hpc } for testing HPC is also provided. We refer to the \href{https://thebuckleylab.github.io/jpc/examples/hybrid_pc/}{example notebook on HPC} for more details.

\subsection{Advanced API} 
\label{ch6:advanced-api}
While convenient and abstracting away many of the details, the basic API can be limiting, for example if one would like to perform some additional computations within the default PC training step \code{ jpc.make\_pc\_step }. Advanced users have therefore the option of accessing all the underlying functions of the basic API as well as additional features.

\paragraph{Custom step function.} A custom PC training step would look like the following.
\begin{jupyterCell}
import jpc

# 1. initialise activities with a feedforward pass
activities (*@\textcolor{codepurple}{=}@*) jpc.init_activities_with_ffwd(model, x)

# 2. run iterative inference (Eq. (*@\ref{pcns:eq:pc-infer}@*))
converged_activities (*@\textcolor{codepurple}{=}@*) jpc.solve_inference(
    params(*@\textcolor{codepurple}{=}@*)(model, None),
    activities(*@\textcolor{codepurple}{=}@*)activities,
    output(*@\textcolor{codepurple}{=}@*)y,
    (*@\textcolor{black}{input}@*)(*@\textcolor{codepurple}{=}@*)x
)

# 3. update parameters at the converged activities (Eq. (*@\ref{pcns:eq:pc-learn}@*))
update_result (*@\textcolor{codepurple}{=}@*) jpc.update_params(
    params(*@\textcolor{codepurple}{=}@*)(model, None),
    activities(*@\textcolor{codepurple}{=}@*)converged_activities,
    optim(*@\textcolor{codepurple}{=}@*)optim,
    opt_state(*@\textcolor{codepurple}{=}@*)opt_state,
    output(*@\textcolor{codepurple}{=}@*)y,
    (*@\textcolor{black}{input}@*)(*@\textcolor{codepurple}{=}@*)x
)
\end{jupyterCell}
This can be embedded in a ``jitted'' function with any other additional computations. One has also the option of using any \href{https://github.com/google-deepmind/optax}{\texttt{Optax}} optimiser, including standard GD, to perform inference. In addition, the user can access (i) other initialisation methods for the activities, (ii) the standard energy functions for PC and HPC, and (iii) the activity as well as parameter gradients used by the update functions. In fact, this is essentially all there is to \textsc{JPC}, providing a simple framework to extend the library for different use cases.
\begin{figure}[t]
    \begin{center}
        \centerline{\includegraphics[width=\textwidth]{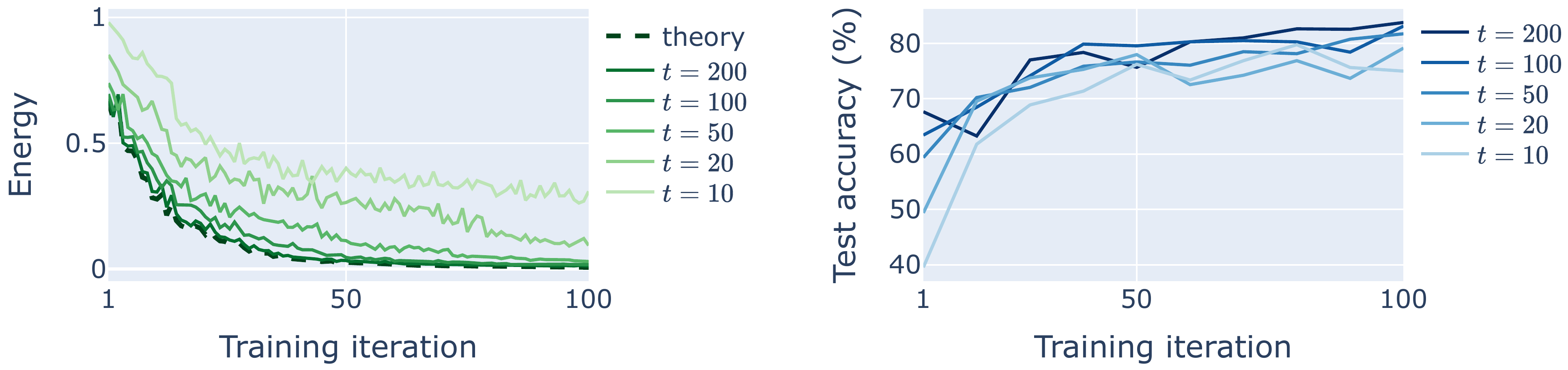}}
        \caption{\textbf{Theoretical PC energy for deep linear networks (Eq.~\ref{ch6:equilib-energy}) can help predict whether more inference could lead to better performance.} We compare the theoretical energy (Eq.~\ref{ch6:equilib-energy}) with the numerical energy for different upper limits $t$ of inference integration, along with test accuracies, for a linear network ($H=10$, $N=300$) trained to classify MNIST with learning rate $1e^{-3}$ and batch size $64$. Results were consistent across different random initialisations.}
        \label{ch6:fig:energies-accs-mnist}
    \end{center}
    \vskip -0.2in
\end{figure}

\paragraph{Theoretical tools.} \textsc{JPC} also comes with some analytical tools that can be used to both study, and potentially diagnose issues with, PCNs. These tools originate from work covered in the previous two chapters related to the analysis of linear PCNs \cite{innocenti2025only, innocenti2025mu}. As an example, in Chapter~\ref{ch:saddles} we saw that for deep linear networks the energy at the inference equilibrium of the activities $\nabla_{\mathbf{z}} \mathcal{F} = \mathbf{0}$ has the following closed-form solution as a rescaled mean squared error loss (Theorem~\ref{ch4:thm1})
\begin{equation}
    \mathcal{F}^* = \frac{1}{2B} \sum_{i=1}^B (\mathbf{y}_i - \matr{W}_{L:1}\mathbf{x}_i)^T \matr{S}^{-1}(\mathbf{y}_i - \matr{W}_{L:1}\mathbf{x}_i)
    \label{ch6:equilib-energy}
\end{equation}
where the rescaling is $\matr{S} = \matr{I}_{N_L} + \sum_{\ell=2}^L (\matr{W}_{L:\ell})(\matr{W}_{L:\ell})^T$, and we use the shorthand $\matr{W}_{k:\ell} = \matr{W}_k \dots \matr{W}_\ell$ for $\ell, k \in 1,\dots, L$.

Experiments in Chapter~\ref{ch:saddles} showed a perfect match between the theory (Eq.~\ref{ch6:equilib-energy}) and the numerical energy of linear PCNs (Figure~\ref{ch4:fig:dln-equilib-energy}). Figure~\ref{ch6:fig:energies-accs-mnist} suggests that the theoretical energy can also help determine whether sufficient inference has been performed, in that more inference steps seem to correlate with higher test accuracy, at least on MNIST. Similar results are observed on Fashion-MNIST (see Figure~\ref{ch6:fig:energies-accs-fashion}). The other theoretical tools provided by \textsc{JPC} include the solution of the activities for linear PCNs (Eq.~\ref{ch5:eq:pc-infer-solution}) and the related Hessian of the energy with respect to the activities (Eq.~\ref{ch5:eq:activity-hessian})—both of which were derived in the previous chapter. As previously mentioned, \textsc{JPC} also includes implementations of $\mu$PC (see the \href{https://thebuckleylab.github.io/jpc/examples/mupc/}{example notebook}), which as demonstrated in the previous chapter allows stable training of 100+ layer PCNs with little tuning and competitive performance on simple tasks.

\section{Runtime efficiency of basic ODE solvers} 
\label{ch6:runtime}
A comprehensive benchmarking of various types of PCN with (discrete-time) gradient descent (GD) as inference optimiser was recently performed by \cite{pinchetti2024benchmarking}. As a preliminary investigation of the ODE solvers' performance, we compared the runtime efficiency of two basic ODE solvers, namely standard Euler integration of the inference gradient flow dynamics and Heun (a second-order Runge-Kutta method). Note that, as a second-order method, Heun has a higher computational cost than Euler; however, it could still be faster if it requires significantly fewer steps to converge.

The solvers were compared on feedforward networks trained to classify simple image datasets with different number of hidden layers $H \in \{3, 5, 10\}$. Because our goal was to specifically test for runtime, we trained each network for only one epoch across different initial step sizes $dt \in \{5e^{-1}, 1e^{-1}, 5e^{-2}\}$, selecting the run with the highest mean test accuracy achieved (see Figures~\ref{ch6:fig:best-test-accs}-\ref{ch6:fig:accs-per-ts-cifar}). Unlike Euler, Heun employed a standard Proportional–Integral–Derivative step size controller. Therefore, to make comparison fair, we also trained networks with a range of upper integration limits $T \in \{5, 10, 20, 50, 100, 200, 500\}$, again reporting the run with the maximum accuracy (Figures~\ref{ch6:fig:best-test-accs}-\ref{ch6:fig:accs-per-ts-cifar}). In cases where the accuracy difference between any $T$ was not significantly different, we selected runs with the smaller $T$.

\begin{figure}[t]
    \begin{center}
        \centerline{\includegraphics[width=\textwidth]{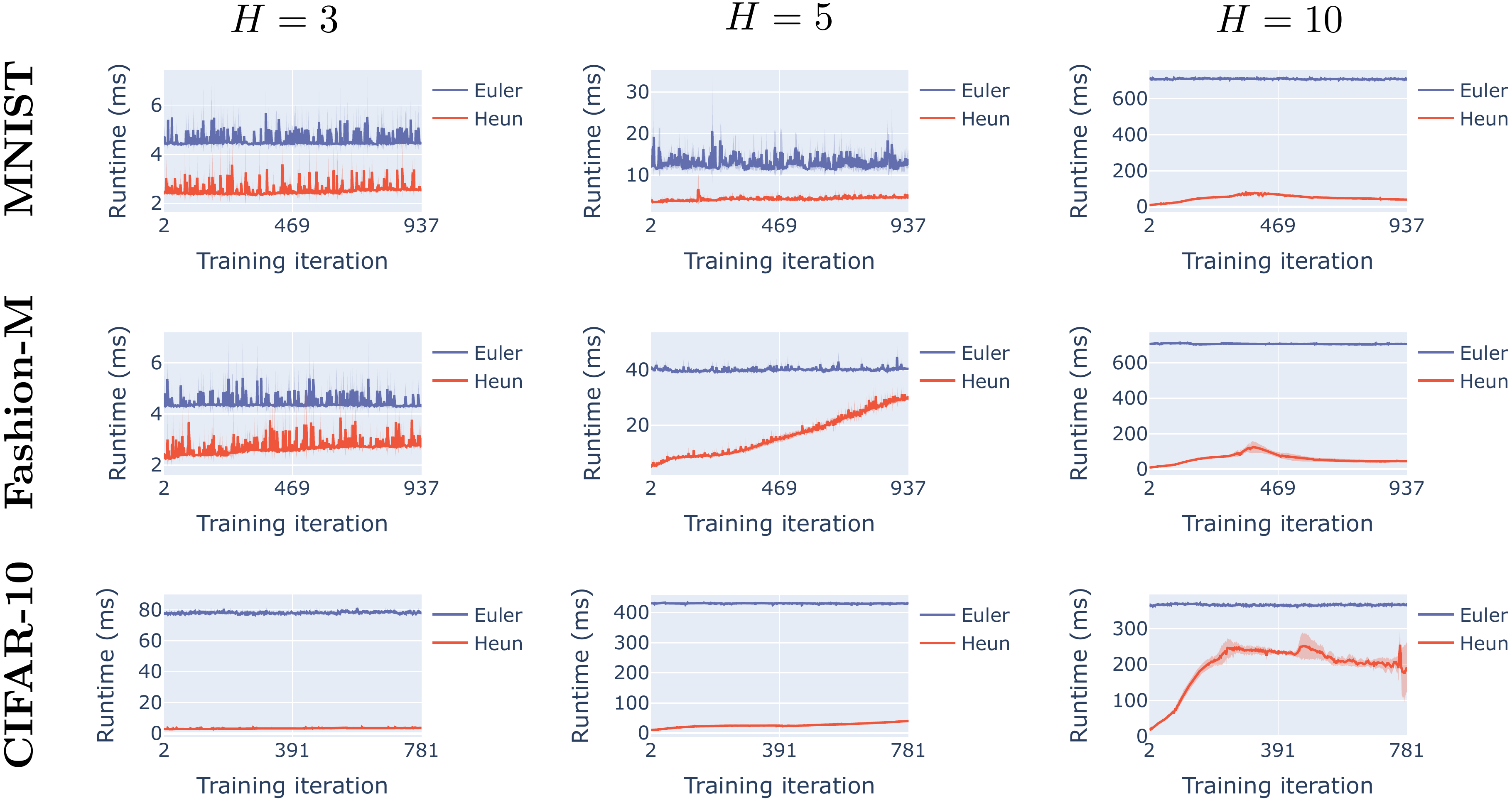}}
        \caption{\textbf{A second-order Runge–Kutta method (Heun) solves PC inference faster than standard Euler on a range of datasets and networks.} We plot the wall-clock time of Euler and Heun at each training step of one epoch for networks with hidden layers $H \in \{3, 5, 10\}$ trained on standard image classification datasets. The runs with the highest mean test accuracy achieved across different hyperparameters were selected (see Figures~\ref{ch6:fig:best-test-accs}-\ref{ch6:fig:accs-per-ts-cifar}). The time of the first training iteration where ``just-in-time'' (jit) compilation occurs is excluded. All networks had 300 hidden units and Tanh as activation function, and were trained with learning rate $1e^{-3}$ and batch size $64$. Shaded regions indicate $\pm1$ standard deviation across 3 different random weight initialisations.}
        \label{ch6:fig:best-infer-runtimes}
    \end{center}
    \vskip -0.2in
\end{figure}

Figure~\ref{ch6:fig:best-infer-runtimes} shows that, despite requiring more computations at each step, Heun tended to converge significantly faster than Euler, and in general more so on deeper networks ($H=10$). However, the convergence behaviour of Euler was more consistent during training across datasets and network depths, with Heun sometimes increasing in runtime. It is also important to note that other optimiser-specific hyperparameters could lead to different results, and we welcome the community to test these and other solvers against other tasks as well as implementations.

\section{Conclusion} 
\label{ch6:conclusion}
We introduced \textsc{JPC}, a new JAX library for training a variety of PCNs. Unlike existing frameworks, \textsc{JPC} is extremely simple (<1000 lines of code), completely functional in design, and offers well-tested ODE solvers to integrate the gradient flow inference dynamics of PCNs. We showed that a second-order solver can provide significant speed-ups in runtime over standard Euler integration across a range of datasets and networks. Importantly, these results should not be taken to mean that ODE solvers will outperform (in speed or performance) standard discrete optimisers, and it is not unlikely that different types of optimiser will be suited to different settings. Indeed, \cite{dherin2025learning} recently evaluated the performance of higher-order ODE solvers for standard neural network training, finding that they can be challenging to scale. As a straightforward extension of \textsc{JPC}, it would be interesting to integrate stochastic differential solvers, which recent work associates with better generation performance \cite{zahid2023sample, oliviers2024learning}. Adding a custom energy function for transformer-based architectures \cite{vaswani2017attention} could also be an interesting direction. We hope that, together with other recent PC libraries \cite{pinchetti2024benchmarking, legrand2024pyhgf}, \textsc{JPC} will help facilitate research on PCNs.

\section*{Author contributions}
FI wrote all the library code, ran the experiments, and wrote the paper. PK originally came up with the idea of using ODE solvers to integrate the gradient flow PC inference dynamics. WYF and MdLV helped test the library, and RS and CLB contributed to conceptual discussions.

%% file: text/mainmatter/ch7-conclusions.tex
\chapter{Conclusions} 
\label{ch:conclusions}
\minitoc

\begin{quote}
    “All models are wrong, but some are useful.”

    \hfill --- George E. P. Box
\end{quote}

\begin{figure}[t]
    %\vskip 0.2in
    \begin{center}
        \centerline{\includegraphics[width=0.9\textwidth]{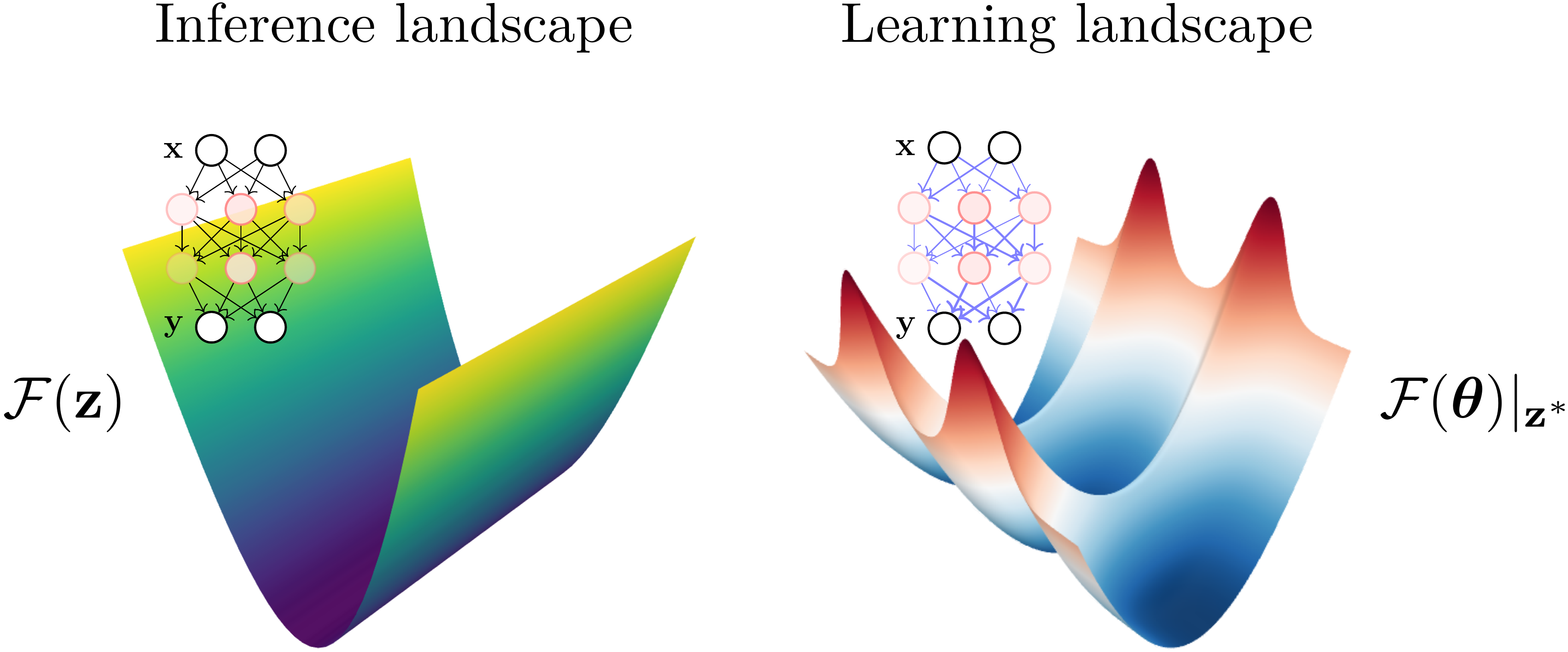}}
        \caption{\textbf{Cartoon depiction of the inference and learning landscapes of PCNs.} Note that the learning landscape is denoted as $\mathcal{F}(\boldsymbol{\theta})|_{\mathbf{z}^*}$ to emphasise that it is a function of the weights evaluated at an equilibrium of the network activities.}
        \label{ch7:fig:cartoon-pc-landscapes}
    \end{center}
    \vskip -0.25in
\end{figure}

In this concluding section, we briefly review the goal and main contributions of this thesis (\S \ref{ch7:summary}), discuss their implications in a unified manner for both neuroscience and AI (\S \ref{ch7:implications}), and speculate on the future of predictive coding (PC) and other local learning algorithms (\S \ref{ch7:speculations}). We also briefly discuss some general limitations of this work (\S \ref{ch7:limitations}). At several points in the discussion, it may be useful to refer to Figure~\ref{ch7:fig:cartoon-pc-landscapes} as a simplified but faithful picture of the inference and learning landscapes of PC networks (PCNs) revealed by previous chapters.

\section{Summary}
\label{ch7:summary}
This thesis studied PC as a biologically plausible and potentially more efficient algorithm than standard backpropagation (BP). We sought to understand how deep neural networks (DNNs) trained with PC work at a fundamental level, with the goal of determining whether PC can be scaled to larger models and datasets as successfully as BP. As reviewed in detail in \textbf{Chapter}~\ref{ch:pcns}, the distinguishing feature of PCNs is the way they perform inference by equilibration of their activities (via gradient-based minimisation) before learning or weight updates. The bulk of this thesis focused on developing theories of the inference and learning landscape and dynamics of practical PCNs, using deep linear networks (DLNs) as a theoretical model.
% (As shown in Figure~\ref{ch7:fig:cartoon-pc-landscapes}, we will refer to the inference landscape as $\mathcal{F}(\mathbf{z})$, and to the learning landscape as $\mathcal{F}(\boldsymbol{\theta})|_{\mathbf{z}^*}$ to emphasise that it is a function of the weights evaluated at the inference solution.)

More specifically, \textbf{Chapter~\ref{ch:trust-region}} showed that the learning dynamics of PC can be implicitly understood as an approximate trust-region method using second-order information, despite explicitly using only first-order information. Leveraging DLNs, \textbf{Chapter~\ref{ch:saddles}} developed a more precise theory and showed that, for feedforward networks, the objective on which PC effectively learns (at inference equilibrium) is equal to a rescaled (mean squared error) loss that is more robust to vanishing gradients and, under certain conditions, much easier to navigate. These works formalised the impact of inference on learning in PCNs. \textbf{Chapter~\ref{ch:mupc}}, on the other hand, focused on the inference dynamics of PCNs, showing (i) that the landscape becomes increasingly ill-conditioned with model size (width and particularly depth) and training time, and (ii) that the forward pass of standard PCNs tends to vanish/explode with depth. Motivated by these findings, we proposed a new parameterisation of PCNs that for the first time allowed stable training of 100+ layer networks with little tuning and competitive performance on simple tasks. Finally, \textbf{Chapter~\ref{ch:jpc}} introduced \textsc{JPC}, a Python library for training a variety of PCNs using JAX. For a breakdown of these contributions, see also Table~\ref{contributions-table}.

\section{Implications}
\label{ch7:implications}
What do the above results, especially related to Chapters~\ref{ch:trust-region}-\ref{ch:mupc}, mean for the neuroscience and machine learning (ML) of PC?

\subsection{Neuroscience}
\label{ch7:implications-neuro}
While this thesis focused primarily on scaling PC for AI, the uncovered learning and inference dynamics of PCNs provide potential insights into the learning and inference problems likely faced by the brain, some of which were already discussed in previous chapters. First, we suggest that alternate, gradient-based optimisation of the same objective with respect to both activities and weights, as in PC, constitutes a biologically plausible way for the brain to deal with an inevitably ill-conditioned learning problem. Second, we argue that the brain must also have mechanisms for dealing with a similar ill-conditioning of the inference landscape, which standard PCNs largely lack at present. Below, we unpack these points.

\paragraph{Learning in the brain.} What does our study of the learning dynamics of PCNs suggest about learning in the brain, if anything? The work in Chapter~\ref{ch:trust-region} suggested that the PC weight update uses second-order (curvature) information about the loss landscape. Chapter~\ref{ch:saddles} showed that this conclusion was limited by the second-order approximation made in the analysis and that, in fact, PC can in principle use \textit{arbitrarily higher-order information} (depending on the degeneracy of the loss saddles and therefore the depth of the network). This is arguably a very surprising and significant result. 

Understanding why requires a brief detour on the learning problem likely faced by the brain. As we have learned from almost two decades of training artificial DNNs, the weight or learning landscape of such networks is extremely ill-conditioned (e.g. full of degenerate saddles as we saw in Chapter~\ref{ch:saddles}) because they are, similarly to the brain, highly overparameterised (i.e. with many more parameters than data points). As Chapters~\ref{ch:saddles}-\ref{ch:mupc} showed, ill-conditioned landscapes are challenging to navigate, especially for first-order methods like SGD. To help with ill-conditioning, deep learning theorists and practitioners therefore developed a variety of techniques, including adaptive optimisers (e.g. Adam \cite{kingma2014adam}), normalisation strategies (e.g. LayerNorm \cite{ba2016layer}) and better-conditioned architectures (e.g. ResNets \cite{he2016deep}), many of which remain the standard for training large-scale models. 

Yet, while individual biological neurons can perform more complex computations than artificial ones \cite{beniaguev2021single}, it is hard to see how the brain could implement any of these techniques without BP, as many of them require computing extra, arguably non-local variables. Adam, for example, requires storing first- and second-moment estimates of the gradients. Moreover, it is implausible for the brain to directly compute second- or higher-order information to help with ill-conditioning, since the Hessian is an inherently global matrix encoding interactions between \textit{all} neurons in the network.

With this context in mind, the significance of our result should now be clearer. In particular, recall that we showed that \textit{higher-order information about an outer optimisation problem (learning) can be implicitly computed by an inner optimisation process (inference) on the same objective (energy) using only first-order, local information}. More succinctly, multiple inference gradient updates allow for a higher-order learning weight update, thus suggesting a biologically plausible mechanism for how the brain could deal with a very ill-conditioned learning problem.

\paragraph{Inference in the brain.} What about the inference dynamics of PCNs? Do they suggest anything about the inference challenges faced by the brain? Here it is important to recall that PCNs perform inference \textit{iteratively} (which is why we can talk about dynamics at all), in contrast to standard neural networks, where inference is typically amortised (with a feedforward pass). Therefore, we need to consider the \textit{inference landscape} in addition to the stability of the forward pass. Perhaps unsurprisingly, Chapter~\ref{ch:mupc} revealed that the inference landscape of deep and wide PCNs is also extremely ill-conditioned (Figure \ref{ch7:fig:cartoon-pc-landscapes}), although more benign than the weight landscape (i.e. convex in the linear case).\footnote{Note that this property (ill-conditioning) is \textit{hardware-independent}, as it was shown to depend only on the network structure and the value of the weights (see \S \ref{ch5:ill-cond-infer}).}
 
This raises a similar question as above: if the brain performs even some degree of iterative inference, how does it deal with ill-conditioning? We saw in Chapter~\ref{ch:mupc} that a stable forward pass seems to help by initialising the activities closer to a solution and that this stability can be achieved with mostly local information.\footnote{The only non-local quantity required was the model depth, but this is a constant and so it is not hard to imagine how the brain could have mechanisms accounting for ``its own depth'' (whatever that is).} However, it is not possible to perform a forward pass in unsupervised settings, and empirically, many more iterations tend to be needed for generative (as opposed to discriminative) tasks. A hybrid strategy, combining iterative and amortised inference as in \cite{tscshantz2023hybrid, oliviers2025bidirectional}, could help. Such hybrid schemes have also increased biological and cognitive plausibility in bottom-up (feedforward) vs top-down (feedback) pathways and fast vs slow inference. 

Another solution might be found in the hardware itself. Recently, \cite{aifer2024thermodynamic} showed that a kind of thermodynamics-based hardware (essentially exploiting the intrinsic noise of the system) could solve ill-conditioned linear problems significantly faster than state-of-the-art digital methods. This could be explained by \textit{fast diffusion dynamics induced by the physics of the hardware}, and it is a basic fact of neuroscience that noise is a feature (rather than a bug) of the brain \cite{faisal2008noise}. The previous study suggests that implementing PC on similar hardware could lead to fast convergence of the inference dynamics despite ill-conditioning and perhaps entirely eschew the need to ensure a stable forward pass (since PC inference converges to the forward pass when the output is free to vary; see e.g. \S \ref{ch5:activity-grad-hess}). This is also consistent with recent studies showing that noisy (Langevin-based) inference updates can lead to some benefits for generation tasks \cite{oliviers2024learning, zahid2023sample}. We return to this point in our speculations below.

\subsection{AI}
\label{ch7:implications-ai}
Having discussed the potential insights that our work might afford for neuroscience, what does it mean for AI? In particular, does PC provide any practical benefits in terms of efficiency or performance for training DNNs compared to standard BP? The short answer is ``no'': \textbf{\textit{while DNNs trained with PC clearly show some advantageous properties over BP, these benefits are negated or become computationally prohibitive at large scale, at least on standard digital hardware (GPUs).}} The rest of this section justifies this conclusion, while the last section speculates on whether a different kind of hardware, potentially more suited to PC, could lead to a different conclusion.

First, let us again revisit the \textit{learning dynamics} of PCNs (Eq.~\ref{pcns:eq:pc-learn}). The landscape theory developed in Chapter~\ref{ch:saddles} suggested that, at or close to an inference equilibrium, deep fully connected networks should be easier to train with PC than BP \textit{under very specific conditions} (since convergence depends on many factors including the optimiser, architecture, initialisation, etc.).\footnote{Indeed, any serious answer to the question of more efficient learning should consider \textit{all} the memory and compute costs involved in training PCNs at both the hardware and software level. We will answer this question below since it is inextricably linked with the cost of PC inference.} In particular, we saw that learning speed-ups with PC should be expected for gradient descent with small step size initialised near saddle points (Figure \ref{ch7:fig:cartoon-pc-landscapes}), as confirmed in \S\ref{ch4:experiments} for models with up to 10 layers. These conditions helped explain conflicting findings in the literature and qualified previous claims about the convergence benefits of PC compared to BP \cite{song2022inferring}. 

As discussed in Chapter~\ref{ch:saddles}, however, these conditions are not realistic: networks are in practice initialised far from the origin saddle, skip connections shift the location of this saddle from the origin \cite{hardt2016identity}, and adaptive (faster saddle-escaping) algorithms like Adam are used \cite{staib2019escaping}. Moreover, even before we begin to compare the computational cost of these techniques with that of PC inference, training very deep (10+ layer) PCNs has proved challenging as we demonstrated in Chapter~\ref{ch:mupc}, negating any potential benefits of PC at large scale.

These observations bring us to the \textit{inference dynamics} of PCNs. Chapter~\ref{ch:mupc} revealed that the challenge of training very deep PCNs was due to a combination of two main factors: (i) an ill-conditioning of the inference landscape with model size and training time (Figure \ref{ch7:fig:cartoon-pc-landscapes}), and (ii) a poor (vanishing/exploding) forward pass initialisation of the activities. We then saw that addressing the forward pass stability by using a specific ResNet parameterisation allowed reliable training of 100+ layer PCNs on simple tasks. 

However, as mentioned above, ResNets effectively shift the origin saddle \cite{hardt2016identity}, making them more robust to vanishing gradients \cite{orvieto2022vanishing}. Therefore, because skip connections are key to the stability of the parameterisation introduced in Chapter~\ref{ch:mupc}—and because this is the only approach to date that allows training of very deep PCNs—any potential convergence benefit of PC inference is unlikely to be realised on modern architectures, including transformers (since ResNets form their backbone). 

Moreover, even if some other way of scaling PC to very deep networks is found, \textit{and} faster learning convergence is determined under realistic conditions, ultimately the speed-up in learning would have to be measured against the slow-down in inference. As suggested by the experiments in Chapter~\ref{ch:mupc}, the cost of PC inference for models with a stable forward pass scales at least linearly with the number of layers, which is about two orders of magnitude more expensive than BP inference on 100+ layer models. This cost could potentially be reduced with a hybrid scheme combining generative and amortiser models \cite{tscshantz2023hybrid, oliviers2025bidirectional}, but at the expense of roughly double the number of parameters and more complex training dynamics. 

As discussed in Chapter~\ref{ch:mupc}, this analysis applies to any algorithm performing some kind of inference optimisation, including equilibrium propagation \cite{scellier2017equilibrium, zucchet2022beyond}. Importantly, it also applies to any other benefit that PC inference might confer (e.g. in continual learning tasks) \cite{song2022inferring}, not just learning convergence, which we mainly focused on. For these reasons, PC (and likely other energy-based algorithms) are currently incapable of providing any \textit{practical} improvements at scale over BP in performance or efficiency.

\section{Limitations}
\label{ch7:limitations}
The main limitations of this thesis arguably have more to do with breadth rather than depth of analysis. In this section, we frame our results in a broader context by briefly discussing some related lines of research.

This thesis studied one among many alternative algorithms to BP, and within these, a particular brain-inspired algorithm. Indeed, even within PC, our experiments (and occasionally theory) were restricted to specific versions or modes of PC (see \S \ref{ch:pcns} for a review), although the conclusions reached do not fundamentally change for PC in general. For example, our theories of the learning dynamics of PCNs (Chapters~\ref{ch:trust-region}-\ref{ch:saddles}) are restricted to supervised settings (generative or discriminative), although an extension to the unsupervised case could possibly be developed. Related experiments focused on the discriminative case (with images as inputs and labels as targets), but similar results can be expected for the generative case. On the other hand, our theory of the inference dynamics of PCNs (Chapter~\ref{ch:mupc}) applies to any setting, but our experiments were again limited to the discriminative case for computational reasons. In general, as previously mentioned, one should expect generative tasks to require more inference than discriminative ones.

As alluded to above, recent hybrid PC schemes combining iterative and amortised inference \cite{tscshantz2023hybrid, oliviers2025bidirectional} also do not fundamentally change the conclusions of this thesis. Two such schemes have been proposed: Hybrid PC (HPC \cite{tscshantz2023hybrid}) and Bidirectional PC (BPC \cite{oliviers2025bidirectional}). HPC augments standard PC with an additional bottom-up network that learns to amortise (or ``shortcut'') the inference process of standard PC. First of all, the stability of the forward pass of the amortiser model would also have to be ensured to avoid the same issue of vanishing/exploding activations discussed in Chapter~\ref{ch:mupc}. This could be achieved with ``$\mu$PC''. Second, as mentioned above, the additional network introduces more parameters and complex training dynamics. BPC differs from HPC in that the inference dynamics are driven by both the top-down and bottom-up models. This impacts the inference conditioning of BPC models which, while it would require a separate analysis, is also likely to be poor at large model size based on the in-depth study of Chapter~\ref{ch:mupc}.

Beyond PC, there are many other alternative algorithms to BP. In addition to previously mentioned schemes such as equilibrium propagation \cite{zucchet2022beyond}, target propagation \cite{meulemans2020theoretical}, and forward learning \cite{hinton2022forward}, there are spike-based learning rules \cite{lagani2023spiking}, promising even greater energy efficiencies. Indeed, a spiking-neuron implementation of PC has been proposed \cite{mikulasch2022dendritic}. There are also, of course, non-bio-inspired alternatives to BP, such as zero\textit{th}-order optimisation \cite{malladi2023fine, chen2023deepzero} and forward gradients using directional derivatives \cite{singhal2023guess, fournier2023can, baydin2022gradients, belouze2022optimization, silver2021learning}.

\section{Speculations}
\label{ch7:speculations}
Having concluded that PC cannot at present provide any practical benefits over BP (\S \ref{ch7:implications-ai}), I believe that there are two major challenges that need to be addressed if PC and similar algorithms are to have a chance of competing with BP at the scale of modern AI applications such as large language models.

First, it still remains to be seen whether very deep PCNs can be trained on more complex datasets and models such as transformers (or equally expressive architectures). Chapter~\ref{ch:mupc} took an important step in this direction by achieving training stability for 100+ layer fully connected ResNets on simple classification tasks. Future work should focus on extending these results to more complicated architectures and datasets, such as convolutional neural networks trained on ImageNet. However, while the modifications we made to PC (``$\mu$PC'') to allow stable training of very deep networks suggest that these results should transfer to more complicated architectures (as discussed in Chapter~\ref{ch:mupc}), other changes might be needed. In particular, it remains unknown whether standard transformers \cite{vaswani2017attention}, shallow or deep, can be trained at all with PC.

Second, even if PC is proved to be capable of training very deep and expressive architectures at scale, it is clear that, to compete with BP, it will need to be implemented on some other hardware than standard GPUs. As explained above, this is because of the high computational cost of PC inference as an inherently sequential process that is slow to simulate on digital hardware. Indeed, this may be key to scaling PC in the first place, since faster simulations would facilitate research and experimentation.

%% file: text/appendices/ch3.tex
\chapter{Appendix for Chapter 3}
\label{ch3:appendix}
\minitoc

\renewcommand{\thefigure}{A.\arabic{figure}}
\setcounter{figure}{0}

\renewcommand{\thetheorem}{A.\arabic{theorem}}
\setcounter{theorem}{0}

\section{Experiment details} 
\label{ch3:exp-details}

\subsection{Toy models} 
\label{ch3:toy-exp}
1MLPs were trained with BP and PC to predict a simple linear function $y = -x$ where $x \sim \mathcal{N}(1, 0.1)$. We used a uniform weight initialisation $\mathbf{w}_i \sim \mathcal{U}(-1, 1)$ and SGD with batch size $64$ and learning rate $\eta = 0.2$ to aid visualisation of the algorithms' learning trajectory. Training was stopped when the test loss reached the tolerance $\mathcal{L}_{\text{test}} < 0.001$. For PC, standard GD was used to solve the inference dynamics (Eq.~\ref{ch3:eq:pc-infer}), with a feedforward pass initialisation, step size $\beta = 0.1$ and $T = 20$ iterations (which were sufficient to reach equilibrium).

In Figure~\ref{ch3:fig:cos-sims}, we computed the cosine similarity between the optimal weight direction $\Delta \mathbf{w}^* = (w_1^* - w_1, w_2^* - w_2)$ and the algorithms' GD update at a given point $\Delta \mathbf{w} = - \nabla_{\mathbf{w}} f$:
\begin{equation}
    cos(\Delta \mathbf{w}^*, \Delta \mathbf{w}) = \frac{\langle \Delta \mathbf{w}^*, \Delta \mathbf{w} \rangle}{\| \Delta \mathbf{w}^* \| \| \Delta \mathbf{w} \|},
\end{equation}
which is simply a normalised dot product. To calculate the optimal direction, at each training batch we solved for the shortest (Euclidean) distance from the current iterate $\mathbf{w} = (w_1, w_2)$ to the manifold of solutions $\mathbf{w}^* = (w_1^*, \frac{y}{w_1^* x}) = (w_1^*, -\frac{1}{w_1^*})$,
\begin{equation}
    D = \sqrt{\left( -\frac{1}{w_1^*} - w_2 \right)^2 + (w_1^* - w_1)^2}.
\end{equation}
To minimise this distance, we set the partial derivative of the distance w.r.t. the optimal weight $w_1^*$ to zero
\begin{equation}
    \frac{\partial D}{\partial w_1^*} = \frac{(w_1^*)^4 - (w_1^*)^3w_1 - w_1^* w_2 - 1}{(w_1^*)^3 \sqrt{\left( - \frac{1}{w_1^*} - w_2 \right)^2 + (w_1^* - w_1)^2}} = 0.
\end{equation}
Finding the roots of this derivative means solving for the quartic polynomial in the numerator, for which we used $\mathtt{numpy}$.

\subsection{Deep chains} 
\label{ch3:chain-exps}
We trained deep chains using SGD with batch size 64. To control for the learning rate $\eta$, we peformed a grid search over $\eta = \{1e^{-4}, 1e^{-3}, 1e^{-2}, 1e^{-1}, 1e^{-0}\}$ and compared the loss dynamics for the learning rate with the minimum training loss for each algorithm. Linear and Tanh chains were trained on the same regression task used for the toy models, $y = -x$ with $x \sim \mathcal{N}(1, 0.1)$, and were initialised with PyTorch default's He initialisation \cite{he2015delving}. ReLU chains were instead trained to predict a positive linear function $y = 2x$ to avoid mapping to zero. For the same reason, weights were initialised from a uniform distribution with positive interval $\mathbf{w}_i \sim \mathcal{U}(0.5, 1)$. 

We recorded the training and test loss on every data batch from initialisation and stopped training if either (i) the training loss on the current batch reached the threshold $\mathcal{L}_{\text{train}} < 0.01$, (ii) the average training loss (estimated every 500 batches) did not decrease, or (iii) the loss diverged to infinity (typically because of high learning rates). For PC, we used an inference schedule similar to that of \cite{song2022inferring}, halving the step size $\beta = 0.1$ up to two times, with maximum $T = 500$ inference iterations.

\subsection{Deep and wide networks} 
\label{ch3:dnn-exps}
Networks of width $N = 500$ and depth $H=10$ were trained on MNIST with SGD, batch size 64, and the same learning rate grid search used for the deep chains. As standard, the MNIST images were normalised. Training was stopped if the training loss did not decrease from the previous epoch or diverged to infinity. For PC, all the hyperparameters were the same as for deep chains (\S \ref{ch3:chain-exps}) except for a maximum of $T = 1000$ inference iterations, used to guard against the possibility that any training failure was due to insufficient inference.

\section{Toy model proofs}
Here we present our two theorems on 1MLPs, showing (i) that PC escapes the saddle point at the origin faster than BP with (S)GD, and (ii) that the 1MLP mimina of the equilibrated energy are flatter than those of the loss. 

\begin{definition}
\label{ch3:def:1mlp-prob}
\textit{1MLP problem.} We define a 1MLP problem as any non-degenerate linear function of the form $y = mx, x, y \neq 0$ that can in principle be learned by a 1MLP $f(x) = w_2w_1x$ where $x, y$ indicate the input and output to the network, respectively.
\end{definition}

\begin{definition}
\label{ch3:def:saddle}
\textit{(Strict) saddle.} A critical point $\mathbf{w}^*$ of $f(\mathbf{w})$ where $\nabla f(\mathbf{w}^*) = 0$ is a saddle if the Hessian at that point has at least one positive and one negative eigenvalue, $\lambda_{\text{max}}(\nabla^2 f(\mathbf{w}^*)) > 0$, $\lambda_{\text{min}}(\nabla^2 f(\mathbf{w}^*)) < 0$. In the literature, these critical points are known as strict or non-degenerate saddles \cite{ge2015escaping, anandkumar2016efficient, jin2017escape}. We will study these and other types of saddle in more detail in Chapter~\ref{ch:saddles}.
\end{definition}

Consider the BP mean squared error loss and PC energy (Eq.~\ref{pcns:eq:pc-energy}) associated with a 1MLP problem (Def. \ref{ch3:def:1mlp-prob}):
\begin{align} 
    \mathcal{L} &= \frac{1}{2} (y - w_2w_1x)^2 \label{ch3:eq:mse-loss} \\
    \mathcal{F} &= \frac{1}{2}(z - w_1x)^2 + \frac{1}{2}(y - w_2z)^2
\end{align} 
where $z$ indicates the value of the hidden unit or latent in PC (which is free to vary). Without loss of generality, we assume a single input-output pair. Note that we can change the sign of the weights without changing the objectives, $f(\mathbf{w}) = f(-\mathbf{w})$. This is known as a ``sign-flip symmetry'' and induces a saddle at the origin of the weight landscape \cite{chen1993geometry, bishop2006pattern}. Now recall that we are interested in how PC inference (Eq.~\ref{ch3:eq:pc-infer}) affects the weight update at convergence of the activities (Eq.~\ref{ch3:eq:pc-learn}). In the linear case, we can analytically solve for the inference equilibrium $\partial \mathcal{F} / {\partial z} = 0, z^* = \frac{w_1x + w_2y}{1 + w_2^2}$ and evaluate the energy at this fixed point
\begin{equation}
    \mathcal{F}^* = \frac{\mathcal{L}}{1 + w_2^2}.
    \label{ch3:eq:equilib-energy}
\end{equation}
where we use $\mathcal{F}^*$ as an abbreviation for $\mathcal{F}(\mathbf{z}^*)$. The origin $\mathbf{w}^* = (0, 0)$ is critical point of both the loss and the equilibrated energy since their gradient is zero, $\nabla_{\mathbf{w}} \mathcal{L}(\mathbf{w}^*) = \nabla_{\mathbf{w}} \mathcal{F}^*(\mathbf{w}^*) = \mathbf{0}$. To confirm that this point is a (strict) saddle (Def. \ref{ch3:def:saddle}), we look at the Hessians
\begin{align}
    \matr{H}_{\mathcal{L}}(\mathbf{w}^*) &= \begin{bmatrix} 0 & -xy \\ -xy & 0 \end{bmatrix} \\
    \matr{H}_{\mathcal{F}^*}(\mathbf{w}^*) &= \begin{bmatrix} 0 & -xy \\ -xy & -y^2 \end{bmatrix}
\end{align} 
and see that indeed they both have positive and negative eigenvalues $\lambda(\matr{H}_{\mathcal{L}}) = \pm xy$, $\lambda(\matr{H}_{\mathcal{F}^*}) = \frac{1}{2}(-y^2 \pm y\sqrt{4x^2 + y^2})$. Crucially, however, the eigenvalues of the energy are smaller than those of the loss
\begin{equation}
    \begin{cases}
        \lambda_{\text{max}}(\matr{H}_{\mathcal{F}^*}) < \lambda_{\text{max}}(\matr{H}_{\mathcal{L}}) \\
        \lambda_{\text{min}}(\matr{H}_{\mathcal{F}^*}) < \lambda_{\text{min}}(\matr{H}_{\mathcal{L}}),
    \end{cases}
    \label{ch3:eq:eigens}
\end{equation}
which can be shown by using the fact that the square root of a sum is always smaller than the sum of the square roots, $\sqrt{a^2 + b^2} < \sqrt{a^2} + \sqrt{b^2}$ for $a, b \neq 0$. This result is sufficient to prove that PC will escape the saddle faster than BP, since the near-saddle (S)GD dynamics are controlled by the local curvature. To see this, consider a second-order Taylor expansion of some objective $f$ around the saddle
\begin{equation}
    f(\mathbf{w}^* + \Delta \mathbf{w}) \approx f(\mathbf{w}^*) + \frac{1}{2} \Delta \mathbf{w}^T \matr{H}_f \Delta \mathbf{w},
    \label{ch3:eq:2nd-taylor-approx}
\end{equation}
where the gradient vanishes. As shown by \cite{lee2016gradient}, taking a gradient descent step of size $\eta$ from this approximation leads to the following recursive update
\begin{align}
    \mathbf{w}_{t+1} &= (I - \eta \matr{H}_f)^{t+1} \mathbf{w}_0 \nonumber \\ 
    &= \sum_{i=1}^{n_w} (1 - \eta \lambda_i)^{t+1} \langle e_i \mathbf{w}_0 \rangle e_i
    \label{ch3:eq:saddle-updates}
\end{align} 
where $\mathbf{w}_0 = (\mathbf{w}^* + \Delta \mathbf{w})$, $n_w = 2$ is the number of parameters, and $\{\lambda_i\}_i^{n_w}$ are the Hessian eigenvalues with corresponding eigenvectors $\{e_i\}_i^{n_w}$. We see that (S)GD will be attracted to, and repelled from, the saddle depending on the degree of curvature along those directions. Because the equilibrated energy has smaller Hessian eigenvalues than the loss at the saddle (Eq.~\ref{ch3:eq:eigens}), PC will be simultaneously less attracted to and more repelled from it than BP. In dynamical systems terms, the energy saddle turns out to be more ``unstable''---and therefore easier to escape---than the loss saddle.

\setcounter{theorem}{0}
\begin{theorem}
\label{ch3:thm:saddle-escape}
Given any 1MLP problem (Def. \ref{ch3:def:1mlp-prob}) which induces a saddle (Def. \ref{ch3:def:saddle}) at the origin in weight space, (S)GD on the equilibrated PC energy (Eq.~\ref{ch3:eq:equilib-energy}) will escape the saddle faster than on the quadratic BP loss (Eq.~\ref{ch3:eq:mse-loss}).
\end{theorem}
\begin{wrapfigure}{r}{0.5\textwidth}
    \vskip 0.1in
    \begin{minipage}{\linewidth}
        \centering
        \centerline{\includegraphics[width=0.9\columnwidth]{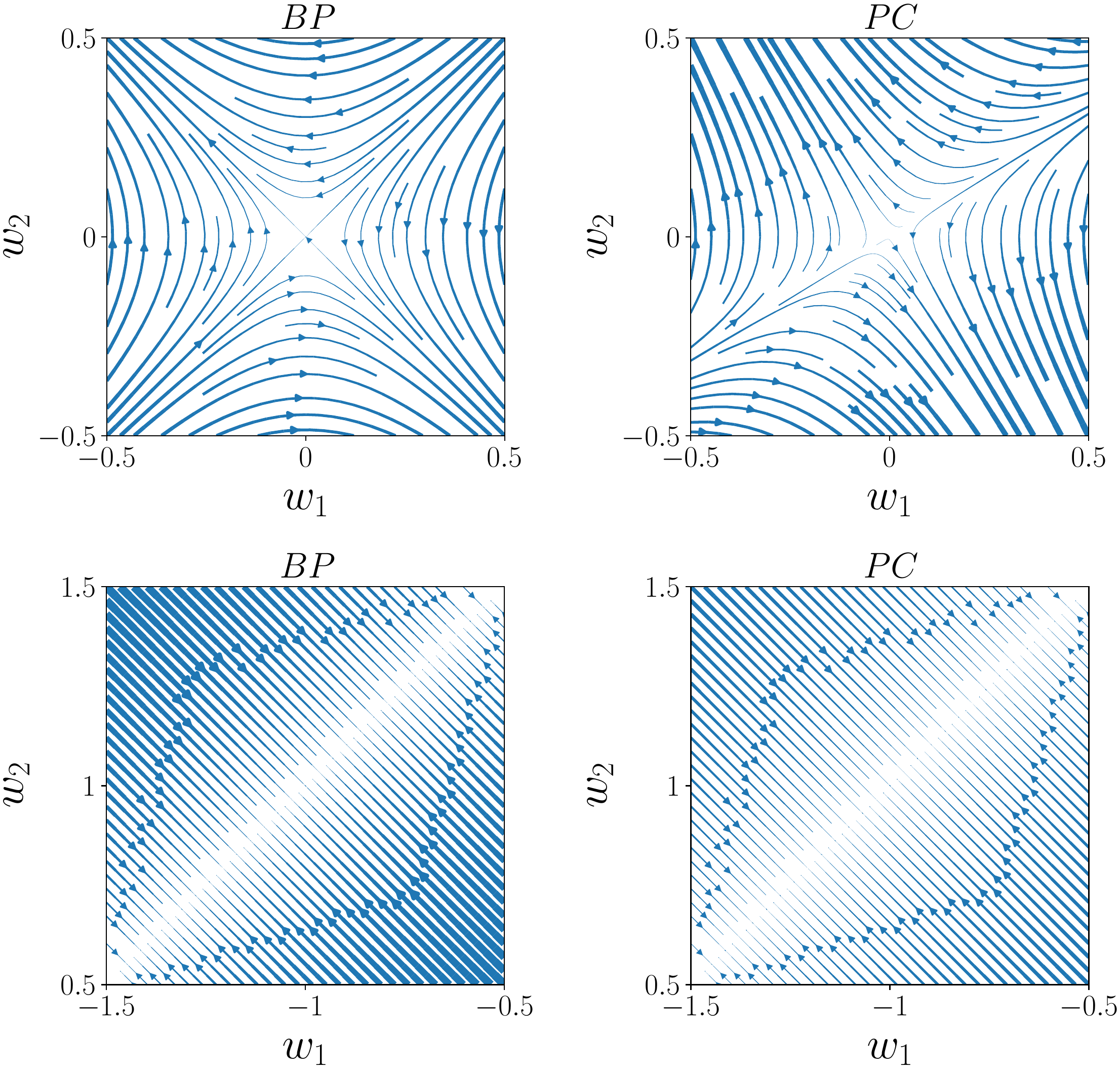}}
        \caption{\textbf{Gradient flow of BP vs PC near different critical points on a toy network.} Continuous-time GD dynamics in the vicinity of the saddle (\textit{top}) and an example minimum (\textit{bottom}) of a 1MLP trained with BP and PC on the same regression problem illustrated in Figure~\ref{ch3:fig:toy-net}. We observe that the continuous dynamics are a good approximation of the discrete ones (Figure~\ref{ch3:fig:toy-net}).}
        \label{ch3:fig:grad-flows}    
    \end{minipage}
    \vskip -0.2in
\end{wrapfigure}
% \begin{figure}[h]
%     \vskip 0.2in
%     \begin{center}
%         \centerline{\includegraphics[width=0.6\columnwidth]{figures/ch3/grad_flows.pdf}}
%         \caption{\textbf{Gradient flow of BP vs PC near different critical points on a toy network.} Continuous-time GD dynamics in the vicinity of the saddle (\textit{top}) and an example minimum (\textit{bottom}) of a 1MLP trained with BP and PC on the same regression problem illustrated in Figure~\ref{ch3:fig:toy-net}. We observe that the continuous dynamics are a good approximation of the discrete ones (Figure~\ref{ch3:fig:toy-net}).}
%         \label{ch3:fig:grad-flows}
%     \end{center}
%     \vskip -0.2in
% \end{figure}

We can also see this by taking the continuous limit of the near-saddle GD dynamics $\eta \rightarrow 0$ (Eq.~\ref{ch3:eq:saddle-updates}, Figure~\ref{ch3:fig:grad-flows}), leading to the linear ordinary differential equation (ODE) system (gradient flow)
\begin{equation}
    \dot{\mathbf{w}}(t) = - \matr{H}_f \mathbf{w}(t)
\end{equation}
with solution $\mathbf{w}(t) = \matr{Q}e^{\matr{\Lambda} t}\matr{Q}^T \mathbf{w}(0)$ and initial condition $\mathbf{w}(0) = (\mathbf{w}^* + \Delta \mathbf{w})$.

Using the same approach, we can also show that any 1MLP global minimum\footnote{It is easy to show that these minima are global since the saddle is the only other type of critical point in this toy example.} of the equilibriated energy is \textit{flatter} than any corresponding minimum of the loss. Formally, the Hessian eigenvalues of equilibrated energy will also be smaller than those of the loss at any minimum. Because 1MLPs already pose an overparameterised (underdetermined) problem, there is no unique solution but rather a manifold. That is, for any value of one weight, there exists only one optimal value of the other, e.g. $\mathbf{w}^* = (\frac{y}{w_2x}, w_2)$. These are also all critical points of both the loss and energy, since their gradient is zero $\nabla_{\mathbf{w}} \mathcal{L}(\mathbf{w}^*) = \nabla_{\mathbf{w}} \mathcal{F}^*(\mathbf{w}^*) = \mathbf{0}$. To verify that this is a manifold of minima, as before we look at the Hessian and see that they both have one zero eigenvalue $\lambda_{\text{min}}(\matr{H}_{\mathcal{L}}) = \lambda_{\text{min}}(\matr{H}_{\mathcal{F}^*}) = 0$ and one positive eigenvalue $\lambda_{\text{max}}(\matr{H}_{\mathcal{L}}) = \frac{w_2^4x^2 + y^2}{w_2^2}$ and $\lambda_{\text{max}}(\matr{H}_{\mathcal{F}^*}) = \frac{w_2^4x^2 + y^2}{w_2^2(1+w_2^2)}$. It is straightforward to see that the positive curvature of the energy is smaller than that of the loss, $\lambda_{\text{max}}(\matr{H}_{\mathcal{F}^*}) < \lambda_{\text{max}}(\matr{H}_{\mathcal{L}})$.

\begin{theorem}
\label{ch3:thm:flat-min}
Given any 1MLP problem (Def. \ref{ch3:def:1mlp-prob}), the minima of the equilibrated PC energy (Eq.~\ref{ch3:eq:equilib-energy}) are flatter than the corresponding minima of the quadratic BP loss (Eq.~\ref{ch3:eq:mse-loss}).
\end{theorem}

\begin{figure}[H]
    \vskip 0.2in
    \begin{center}
        \centerline{\includegraphics[width=0.5\columnwidth]{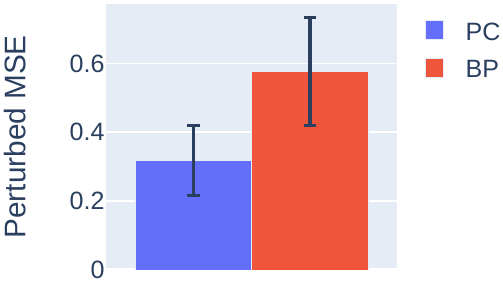}}
        \caption{\textbf{PC is more robust to near-minimum weight perturbations than BP on a toy network.} Mean squared error (MSE) between output target and weight-perturbed prediction $(y - \hat{y})^2$ of BP and PC trained on the same 1MLP problem illustrated in Figure~\ref{ch3:fig:toy-net}. Weights were perturbed with i.i.d. Gaussian noise $\xi \sim \mathcal{N}(0, 0.5)$. Error bars indicate the standard error of the mean across 10 different seeds.}
        \label{ch3:fig:weight-perturb}
    \end{center}
    \vskip -0.2in
\end{figure}

Performing the same quadratic approximation and GD analysis as above (Eqs.~\ref{ch3:eq:2nd-taylor-approx}-\ref{ch3:eq:saddle-updates}) around this manifold of minima leads to the conclusion that GD will converge slower than BP in the vicinity of a minimum but also be more robust to random weight perturbations where the local approximation holds (Figure~\ref{ch3:fig:weight-perturb}). As before we can make a similar argument for the continuous case, which is illustrated in Figure~\ref{ch3:fig:grad-flows}.

\section{Derivations of theoretical results} 
\label{ch3:derivations}
\textbf{Free energy expansion}.
Recall from Chapter~\ref{ch:pcns} that the PC energy is a sum of local prediction errors at every layer
\begin{equation}
    \mathcal{F} = \frac{1}{2}||\mathbf{y} - \mathbf{z}_L||^2 + \sum_{\ell=1}^{L-1} \frac{1}{2} ||\mathbf{z}_\ell - h_\ell(\mathbf{z}_{\ell-1}; \boldsymbol{\theta}_\ell) ||^2
    \label{ch3:eq:pc-energy}
\end{equation}
where $h_\ell(\mathbf{z}_{\ell-1}; \boldsymbol{\theta}_\ell)$ is some (potentially nonlinear) parameterised function of the activities of the previous layer. We choose such a general formulation because our results below apply in principle to any model for which a feedforward pass can be defined. Let $\{ \hat{\mathbf{z}}_\ell = h_\ell(\dots h_1(\mathbf{x})) \}_{\ell=1}^L$ represent the forward activations. We perform a second-order Taylor expansion of the PC energy (Eq.~\ref{ch3:eq:pc-energy})
\begin{align}
    \mathcal{F}(\mathbf{z}) =\mathcal{F}(\hat{\mathbf{z}}) &+ \matr{J}_\mathcal{F}^T(\hat{\mathbf{z}})\Delta \mathbf{z} \nonumber \\
    &+ \frac{1}{2}\Delta \mathbf{z}^T \matr{H}_{\mathcal{F}}(\hat{\mathbf{z}}) \Delta \mathbf{z} + \mathcal{O}(\Delta \mathbf{z}^3),
\end{align}
where $\Delta \mathbf{z} = (\mathbf{z} - \hat{\mathbf{z}})$, and $\matr{J}_\mathcal{F}^T(\hat{\mathbf{z}})$ and $\matr{H}_{\mathcal{F}}(\hat{\mathbf{z}})$ are the Jacobian and Hessian of the energy with respect to the forward pass values, respectively. We now observe (i) that the energy is equal to the (mean squared error) loss at the forward pass $\mathcal{F}(\hat{\mathbf{z}}) = \mathcal{L}(\hat{\mathbf{z}})$, and (ii) that the Jacobian term is equal to the gradient of the loss with respect to the activations $\matr{J}_\mathcal{F}^T(\hat{\mathbf{z}})=\mathbf{g}_{\mathcal{L}}(\hat{\mathbf{z}})$, since in both cases the terms in the sum collapse at the forward values. In addition, $\matr{H}_{\mathcal{F}}(\hat{\mathbf{z}}) \approx - \frac{\partial^2 \mathbb{E}_{y, x}\ln p(y, \mathbf{z}, x)}{\partial \mathbf{z}^2 } \mid_{\hat{\mathbf{z}}} = \mathcal{I}(\hat{\mathbf{z}})$ can be seen as the Fisher information of the forward values with respect to the model $p$. Hence
\begin{align}
    \mathcal{F}(\mathbf{z}) = \mathcal{L}(\hat{\mathbf{z}}) &+ \mathbf{g}_\mathcal{L}^T(\hat{\mathbf{z}})\Delta \mathbf{z} \nonumber \\ 
    &+ \frac{1}{2}\Delta \mathbf{z}^T \mathcal{I}(\hat{\mathbf{z}}) \Delta \mathbf{z} + \mathcal{O}(\Delta \mathbf{z}^3).
\end{align}
\textbf{Approximate inference solution}. 
If we assume that $\mathcal{O}(\Delta \mathbf{z}^3)$ is a small contribution, we can approximate the inference equilibrium by finding the stationary point of the second-order expansion, yielding
\begin{equation}
    \mathbf{z}^* \approx \hat{\mathbf{z}} - \mathcal{I}(\hat{\mathbf{z}})^{-1}\mathbf{g}_{\mathcal{L}}(\hat{\mathbf{z}}).
\end{equation}
\textbf{Approximate weight update}.
As reviewed in \S \ref{ch3:pc}, after the activities converge (at an inference equilibrium), PC takes a gradient step on the energy with respect to the weights. In order to find this, we first calculate $\frac{\partial \mathcal{F}}{\partial \boldsymbol{\theta}} = \frac{\partial \hat{\mathbf{z}}}{\partial \boldsymbol{\theta}}\frac{\partial \mathcal{F}}{\partial \hat{\mathbf{z}}}$:
\begin{equation}
    \frac{\partial \mathcal{F}}{\partial \boldsymbol{\theta}} = \frac{\partial \hat{\mathbf{z}}}{\partial \boldsymbol{\theta}}\left[-\Delta \mathbf{z} - \mathcal{I}(\hat{\mathbf{z}})^T\Delta \mathbf{z} + \mathcal{O}(\Delta \mathbf{z}^2)\right].
\end{equation}
Finally, plugging in the equilibrium value $\mathbf{z}^*$, we obtain
\begin{align}
    \frac{\partial \mathcal{F}(\mathbf{z}^*)}{\partial \boldsymbol{\theta}} &\approx \frac{\partial \hat{\mathbf{z}}}{\partial \boldsymbol{\theta}}\left[\mathcal{I}(\hat{\mathbf{z}})^{-1}\mathbf{g}_{\mathcal{L}}(\hat{\mathbf{z}}) + \mathbf{g}_{\mathcal{L}}(\hat{\mathbf{z}})\right] \nonumber \\
    &\approx \frac{\partial \hat{\mathbf{z}}}{\partial \boldsymbol{\theta}}\mathcal{I}(\hat{\mathbf{z}})^{-1}\mathbf{g}_{\mathcal{L}}(\hat{\mathbf{z}}) + \mathbf{g}_{\mathcal{L}}(\boldsymbol{\theta}).
\end{align}

\section{Supplementary figures}
\begin{figure}[H]
    \vskip 0.2in
    \begin{center}
        \centerline{\includegraphics[width=0.9\columnwidth]{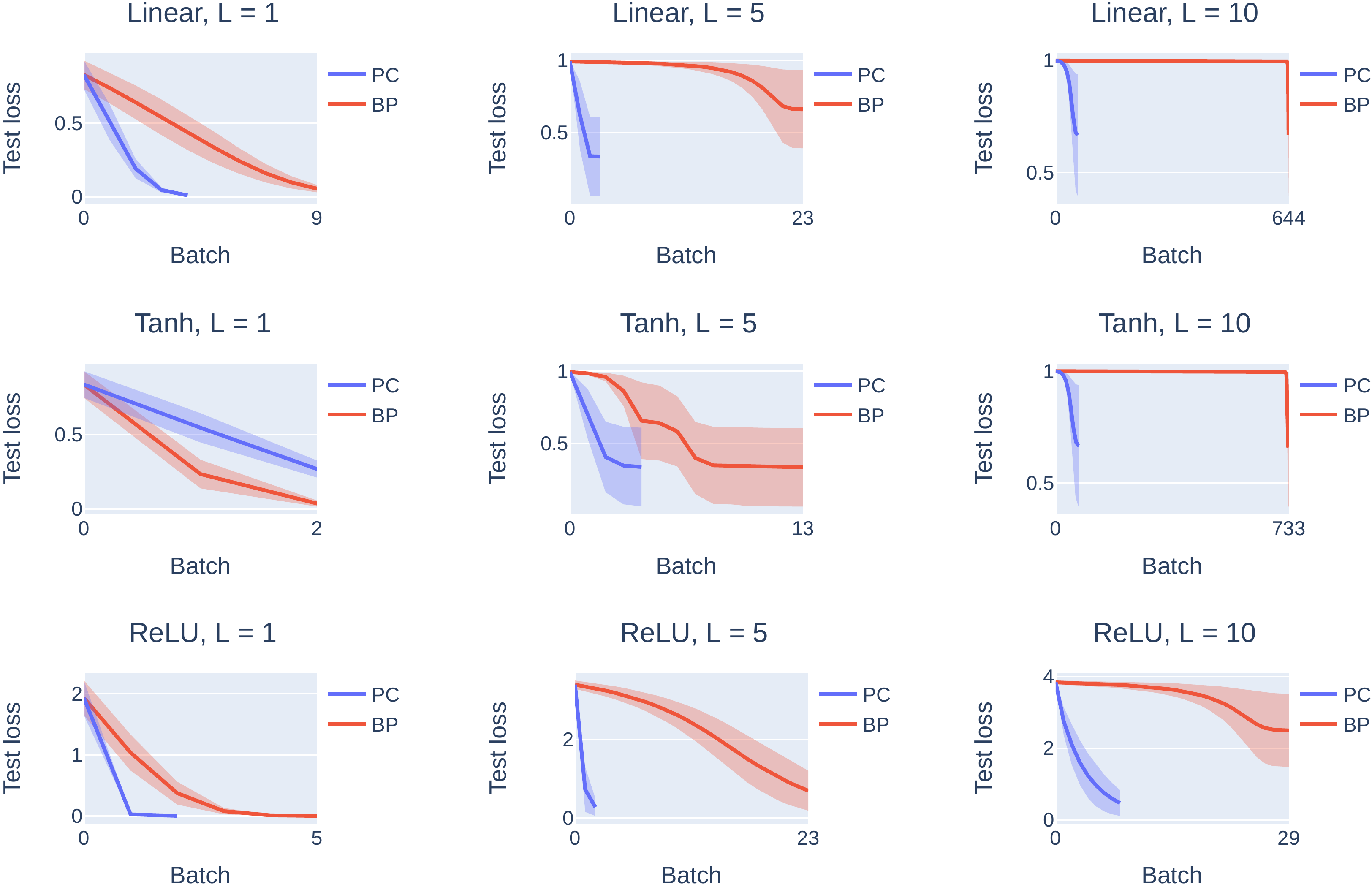}}
        \caption{\textbf{Mean test losses for the deep chain experiments in \S\ref{ch3:experiments}.}}
        \label{ch3:fig:chain-test-losses}
    \end{center}
    \vskip -0.2in
\end{figure}
\begin{figure}[H]
    \vskip 0.2in
    \begin{center}
        \centerline{\includegraphics[width=\columnwidth]{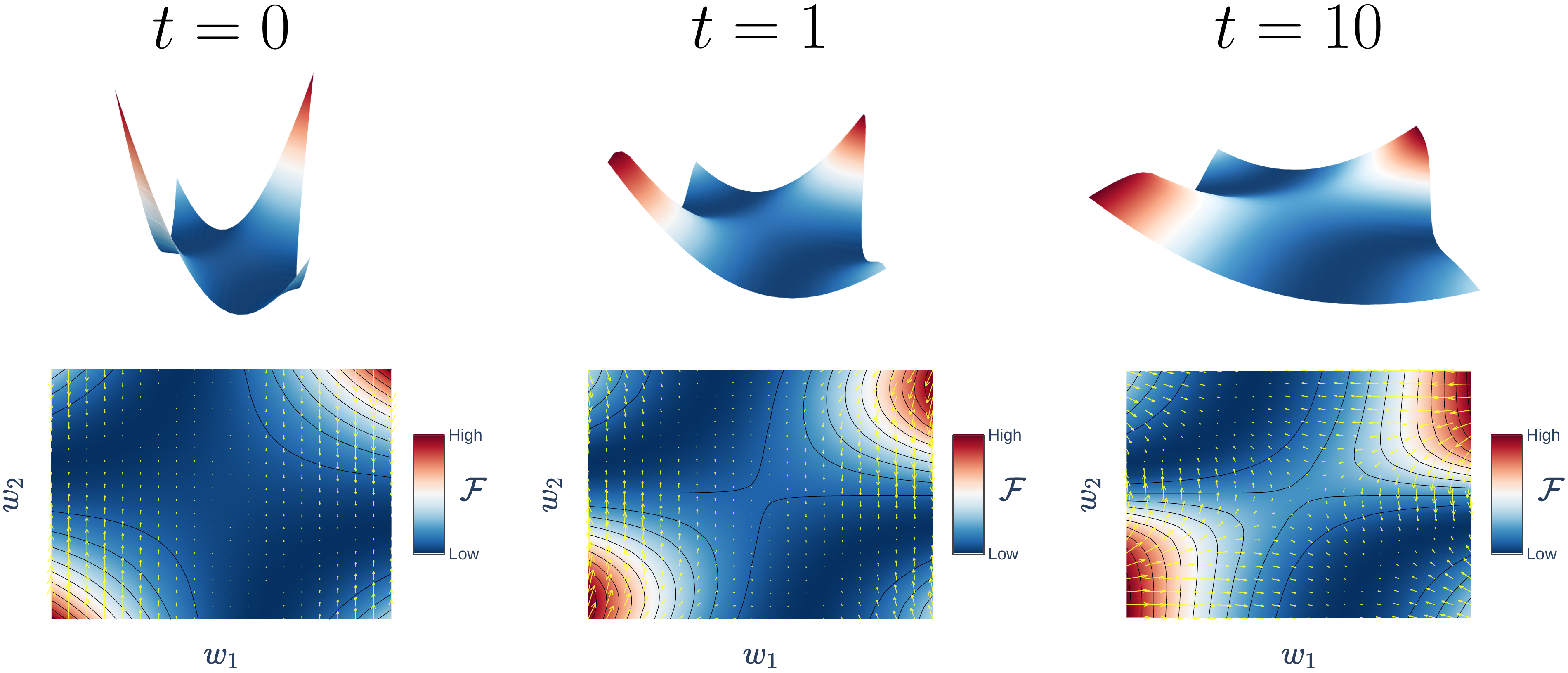}}
        \caption{\textbf{Inference dynamics of PC energy landscape of a toy network.} Evolution of the free energy landscape as a function of the 1MLP weights over inference, plotted at initialisation ($t=0$), the first inference step ($t=1$), and equilibrium ($t=10$) for the same problem illustrated in \S\ref{ch3:fig:toy-net}.}
        \label{ch3:fig:energy-land-infer-dynamics}
    \end{center}
    \vskip -0.2in
\end{figure}
\begin{figure}[H]
    \vskip 0.2in
    \begin{center}
        \centerline{\includegraphics[width=\columnwidth]{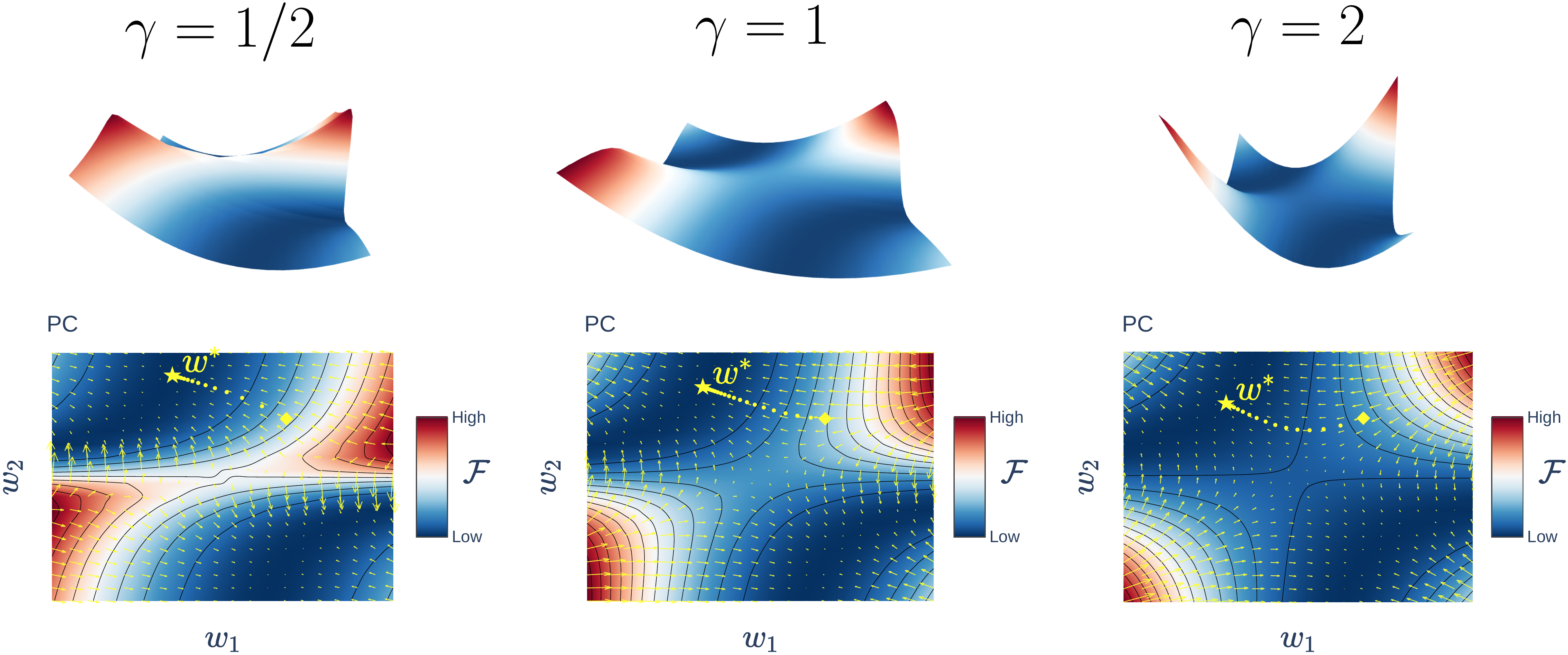}}
        \caption{\textbf{Equilibrated PC energy landscape as a function of the ratio of bottom-up vs top-down information in a toy network.} If we assume a generative model with non-identity scalar covariances (see Chapter~\ref{ch:pcns}), we can rewrite the PC energy for our toy model  as $\mathcal{F} = p_1(z - w_1x)^2/2 + p_2(y - w_2z)^2/2$, where the scalars $p_i$ weigh each energy term. Let $\gamma = p_1 / p_2$ be the ratio of these precisions, thus quantifying the degree of bottom-up vs top-down information. Varying $\gamma$ can be seen as adjusting the size of the trust region or per-parameter learning rates. Increasing the relative influence of the input ($\gamma = 2$) recovers BP, while increasing that of the output ($\gamma = 1/2$) recovers target propagation.}
        \label{ch3:fig:energy-land-precision-ratio}
    \end{center}
    \vskip -0.2in
\end{figure}

%% file: text/appendices/ch4.tex
\chapter{Appendix for Chapter 4}
\label{ch4:appendix}
\minitoc

\renewcommand{\thefigure}{B.\arabic{figure}}
\setcounter{figure}{0}

\section{General notation and definitions} 
\label{ch4:notation}
Matrices, vectors and scalars are denoted with bold capitals $\matr{A}$, bold lower-case characters $\mathbf{v}$ and non-bold characters $u$ or $U$, respectively. All vectors are by default column vectors $[\cdot] \in \mathbb{R}^{n \times 1}$, and $\vect(\cdot)$ denotes the row-wise vec operator. Following \citep{singh2021analytic}, unless otherwise stated we define matrix-by-matrix derivatives by row-vectorisation, using the numerator or Jacobian layout
\begin{equation}
    \left(\frac{\partial \matr{A}}{\partial \matr{B}}\right)_{ij} \coloneq \frac{[\partial \vect_r(\matr{A})]_i}{[\partial \vect_r(\matr{B})^T]_j},
\end{equation}
such that the result is a matrix rather than a 4D tensor. Following from this, we will also use the rules
\begin{align}
    \frac{\partial \matr{A}\matr{B}\matr{C}}{\partial \matr{B}} &= \mathbf{A} \otimes \mathbf{C}^T \\
    \frac{\partial \matr{A}\matr{B}}{\partial \matr{A}} &= \matr{I}_m \otimes \matr{B}^T, \quad \matr{A} \in \mathbb{R}^{m \times n}, \matr{B} \in \mathbb{R}^{n \times p}.
\end{align}

\section{Related work} 
\label{ch4:related-work}
\subsection{Theories of predictive coding} 
\label{ch4:pc-theory}
\textbf{PC and BP.} As reviewed in Chapter~\ref{ch:trust-region}, \citep{whittington2017approximation} where the first to show that PC can approximate BP on multi-layer perceptrons when the influence of the input is upweighted relative to that of the output. \citep{millidge2022predictive} generalised this result to arbitrary computational graphs including convolutional and recurrent neural networks under the so-called ``fixed prediction assumption''. A variation of PC where weights are updated at precisely timed inference steps was later shown to compute exactly the same gradients as BP on multi-layer perceptrons \citep{song2020can}, a result further generalised by \citep{salvatori2021predictive} and \citep{rosenbaum2022relationship}. \citep{millidge2022backpropagation} unified these and other approximation results from an energy-based modelling perspective. \citep{zahid2023predictive} proved that the time complexity of all of these PC variants is lower-bounded by BP.

\textbf{PC and other algorithms.} \citep{frieder2022non} provided an in-depth dynamical systems analysis of the PC inference dynamics for variants approximating BP. As reviewed in Chapter~\ref{ch:trust-region}, \citep{millidge2022theoretical} showed that for linear networks the PC inference equilibrium can be interpreted as an average of BP's feedforward pass values and the local targets computed by target propagation \citep{meulemans2020theoretical}. \citep{song2022inferring} proposed that PC and other energy-based algorithms implement a fundamentally different principle of credit assignment called ``prospective configuration''. For mini-batches of size one, \citep{alonso2022theoretical} proved that PC approximates implicit gradient descent under specific layer-wise rescalings of the activities and parameter learning rates. \citep{alonso2023understanding} further showed that when this approximation holds, PC can be sensitive to Hessian information. Similarly, Chapter~\ref{ch:trust-region} casts PC as a second-order trust-region method \citep{innocenti2023understanding}.

\subsection{Saddle points and neural networks} 
\label{ch4:saddles}
Here we review some relevant theoretical and empirical work on (i) saddle points in the loss landscape of neural networks and (ii) the behaviour of different learning algorithms, especially (S)GD, near saddles. For more general reviews on the loss landscape and optimisation of neural networks, see \citep{sun2019optimization} and \citep{sun2020global}.

\textbf{Saddles in the neural loss landscape}. This work began with \citep{baldi1989neural} showing that for linear networks with one hidden layer, all critical points of the MSE loss are either global minima or strict saddle points (Def. \ref{ch4:def:strict-saddle}). For the same model, \citep{saxe2013exact} later showed saddle-to-saddle learning transitions for small initialisation and characterised the GD dynamics under specific assumptions on the data. \citep{dauphin2014identifying} highlighted the prevalence of saddles, relative to local minima, in the high-dimensional non-convex loss of neural networks. In particular, they empirically demonstrated a qualitative similarity between the landscape of networks and random Gaussian error functions, where the higher the error a critical point is associated with, the more exponentially likely it is to be a saddle \citep{bray2007statistics}. 

\citep{kawaguchi2016deep} famously generalised the \citep{baldi1989neural} result that all local minima are global to arbitrarily DLNs under some weak assumptions on the data. This was simplified as well as extended under less strict assumptions by \citep{lu2017depth}. Importantly, \citep{kawaguchi2016deep} was the first to show that for neural networks with one hidden layer $H = 1$ all saddle points are strict (or first-order), while deeper networks have non-strict ($H$-order) saddles (for example at the origin). Several variations and extensions of this set of results have since been formulated \citep{yun2017global, zhou2018critical, laurent2018deep, zhu2020global, nouiehed2022learning, ziyin2022exact}. For our purposes, one important extension was made by \citep{achour2021loss}, who characterised all the critical points of the MSE loss for DLNs to second-order, including strict and non-strict saddles.

\textbf{Learning near saddles}. This work can be traced back to \citep{ge2015escaping} who showed that SGD with added noise can converge in polynomial time on strict saddle functions. \citep{lee2016gradient} 
proved a similar result that GD without any noise asymptotically escapes strict saddles for almost all initialisations. This was later generalised to other first-order methods \citep{lee2019first}. \citep{jin2017escape} proved that another noisy version of GD converges with high probability to a second-order critical point in poly-logarithmic time depending on the dimension. For a review of these and other convergence results of GD and its variants, see \citep{jin2021nonconvex}. \citep{anandkumar2016efficient} showed (i) that a further GD variant can be proved to converge to a third-order critical point and escape second-order saddles but at a high computational cost and (ii) that finding higher-order critical points is NP-hard.

\citep{du2017gradient} proved the important result that while standard GD with common initialisations will eventually escape strict saddles, it can take an exponential time to do so. This is in contrast to the perturbed GD versions mentioned above, which converge in polynomial time. Similarly, \citep{shamir2019exponential} proved that for linear chains or one-dimensional networks with unit width, the convergence time of GD scales exponentially with the depth. \citep{orvieto2022vanishing} analysed similar models and showed that both the gradients and the curvature vanish with network depth unless the width is appropriately scaled. \citep{orvieto2022vanishing}  suggested that this in part explains the success of adaptive gradient optimisers like Adam \citep{kingma2014adam} which can adapt to flat curvature. Similarly, \citep{staib2019escaping} showed that adaptive methods can escape saddle points faster by rescaling the gradient noise near critical points to be isotropic.

\citep{jacot2021saddle} conjectured a saddle-to-saddle dynamic where GD visits a sequence of saddles of increasing rank before converging to a sparse global minimum. A few works have also shown that in practice SGD can converge to second-order critical points that are non-strict saddles rather than minima \citep{sankar2018saddles, bottcher2024visualizing}.

\section{Proofs and derivations} 
\label{ch4:derivations}
\subsection{Loss Hessian for DLNs} 
\label{ch4:loss-hessian}
Here we derive the Hessian of the MSE loss (Eq.~\ref{ch4:eq:mse-loss}) with respect to the weights of arbitrary DLNs (\S\ref{ch4:dln}); this is essentially a re-derivation of \citep{singh2021analytic} with slightly different notation.\footnote{In particular, unlike \citep{singh2021analytic} we make transposes of weight matrix products explicit.} We then show how the Hessian and its eigenspectrum at the origin ($\boldsymbol{\theta} = \mathbf{0}$) changes as a function of the number of hidden layers $H$. We start from the gradient of the loss for a given weight matrix
\begin{align}
    \frac{\partial \mathcal{L}}{\partial \matr{W}_\ell} &= (\matr{W}_{L:\ell+1})^T (\matr{W}_{L:1}\mathbf{x} - \mathbf{y}) (\matr{W}_{\ell-1:1}\mathbf{x})^T \\
    &= (\matr{W}_{L:\ell+1})^T (\matr{W}_{L:1}\widetilde{\matr{\Sigma}}_{\mathbf{xx}} - \widetilde{\matr{\Sigma}}_{\mathbf{yx}}) (\matr{W}_{\ell-1:1})^T \in \mathbb{R}^{N_\ell \times N_{\ell-1}},
\end{align}
where following previous works we take the empirical mean over the data matrices $\widetilde{\matr{\Sigma}}_{\mathbf{xx}} \coloneq \frac{1}{B} \sum_i^B \mathbf{x}_i\mathbf{x}_i^T$ and $\widetilde{\matr{\Sigma}}_{\mathbf{yx}} \coloneq \frac{1}{B} \sum_i^B \mathbf{y}_i\mathbf{x}_i^T$. For networks with at least one hidden layer, the origin is a critical point since the gradient is zero $\mathbf{g}_{\mathcal{L}}(\boldsymbol{\theta} = \mathbf{0}) = \mathbf{0}$. To characterise this point to second order, we look at the Hessian. Starting with the diagonal blocks of size $(N_\ell N_{\ell-1})\times(N_\ell N_{\ell-1})$,
\begin{equation}
    \frac{\partial^2 \mathcal{L}}{\partial \matr{W}_\ell^2} = (\matr{W}_{L:\ell+1})^T \matr{W}_{L:\ell+1} \otimes \matr{W}_{\ell-1:1} \widetilde{\matr{\Sigma}}_{\mathbf{xx}} (\matr{W}_{\ell-1:1})^T,
\end{equation}
it is straightforward to see that at the origin this term collapses to the null matrix for any $l$.\footnote{To be precise, this is true for any network with at least one hidder layer $H \geq 1$. For zero-hidden-layer networks $H = 0$---which are equivalent to a linear regression problem---the origin is a not a critical point, $\mathbf{g}_{\mathcal{L}}(\boldsymbol{\theta} = \mathbf{0}) = - \widetilde{\matr{\Sigma}}_{\mathbf{yx}}$, and the Hessian is constant $\matr{H}_{\mathcal{L}} = \matr{I}_{N_l} \otimes \widetilde{\matr{\Sigma}}_{\mathbf{xx}}$.} To compute the $(N_k N_{k-1})\times(N_\ell N_{\ell-1})$ off-diagonal blocks, we follow \citep{singh2021analytic} and write the distinct contributions as follows 
\begin{align}
    \forall k \neq \ell, \quad \widetilde{\matr{H}}_{\mathcal{L}} \coloneq \frac{\partial^2 \mathcal{L}}{\partial \matr{W}_k \partial \matr{W}_\ell} &= (\matr{W}_{L:\ell+1})^T \matr{W}_{L:k+1} \otimes \matr{W}_{\ell-1:1} \widetilde{\matr{\Sigma}}_{\mathbf{xx}} (\matr{W}_{k-1:1})^T \\ 
    \forall k > \ell, \quad \widehat{\matr{H}}_{\mathcal{L}} \coloneq \frac{\partial^2 \mathcal{L}}{\partial \matr{W}_k^T \partial \matr{W}_\ell} &= (\matr{W}_{k-1:\ell+1})^T \otimes \matr{W}_{\ell-1:1} (\matr{W}_{L:1}\widetilde{\matr{\Sigma}}_{\mathbf{xx}} - \widetilde{\matr{\Sigma}}_{\mathbf{yx}})^T \matr{W}_{L:k+1} \\ 
    \forall k < \ell, \quad \widehat{\matr{H}}_{\mathcal{L}} \coloneq \frac{\partial^2 \mathcal{L}}{\partial \matr{W}_k^T \partial \matr{W}_\ell} &= (\matr{W}_{L:\ell+1})^T (\matr{W}_{L:1}\widetilde{\matr{\Sigma}}_{\mathbf{xx}} - \widetilde{\matr{\Sigma}}_{\mathbf{yx}}) (\matr{W}_{k-1:1})^T \otimes (\matr{W}_{\ell-1:k+1})^T.
\end{align}
At the origin, these blocks always vanish except for networks with one hidden layer, where as shown by \citep{saxe2013exact} they are characterised by the empirical input-output covariance, e.g. for $k<\ell, \partial^2 \mathcal{L}/\partial \matr{W}_k \partial \matr{W}_\ell(\boldsymbol{\theta} = \mathbf{0}) = - \widetilde{\matr{\Sigma}}_{\mathbf{xy}} \otimes \matr{I}_N, H = 1$. Putting the above facts together, we can now write the full loss Hessian at the origin for different number of hidden layers:
\begin{align}
    \matr{H}_{\mathcal{L}} (\boldsymbol{\theta} = \mathbf{0}) = 
    \begin{cases} 
    \begin{bmatrix} \mathbf{0} & - \widetilde{\matr{\Sigma}}_{\mathbf{xy}} \otimes \matr{I}_{N_1} \\ -\matr{I}_{N_1} \otimes \widetilde{\matr{\Sigma}}_{\mathbf{yx}} & \mathbf{0} \end{bmatrix}, & H = 1 \quad \text{[strict saddle]} \\ \\
    \begin{bmatrix} \mathbf{0} & \dots & \mathbf{0} \\ \vdots & \ddots & \vdots \\ \mathbf{0} & \dots & \mathbf{0} \end{bmatrix} = \mathbf{0}_p, & H > 1 \quad \text{[non-strict saddle]}
    \end{cases}.
    \label{ch4:eq:loss-hess}
\end{align}
For one-hidden-layer networks, the Hessian is indefinite, with positive and negative eigenvalues given by the empirical input-output covariance, as described by \citep{saxe2013exact}. For any DLN with more than one hidden layer, the Hessian is null, and the origin is therefore a second-order critical point. In the general case, this point is a non-strict saddle because some higher-order derivative of the loss depending on the network depth will contain at least one negative escape direction. More specifically, for a network with $L$ layers, all the $L-1$ derivatives vanish, and negative directions will be found in the derivatives of order $\geq L$.

\subsection{Equilibrated energy for DLNs} 
\label{ch4:equilib-energy}
Here we derive an exact solution to the PC energy (Eq.~\ref{ch4:eq:pc-energy}) of DLNs at the inference equilibrium $\mathcal{F}(\boldsymbol{\theta}, \mathbf{z}^*)$ (Theorem~\ref{ch4:thm1}, Eq.~\ref{ch4:eq:dln-equilib-energy}), which we will abbreviate as $\mathcal{F}^*$. This turns out to be a non-trivial rescaled MSE loss where the rescaling depends on covariances of the network weight matrices. To highlight the difference with the loss, recall that the standard MSE (Eq.~\ref{ch4:eq:mse-loss}) for a DLN implicitly defines the following generative model
\begin{equation}
    \mathbf{y} \sim \mathcal{N}(\matr{W}_{L:1}\mathbf{x}, \matr{\Sigma})
    \label{ch4:eq:loss-gauss-model}
\end{equation}
where the target is modelled as a Gaussian with a mean given by the network map (i.e. forward pass) and some covariance $\matr{\Sigma}$. In a PC network, by contrast, the activity of \textit{each hidden layer}--and not just the output--is modelled as a Gaussian (see \S \ref{ch4:pc})
\begin{equation}
    \mathbf{z}_\ell \sim \mathcal{N}(\matr{W}_\ell \mathbf{z}_{\ell-1}, \matr{I}_\ell),
\end{equation}
where $\mathbf{z}_0 \coloneq \mathbf{x}$ and $\mathbf{z}_L \coloneq \mathbf{y}$. Now, to work out the generative model for the target implied by this hierarchical Gaussian model, we can simply ``unfold'' the model at each layer. Specifically, we can reparameterise the activity of each hidden layer as a noisy function of the previous layer and so on recursively up to the first layer
\begin{align}
    \mathbf{z}_1 &= \matr{W}_1\mathbf{z}_0 + \boldsymbol{\xi}_1 \\
    \mathbf{z}_2 &= \matr{W}_2\mathbf{z}_1 + \boldsymbol{\xi}_2 = \matr{W}_2\matr{W}_1\mathbf{x} + \matr{W}_2\boldsymbol{\xi}_1 + \boldsymbol{\xi}_2 \\
    \mathbf{z}_3 &= \matr{W}_3\mathbf{z}_2 + \boldsymbol{\xi}_3 = \matr{W}_3\matr{W}_2\matr{W}_1\mathbf{x} + \matr{W}_3\matr{W}_2\boldsymbol{\xi}_1 + \matr{W}_3\boldsymbol{\xi}_2 + \boldsymbol{\xi}_3 \\
    \vdots \nonumber
\end{align}
where $\boldsymbol{\xi}_\ell \sim \mathcal{N}(\mathbf{0}, \matr{I}_\ell)$ is white Gaussian noise. The last layer can then be written as
\begin{align}
    \mathbf{z}_L &= \matr{W}_L \mathbf{z}_{L-1} + \boldsymbol{\xi}_L \\
    &= \matr{W}_{L:1} \mathbf{z}_0 + \sum^L_{\ell=2} \matr{W}_{L:\ell} \boldsymbol{\xi}_{\ell-1} + \boldsymbol{\xi}_L.
    \label{ch4:eq:last-layer-reparam}
\end{align}
We can now derive the implicit generative model for the target by taking the expectation and covariance of Eq.~\ref{ch4:eq:last-layer-reparam} with respect to the random noise:
\begin{equation}
    \mathbf{y} \sim \mathcal{N} \left( \matr{W}_{L:1}\mathbf{x}, \matr{I}_L + \sum_{\ell=2}^L (\matr{W}_{L:\ell})(\matr{W}_{L:\ell})^T \right).
\end{equation}
We therefore observe that, in contrast to the loss (Eq.~\ref{ch4:eq:loss-gauss-model}), PC implicitly models the target with a non-identity covariance depending on a chained covariance of the previous layers which in turns depends only on the weights. It follows that, at the exact inference equilibrium where that implicit generative model holds, the PC energy is simply the following rescaled MSE loss
\begin{align}
    \mathcal{F}^* &= \frac{1}{2B} \sum_{i}^B (\mathbf{y}_i - \matr{W}_{L:1}\mathbf{x}_i)^T \matr{S}(\boldsymbol{\theta})^{-1}(\mathbf{y}_i - \matr{W}_{L:1}\mathbf{x}_i),
    \label{ch4:eq:appendix-dln-equilib-energy}
\end{align}
where the rescaling is $\matr{S}(\boldsymbol{\theta}) = \matr{I}_{N_L} + \sum_{\ell=2}^L (\matr{W}_{L:\ell})(\matr{W}_{L:\ell})^T$. One can also arrive at this expression by explicitly solving for the activities $\partial \mathcal{F} / \partial \mathbf{z} = 0$ and plugging the solution back into the energy, although the calculation becomes much more involved. Note that a generative model with non-identity covariances at each layer would lead to a different rescaling, but we do not consider this case here to remain as close as possible to what is done in practice.

\subsection{Hessian of the equilibrated energy for DLNs} 
\label{ch4:energy-hessian}
Here we derive the Hessian at the origin of the equilibrated energy for DLNs, following the calculation of the loss Hessian (\S\ref{ch4:loss-hessian}). Section~\ref{ch4:chains-hessian} shows an equivalent derivation for one-dimensional linear networks, which preserves all the key the intuitions and is easier to follow. We start from the equilibrated energy we derived previously for DLNs (\S\ref{ch4:equilib-energy}, Eq.~\ref{ch4:eq:appendix-dln-equilib-energy}), which turned out to be the following rescaled MSE loss
\begin{equation}
    \mathcal{F}^* = \frac{1}{2B} \sum_{i}^B \mathbf{r}_i^T \matr{S}(\boldsymbol{\theta})^{-1} \mathbf{r}_i
\end{equation}
where $\matr{S}(\boldsymbol{\theta}) = \matr{I}_{N_L} + \sum_{\ell=2}^L (\matr{W}_{L:\ell})(\matr{W}_{L:\ell})^T$, and we denote the residual error for a given data sample as $\mathbf{r}_i \coloneq \mathbf{y}_i - \matr{W}_{L:1}\mathbf{x}_i$. In the general case, both the residual and the rescaling depend on $\matr{W}_\ell$, so to take the gradient of the equilibrated energy we need the product rule. For simplicity, and similar to the characterisation of the off-diagonal blocks of the loss Hessian (\S\ref{ch4:loss-hessian}), we write the two contributions separately, as follows
\begin{align}
    \matr{A} &\coloneq \frac{1}{2N} \sum_{i}^N \frac{\partial \mathbf{r}_i^T}{\partial \matr{W}_\ell} \matr{S}^{-1} \frac{\partial \mathbf{r}_i}{\partial \matr{W}_\ell} = (\matr{W}_{L:\ell+1})^T \matr{S}^{-1} (\matr{W}_{L:1}\widetilde{\matr{\Sigma}}_{\mathbf{xx}} - \widetilde{\matr{\Sigma}}_{\mathbf{yx}}) (\matr{W}_{\ell-1:1})^T \label{ch4:eq:first-grad-equilib-energy-term} \\
    \matr{B} &\coloneq \frac{1}{2N} \sum_{i}^N \mathbf{r}_i^T\frac{\partial \matr{S}^{-1}}{\partial \matr{W}_\ell} \mathbf{r}_i = - \frac{1}{N}\sum_i^N \matr{S}^{-1}\mathbf{r}_i \mathbf{r}_i^T \matr{S}^{-1} \frac{\partial \matr{S}}{\partial \matr{W}_\ell}, \label{ch4:eq:second-grad-equilib-energy-term}
\end{align}
where in Eq.~\ref{ch4:eq:second-grad-equilib-energy-term} $\partial \matr{S}/\partial \matr{W}_\ell$ is a 4D tensor, and we use the rule $\partial \mathbf{a}^T\matr{X}\mathbf{b} / \partial \matr{X} = - \matr{X}^{-T}\mathbf{a}\mathbf{b}^T\matr{X}^{-T}$. The first term $\matr{A}$ is simply a rescaled loss gradient, while the second term $\matr{B}$ depends on the derivative of the rescaling. Note that for $\matr{W}_1$ the gradient collapses to the first term since the rescaling does not depend on it, $\partial \mathcal{F}^*/\partial \matr{W}_1 = (\matr{W}_{L:2})^T \matr{S}^{-1} (\matr{W}_{L:1}\widetilde{\matr{\Sigma}}_{\mathbf{xx}} - \widetilde{\matr{\Sigma}}_{\mathbf{yx}})$. 

As an aside relevant to the zero-rank saddles analysed in \S \ref{ch4:other-saddles}, we note that, in contrast to the loss, $\matr{W}_L = \mathbf{0}$ is a necessary (though not sufficient) condition for the energy gradient to be zero. This is because the derivative of the rescaling $\partial \matr{S}/\partial \matr{W}_\ell$ needs to be zero in order for the gradient term $\matr{B}$ to vanish, and it has one term linear in the last weight matrix.

As for the loss (\S\ref{ch4:loss-hessian}), the origin is a critical point of the energy since $\mathbf{g}_{\mathcal{F}^*}(\boldsymbol{\theta} = \mathbf{0}) = \mathbf{0}$. For $\matr{B}$, this is because while the rescaling at zero is the identity, the derivative of the rescaling vanishes since it is linear with respect to any weight matrix:
\begin{align}
    \matr{S}^{-1}(\boldsymbol{\theta} = \mathbf{0}) &= \matr{I}_{N_L} \label{ch4:eq:inverse-rescaling-origin} \\
    \frac{\partial \matr{S}}{\partial \matr{W}_\ell}(\boldsymbol{\theta} = \mathbf{0}) &= \mathbf{0}. \label{ch4:eq:deriv-rescaling-origin}
\end{align}
Calculating the Hessian involves multiple application of the product rule, so for simplicity we analyse the contribution of the derivative of each term (Eqs.~\ref{ch4:eq:first-grad-equilib-energy-term} \& \ref{ch4:eq:second-grad-equilib-energy-term}) at the origin. Because the first term is simply a rescaling of the loss, and given Eq.~\ref{ch4:eq:inverse-rescaling-origin}, its second derivative at zero is always zero with respect to the same weight matrix,
\begin{equation}
    k = \ell, \quad \frac{\partial \matr{A}}{\partial \matr{W}_k}(\boldsymbol{\theta} = \mathbf{0}) = \mathbf{0}, \quad H \geq 1.
\end{equation}
As for the loss, this term is also zero with respect to some other weight matrix $k \neq \ell$ except for the case of a one-hidden-layer network
\begin{align}
    k \neq \ell, \quad \frac{\partial \matr{A}}{\partial \matr{W}_k}(\boldsymbol{\theta} = \mathbf{0}) =
    \begin{cases} 
    -\matr{I}_{N_1} \otimes \widetilde{\matr{\Sigma}}_{\mathbf{yx}}, & k > \ell, H = 1 \\ \\
    - \widetilde{\matr{\Sigma}}_{\mathbf{xy}} \otimes \matr{I}_{N_1}, & k < \ell, H = 1 \\ \\
    \mathbf{0}, & H > 1
    \end{cases}.
\end{align}
The second derivative of $\matr{B}$ requires a 5-fold application of the product rule, involving the first derivative of the residual (and its transpose) and the first and second derivatives of the rescaling. As shown above (Eq.~\ref{ch4:eq:deriv-rescaling-origin}), the first derivative of the rescaling at the origin is zero, and the derivative of the residual with respect to any weight matrix at zero is always zero for any network with one or more hidden layers, $\partial \mathbf{r}/\partial \matr{W}_k(\boldsymbol{\theta} = \mathbf{0}) = \mathbf{0}, H \geq 1$. The second derivative of the rescaling, however, is non-zero for the special case of the last weight matrix with respect to itself:
\begin{align}
    \frac{\partial^2 \matr{S}}{\partial \matr{W}_k \partial \matr{W}_\ell}(\boldsymbol{\theta} = \mathbf{0}) =
    \begin{cases} 
    \matr{I}_{N_{L-1}}, & \ell = k = L \\ \\
    \mathbf{0}, & \text{else}
    \end{cases},
\end{align}
which means that at zero $\matr{B}$ takes the following form
\begin{align}
    \frac{\partial \matr{B}}{\partial \matr{W}_k}(\boldsymbol{\theta} = \mathbf{0}) =
    \begin{cases} 
    -\widetilde{\matr{\Sigma}}_{\mathbf{yy}} \otimes I_{N_{L-1}}, & \ell = k = L \\ \\
    \mathbf{0}, & \text{else}
    \end{cases}
\end{align}
where $\widetilde{\matr{\Sigma}}_{\mathbf{yy}} \coloneq \frac{1}{B} \sum_i^B \mathbf{y}_i\mathbf{y}_i^T$ is the empirical output covariance matrix. Drawing all these observations together, we can write the full Hessian at the origin of the equilibrated energy for different number of hidden layers:
\begin{align}
    \matr{H}_{\mathcal{F}^*} (\boldsymbol{\theta} = \mathbf{0}) = 
    \begin{cases} 
    \begin{bmatrix*}[l] \mathbf{0} & - \widetilde{\matr{\Sigma}}_{\mathbf{xy}} \otimes \matr{I}_{N_1} \\ -\matr{I}_{N_1} \otimes \widetilde{\matr{\Sigma}}_{\mathbf{yx}} & -\widetilde{\matr{\Sigma}}_{\mathbf{yy}} \otimes I_{N_{L-1}} \end{bmatrix*}, & H = 1 \quad \text{[strict saddle]} \\ \\
    \begin{bmatrix} \mathbf{0} & \dots & \mathbf{0} \\ \vdots & \ddots & \vdots \\ \mathbf{0} & \dots & -\widetilde{\matr{\Sigma}}_{\mathbf{yy}} \otimes I_{N_{L-1}} \end{bmatrix}, \label{eq41} & H > 1 \quad \text{[strict saddle]}
    \end{cases}.
\end{align}
We see that, compared to the loss Hessian (Eq.~\ref{ch4:eq:loss-hess}), the energy Hessian has a non-zero last diagonal block given for any $H$. We note, but do not investigate in any depth, the potential connection with target propagation \citep{meulemans2020theoretical, millidge2022theoretical}. The one-hidden-layer case is fully derived in the next section (\S \ref{ch4:one-hidden-example}). It is straightforward to show that these matrices have negative eigenvalues
\begin{equation}
    H \geq 1, \quad \lambda_{\text{min}}(\matr{H}_{\mathcal{F}^*}(\boldsymbol{\theta} = \mathbf{0})) < 0, \quad \forall y_i \neq 0
\end{equation}
since $\matr{A}\matr{A}^T$ is positive definite $\forall \matr{A} \neq \mathbf{0}$. The origin is therefore a strict saddle (Def. \ref{ch4:def:strict-saddle}) of the equilibrated energy. This is in stark contrast to the MSE loss, which at the origin has a strict saddle only for one-hidden-layer networks and a non-strict saddle of order $H$ for any deeper network. For the general case $H > 1$, the negative curvature of the energy Hessian is given only by the output-output covariance $\widetilde{\matr{\Sigma}}_{\mathbf{yy}}$. This means that, in the vicinity of the origin saddle, GD steps of equal size on the equilibrated energy will escape the saddle faster (at a rate depending on the output structure) than on the loss, and increasingly so as a function of depth. In \S \ref{ch4:experiments}, we empirically verify this prediction experimentally on linear as well as non-linear architectures (including convolutional) trained on different datasets.

\subsection{Example: 1-hidden layer linear network} 
\label{ch4:one-hidden-example}
Here we show an example calculation comparing the Hessian at the origin of the loss and equilibrated energy for DLNs with a single hidden layer $H = 1$. For this case, the MSE loss and equilibrated energy are
\begin{align}
    \mathcal{L} &= \frac{1}{2B}\sum_i^{B}||\mathbf{y}_i - \matr{W}_2 \matr{W}_1 \mathbf{x}_i||^2 \\
    \mathcal{F}^* &= \frac{1}{2B}\sum_i^{B}(\mathbf{y}_i - \matr{W}_2 \matr{W}_1 \mathbf{x}_i)^T (\matr{I}_{N_L} + \matr{W}_2\matr{W}_2^T)^{-1} (\mathbf{y}_i - \matr{W}_2 \matr{W}_1 \mathbf{x}_i)
\end{align}
where $\mathbf{x} \in \mathbb{R}^{N_0}, \mathbf{y} \in \mathbb{R}^{N_L}, \matr{W}_1 \in \mathbb{R}^{N_1 \times N_0}, \matr{W}_2 \in \mathbb{R}^{N_L \times N_1}$. We now show the weight gradients, first of the loss
\begin{align}
    \frac{\partial \mathcal{L}}{\partial \matr{W}_1} &= \matr{W}_2^T\matr{W}_2\matr{W}_1\widetilde{\matr{\Sigma}}_{\mathbf{xx}} -\matr{W}_2^T\widetilde{\matr{\Sigma}}_{\mathbf{yx}} \\
    \frac{\partial \mathcal{L}}{\partial \matr{W}_2} &= \matr{W}_2\matr{W}_1\widetilde{\matr{\Sigma}}_{\mathbf{xx}}\matr{W}_1^T -\widetilde{\matr{\Sigma}}_{\mathbf{yx}}\matr{W}_1^T,
\end{align}
and then of the equilibrated energy
\begin{align}
    \frac{\partial \mathcal{F}^*}{\partial \matr{W}_1} &= \matr{W}_2^T\matr{S}^{-1}\matr{W}_2\matr{W}_1\widetilde{\matr{\Sigma}}_{\mathbf{xx}} - \matr{W}_2^T \matr{S}^{-1}\widetilde{\matr{\Sigma}}_{\mathbf{yx}} \\
    \frac{\partial \mathcal{F}^*}{\partial \matr{W}_2} &= \matr{S}^{-1}(\matr{W}_2\matr{W}_1\widetilde{\matr{\Sigma}}_{\mathbf{xx}} - \widetilde{\matr{\Sigma}}_{\mathbf{yx}})\matr{W}_1^T - \matr{S}^{-1}\matr{\Psi}\matr{S}^{-1}\matr{W}_2,
\end{align}
where we denote the empirical mean of the residual as $\matr{\Psi} \coloneq \frac{1}{N} \sum_i^N \mathbf{r}_i \mathbf{r}_i^T$. The origin is a critical point of the both the loss and the equilibrated energy since $\mathbf{g}_{\mathcal{L}}(\boldsymbol{\theta} = \mathbf{0}) = \mathbf{g}_{\mathcal{F}^*}(\boldsymbol{\theta} = \mathbf{0}) = \mathbf{0}$. We now compute the Hessian blocks, evaluating the off-diagonals at the origin for simplicity, again first for the loss
\begin{align}
    \frac{\partial^2 \mathcal{L}}{\partial \matr{W}_1^2} &= \matr{W}_2^T\matr{W}_2 \otimes \widetilde{\matr{\Sigma}}_{\mathbf{xx}} \\
    \frac{\partial^2 \mathcal{L}}{\partial \matr{W}_2^2} &= \matr{I}_{N_0} \otimes \matr{W}_1\widetilde{\matr{\Sigma}}_{\mathbf{xx}}\matr{W}_1^T \\
    \frac{\partial^2 \mathcal{L}}{\partial \matr{W}_2 \partial \matr{W}_1}(\boldsymbol{\theta} = \mathbf{0}) &= -\matr{I}_{N_1} \otimes \widetilde{\matr{\Sigma}}_{\mathbf{yx}},
\end{align}
and then for the energy
\begin{align}
    \frac{\partial^2 \mathcal{F}^*}{\partial \matr{W}_1^2} &= \matr{W}_2^T\matr{S}^{-1}\matr{W}_2 \otimes \widetilde{\matr{\Sigma}}_{\mathbf{xx}} \\
    \frac{\partial^2 \mathcal{F}^*}{\partial \matr{W}_2^2} &= \matr{S}^{-1} \otimes \matr{W}_1\widetilde{\matr{\Sigma}}_{\mathbf{xx}}\matr{W}_1^T - \matr{S}^{-1}\matr{\Psi}\matr{S}^{-1} \otimes \matr{I}_{N_1} \\
    \frac{\partial^2 \mathcal{F}^*}{\partial \matr{W}_2 \partial \matr{W}_1}(\boldsymbol{\theta} = \mathbf{0}) &= -\matr{I}_{N_1} \otimes \widetilde{\matr{\Sigma}}_{\mathbf{yx}}.
\end{align}
At the origin, the Hessians become
\begin{align}
    \matr{H}_{\mathcal{L}}(\boldsymbol{\theta} = \mathbf{0}) = 
    \begin{bmatrix} \mathbf{0} & -\widetilde{\matr{\Sigma}}_{\mathbf{xy}} \otimes \matr{I}_{N_1} \\ - \matr{I}_{N_1} \otimes \widetilde{\matr{\Sigma}}_{\mathbf{yx}} & \mathbf{0} \end{bmatrix} \\ \nonumber \\
    \matr{H}_{\mathcal{F}^*}(\boldsymbol{\theta} = \mathbf{0}) = 
    \begin{bmatrix} \mathbf{0} & -\widetilde{\matr{\Sigma}}_{\mathbf{xy}} \otimes \matr{I}_{N_1} \\ - \matr{I}_{N_1} \otimes \widetilde{\matr{\Sigma}}_{\mathbf{yx}} & - \widetilde{\matr{\Sigma}}_{\mathbf{yy}} \otimes \matr{I}_{N_1} \end{bmatrix}.
\end{align}

\subsection{Hessian of the equilibrated energy for linear chains} 
\label{ch4:chains-hessian}
Here we include a derivation the Hessian of the equilibrated energy (as well as its eigenstructure at the origin) for linear chains or networks of unit width $w_{L:1}x$ where $N_0 = \dots = N_L = 1$. This follows the derivation for the wide case (\S\ref{ch4:energy-hessian}), but it is easier to follow and reveals all the key insights. For the scalar case, the implicit generative model of the target defined by PC (see \S\ref{ch4:equilib-energy}) is
\begin{equation}
    y \sim \mathcal{N} \left( w_{L:1}x, 1 + \sum_{\ell=2}^L (w_{L:\ell})^2 \right),
\end{equation}
leading to the following rescaled loss 
\begin{equation}
    \mathcal{F}^* = \mathcal{L}/s, \quad s = 1 + \sum_{\ell=2}^{L} (w_{L:\ell})^2
    \label{ch4:eq:equilib-energy-linear-chain}
\end{equation}
where $\mathcal{L} = \frac{1}{2N} \sum_i^N (y_i -w_{L:1}x_i)^2$. The weight gradient of the equilibrated energy is
\begin{align}
    \frac{\partial \mathcal{F}^*}{\partial w_i} = 
    \begin{cases} 
    \large \frac{1}{s}\frac{\partial \mathcal{L}}{\partial w_i}, & i = 1  \\ \\
    \large \frac{1}{s} \frac{\partial \mathcal{L}}{\partial w_i} - \frac{1}{s^2} \mathcal{L} \frac{\partial s}{\partial w_i}, & i > 1
    \end{cases}
\end{align}
where the loss gradient is $\partial \mathcal{L}/\partial w_i = - w_{L:1 \neq i}xr$ with residual error $r = (y - w_{L:1} x)$. As shown in \S\ref{ch4:equilib-energy}, The origin is a critical point of both the loss and the equilibrated energy since their gradients are zero $\mathbf{g}_{\mathcal{L}}(\boldsymbol{\theta} = \mathbf{0}) = \mathbf{0}, \mathbf{g}_{\mathcal{F}^*} (\boldsymbol{\theta} = \mathbf{0}) = \mathbf{0}$. We now show the Hessians, first of the loss
\begin{align}
    \frac{\partial^2 \mathcal{L}}{\partial w_i \partial w_j} = 
    \begin{cases} 
    (w_{L:1 \neq i})^2 x^2, & i = j  \\ \\
    (w_{L:1 \neq i, j}) (2 w_{L:1} x^2 - xy), & i \neq j
    \end{cases},
    \label{ch4:eq:loss-hess-chain}
\end{align}
and then of the energy
\begin{align}
    \frac{\partial^2 \mathcal{F}^*}{\partial w_i \partial w_j} = 
    \begin{cases} 
    \Large \frac{1}{s}\frac{\partial^2 \mathcal{L}}{\partial w_i \partial w_j}, & i = j = 1  \\ \\
    \Large \frac{1}{s} \frac{\partial^2 \mathcal{L}}{\partial w_i \partial w_j} - \frac{1}{s^2} \frac{\partial \mathcal{L}}{\partial w_i} \frac{\partial s}{\partial w_j}, & i = 1, \quad j > 1 \\ \\ 
    \Large \frac{1}{s} \frac{\partial^2 \mathcal{L}}{\partial w_i \partial w_j} - \frac{1}{s^2} \frac{\partial \mathcal{L}}{\partial w_i} \frac{\partial s}{\partial w_j} + \frac{1}{s^2} \frac{\partial \mathcal{L}}{\partial w_j} \frac{\partial s}{\partial w_i} + \frac{1}{s^2} \mathcal{L} \frac{\partial^2 s}{\partial w_i \partial w_j} - \frac{2}{s^3} \frac{\partial s}{\partial w_j} \mathcal{L} \frac{\partial s}{\partial w_i}, & i, j > 1 \\
    \end{cases}.
    \label{ch4:eq:equilib-energy-hess-chain}
\end{align}
Generalising the one-hidden-unit case shown by \citep{innocenti2023understanding}, at the origin the Hessians reduce to
\begin{alignat}{2}
    \matr{H}_{\mathcal{L}}(\boldsymbol{\theta} = \mathbf{0}) &= 
    \begin{cases} 
    \begin{bmatrix} 0 & -xy \\ -xy & 0 \end{bmatrix}, & H = 1 \quad \text{[strict saddle]} \\ \\
    \begin{bmatrix} 0 & \dots & 0 \\ \vdots & \ddots & \vdots \\ 0 & \dots & 0 \end{bmatrix} = \mathbf{0}_p, & H > 1 \quad \text{[non-strict saddle]}
    \end{cases} \\ \nonumber \\
    \matr{H}_{\mathcal{F}^*}(\boldsymbol{\theta} = \mathbf{0}) &=
    \begin{cases} 
    \begin{bmatrix} 0 & -xy \\ -xy & -y^2 \end{bmatrix}, & H = 1 \quad \text{[better-conditioned strict saddle]} \\ \\
    \begin{bmatrix} 0 & \dots & 0 \\ \vdots & \ddots & \vdots \\ 0 & \dots & -y^2 \end{bmatrix}, & H > 1 \quad \text{[strict saddle]}
    \end{cases}.
    \label{ch4:eq:equilib-energy-hess-chain-origin}
\end{alignat}
For one-hidden-layer networks $H = 1$, the Hessian eigenvalues of the loss and energy are $\lambda(\matr{H}_{\mathcal{L}}(\boldsymbol{\theta} = \mathbf{0})) = \pm xy, \lambda(\matr{H}_{\mathcal{F}^*}(\boldsymbol{\theta} = \mathbf{0})) = (-y^2 \pm y \sqrt{4x^2 + y^2})/2$, respectively. In this case, the eigenvalues of the energy turn out to be smaller than those of the loss,
\begin{equation}
    H = 1, \quad \lambda(\matr{H}_{\mathcal{F}^*}(\boldsymbol{\theta} = \mathbf{0})) < \lambda(\matr{H}_{\mathcal{L}}(\boldsymbol{\theta} = \mathbf{0})), \quad \forall x, y \neq 0 \\
\end{equation}
following from the fact that the square root of a sum is smaller than the sum of the square roots, $\sqrt{a^2 + b^2} < \sqrt{a^2} + \sqrt{b^2}, \forall a, b \neq 0$. This means that, at the origin, the strict saddle of the equilibrated energy is better conditioned (i.e. easier to escape) than that of the loss. For deeper networks, the Hessian of the loss is zero, and it is easy to see that that of the energy has zero eigenvalues of multiplicity $L-1$ and a single negative eigenvalue given by the target squared
\begin{equation}
    H > 1, \quad \lambda_{\text{min}}(\matr{H}_{\mathcal{F}^*}(\boldsymbol{\theta} = \mathbf{0})) = -y^2.
\end{equation}

\subsection{Strictness of zero-rank saddles of the equilibrated energy} 
\label{ch4:strict-zero-rank}
Here we prove the strictness of the zero-rank saddles of the equilibrated energy (Theorem~\ref{ch4:thm3}). 
It is easy to check via Eqs. \ref{ch4:eq:first-grad-equilib-energy-term} \& \ref{ch4:eq:second-grad-equilib-energy-term} that any point $\boldsymbol{\theta}^*$ such that $(\matr{W}_L = \mathbf{0}, \matr{W}_{L-1:1} = \mathbf{0})$ is a critical point. Now let us prove that the Hessian at $\boldsymbol{\theta}^*$ has a negative eigenvalue. To do this, we rely on the Taylor expansion of the function around $\boldsymbol{\theta}^*$. Since $\mathbf{g}_{\mathcal{F}^*}(\boldsymbol{\theta}^*) = \mathbf{0}$, we have for any $\boldsymbol{\hat{\theta}}$ and any $\delta \to 0$,
\begin{align}
    \mathcal{F}^*(\boldsymbol{\theta}^* + \delta \boldsymbol{\hat{\theta}}) = \mathcal{F}^*(\boldsymbol{\theta}^*) + \frac{1}{2} \delta^2 \boldsymbol{\hat{\theta}}^T \matr{H}_{\mathcal{F}^*}(\boldsymbol{\theta}^*)\boldsymbol{\hat{\theta}} + \mathcal{O}(\delta^3).
\end{align}
Hence by unicity of the Taylor expansion, if we can find $\boldsymbol{\hat{\theta}}$ such that $\mathcal{F}^*(\boldsymbol{\theta}^* + \delta \boldsymbol{\hat{\theta}}) = \mathcal{F}^*(\boldsymbol{\theta}^*) -c \delta^2 + \mathcal{O}(\delta^3)$ where $c>0$, this would mean that $\boldsymbol{\hat{\theta}}^T \matr{H}_{\mathcal{F}^*}(\boldsymbol{\theta}^*)\boldsymbol{\hat{\theta}}=-2c <0$ and therefore that it is a strict saddle point. By considering the direction of perturbation $\boldsymbol{\hat{\theta}} = ( \matr{I}, \mathbf{0}, \dots, \mathbf{0})$, %\textcolor{blue}{to be continued}
% Consider a quadratic approximation at any critical point $\boldsymbol{\theta}^*$ of a twice differentiable function $f$ where $\mathbf{g}_{f}(\boldsymbol{\theta}^*) = \mathbf{0}$.
% \begin{equation}
%     f(\boldsymbol{\theta}^* + \delta \boldsymbol{\hat{\theta}}) = f(\boldsymbol{\theta}^*) + \frac{1}{2} \delta^2 \boldsymbol{\hat{\theta}}^T \matr{H}_f \boldsymbol{\hat{\theta}} + \mathcal{O}(\delta \boldsymbol{\hat{\theta}}^3)
%     \label{eq66}
% \end{equation}
% The change in $f$ is therefore approximately given by the quadratic form.
% \begin{equation}
%     \Delta f \approx \frac{1}{2} \delta^2 \boldsymbol{\hat{\theta}}^T \matr{H}_f \boldsymbol{\hat{\theta}}
%     \label{eq67}
% \end{equation}
% If we can find a direction $\boldsymbol{\hat{\theta}}$ that leads to a decrease in $f$, then that implies that the Hessian at that point has at least one negative eigenvalue.
% \begin{equation}
%     \Delta f < 0 \implies \exists \lambda(\matr{H}_f(\boldsymbol{\theta}^*)) < 0
%     \label{eq68}
% \end{equation}
% Now, consider the set of critical points of the equilibrated energy (Eq. \ref{eq5}) where the rank product of the weight matrices is zero and the last weight matrix is null, namely $\boldsymbol{\theta}^*(\matr{W}_L = \mathbf{0}, \matr{W}_{L-1:1} = \mathbf{0})$ where $\mathbf{g}_{\mathcal{F}^*}(\boldsymbol{\theta}^*) = \mathbf{0}$. Performing a second-order perturbation (as in Eqs. \ref{eq66}-\ref{eq67}) at these critical points along the direction $\boldsymbol{\hat{\theta}} = ( \matr{I}, \mathbf{0}, \dots, \mathbf{0})$ we obtain
%\textcolor{blue}{TODO Mehdi: add full proof here}
we have
\begin{align}
    \mathcal{F}^*(\boldsymbol{\theta}^* + \delta \boldsymbol{\hat{\theta}}) &= \mathcal{F}^* (\delta \matr{I}, \matr{W}_{L-1}, \ldots, \matr{W}_1) \\
    &= \sum_{i=1}^N \mathbf{y}_i^T \left(\matr{I} + \delta^2 \left(\matr{I} + \sum_{\ell=2}^{L-1} \matr{W}_{L-1:\ell} \matr{W}_{L-1:\ell}^T\right)\right)^{-1} \mathbf{y}_i.
\end{align}
Denoting by $\matr{A}:= \matr{I} + \sum_{\ell=2}^{L-1} \matr{W}_{L-1:\ell} \matr{W}_{L-1:\ell}^T$, we have when $\delta \to 0$
\begin{align}
    \matr{S}^{-1} &= (\matr{I} + \delta^2 \matr{A})^{-1} = \matr{I} - \delta^2  \matr{A} + \mathcal{O}(\delta^3).
\end{align}
Hence
\begin{align}
    \mathcal{F}^* (\delta \matr{I},\matr{W}_{L-1},\ldots,\matr{W}_1) &= \sum_{i=1}^N \mathbf{y}_i^T  (\matr{I} - \delta^2 \matr{A} + \mathcal{O}(\delta^3))\mathbf{y}_i \\
    &= \sum_{i=1}^N \mathbf{y}_i^T \mathbf{y}_i - \delta^2 \sum_{i=1}^L \mathbf{y}_i^T\matr{A}\mathbf{y}_i + \mathcal{O}(\delta^3) \\
    &= \mathcal{F}^*(\matr{W}_L,\matr{W}_{L-1},\ldots,\matr{W}_1) -c\delta^2 + \mathcal{O}(\delta^3),
\end{align}
where $c=\sum_{i=1}^L y_i^T\matr{A}y_i>0$ because $\matr{A}$ is symmetric definite positive and there exists $j$ such that $y_j \neq 0$. Hence 
\begin{align}
    \mathcal{F}^*(\boldsymbol{\theta}^* + \delta \boldsymbol{\hat{\theta}}) = \mathcal{F}^*(\boldsymbol{\theta}^*) - c \delta^2 + \mathcal{O}(\delta^3),
\end{align}
which concludes the proof.
%\textcolor{blue}{to be continued ... end of proof}

%  \begin{align}
%     \Delta \mathcal{F}^* &\approx \frac{1}{2} \delta^2 \boldsymbol{\hat{\theta}}^T \matr{H}_{\mathcal{F}^*} \boldsymbol{\hat{\theta}} \\
%     &\approx - \delta^2 \sum_{i=1}^N \mathbf{y}_i^T \matr{C} \mathbf{y}_i < 0
% \end{align}
% where $\matr{C} = I + \sum_{\ell=2}^{L-1} (W_{L-1:\ell}) (W_{L-1:\ell})^T$. Since $\matr{C}$ is symmetric positive definite, then the energy decreases along the considered direction $\boldsymbol{\hat{\theta}}$ and so there exists negative curvature (the Hessian has at least one negative eigenvalue at the given critical points, $\exists \lambda(\matr{H}_{\mathcal{F}^*}(\boldsymbol{\theta}^*)) < 0$ as per Eq. \ref{eq68}).

\subsection{Flatter global minima of the equilibrated energy (linear chains)}
\label{ch4:energy-minima}
Here we present a preliminary investigation into the minima of the equilibrated energy compared to the MSE loss. For linear chains (\S\ref{ch4:chains-hessian}), we show that global minima of the equilibrated energy are flatter than those of the MSE loss. More precisely, the energy global minima turn out be scaled down versions of those of the loss by the same rescaling factor of the equilibrated energy (\S\ref{ch4:equilib-energy}). This generalises the result of \citep{innocenti2023understanding} for linear chains with a single hidden unit.

The proof has only two steps and does not require explicit calculation of the Hessian. First, we know that we are at a global minimum of loss when we perfectly fit the data $w_{L:1}x = y$, since $\mathcal{L}(w_{L:1}x = y) = 0$. This is also true of the equilibrated energy, $\mathcal{F}^*(w_{L:1}x = y) = 0$. We can check that these are critical points by seeing that the weight gradient of the loss is zero $\nabla_{\boldsymbol{\theta}} \mathcal{L} (w_{L:1}x = y) = \mathbf{0}$, which follows from the fact that the residual vanishes when we perfectly fit the data. Again, the same is true of the energy, $\nabla_{\boldsymbol{\theta}} \mathcal{F}^* (w_{L:1}x = y) = \mathbf{0}$.

The second and last step is to realise that, at these minima, the terms of the energy Hessian (Eq.~\ref{ch4:eq:equilib-energy-hess-chain}) collapse to those of a rescaled loss Hessian (Eq.~\ref{ch4:eq:loss-hess-chain}):
\begin{align}
    \frac{\partial^2 \mathcal{F}^*}{\partial w_i \partial w_j}(w_{L:1}x = y) = 
    \begin{cases} 
    \Large \frac{1}{s}\frac{\partial^2 \mathcal{L}}{\partial w_i \partial w_j}, & i = j = 1  \\ \\
    \Large \frac{1}{s} \frac{\partial^2 \mathcal{L}}{\partial w_i \partial w_j}, & i = 1, \quad j > 1 \\ \\ 
    \Large \frac{1}{s} \frac{\partial^2 \mathcal{L}}{\partial w_i \partial w_j}, & i, j > 1 \\
    \end{cases},
\end{align}
where the rescaling is the same as that of the equilibrated energy (Eq.~\ref{ch4:eq:equilib-energy-linear-chain}). Factoring out the rescaling
\begin{align}
    \matr{H}_{\mathcal{F}*}(w_{L:1}x = y) &= \matr{H}_{\mathcal{L}}(w_{L:1}x = y) / s \\
    \implies \matr{H}_{\mathcal{F}*}(w_{L:1}x = y) &< \matr{H}_{\mathcal{L}}(w_{L:1}x = y),
\end{align}
we observe that the minima of the equilibrated energy are simply a rescaled version of those of the loss. As we saw in \S\ref{ch4:equilib-energy}, the rescaling is positive, so it follows that the global minima of the equilibrated energy are flatter than those of the loss. In other words, PC inference has the effect of flattening the global minima of the MSE loss (at least for linear chains).

\section{Experimental details} 
\label{ch4:exp-details}
Code to reproduce all the experiments is available at \sloppy{\url{https://github.com/francesco-innocenti/pc-saddles}}. Unless otherwise stated, for all PC networks standard GD with step size $\beta = 0.1$ was used to converge the inference dynamics (\S\ref{ch4:pc}, Eq.~\ref{ch4:eq:pc-infer}), with the number of iterations depending on the problem.

\paragraph{Theoretical energy (Figure~\ref{ch4:fig:dln-equilib-energy}).} We trained DLNs with different number of hidden layers $H \in \{2, 5, 10 \}$ on standard image classification datasets (MNIST, Fashion-MNIST and CIFAR10). At every training step, we compared the total energy (Eq.~\ref{ch4:eq:pc-energy}) at the numerical inference equilibrium $\mathcal{F}|_{\nabla_{\mathbf{z}} \mathcal{F}\approx0}$ with the theoretical prediction (Eq.~\ref{ch4:eq:dln-equilib-energy}). The following hyperparameters were used for all networks: 300 hidden units and SGD with learning rate $\eta = 1e^{-3}$ and batch size 64. We used a second-order explicit Runge--Kutta ordinary differential equation solver (Heun) with a maximum upper integration limit $T = 300$ and an adaptive Proportional-Integral-Derivative controller (absolute and relative tolerances: $1e^{-3}$) to ensure convergence of the PC inference dynamics (Eq.~\ref{ch4:eq:pc-infer}). Results were consistent across different random initialisations.

\paragraph{Toy examples (Figure~\ref{ch4:fig:toy-examples}).} All networks were linear and trained on a toy regression problem using the MSE loss (Eq.~\ref{ch4:eq:mse-loss}) and energy (Eq.~\ref{ch4:eq:pc-energy}) with output $\mathbf{y} = -\mathbf{x}, \mathbf{x}_i \sim \mathcal{N}(1, 0.1)$. Weights were initialised close to the origin $\matr{W}_{ij} \sim \mathcal{N}(0, \sigma^2)$ with $\sigma \ll 1$. For the chains, the initialisation scale was chosen to be $\sigma = 5e^{-2}$, while for the wide network it was increased to $\sigma = 1e^{-1}$ in order to make escape from the saddle faster but still visible. For PC, $T = 20$ inference iterations were used for chains and $50$ for the wide network. All networks were trained with SGD and batch size $64$. Learning rate $\eta = 0.4$ was used for the chains and $1e^{-3}$ for the wide network. Training was stopped when it was determined that convergence had been effectively reached, to allow for intuitive visualisation of the loss dynamics. 

\begin{figure}[H]
    \begin{center}
        \centerline{\includegraphics[width=0.8\textwidth]{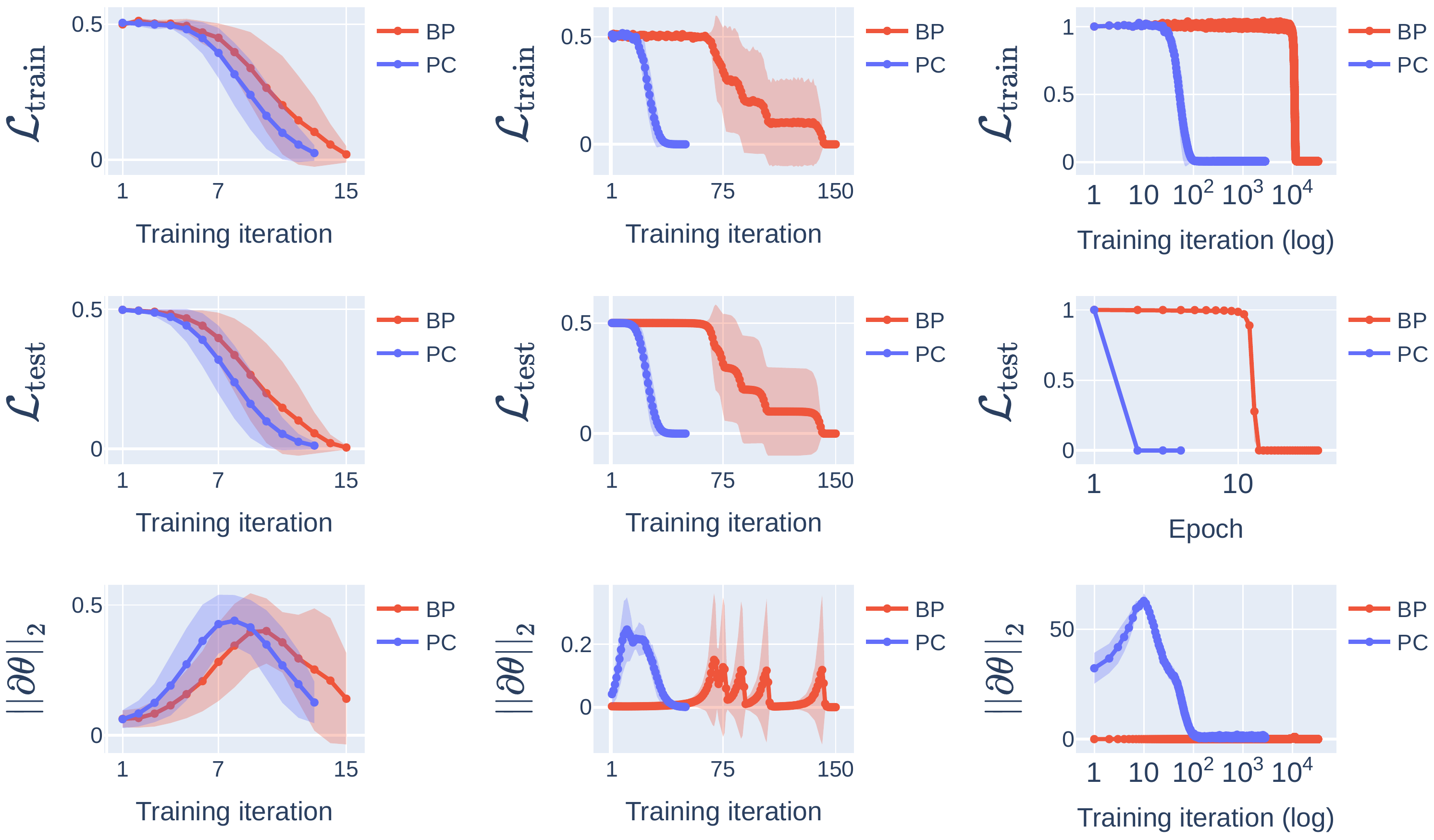}}
        \caption{\textbf{Training and test statistics for linear networks of Figure~\ref{ch4:fig:toy-examples}.} For each network, we plot the mean and $\pm1$ standard deviation of the training loss, test loss and gradient norm over 5 random initialisations. For the wide network, the test loss is evaluated once every epoch (rather than for each batch), and the training metrics are plotted on a log axis for easier visualisation. For the chain with two hidden units, the multiple loss plateaus and corresponding gradient spikes are due to different escape times from the saddle for different runs.}
        \label{ch4:fig:toy-examples-stats}
    \end{center}
\end{figure}

The landscapes were sampled on the training loss or energy with a $30\times30$ resolution and domain $\in [-2, 2]$ for the 2-hidden node chain and $\in [-1, 1]$ for the other networks. For the wide network, the landscape was projected onto the maximum and minimum eigenvectors of the Hessian at the origin $\boldsymbol{\theta}^* = \mathbf{0}$, $f(\boldsymbol{\theta}^* + \alpha \hat{\mathbf{v}}_{\text{min}} + \beta \hat{\mathbf{v}}_{\text{max}})$ since as shown by \citep{bottcher2024visualizing} random directions \citep{li2018visualizing} can fail to identify saddle points. The energy landscape was plotted at the numerical equilibrium $\mathcal{F}(\boldsymbol{\theta})|_{\nabla_{\mathbf{z}}\mathcal{F} \approx 0}$. Figure~\ref{ch4:fig:toy-examples} displays results for an example run, and Figure~\ref{ch4:fig:toy-examples-stats} shows the statistics of the training and test losses as well as the weight gradient norms for 5 random initialisations.

\begin{figure}[H]
    \begin{center}
        \centerline{\includegraphics[width=\textwidth]{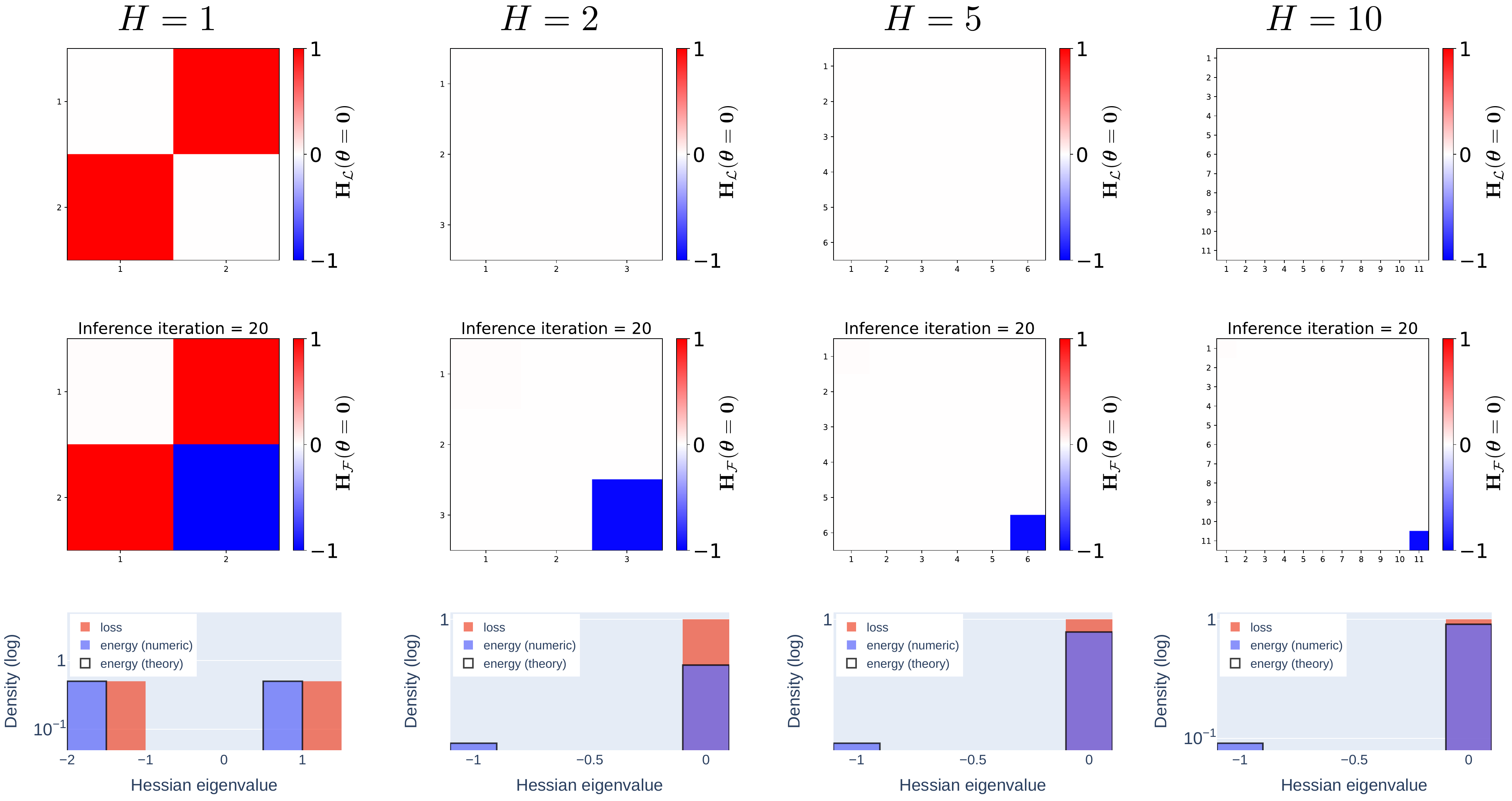}}
        \caption{\textbf{Empirical verification of the Hessian at the origin of the equilibrated energy for linear chains.} This shows the same results of Figure~\ref{ch4:fig:hess-origin-toy} for networks of unit width $N_0 = \dots = N_L = 1$ (see \S\ref{ch4:exp-details} for details). Again, we observe a perfect match between theory and experiment (see in particular Eq.~\ref{ch4:eq:equilib-energy-hess-chain-origin}).}
        \label{ch4:fig:hess-origin-toy-chain}
    \end{center}
\end{figure}

\begin{figure}[H]
    \begin{center}
        \centerline{\includegraphics[width=0.8\textwidth]{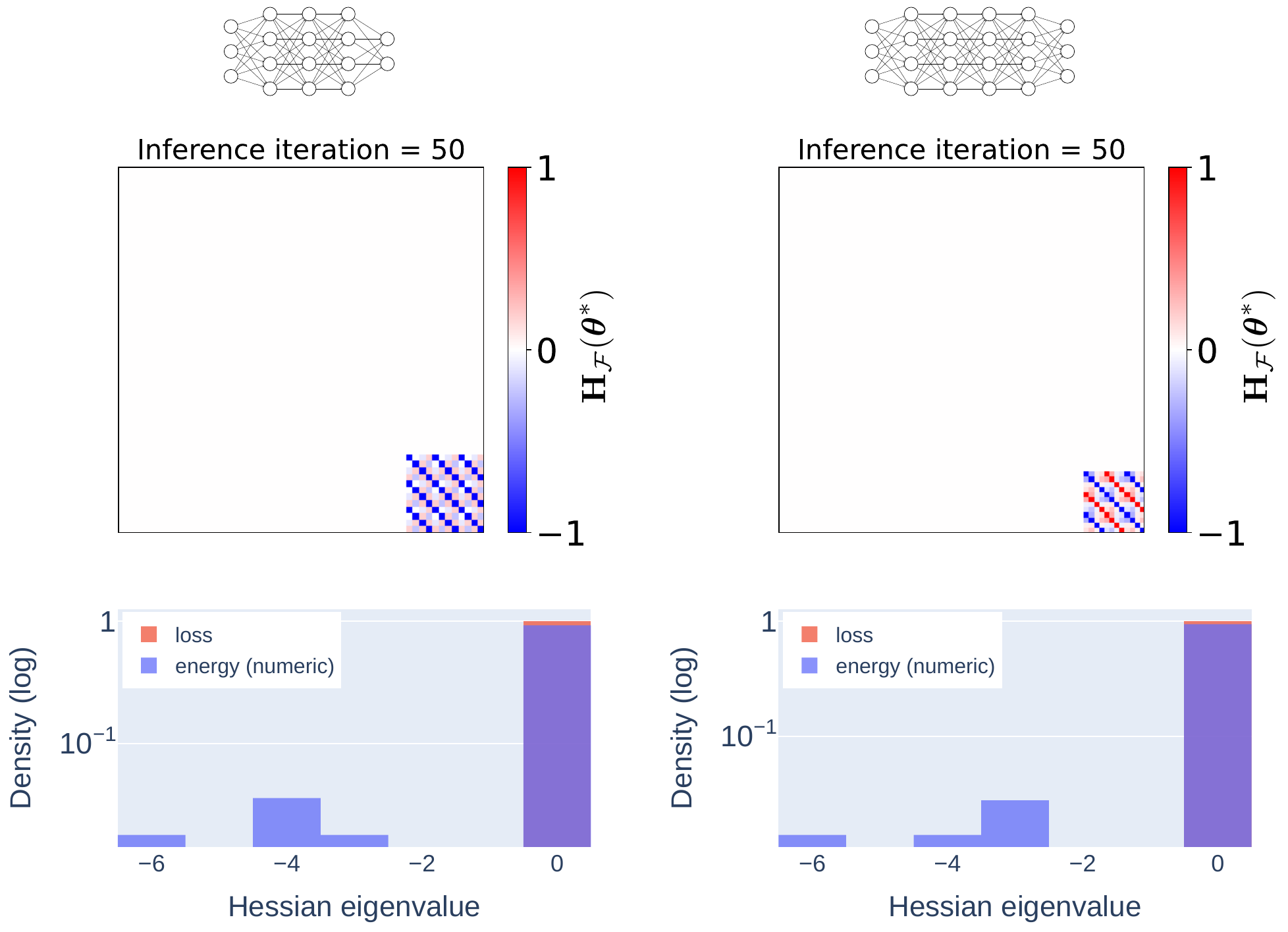}}
        \caption{\textbf{Empirical verification of a strict zero-rank saddle of the equilibrated energy other than the origin for DLNs tested on a toy dataset.} We show the Hessian eigenspectrum of the MSE loss and equilibrated energy at a strict saddle other than the origin covered by Theorem~\ref{ch4:thm3}, specifically for the critical point where all weight matrices except the penultimate are zero $\boldsymbol{\theta}^*(\matr{W}_\ell = \mathbf{0}, \forall \ell \neq L-1)$. We do not show the loss Hessians because they are zero for $H > 1$ (Eq.~\ref{ch4:eq:loss-hess-origin}). The target is the same as used for Figure~\ref{ch4:fig:hess-origin-toy}, and in the right panel one of the output dimensions is varied to be $y_2 = x_2$. Figure~\ref{ch4:fig:hess-other-saddle-mnist} shows results for the same critical point on MNIST and MNIST-1D.}
        \label{ch4:fig:hess-other-saddle-toy}
    \end{center}
\end{figure}

\begin{figure}[H]
    \begin{center}
        \centerline{\includegraphics[width=0.8\textwidth]{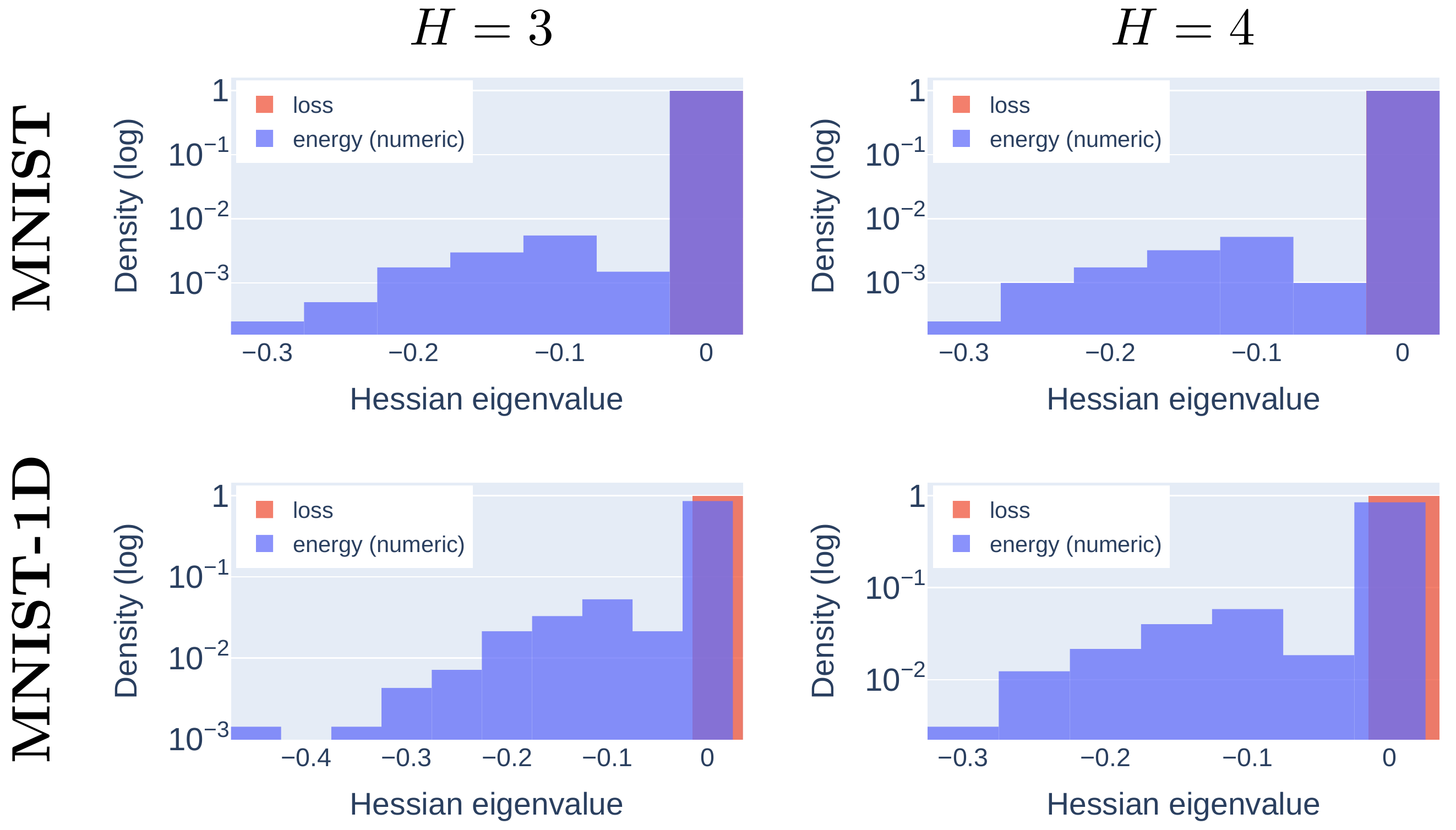}}
        \caption{\textbf{Empirical verification of a strict zero-rank saddle of the equilibrated energy other than the origin for DLNs tested on more realistic datasets.} This shows similar results to Figure~\ref{ch4:fig:hess-other-saddle-toy} for the more realistic datasets MNIST and MNIST-1D.}
        \label{ch4:fig:hess-other-saddle-mnist}
    \end{center}
\end{figure}

\paragraph{Hessian eigenspectra (Figure~\ref{ch4:fig:hess-origin-toy}-\ref{ch4:fig:hess-origin-mnist}).} For different linear network architectures, we computed the Hessian of the loss and equilibrated energy at the origin on a random batch (of size $64$) of a given dataset. The datasets used were (i) a toy Gaussian with 3D input and output with the same statistics used for experiments in Figure~\ref{ch4:fig:toy-examples}, (ii) MNIST and (iii) MNIST-1D \citep{greydanus2020scaling}, a procedurally generated, one-dimensional dataset smaller than MNIST with higher model discriminability. The depth, width and data dimensions of the networks tested on the Gaussian data are clear from the vignettes in Figure~\ref{ch4:fig:hess-origin-toy}. Figure~\ref{ch4:fig:hess-origin-toy-chain} shows the same results for linear chains. For MNIST and MNIST-1D, networks with $H$ hidden layers $\{1, 2, 3\}$ had $N_\ell$ widths $\{10, 10, 5\}$ and $\{100, 50, 10\}$, respectively. Note that the MNIST networks were relatively narrow to allow for tractable computation of the Hessian. The Hessian matrices for the Gaussian data were normalised between 1 and -1, and the Hessian of the energy was computed after $T = 50$ inference iterations. For the theoretical eigenspectra of the energy Hessian, we computed the eigenvalues of Eq.~\ref{ch4:eq:equilib-energy-hess-origin}. Figures~\ref{ch4:fig:hess-origin-toy} and \ref{ch4:fig:hess-origin-mnist} show results for an example run, and we found practically indistinguishable results for different seeds. Figures~\ref{ch4:fig:hess-other-saddle-toy} \& \ref{ch4:fig:hess-other-saddle-mnist} show a similar analysis for a zero-rank saddle covered by Theorem~\ref{ch4:thm3} other than the origin.

\paragraph{Experiments (Figure~\ref{ch4:fig:origin-escape}-\ref{ch4:fig:matrix-completion}).} For the first set of experiment, we trained and tested linear, Tanh and ReLU networks on standard image classification tasks. Networks tested on MNIST and Fashion-MNIST had 5 fully connected (FC) layers with 500 hidden units, while those trained on CIFAR-10 had a convolutional architecture consisting of 3 blocks (with a convolution and max pooling operation) followed by two FC layers (with the last one always being linear). For PC, $T=50$ inference iterations were used. Similar to the experiments for Figure~\ref{ch4:fig:toy-examples}, all networks were initialised close to the origin $\matr{W}_{ij} \sim \mathcal{N}(0, \sigma^2)$ with $\sigma = 5e^{-3}$. SGD with batch size $64$ and learning rate $\eta = 1e^{-3}$ was used for all networks. PC networks were trained until the training loss reached the tolerance $\mathcal{L}_{\text{train}} < 1e^{-3}$. For computational reasons, the BP-trained networks were not trained until convergence. Instead, training was stopped at as many iterations as it took PC to converge. We do report the full saddle escape dynamic for the toy examples in Figure~\ref{ch4:fig:toy-examples} and the matrix completion experiment in Figure~\ref{ch4:fig:matrix-completion}. All hyperparameters except for the initialisation remained unchanged for the other (zero-rank) saddle experiment shown in Figure~\ref{ch4:fig:other-saddle-escape}. 

For the matrix completion task (Figure~\ref{ch4:fig:matrix-completion}), we attempted to replicate the experiment by \citep[Figure~1]{jacot2021saddle} as closely as possible. Networks of depth $H=3$ and width $N=100$ were trained with GD and learning rate $\eta = 1e^{-2}$ to fit a 10x10 matrix of rank 3. The target matrix was generated by multiplying two i.i.d. matrices of size 10x3 with standard Gaussian entries, and 20\% of these entries were masked during training. The networks trained with PC were initialised at each saddle visited by BP, which was determined numerically by computing the rank of the network map. The origin initialisation had the same scale $\sigma = 5e^{-3}$ used in the previous experiments.

\section{Supplementary figures} 
\label{ch4:supp-figs}
\begin{figure}[H]
    \begin{center}
        \centerline{\includegraphics[width=0.9\textwidth]{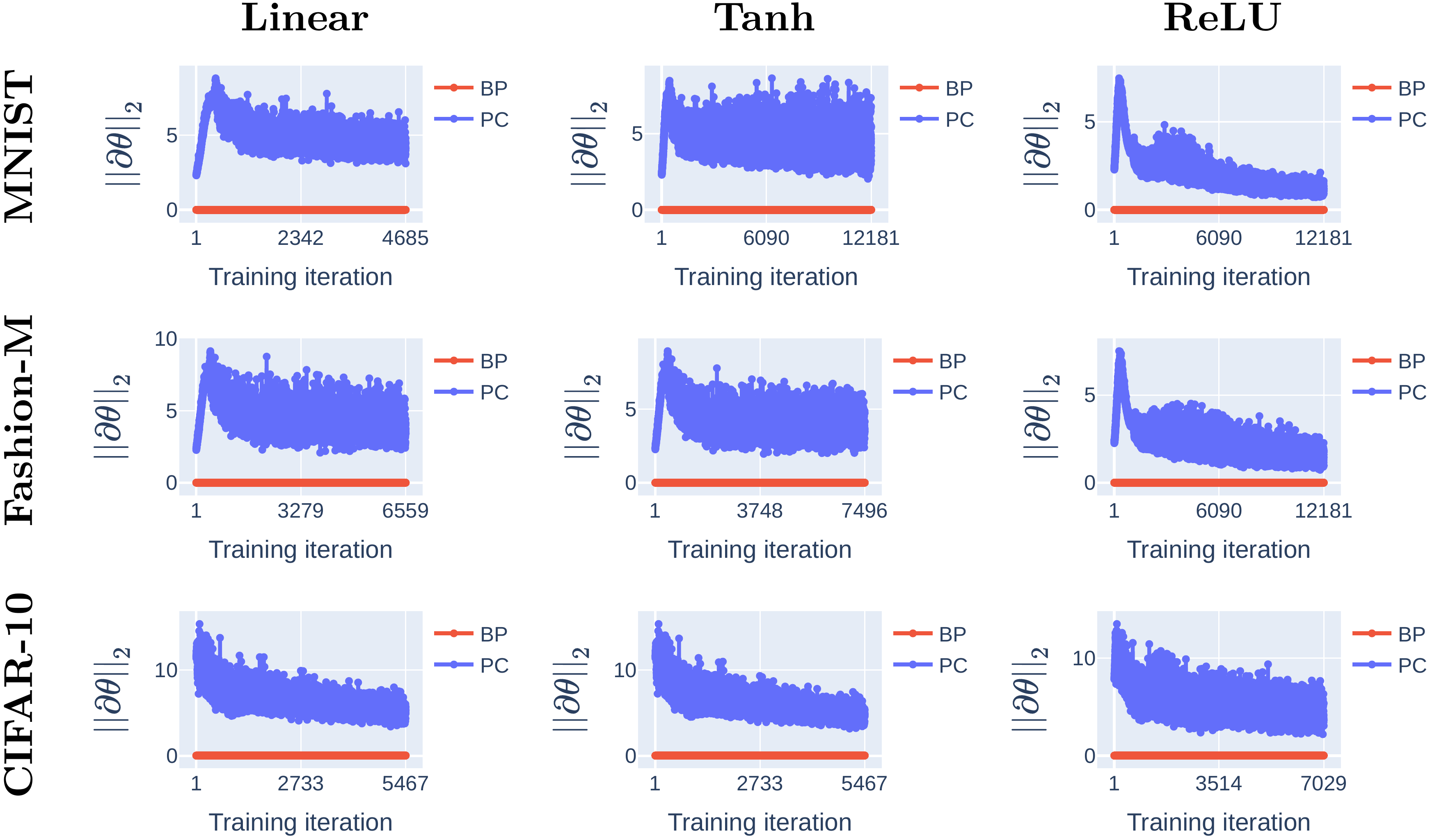}}
        \caption{\textbf{No vanishing gradients for PC starting near the origin.} Weight gradient norms of the loss $||\nabla_{\boldsymbol{\theta}} \mathcal{L}||$ (BP) and equilibrated energy $||\nabla_{\boldsymbol{\theta}} \mathcal{F}^*||$ (PC) for the experiments in Figure~\ref{ch4:fig:origin-escape}.}
        \label{ch4:fig:origin-grad-norms}
    \end{center}
\end{figure}
\begin{figure}[H]
    \begin{center}
        \centerline{\includegraphics[width=0.9\textwidth]{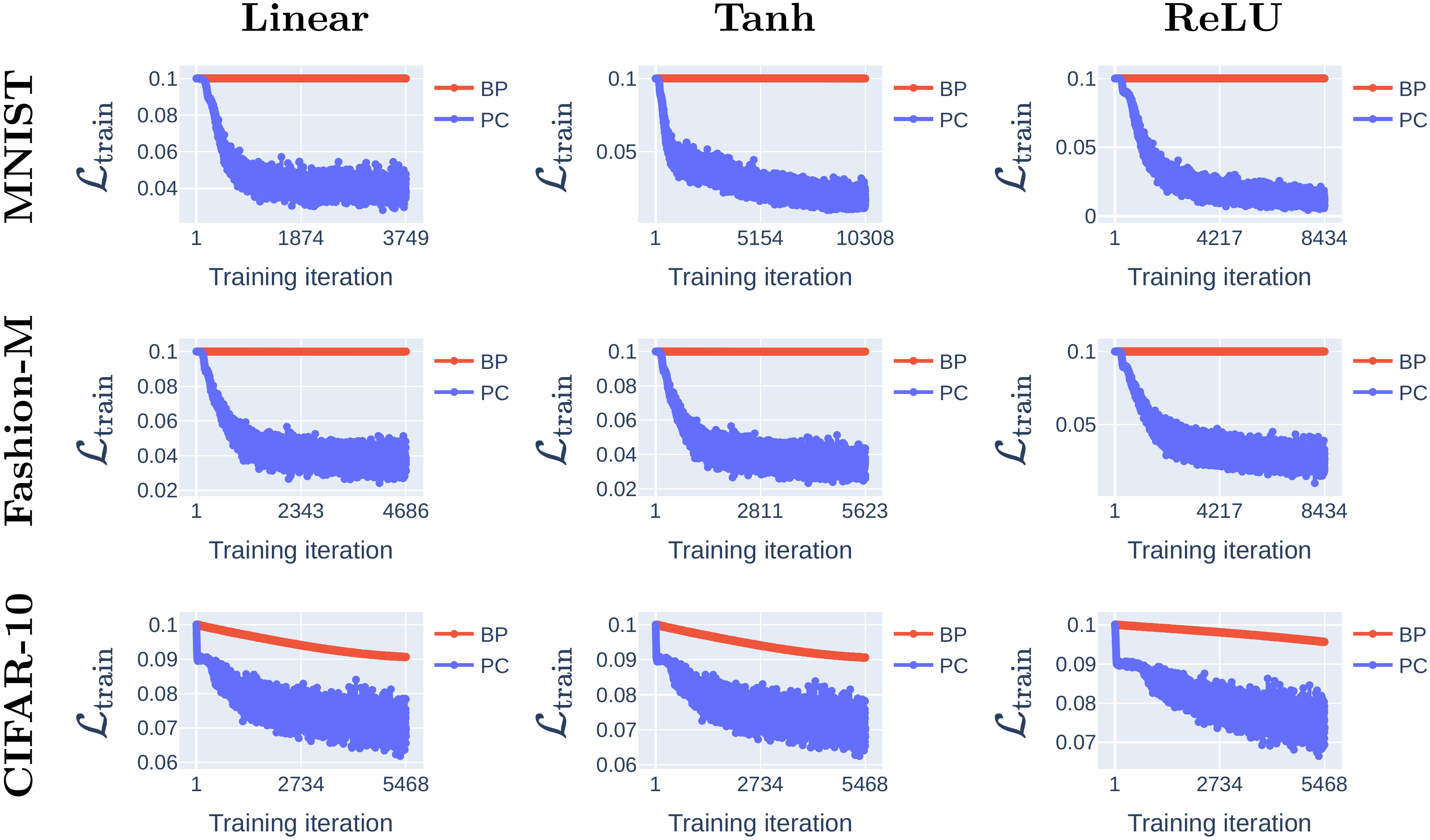}}
        \caption{\textbf{PC escapes another non-strict saddle of the loss much faster than BP with SGD on non-linear networks.} This shows the same results as Figure~\ref{ch4:fig:origin-escape} for the same saddle analysed in Figures~\ref{ch4:fig:hess-other-saddle-toy} \& \ref{ch4:fig:hess-other-saddle-mnist} (see \S\ref{ch4:exp-details} for details). We show results for an example run as they were practically indistinguishable across different random seeds.}
        \label{ch4:fig:other-saddle-escape}
    \end{center}
\end{figure}

%% file: text/appendices/ch5.tex
\chapter{Appendix for Chapter 5}
\label{ch5:appendix}
\minitoc

\renewcommand{\thefigure}{C.\arabic{figure}}
\setcounter{figure}{0}

\section{Related work} 
\label{ch5:related-work}

\paragraph{$\mu$P for PC \cite{ishikawa2024local}.} The study closest to our work is \cite{ishikawa2024local}, who derived a $\mu$P parameterisation for PC (as well as target propagation), also showing hyperparameter transfer across widths. This work differs from ours in the following three important aspects: (i) it derives $\mu$P for PC only for the width, (ii) it focuses on regimes where PC approximates or is equivalent to other algorithms (including BP) so that all the $\mu$P theory can be applied, and (iii) it considers layer-wise scalar precisions $\gamma_\ell$ for each layer energy term, which are not standard in how PCNs are trained (but are nevertheless interesting to study). By contrast, we propose to apply Depth-$\mu$P to PC, showing transfer for depth as well as width (Figs.~\ref{ch5:fig:mupc-hyperparam-transfer-tanh} \& \ref{ch5:fig:mupc-hyperparam-transfer-linear}-\ref{ch5:fig:mupc-hyperparam-transfer-relu}). We also study a regime where this parameterisation reduces to BP (Fig.~\ref{ch5:fig:mupc-loss-energy-ratios-init-1e-1}) while showing that successful training is still possible far from this regime (Fig.~\ref{ch5:fig:mupc-vs-pc-mnist-accs}).

\paragraph{Training deep PCNs \cite{qi2025training, pinchetti2024benchmarking}.} Our work is also related to \cite{qi2025training}, who following \cite{pinchetti2024benchmarking} showed that the PC energy (Eq.~\ref{ch5:eq:pc-energy}) is disproportionately concentrated at the output layer $\mathcal{F}_L$ (closest to the target) for deep PCNs. They conjecture that this is problematic for two reasons: first, it does not allow the model to use (i.e. update) all of its layers; and second, it makes the latents diverge from the forward pass, which they claim leads to suboptimal weight updates. The first point is consistent with our theory and experiments. In particular, because the activities of standard PCNs vanish/explode with the depth (\S \ref{ch5:vanish-explode-ffwd}) and stay almost constant during inference due to the ill-conditioning of the landscape (\S \ref{ch5:ill-cond-infer}) (Figs.~\ref{ch5:fig:sp-activity-inits}-\ref{ch5:fig:mupc-vs-pc-ffwd-lrs} \& \ref{ch5:fig:mupc-vs-pc-ffwd-relu-lrs}), the weight updates are likely to be imbalanced across layers. However, the ill-conditioning contradicts the second point, in that the activities barely move during inference and stay close to the forward pass (see \S \ref{ch5:activity-inits} for relevant experiments). Moreover, divergence from the forward pass does not necessarily lead to suboptimal weight updates and worse performance. For standard PC, deep networks cannot achieve good performance regardless of whether one stays close to the forward pass (see \S \ref{ch5:infer-converge-sufficient}). For $\mu$PC, on the other hand, as many steps as the number of hidden layers (e.g. Fig.~\ref{ch5:fig:mupc-vs-pc-mnist-accs}) leads to depth-stable and much better accuracy than a single step (e.g. Fig.~\ref{ch5:fig:mupc-one-step-mnist}). Another recent study that investigated the problem of training deep PCNs is \cite{goemaere2025error}, which we discuss in \S \ref{ch5:discussion}.

\paragraph{PC and BP.} Our theoretical result about the convergence of $\mu$PC to BP (Theorem~\ref{ch5:thm1}) relates to a relatively well-established series of correspondences between PC and BP \cite{whittington2017approximation, millidge2022predictive, song2020can, rosenbaum2022relationship, salvatori2021predictive, millidge2022backpropagation}. In brief, if one makes some rather biologically implausible assumptions (such as precisely timed inference updates), it can be shown that PC can approximate or even compute exactly the same gradients as BP. In stark contrast to these results and also the work of \cite{ishikawa2024local} (which requires arbitrarily specific precision values at different layers), Theorem~\ref{ch5:thm1} applies to standard PC, with arguably interpretable width- and depth-dependent scalings.\footnote{The width scaling is inherently local, while the depth scaling is more global but could be perhaps argued to be bio-plausible based on a notion of the brain ``knowing its own depth''.}

\paragraph{Theory of PC inference (Eq.~\ref{ch5:eq:pc-infer}) \& learning (Eq.~\ref{ch5:eq:pc-learn}).} Finally, our work can be seen as a companion to the study presented in the previous chapter \cite{innocenti2025only}, where we provided the first rigorous, explanatory and predictive theory of the learning landscape and dynamics of practical PCNs (Eq.~\ref{ch5:eq:pc-learn}). Recall that we first showed that for DLNs the energy at the inference equilibrium is a rescaled MSE loss with a weight-dependent rescaling, a result that we build on here for Theorem~\ref{ch5:thm1}. We then characterised the geometry of the equilibrated energy (the effective landscape on which PC learns), showing that many highly degenerate saddles of the loss including the origin become much easier to escape in the equilibrated energy. Here, by contrast, we focus on the geometry of the \textit{inference landscape and dynamics} (Eq.~\ref{ch5:eq:pc-infer}). As an aside, we note that the origin saddle result of the previous chapter probably breaks down for ResNets, where for the linear case it has been shown that the saddle is effectively shifted and the origin becomes locally convex \cite{hardt2016identity}. We suspect that the results generalise, but it could still be interesting to extend the theory of the previous chapter to ResNets, especially by also looking at the geometry of minima.

\paragraph{$\mu$P.} For a full treatment of $\mu$P and its extensions, we refer the reader to key works of the ``Tensor Programs'' series \cite{yang2021tensor, yang2021tuning, yang2023tensoradaptive, yang2023tensorinfdepth}. $\mu$P effectively puts feature learning back into the infinite-width limit of neural networks, lacking from the neural tangent kernel (NKT) or ``lazy'' regime \cite{jacot2018neural, chizat2019lazy, lee2019wide}. In particular, in the NTK the layer preactivations evolve in $\mathcal{O}(N^{-1/2})$ time. In $\mu$P, the features instead change in a ``maximal'' sense (hence ``$\mu$''), in that they vary as much as possible without diverging with the width, which occurs for the output predictions under SP \cite{yang2021tensor}. More formally, $\mu$P can be derived from the 3 desiderata stated in \S \ref{ch5:mup}. $\mu$P was extended to depth (Depth-$\mu$P) for ResNets by mainly introducing a $1/\sqrt{L}$ scaling before each residual block \cite{yang2023tensorinfdepth, bordelon2023depthwise}. This breakthrough was enabled by the commutativity of the infinite-width and infinite-depth limit of ResNets \cite{hayou2023width, hayou2024commutative}. Standard $\mu$P has also been extended to local algorithms including PC \cite{ishikawa2024local} (see \textbf{$\mu$P for PC} above), sparse networks \cite{dey2024sparse}, second-order methods \cite{ishikawa2023parameterization}, and sharpness-aware minimisation \cite{haas2024effective}.

\section{Proofs and derivations}
All the theoretical results below are derived for linear networks of some form.

\subsection{Activity gradient (Eq.~\ref{ch5:eq:pc-infer-solution}) and Hessian (Eq.~\ref{ch5:eq:activity-hessian}) of DLNs} \label{ch5:activity-grad-hess}
The gradient of the energy with respect to all the PC activities of a DLN (Eq.~\ref{ch5:eq:pc-infer-solution}) can be derived by simple rearrangement of the partials with respect to each layer, which are given by
\begin{align}
    \partial \mathcal{F}/\partial \mathbf{z}_1 &= \mathbf{z}_1 - a_1\matr{W}_1\mathbf{x} - a_2\matr{W}_2^T\mathbf{z}_2 + a_2^2\matr{W}_2^T\matr{W}_2\mathbf{z}_1 \\
    \partial \mathcal{F}/\partial \mathbf{z}_2 &= \mathbf{z}_2 - a_2\matr{W}_2\mathbf{z}_1 - a_3\matr{W}_3^T\mathbf{z}_3 + a_3^2\matr{W}_3^T\matr{W}_3\mathbf{z}_2 \\
    \quad \quad \vdots \\
    \partial \mathcal{F}/\partial \mathbf{z}_H &= \mathbf{z}_H - a_{L-1}\matr{W}_{L-1}\mathbf{z}_{H-1} - a_L\matr{W}_L^T\mathbf{y} + a_L^2\matr{W}_L^T\matr{W}_L\mathbf{z}_H.
\end{align}
Factoring out the activity of each layer
\begin{align}
    \partial \mathcal{F}/\partial \mathbf{z}_1 &= \mathbf{z}_1(\mathbf{1} + a_2^2\matr{W}_2^T\matr{W}_2) - a_1\matr{W}_1\mathbf{x} - a_2\matr{W}_2^T\mathbf{z}_2 \\
    \partial \mathcal{F}/\partial \mathbf{z}_2 &= \mathbf{z}_2 (\mathbf{1} + a_3^2\matr{W}_3^T\matr{W}_3) - a_2\matr{W}_2\mathbf{z}_1 - a_3\matr{W}_3^T\mathbf{z}_3 \\
    \quad \quad \vdots \\
    \partial \mathcal{F}/\partial \mathbf{z}_H &= \mathbf{z}_H (\mathbf{1} + a_L^2\matr{W}_L^T\matr{W}_L) - a_{L-1}\matr{W}_{L-1}\mathbf{z}_{H-1} - a_L\matr{W}_L^T\mathbf{y},
\end{align}
one realises that this can be rearranged in the form of a linear system
\begin{align}
    \nabla_\mathbf{z} \mathcal{F} &= \underbrace{\begin{bsmallmatrix} 
        \matr{I} + a_2^2\matr{W}_2^T \matr{W}_2 & -a_2\matr{W}_2^T & \mathbf{0} & \dots & \mathbf{0} \\ -a_2\matr{W}_2 & \matr{I} + a_3^2\matr{W}_3^T \matr{W}_3 & -a_3\matr{W}_3^T & \dots & \mathbf{0} \\ \mathbf{0} & -a_3\matr{W}_3 & \matr{I} + a_4^2\matr{W}_4^T \matr{W}_4 & \ddots & \mathbf{0} \\ \vdots & \vdots & \ddots & \ddots & -a_{L-1}\matr{W}_{L-1}^T \\ \mathbf{0} & \mathbf{0} & \mathbf{0} & -a_{L-1}\matr{W}_{L-1} & \matr{I} + a_L^2\matr{W}_L^T \matr{W}_L
    \end{bsmallmatrix}}_{\matr{H}_{\mathbf{z}}}
    \underbrace{\begin{bmatrix} 
        \mathbf{z}_1 \\
        \mathbf{z}_2 \\
        \vdots \\
        \mathbf{z}_{H-1} \\
        \mathbf{z}_H
    \end{bmatrix}}_{\mathbf{z}} - 
    \underbrace{\begin{bmatrix} 
        a_1\matr{W}_1\mathbf{x} \\
        \mathbf{0} \\
        \vdots \\
        \mathbf{0} \\
        a_L\matr{W}_L^T\mathbf{y}
    \end{bmatrix}}_{\mathbf{b}}
    \label{eq15}
\end{align}
where the matrix of coefficients corresponds to the Hessian of the energy with respect to the activities $(\matr{H}_{\mathbf{z}})_{\ell k} \coloneqq \partial^2 \mathcal{F}/\partial \mathbf{z}_\ell \partial \mathbf{z}_k$. We make the following side remarks about how different training and architecture design choices impact the structure of the activity Hessian:
\begin{itemize}
    \item In the unsupervised case where $\mathbf{z}_0$ is left free to vary like any other hidden layer, the Hessian gets the additional terms $a_1^2\matr{W}_1^T\matr{W}_1$ as the first diagonal block, $-a_1\matr{W}_1$ as the superdiagonal block (and its transpose as the subdiagonal block), and $\mathbf{b}_1 = \mathbf{0}$.\footnote{Note that the lack of an identity term in the block diagonal term comes from the fact that the first layer is not directly predicted by any other layer.} This does not fundamentally change the structure of the Hessian; in fact, in the next section we show that convexity holds for both the unsupervised and supervised cases.
    \item Turning on biases at each layer such that $\mathcal{F}_\ell = \frac{1}{2}||\mathbf{z}_\ell - a_\ell\matr{W}_\ell \mathbf{z}_{\ell-1} - \mathbf{b}_\ell||^2$ does not impact the Hessian and simply makes the constant vector of the linear system more dense: $\mathbf{b} = [a_1\matr{W_1}\mathbf{x} + \mathbf{b_1} - a_2\matr{W}_2^T \mathbf{b}_2, \mathbf{b}_2 - a_3\matr{W}_3^T \mathbf{b}_3, \dots, a_L\matr{W}_L^T \mathbf{y} + \mathbf{b}_{L-1} - a_L\matr{W}^T_L \mathbf{b}_L]^T$.
    \item Adding an $\ell^2$ norm regulariser to the activities $\frac{1}{2}||\mathbf{z}_\ell||^2$ scales the identity in each diagonal block by 2. This induces a unit shift in the Hessian eigenspectrum such that the minimum eigenvalue is lower bounded at one rather than zero (see \S \ref{ch5:rand-matrix}), as shown in Fig.~\ref{ch5:fig:activity-decay-exp}.
    \item Adding ``dummy'' latents at either end of the network, such that $\mathcal{F}_0 = \frac{1}{2}||\mathbf{x} - \mathbf{z}_0||^2$ or $\mathcal{F}_L = \frac{1}{2}||\mathbf{y} - \mathbf{z}_L||^2$, simply adds one layer to the Hessian with a block diagonal given by $2\matr{I}$.
    \item Compared to fully connected networks, the activity Hessian of convolutional networks is sparser in that (dense) weight matrices are replaced by (sparser) Toeplitz matrices. The activity Hessian of ResNets is derived and discussed in \S \ref{ch5:resnets-hessian}.
\end{itemize}
We also note that Eq.~\ref{eq15} can be used to provide an alternative proof of the known convergence of PC inference to the feedforward pass \cite{millidge2022theoretical} $\mathbf{z}^* = \matr{H}_{\mathbf{z}}^{-1} \mathbf{b} = f(\mathbf{x}) = a_L\matr{W}_L \dots a_1\matr{W}_1 \mathbf{x}$ when the output layer is unclamped or free to vary with $\partial^2 \mathcal{F}/\partial \mathbf{z}_L^2 = \matr{I}$ and $\mathbf{b}_H = \mathbf{0}$.

\subsection{Positive definiteness of the activity Hessian} 
\label{ch5:pos-def}
Here we prove that the Hessian of the energy with respect to the activities of arbitrary DLNs (Eq.~\ref{ch5:eq:activity-hessian}) is positive definite (PD), $\matr{H}_{\mathbf{z}}\succ 0$. The result is empirically verified for DLNs in \S \ref{ch5:rand-matrix} and also appears to generally hold for nonlinear networks, where we observe small negative Hessian eigenvalues only for very shallow Tanh networks with no skip connections (see Figs.~\ref{ch5:fig:max-min-eigens} \& \ref{ch5:fig:resnets-cond-nums-init}).
\begin{tcolorbox}[width=\linewidth, sharp corners=all, colback=white!95!black, colframe=white!95!black]
    \begin{mythm}{A.1}[Convexity of the PC inference landscape of DLNs.]\label{ch5:thmA1}
        For any DLN parameterised by $\boldsymbol{\theta} \coloneq (\matr{W}_1, \dots, \matr{W}_L)$ with input and output $(\mathbf{x}, \mathbf{y})$, the activity Hessian of the PC energy (Eq.~\ref{ch5:eq:pc-energy}) is positive definite
            \begin{equation}
                \matr{H}_{\mathbf{z}}(\boldsymbol{\theta}) \succ 0,
            \end{equation}
        showing that the inference or activity landscape $\mathcal{F}(\mathbf{z})$ is strictly convex.
    \end{mythm}
\end{tcolorbox}
To prove this, we will show that the Hessian satifies \textit{Sylvester's criterion}, which states that a Hermitian matrix is PD if all of its leading principal minors (LPMs) are positive, i.e. if the determinant of all its square top-left submatrices is positive \cite{horn2012matrix}. Recall that an $n\times n$ square matrix $\matr{A}$ has $n$ LPMs $\matr{A}_h$ of size $h\times h$ for $h = 1, \dots, n$. For a Hermitian matrix, showing that the determinant of all its LPMs is positive is a necessary and sufficient condition to determine whether the matrix is PD ($\matr{A} \succ 0$), and this result can be generalised to block matrices.

We now show that the activity Hessian of arbitrary DLNs (Eq.~\ref{ch5:eq:activity-hessian}) satisfies Sylvester's criterion. We drop the Hessian subscript $\matr{H}$ for brevity of notation. The proof technique lies in a Laplace or cofactor expansion of the LPMs along the last row. This has an intuitive interpretation in that it starts by proving that the inference landscape of one-hidden-layer PCNs is (strictly) convex, and then proceeds by induction to show that adding layers does not change the result.

The activity Hessian has $NH$ LPMs of size $N\ell \times N\ell$ for $\ell = 1, \dots, H$. Let $[\matr{H}]_\ell$ denote the $\ell$th LPM of $\matr{H}$, $\Delta_\ell = |[\matr{H}]_\ell|$ its determinant, and $\matr{D}_\ell$ and $\matr{O}_\ell$ the $\ell$th diagonal and off-diagonal blocks of $\matr{H}$, respectively. Now note that $\matr{H}$ is a block tridiagonal symmetric matrix, as can be clearly seen from Eq.~\ref{eq15}. There is a known two-term recurrence relation that can be used to calculate the determinant of such matrices through their LPMs \cite{salkuyeh2006comments}
\begin{align}
    \Delta_\ell = |\matr{D}_\ell|\Delta_{\ell-1} - |\matr{O}_{\ell-1}|^2 \Delta_{\ell-2}, \quad \ell = 2, \dots, H
    \label{eq17}
\end{align}
with $\Delta_0 = 1$ and $\Delta_1 = |\matr{D}_1|$. The first LPM is clearly PD and so its determinant is positive, $\matr{D}_1 = \matr{I} + a_2^2\matr{W}_2^T\matr{W}_2 \succ 0 \implies \Delta_1 > 0$, showing that the inference landscape of one-hidden-layer linear PCNs is strictly convex. For $\ell=2$, the first term of the recursion (Eq.~\ref{eq17}) is positive, since $|\matr{D}_2| = |\matr{I} + a_3^2\matr{W}_3^T\matr{W}_3|>0$, and $\Delta_1 >0$ as we just saw. The second term is negative, but it is strictly less than the positive term, $|a_2\matr{W}_2|^2 < |\matr{I} + a_3^2\matr{W}_3^T\matr{W}_3| |\matr{I} + a_2^2\matr{W}_2^T\matr{W}_2|$ and so $\Delta_2>0$. Hence, the activity landscape of 2-hidden-layer linear PCNs remains convex. The same holds for three hidden layers where $|\matr{O}_2|\Delta_1 < |\matr{D}_3|\Delta_2 \implies \Delta_3>0$. 

We can keep iterating this argument, showing by induction that the inference landscape is (strictly) convex for arbitrary DLNs. More formally, the positive term of the recurrence relation is always strictly greater than the negative term,
\begin{align}
    |\matr{D}_\ell|\Delta_{\ell-1} &> 0 \\
    |\matr{D}_\ell|\Delta_{\ell-1} &> |\matr{O}_{\ell-1}|^2 \Delta_{\ell-2}
\end{align}
and so $\Delta_\ell>0$ and $\matr{H} \succ 0$ for all $\ell$. Convexity holds for the unsupervised case, where the activity Hessian is now positive \textit{semidefinite} since the term $a_1^2\matr{W}_1^T\matr{W}_1$ is introduced (see \S \ref{ch5:activity-grad-hess}). The result can also be extended to any other linear layer transformation $\matr{B}_\ell$ including ResNets where $\matr{B}_\ell = \matr{I} + \matr{W}_\ell$.

\subsection{Random matrix theory of the activity Hessian} 
\label{ch5:rand-matrix}
Here we analyse the Hessian of the energy with respect to the activities of DLNs (Eq.~\ref{ch5:eq:activity-hessian}) using random matrix theory (RMT). This analysis follows a line of work using RMT to study the Hessian of neural networks, specifically the Hessian of the loss with respect to the parameters \cite{choromanska2015loss, pennington2017geometry, granziol2020beyond, liao2021hessian, baskerville2022universal}. We note that the structure of the activity Hessian is much simpler than the weight or parameter Hessian, in that for linear networks the former is positive definite (Theorem~\ref{ch5:thmA1}, \S \ref{ch5:pos-def}), while for the latter this is only true for one hidden layer as we saw in the previous chapter.
\begin{wrapfigure}{r}{0.5\textwidth}
    \vskip 0.1in
    \begin{minipage}{\linewidth}
        \centering
        \includegraphics[width=0.9\textwidth]{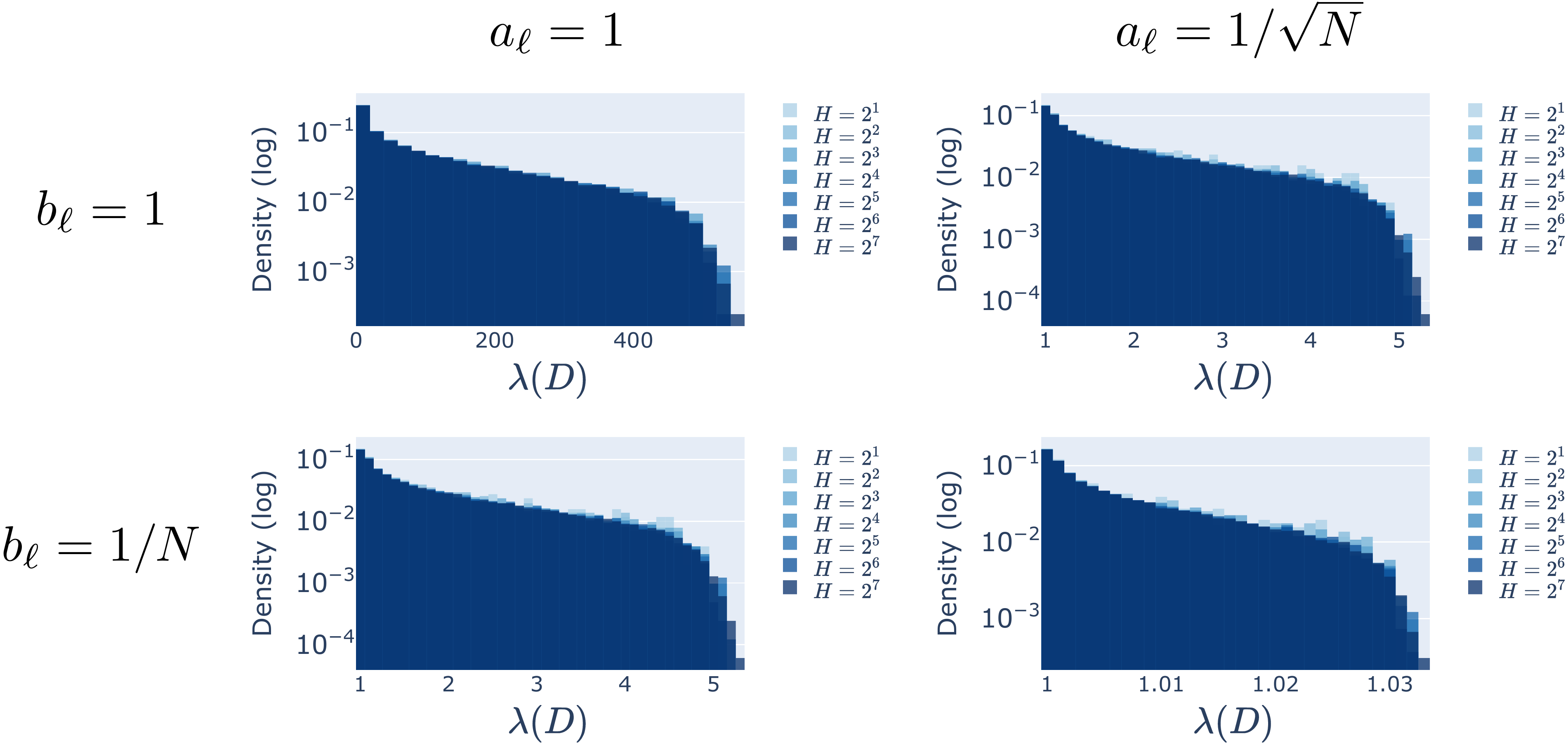}
        \caption{\textbf{Empirical eigenspectra of $\matr{D}$ at initialisation, holding the network width constant ($N=128$) and varying the depth $H$.} $a_\ell$ indicates the premultiplier at each network layer (Eq.~\ref{ch5:eq:pc-energy}), while $b_\ell$ is the variance of Gaussian initialisation, with $a_\ell=1$ and $b_\ell =1/N$ corresponding to the ``standard parameterisation '' (SP).}
        \label{ch5:fig:D_eigens_N_slice}
        \vspace{0.5cm}
        \includegraphics[width=0.9\textwidth]{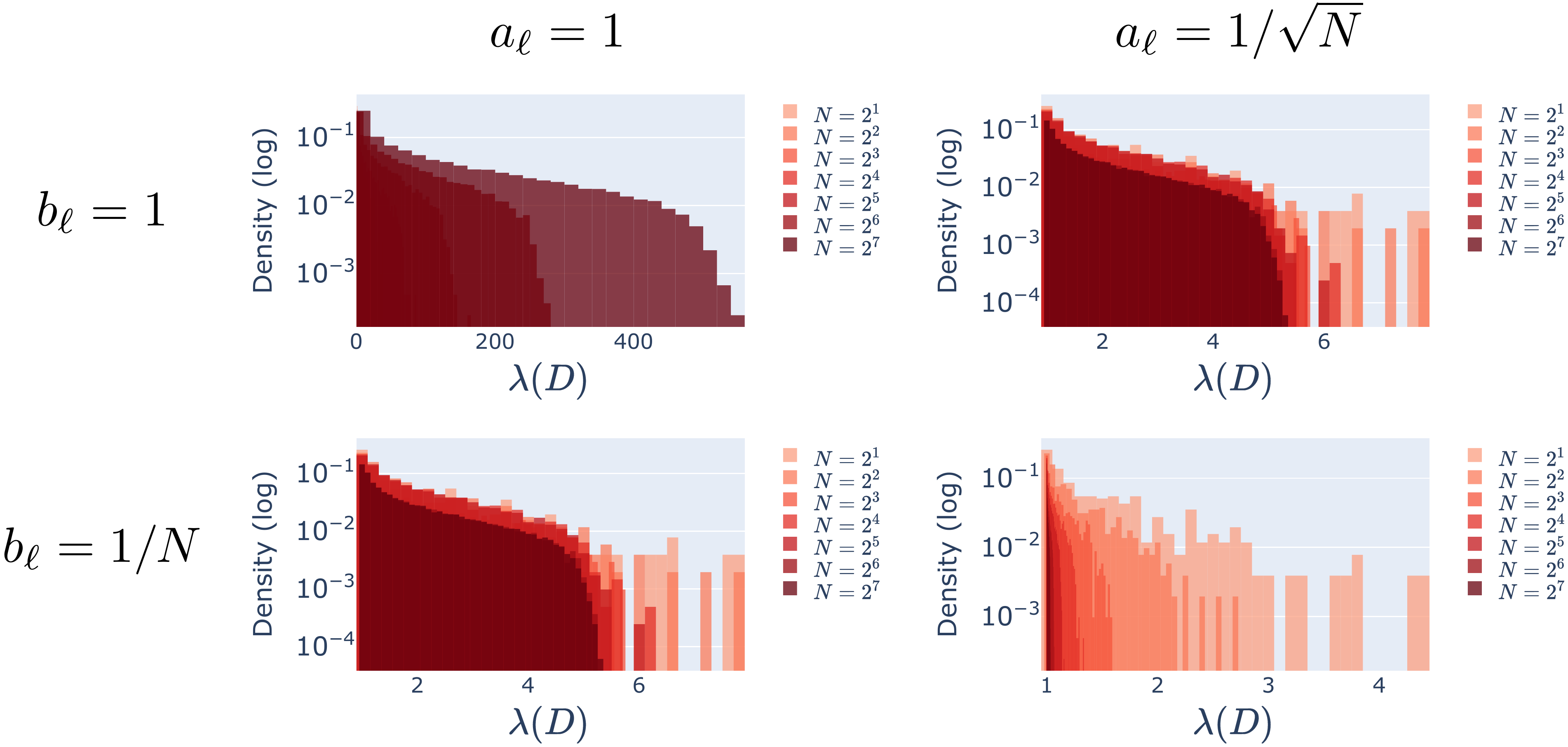}
        \caption{\textbf{Empirical eigenspectra of $\matr{D}$ at initialisation, holding the network depth constant ($H=128$) and varying the width $N$.}}
        \label{ch5:fig:D_eigens_H_slice}
    \end{minipage}
    %\vskip -0.25in
\end{wrapfigure}

In what follows, we recall from \S \ref{ch5:pcns} that the PC energy (Eq.~\ref{ch5:eq:pc-energy}) has layer-wise scalings $a_\ell$ for all $\ell$, and the weights are assumed to be drawn from a zero-mean Gaussian $(\matr{W}_\ell)_{ij} \sim \mathcal{N}(0, b_\ell)$ with variance set by $b_\ell$.

\paragraph{Hessian decomposition.} The activity Hessian (Eq.~\ref{ch5:eq:activity-hessian}) is a challenging matrix to study theoretically as its entries are not i.i.d. even at initialization due to the off-diagonal couplings between layers. However, we can decompose the matrix into its diagonal and off-diagonal components:
\begin{align}
    \matr{H}_{\mathbf{z}} = \matr{D} + \matr{O}
\end{align}
with $\matr{D} \coloneqq \diag(\matr{I} + a_2^2\matr{W}_2^T \matr{W}_2, \dots, \matr{I} + a_L^2\matr{W}_L^T \matr{W}_L)$ and $\matr{O} \coloneqq \offdiag(-a_2\matr{W}_2, \dots, -a_{L-1}\matr{W}_{L-1})$, where the off-diagonal part can be seen as a perturbation. Since these matrices are on their own i.i.d. at initialisation, we can use standard RMT results to analyse their respective eigenvalue distributions in the regime of large width $N$ and depth $H$ we are interested in. We will then use these results to gain some qualitative insights into the overall spectrum of $\matr{H}_{\mathbf{z}}$.

\paragraph{Analysis of $\matr{D}$.} As a block diagonal matrix, the eigenvales of $\matr{D}$ are given by those of its blocks $\matr{D}_\ell = \matr{I} + a_{\ell+1}^2\matr{W}_{\ell+1}^T \matr{W}_{\ell+1} \in \mathbb{R}^{N\times N}$ for $\ell = 1, \dots, H$. Note that the size of each block depends only on the network width $N$. It is easy to see that each block is a positively shifted Wishart matrix. As $N \rightarrow \infty$, the eigenspectrum of such matrices converges to the well-known Marčhenko-Pastur (MP) distribution \cite{marchenko1967distribution} if properly normalised such that $a_{\ell+1}^2\matr{W}_{\ell+1}^T \matr{W}_{\ell+1}\sim \mathcal{O}(1/N)$. 
\begin{wrapfigure}{r}{0.5\textwidth}
    \vskip 0.1in
    \begin{minipage}{\linewidth}
        \centering
        \includegraphics[width=0.9\textwidth]{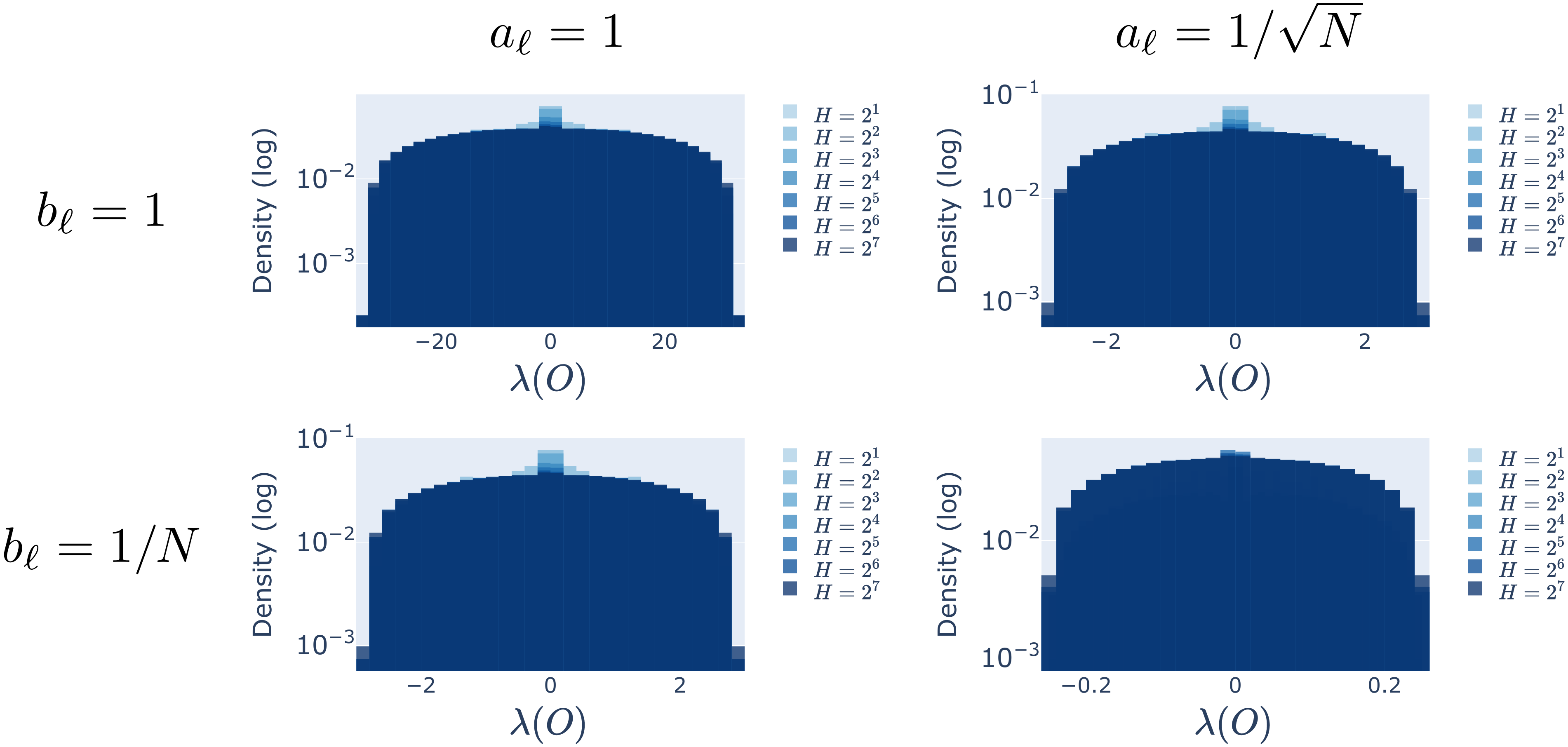}
        \caption{\textbf{Empirical eigenspectra of $\matr{O}$ at initialisation, holding the network width constant ($N=128$) and varying the depth $H$.}}
        \label{ch5:fig:O_eigens_N_slice}
        \vspace{0.5cm}
        \includegraphics[width=0.9\textwidth]{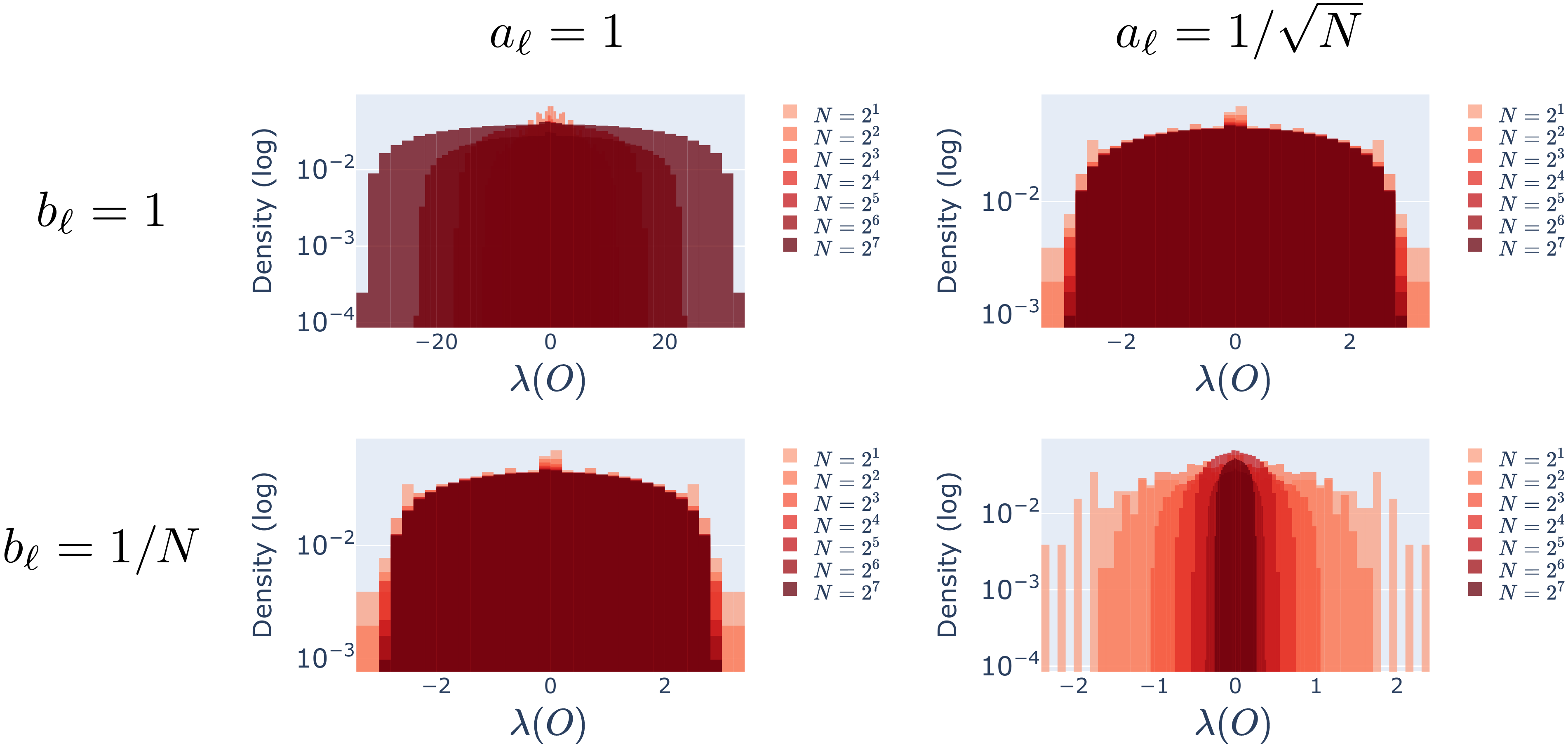}
        \caption{\textbf{Empirical eigenspectra of $\matr{O}$ at initialisation, holding the network depth constant ($H=128$) and varying the width $N$.}}
        \label{ch5:fig:O_eigens_H_slice}
    \end{minipage}
\end{wrapfigure}

As shown in Figs.~\ref{ch5:fig:D_eigens_N_slice}-\ref{ch5:fig:D_eigens_H_slice}, this normalisation can be achieved in two distinct but equivalent ways: (i) by initialising from a standard Gaussian with $b_\ell=1$ and setting the layer scaling to $a_\ell=1/\sqrt{N}$, or (ii) by setting $a_\ell=1$ and $b_\ell=1/N$ as done by standard initialisations \cite{lecun2002efficient, glorot2010understanding, he2015delving}. In either case, in the infinite-width limit the eigenvalues of each diagonal block will converge to a unit-shifted MP density with extremes 
\begin{align}
    \lim_{N \to \infty} \quad \lambda_{\pm}(\matr{D}_\ell) &=  1 + (1 \pm \sqrt{N/N})^2  \\
    &= \{1, 5\}.
\end{align}
While the spectrum of $\matr{D}$ will be a combination of these independent MP densities, its extremes will be the same of $\matr{D}_\ell$ since all of the blocks are i.i.d. and grow at the same rate as $N \rightarrow \infty$. This is empirically verified in Figs.~\ref{ch5:fig:D_eigens_N_slice}-\ref{ch5:fig:D_eigens_H_slice}, which also confirm that the spectrum of $\matr{D}$ is only affected by the width and not the depth.

\paragraph{Analysis of $\matr{O}$.} The off-diagonal component of the Hessian $\matr{O}$ is a sparse Wigner matrix whose size depends on both the width and the depth and so the correct limit should take both $N, H \rightarrow \infty$ at some constant ratio. Note that the sparsity of $\matr{O}$ grows much faster with the depth. Because sparse Wigner matrices are poorly understood and still an active area of research \cite{van2017structured}, we make the simplifying assumption that $\matr{O}$ is dense. 

If properly normalised as above, we know that in the limit the eigenspectrum of dense Wigner matrices converges the classical Wigner semicircle distribution \cite{wigner1993characteristic} with extremes
\begin{align}
    \lim_{H/N \to \infty} \quad \lambda_{\pm}(\matr{O}) \pm 2.
\end{align}
We find that the empirical eigenspectrum of $\matr{O}$ is slightly broader than the semicircle and, as expected, is affected by both the width and the depth (Figs.~\ref{ch5:fig:O_eigens_N_slice}-\ref{ch5:fig:O_eigens_H_slice}).

\begin{figure}[htbp]
    \centering
    \begin{minipage}{0.48\textwidth}
        \centering
        \includegraphics[width=\textwidth]{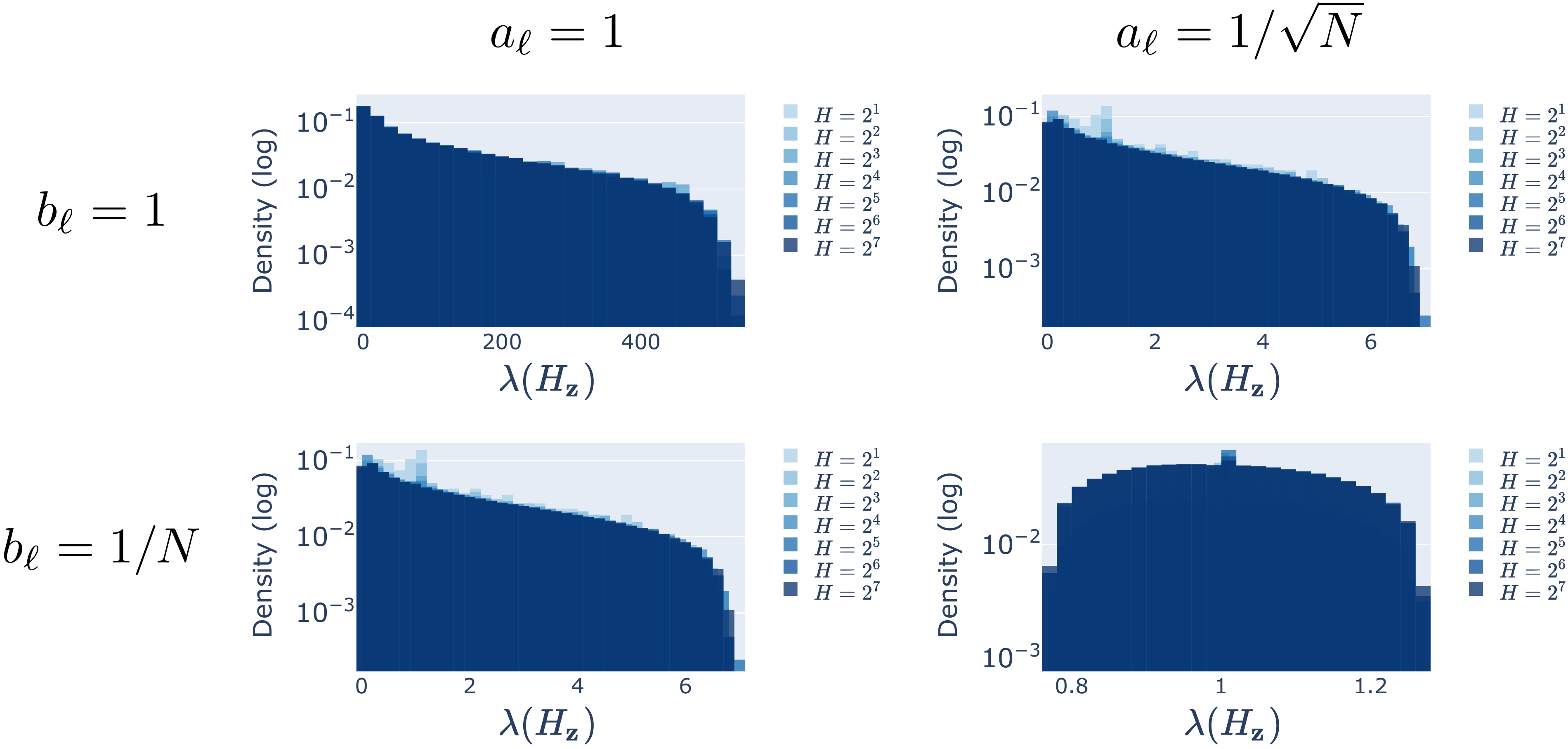}
        \caption{\textbf{Empirical eigenspectra of $\matr{H}$ at initialisation, holding the network width constant ($N=128$) and varying the depth $H$.}}
        \label{ch5:fig:H_eigens_N_slice}
    \end{minipage}
    \hfill % Add horizontal fill between the minipages to push them apart
    \begin{minipage}{0.48\textwidth}
        \centering
        \includegraphics[width=\textwidth]{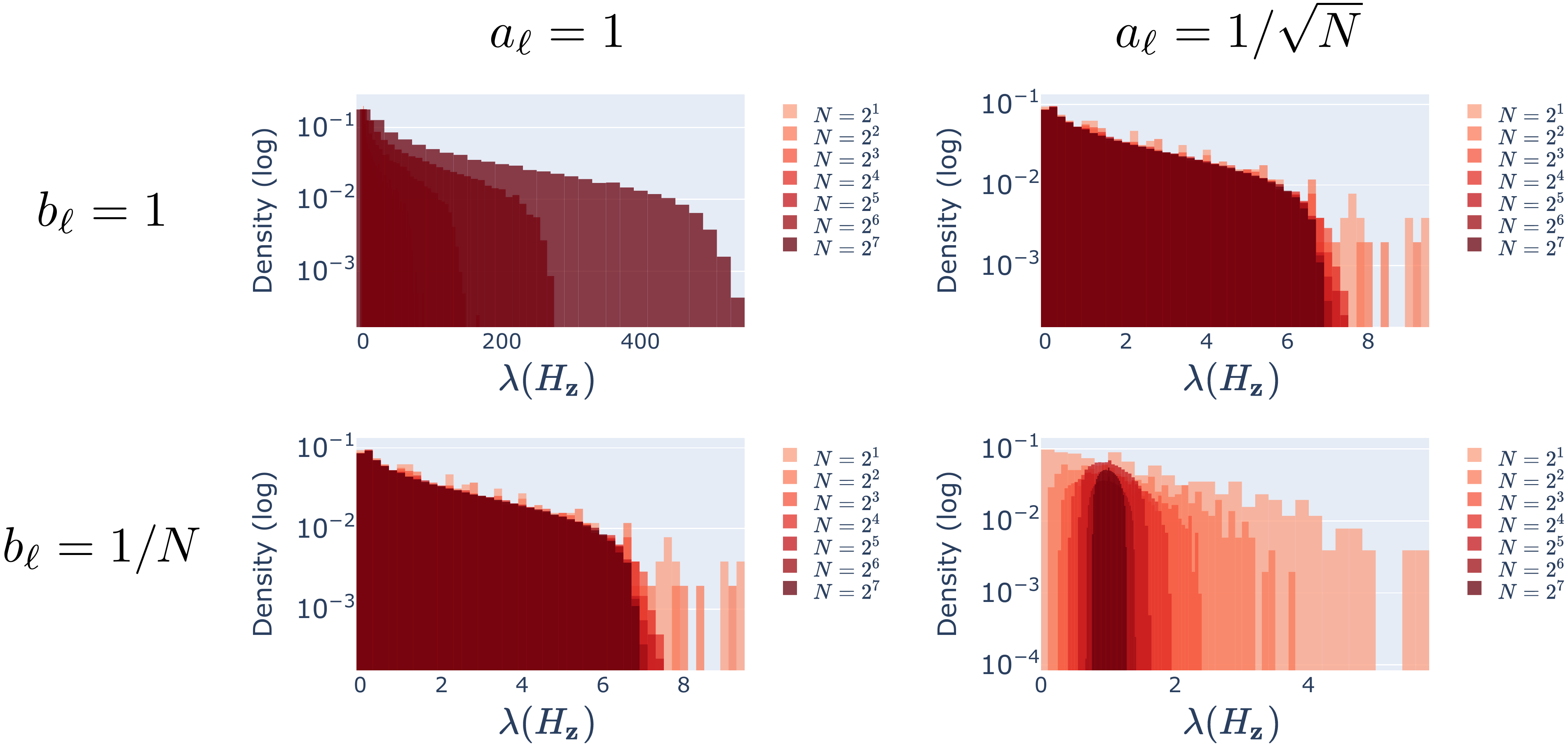}
        \caption{\textbf{Empirical eigenspectra of $\matr{H}$ at initialisation, holding the network depth constant ($H=128$) and varying the width $N$.}}
        \label{ch5:fig:H_eigens_H_slice}
    \end{minipage}
\end{figure}

\begin{figure}[h!]
    \vskip 0.2in
    \begin{center}
        \centerline{\includegraphics[width=0.9\textwidth]{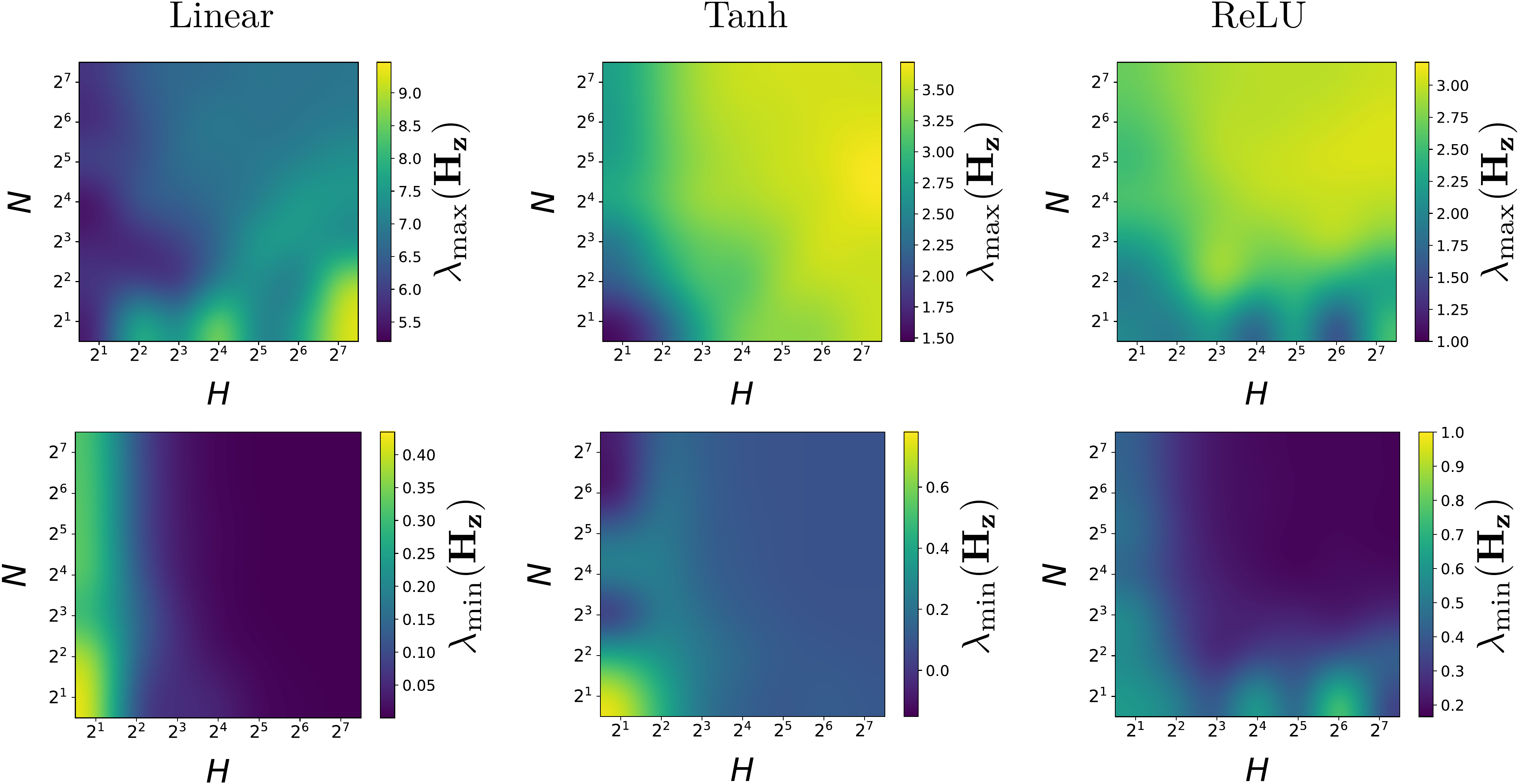}}
        \caption{\textbf{Maximum and minimum eigenvalues of $\matr{H}_{\mathbf{z}}$ at initialisation as a function of network width $N$ and depth $L$.}}
        \label{ch5:fig:max-min-eigens}
    \end{center}
    \vskip -0.25in
\end{figure}

\paragraph{Analysis of $\matr{H}_{\mathbf{z}}$.} Given the above asymptotic results on $\matr{D}$ and $\matr{O}$, we can use Weyl's inequalities \cite{weyl1912asymptotische} to lower and upper bound the minimum and maximum eigenvalues (and so the condition number) of the overall Hessian at initialisation: $\lambda_{\text{max}}(\matr{D} + \matr{O}) \leq \lambda_{\text{max}}(\matr{D}) + \lambda_{\text{max}}(\matr{O})$ and $\lambda_{\text{min}}(\matr{D} + \matr{O}) \geq \lambda_{\text{min}}(\matr{D}) + \lambda_{\text{min}}(\matr{O})$. The upper bound ($\tilde{\lambda}_{\text{max}}=7$) appears tight, as shown in Figs.~\ref{ch5:fig:H_eigens_N_slice}-\ref{ch5:fig:max-min-eigens}. However, the lower bound predicts a negative minimum eigenvalue ($\tilde{\lambda}_{\text{min}}=-1$), which is not possible since the Hessian is positive definite as we proved in \S \ref{ch5:pos-def}.

Nevertheless, we can still gain some insights into the interaction between $\matr{D}$ and $\matr{O}$ by looking at the empirical eigenspectrum of $\matr{H}_{\mathbf{z}}$. In particular, we observe that the maximum and especially the minimum eigenvalue of the Hessian scale with the network depth (Figs.~\ref{ch5:fig:max-min-eigens} \& \ref{ch5:fig:resnets-cond-nums-init}), thus driving the growth of the condition number.

\subsection{Activity Hessian of linear ResNets}
\label{ch5:resnets-hessian}
Here we derive the activity Hessian for linear ResNets \cite{he2016deep}, extending the derivation in \S \ref{ch5:activity-grad-hess} for DLNs. Following the Depth-$\mu$P parameterisation \cite{yang2023tensorinfdepth, bordelon2023depthwise}, we consider ResNets with identity skip connections at every layer except from the input and to the output. The PC energy for such ResNets is given by
\begin{equation}
    \mathcal{F}_{\text{1-skip}} =  \frac{1}{2}||\boldsymbol{\epsilon}_L||^2 + \frac{1}{2}||\boldsymbol{\epsilon}_1||^2 + \sum_{\ell=2}^H \frac{1}{2}||\mathbf{z}_\ell - a_\ell\matr{W}_\ell \mathbf{z}_{\ell-1} - \underbrace{\mathbf{z}_{\ell-1}}_{\text{1-skip}}||^2,
    \label{ch5:eq:resnet-energy}
\end{equation}
\begin{center}
    \begin{tikzpicture}[x=1.5cm,y=0.1cm]
      \foreach \N [count=\lay,remember={\N as \Nprev (initially 0);}]
                   in {1,1,1,1,1,1}{ % loop over layers
        \foreach \i [evaluate={\y=\N/2-\i; \x=\lay; \prev=int(\lay-1);}]
                     in {1,...,\N}{ % loop over nodes
          \node[mynode] (N\lay-\i) at (\x,\y) {};
          \ifnum\Nprev>0 % connect to previous layer 
            \foreach \j in {1,...,\Nprev}{ % loop over nodes in previous layer
              \draw[->] (N\prev-\j) -- (N\lay-\i);
            }
          \fi
        }
      }
      % Add skip connections
      \draw[->] (N2-1) to[bend left=60] (N3-1); % From 2nd to 3rd hidden unit
      \draw[->] (N3-1) to[bend left=60] (N4-1); % From 3rd to 4th hidden unit
      \draw[->] (N4-1) to[bend left=60] (N5-1); % From 4th to 5th hidden unit
    \end{tikzpicture}
\end{center}
where recall that $\boldsymbol{\epsilon}_\ell = \mathbf{z}_\ell - a_\ell\matr{W}_\ell \mathbf{z}_{\ell-1}$ and $\mathbf{z}_0 \coloneqq \mathbf{x}$, $\mathbf{z}_L \coloneqq \mathbf{y}$. We refer to this model as ``1-skip'' since the residual is added to every layer. Its activity Hessian is given by
\begin{align} 
    \matr{H}_{\mathbf{z}}^{\text{1-skip}} \coloneqq \frac{\partial^2 \mathcal{F}_{\text{1-skip}}}{\partial \mathbf{z}_\ell \partial \mathbf{z}_k} =
    \begin{cases} 
    2\matr{I} + a_{\ell+1}^2\matr{W}_{\ell+1}^T \matr{W}_{\ell+1} + a_{\ell+1}(\matr{W}_{\ell+1}^T + \matr{W}_{\ell+1}), & \ell = k \neq H \\
    \matr{I} + a_{\ell+1}^2\matr{W}_{\ell+1}^T \matr{W}_{\ell+1}, & \ell = k = H \\
    -a_{k+1}\matr{W}_{k+1} - \matr{I}, & \ell - k = 1 \\
    -a_{\ell+1}\matr{W}_{\ell+1}^T - \matr{I}, & \ell - k = -1 \\
    \mathbf{0}, & \text{else}
    \end{cases}.
    \label{ch5:eq:resnets-activity-hessian}
\end{align}
We find that this Hessian is much more ill-conditioned (Fig.~\ref{ch5:fig:resnets-cond-nums-init}) than that of networks without skips (Fig.~\ref{ch5:fig:sp-cond-nums-init}), across different parameterisations (Fig.~\ref{ch5:fig:des-trade-off}). We note that one can extend these results to $n$-skip linear ResNets with energy
\begin{equation}
    \mathcal{F}_{n\text{-skip}} = \frac{1}{2}||\boldsymbol{\epsilon}_L||^2 + \sum_{\ell=1}^n \frac{1}{2}||\boldsymbol{\epsilon}_\ell||^2 + \sum_{\ell=n+1}^H \frac{1}{2}||\mathbf{z}_\ell - a_\ell\matr{W}_\ell \mathbf{z}_{\ell-1} - \underbrace{\mathbf{z}_{\ell-n}}_{n\text{-skip}}||^2
\end{equation}
or indeed arbitrary computational graphs \cite{salvatori2022learning}. It could be interesting to investigate whether there exist architectures with better conditioning of the inference landscape that do not sacrifice the stability of the forward pass (see \S \ref{ch5:desiderata}, Fig. \ref{ch5:fig:des-trade-off}).

\subsection{Extension to other energy-based algorithms} 
\label{ch5:other-algos}
Here we include a preliminary investigation of the inference dynamics of other energy-based local learning algorithms. As an example, we consider equilibrium propagation (EP) \cite{scellier2017equilibrium, zucchet2022beyond}, whose energy for a DLN is given by
\begin{equation}
    E = \frac{1}{2}||\mathbf{z}_\ell||^2 - \sum_{\ell=1}^L \mathbf{z}_\ell^T \matr{W}_\ell \mathbf{z}_{\ell-1} + \frac{\beta}{2}||\mathbf{y}-\mathbf{z}_L||^2,
    \label{ch5:eq:equilib-prop-energy}
\end{equation}
where $\mathbf{z}_0 \coloneqq \mathbf{x}$ for supervised learning (as for PC), and it is also standard to include an $\ell^2$ regulariser on the activities. Unlike PC, EP has two inference phases: a \textit{free} phase where the output layer $\mathbf{z}_L$ is free to vary like any other hidden layer with $\beta=0$; and a \textit{clamped} or  \textit{nudged} phase where the output is fixed to some target $\mathbf{y}$ with $\beta>0$. The activity gradient and Hessian of the EP energy (Eq.~\ref{ch5:eq:equilib-prop-energy}) are given by
\begin{align}
    \frac{\partial E}{\partial \mathbf{z}_\ell} =
    \begin{cases} 
    \mathbf{z}_\ell - \matr{W}_\ell \mathbf{z}_{\ell-1} - \mathbf{z}_{\ell+1}^T \matr{W}_{\ell+1}, & \ell \neq L \\
    \mathbf{z}_\ell - \matr{W}_\ell \mathbf{z}_{\ell-1} - \beta(\mathbf{y} - \mathbf{z}_\ell), & \ell = L
    \end{cases}
    \label{ch5:eq:equilib-prop-grad}
\end{align}
and
\begin{align}
    \matr{H}_{\mathbf{z}} \coloneqq \frac{\partial^2 E}{\partial \mathbf{z}_\ell \partial \mathbf{z}_k} =
    \begin{cases}
    \matr{I}, & \ell = k \neq L \\
    \matr{I} + \beta, & \ell = k = L \\
    - \matr{W}_{\ell+1}, & \ell - k = 1 \\
    - \matr{W}_{k+1}^T, & \ell - k = -1 \\
    \mathbf{0}, & \text{else}
    \end{cases}
    \label{ch5:eq:equilib-prop-hess}
\end{align}
where we abuse notation by denoting the Hessian in the same way as that of the PC energy. We observe that the off-diagonal blocks are equal to those of the PC activity Hessian (Eq.~\ref{ch5:eq:activity-hessian}). Similar to PC, one can also rewrite the EP activity gradient (Eq.~\ref{ch5:eq:equilib-prop-grad}) as a linear system
\begin{align}
    \nabla_\mathbf{z} E &= \underbrace{\begin{bmatrix} 
        \matr{I} & -\matr{W}_2^T & \mathbf{0} & \dots & \mathbf{0} \\ -\matr{W}_2 & \matr{I} & -\matr{W}_3^T & \dots & \mathbf{0} \\ \mathbf{0} & -\matr{W}_3 & \matr{I} & \ddots & \mathbf{0} \\ \vdots & \vdots & \ddots & \ddots & -\matr{W}_L^T \\ \mathbf{0} & \mathbf{0} & \mathbf{0} & -\matr{W}_L & \matr{I} + \beta
    \end{bmatrix}}_{\matr{H}_{\mathbf{z}}}
    \underbrace{\begin{bmatrix} 
        \mathbf{z}_1 \\
        \mathbf{z}_2 \\
        \vdots \\
        \mathbf{z}_{L-1} \\
        \mathbf{z}_L
    \end{bmatrix}}_{\mathbf{z}} - 
    \underbrace{\begin{bmatrix} 
        \matr{W}_1\mathbf{x} \\
        \mathbf{0} \\
        \vdots \\
        \mathbf{0} \\
        \beta \mathbf{y}
    \end{bmatrix}}_{\mathbf{b}}
\end{align}
with solution $\mathbf{z}^* = \matr{H}_{\mathbf{z}}^{-1} \mathbf{b}$. Interestingly, unlike for PC, the EP inference landscape is not necessarily convex, which can be easily seen for a shallow 2-layer scalar network where $\exists \lambda(\matr{H}_{\mathbf{z}}(w_2 > 1)) < 0$. This is always true without the activity regulariser, in which case the identity in each diagonal block vanishes.

\subsection{Limit convergence of $\mu$PC to BP (Thm. \ref{ch5:thm1})}
\label{ch5:limit-converge-proof}
Here we provide a simple proof of Theorem~\ref{ch5:thm1}. Consider a slight generalisation to linear ResNets (Eq.~\ref{ch5:eq:resnet-energy}) of the PC energy at the inference equilibrium derived in the previous chapter for DLNs (Eq.~\ref{ch4:eq:dln-equilib-energy}):
\begin{align}
    \mathcal{F}(\mathbf{z}^*) &= \frac{1}{2B} \sum_{i=1}^B  \mathbf{r}_i^T \matr{S}^{-1} \mathbf{r}_i, \label{ch5:eq:resnet-equilib-energy} \\ 
    \text{where} &\quad \matr{S} = \matr{I}_{d_y} + a_L^2 \matr{W}_L\matr{W}_L^T + \sum_{\ell=2}^H \left( a_L \matr{W}_L \prod_\ell^H \matr{I} + a_\ell \matr{W}_\ell \right) \left( a_L \matr{W}_L \prod_\ell^H \matr{I} + a_\ell \matr{W}_\ell \right)^T, \label{ch5:eq:resnet-equilib-energy-rescaling}
\end{align}
the residual error is $\mathbf{r}_i = \mathbf{y}_i - a_L \matr{W}_L \left( \prod_{\ell=2}^H \matr{I} + a_\ell \matr{W}_\ell \right) a_1 \matr{W}_1\mathbf{x}_i$, and $B$ is the batch or dataset size. Note that, as for non-residual DLNs, Eq.~\ref{ch5:eq:resnet-equilib-energy} is an MSE loss with a weight-dependent rescaling (Eq.~\ref{ch5:eq:resnet-equilib-energy-rescaling}). Now, we know that for Depth-$\mu$P the forward pass of this model has $\mathcal{O}_{N, H}(1)$ preactivations at initialisation and so the residual will also be of order 1. Note that, by contrast, for SP ($a_\ell = 1$ for all $\ell$ and $b_\ell = 1/N_{\ell-1}$) the preactivations explode with the depth (Fig.~\ref{ch5:fig:fwd-pass-stability-depth-params}). 

The key question, then, is what happens to the rescaling $\matr{S}$ in the limit of large depth $L$ and width $N$. Recall that for $\mu$PC, $a_L = 1/N$ and $a_\ell = 1/\sqrt{NL}$ for $\ell = 2, \dots, H$ (see Table~\ref{ch5:params-table}). Because the output weights factor in every term of the rescaling $\matr{S}$ except for the identity, these terms will all vanish at a $1/N$ rate as $N \rightarrow \infty$, i.e. $\matr{W}_L\matr{W}_L^T/N^2 \sim \mathcal{O}(1/N)$. The depth, on the other hand, scales the number of terms in $\matr{S}$. Therefore, the width will have to grow with the depth at some constant ratio $L/N$---which can be thought of as the aspect ratio of the network \cite{roberts2022principles}---to make the contribution of each term as small as possible. In the limit of this ratio $r \rightarrow 0$, the energy rescaling (Eq.~\ref{ch5:eq:resnet-equilib-energy-rescaling}) approaches the identity $\matr{S} = \matr{I}$, the equilibrated energy converges to the MSE $\mathcal{F}_{\mu \text{PC}}(\mathbf{z}^*, \boldsymbol{\theta}) = \mathcal{L}_{\mu \text{P}}(\boldsymbol{\theta})$, and so PC computes the same gradients as BP.
% Note that for a scalar PCN parameterised with Depth-$\mu$P, the energy reduces to
% \begin{align}
%     \mathcal{F}^* &= \frac{1}{2N} \sum_{i=1}^N \frac{\left( y_i - w_L \left( \prod_{\ell=2}^H 1 + \frac{1}{\sqrt{L}} w_\ell \right) w_1 x_i \right)^2}{1 + w_L^2 + \sum_{\ell=2}^H \left( w_L \prod_\ell^H 1 + \frac{1}{\sqrt{L}} w_\ell \right)^2}
% \end{align}

\section{Additional experiments} 
\label{ch5:add-exps}

\subsection{Ill-conditioning with training}
For the setting in Fig.~\ref{ch5:fig:sp-train-cond-nums-GD-mnist}, we also ran experiments with Adam as inference algorithm and ResNets with standard GD. All the results were tuned for the weight learning rate (see \S\ref{ch5:exp-details} for more details). We found that Adam led to more ill-conditioned inference landscapes associated with significantly lower and more unstable performance than GD (Figs.~\ref{ch5:fig:sp-train-cond-nums-GD-mnist} \& \ref{ch5:fig:sp-train-cond-nums-GD-fashion}).
\begin{figure}[H]
    \vskip 0.2in
    \begin{center}
        \centerline{\includegraphics[width=\textwidth]{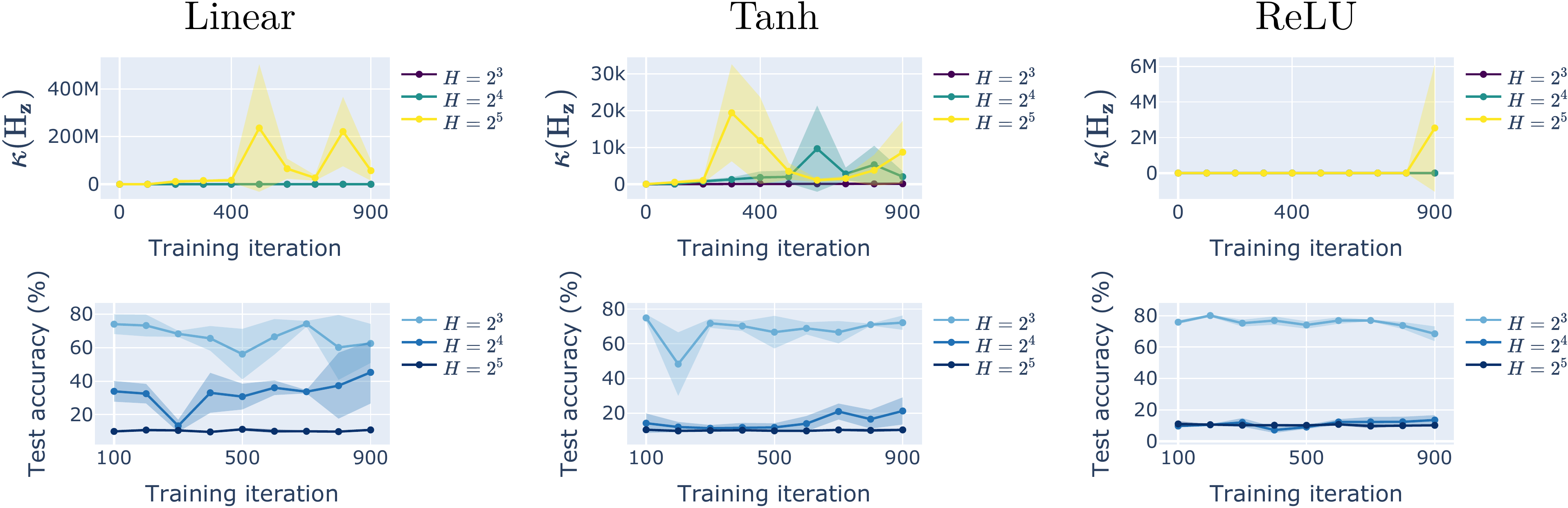}}
        \caption{\textbf{Same results as Fig.~\ref{ch5:fig:sp-train-cond-nums-GD-mnist} with Adam as inference algorithm (MNIST).}}
        \label{ch5:fig:sp-train-cond-nums-adam-mnist}
    \end{center}
    \vskip -0.25in
\end{figure}

\begin{figure}[H]
    \vskip 0.2in
    \begin{center}
        \centerline{\includegraphics[width=\textwidth]{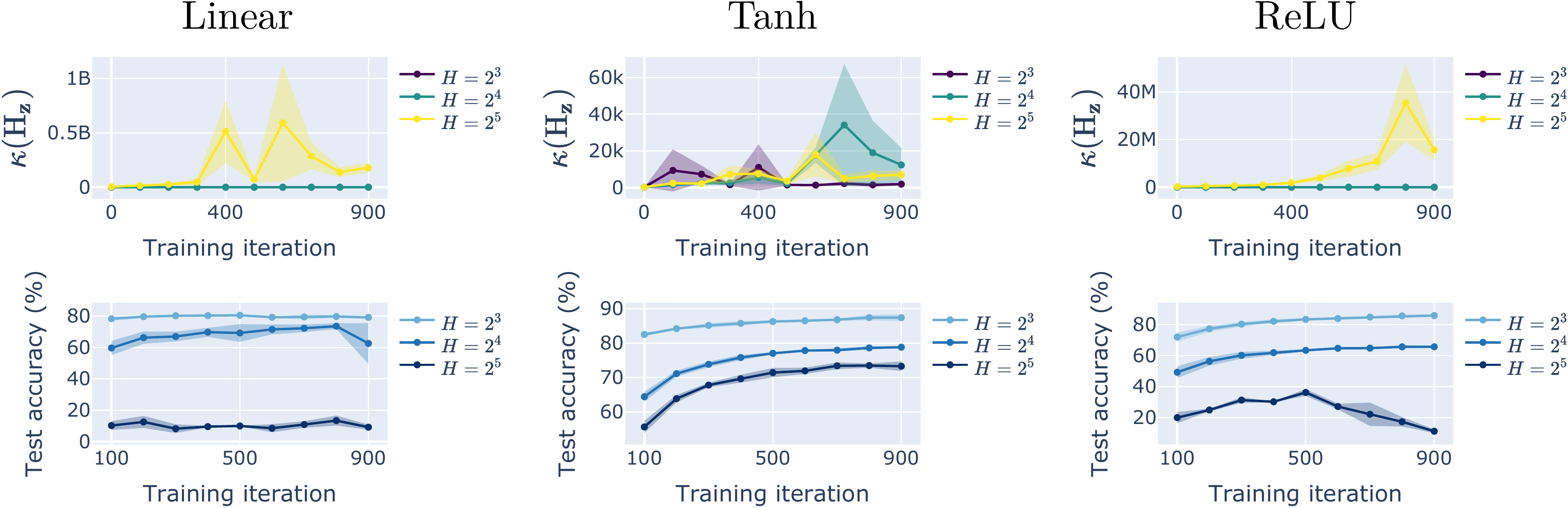}}
        \caption{\textbf{Same results as Fig.~\ref{ch5:fig:sp-train-cond-nums-GD-mnist} with skip connections (MNIST).}}
        \label{ch5:fig:sp-train-cond-nums-skips-mnist}
    \end{center}
    \vskip -0.25in
\end{figure}

Interestingly, while skip connections induced much more extreme ill-conditioning (Fig.~\ref{ch5:fig:resnets-cond-nums-init}), performance was equal to, and sometimes significantly better than, networks without skips (Figs.~\ref{ch5:fig:sp-train-cond-nums-skips-mnist} \& \ref{ch5:fig:sp-train-cond-nums-skips-fashion}), suggesting a complex relationship between trainability and the geometry of the inference landscape which we return to in \S \ref{ch5:infer-converge-sufficient}. 

\subsection{Activity initialisations}
\label{ch5:activity-inits}
\begin{figure}[H]
    \vskip 0.2in
    \begin{center}
        \centerline{\includegraphics[width=\textwidth]{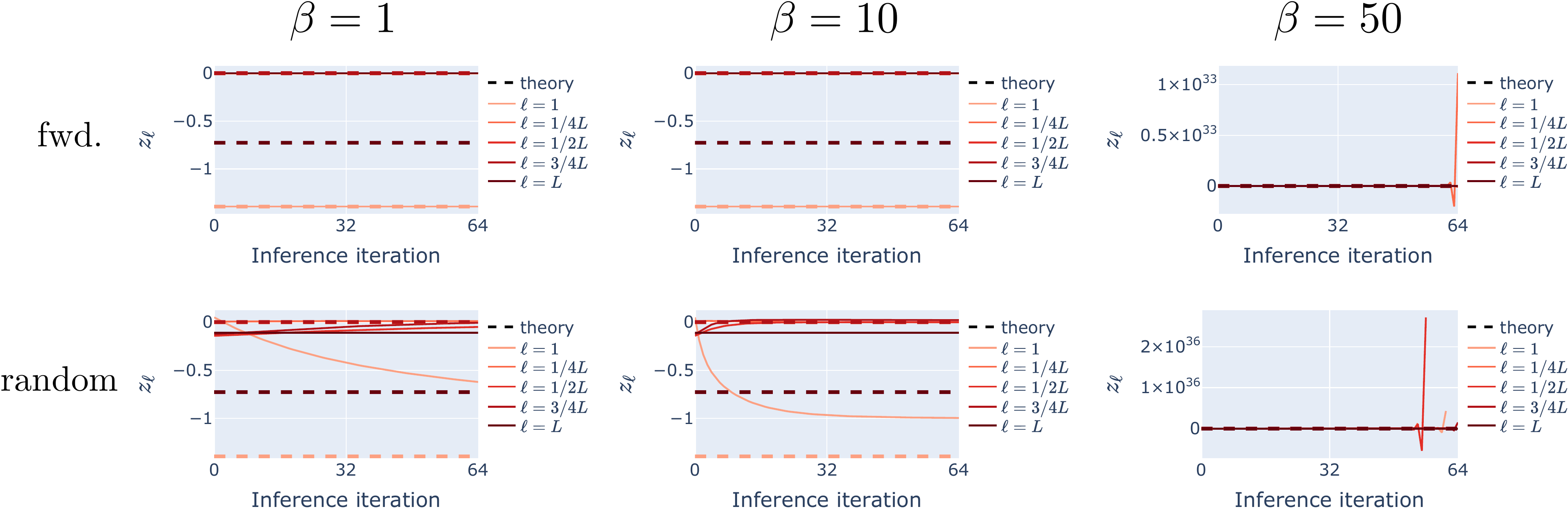}}
        \caption{\textbf{Ill-conditioning of the inference landscape prevents convergence to the analytical solution regardless of initialisation.} For different initialisations (forward and random) and activity learning rates $\beta$, we plot the activities of a 64-layer scalar PCN over inference at the start of training. The theoretical activities were computed using Eq.~\ref{ch5:eq:pc-infer-solution}. The task was a simple toy regression with $y = -x + \epsilon$ with $x \sim \mathcal{N}(1, 1)$ and $\epsilon \sim \mathcal{N}(0, 0.5)$. A standard Gaussian was used for random initialisation, $z_\ell \sim \mathcal{N}(0, 1)$. Results were similar across different random seeds.}
        \label{ch5:fig:sp-activity-inits}
    \end{center}
    \vskip -0.25in
\end{figure}

\begin{figure}[H]
    \vskip 0.2in
    \begin{center}
        \centerline{\includegraphics[width=\textwidth]{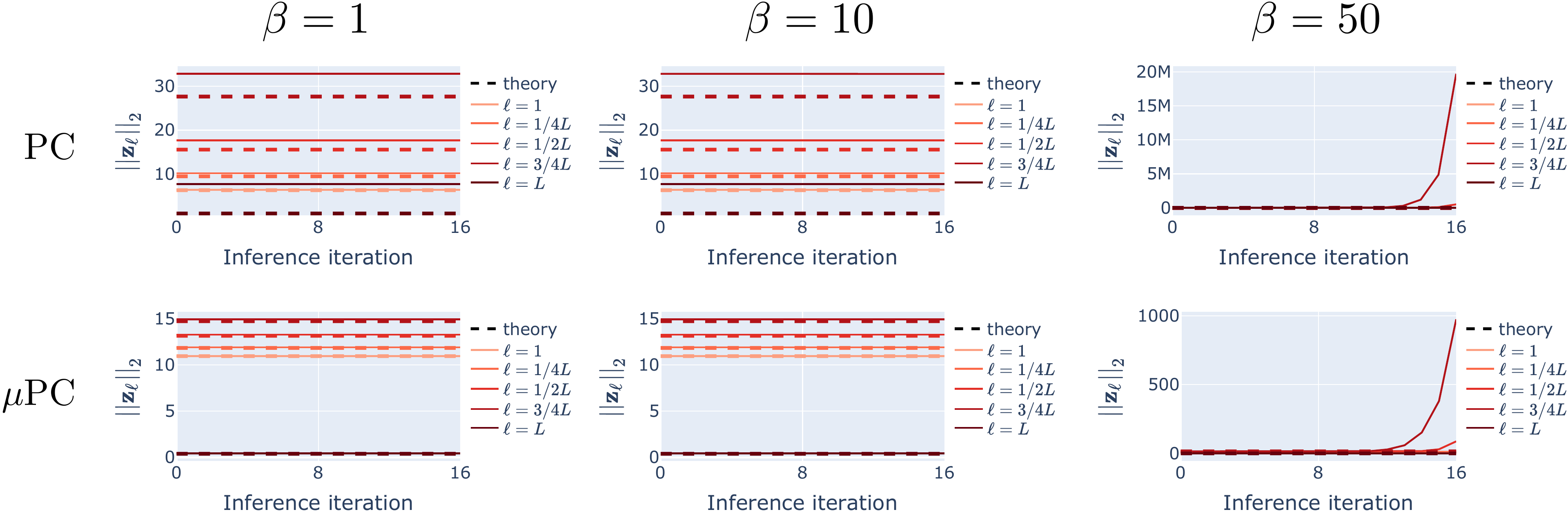}}
        \caption{\textbf{The forward pass of $\mu$PC seems to initialise the activities close to the analytical solution (Eq.~\ref{ch5:eq:pc-infer-solution}).} Similar to Fig.~\ref{ch5:fig:sp-activity-inits}, we plot the $\ell^2$ norm of the activities over inference of 16-layer linear PCNs ($N = 128$) at the start of training (MNIST). Again, results were similar across different random initialisations.}
        \label{ch5:fig:mupc-vs-pc-ffwd-lrs}
    \end{center}
    \vskip -0.25in
\end{figure}

Here we present some additional results on the initialisation of the activities of PCNs. All experiments used fully connected ResNets, GD as activity optimiser, and as many inference steps as the number of hidden layers. For intuition, we start with linear scalar PCNs or chains. First, we verify that the ill-conditioning of the inference landscape (\S \ref{ch5:ill-cond-infer}) causes the activities to barely move during inference, and increasing the activity learning rate leads to divergence for both forward and random initialisation (Fig.~\ref{ch5:fig:sp-activity-inits}). Similar results are observed for $\mu$PC (see Fig.~\ref{ch5:fig:mupc-activity-inits}).
For wide linear PCNs with forward initialisation, we find similar results except that $\mu$PC seems to initialise the activities close to the analytical solution (Fig.~\ref{ch5:fig:mupc-vs-pc-ffwd-lrs}). The same pattern of results is observed for nonlinear networks (Fig.~\ref{ch5:fig:mupc-vs-pc-ffwd-relu-lrs}), although note that in this case we do have an analytical solution. These results might suggest that one does not need to perform many inference steps to achieve good performance with $\mu$PC. However, we found that one inference step led to worse performance (including as a function of depth) (Figs.~\ref{ch5:fig:mupc-one-step-mnist} \& \ref{ch5:fig:mupc-one-step-fashion}) compared to as many steps as number of hidden layers (Figs.~\ref{ch5:fig:mupc-vs-pc-mnist-accs-all-act-fns} \& \ref{ch5:fig:mupc-fashion-15-epochs}).

\subsection{Activity decay}
\label{ch5:activity-decay}
\begin{figure}[H]
    \vskip 0.2in
    \begin{center}
        \centerline{\includegraphics[width=0.7\textwidth]{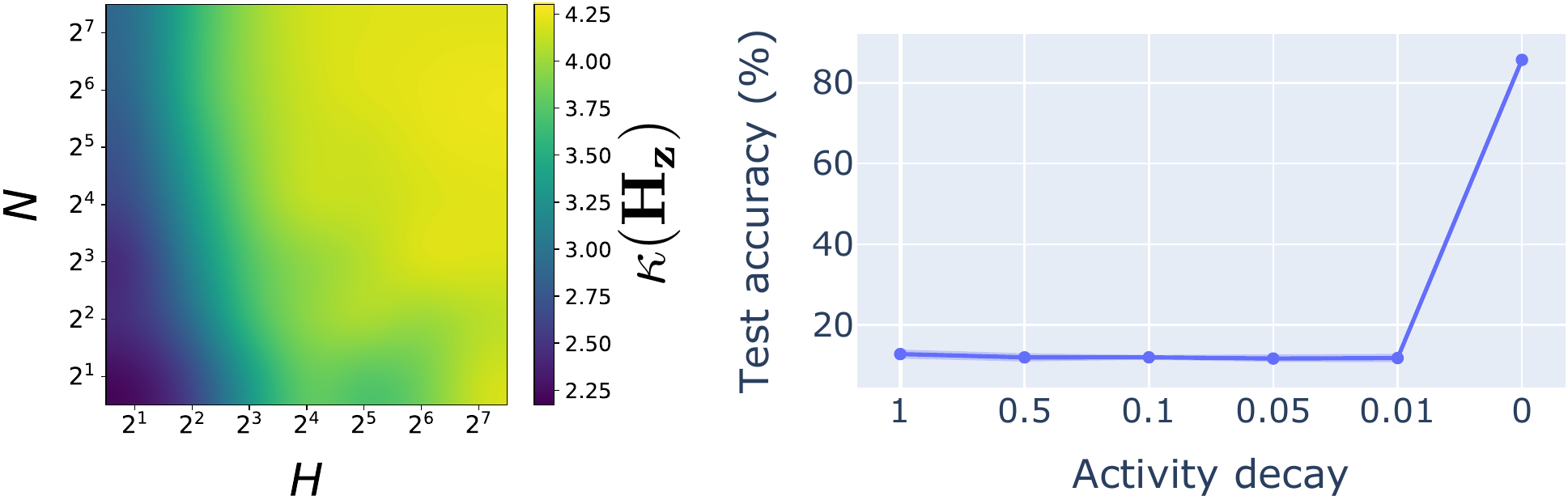}}
        \caption{\textbf{Activity decay induces well-conditioned inference at the cost of performance.} \textit{Left}: Same plot as Fig.~\ref{ch5:fig:sp-cond-nums-init} with an added activity regulariser $\frac{\alpha}{2}||\mathbf{z}_\ell||^2$ with $\alpha=1$. \textit{Right}: Maximum test accuracy on MNIST achieved by a linear PCN with $N = 128$ and $H=8$ over activity regularisers of varying strength $\alpha$. Solid lines and (barely visible) shaded regions indicate the mean and standard deviation across 3 random seeds, respectively.}
        \label{ch5:fig:activity-decay-exp}
    \end{center}
    \vskip -0.25in
\end{figure}

In \S \ref{ch5:desiderata}, we discussed how it seems impossible to achieve good conditioning of the inference landscape without making the forward pass unstable (e.g. by zeroing out the weights). We identified one way of inducing relative well-conditionness at initialisation without affecting the forward pass, namely adding an $\ell^2$ norm regulariser on the activities $\frac{\alpha}{2}\sum_\ell^H||\mathbf{z}_\ell||^2$ with $\alpha = 1$. This effectively induces a unit shift in the Hessian spectrum and bounds the minimum eigenvalue at one rather than zero (see \S \ref{ch5:rand-matrix}). However, we find that PCNs with \textit{any degree of activity regularisation} $\alpha$ are untrainable (Fig.~\ref{ch5:fig:activity-decay-exp}).

\subsection{Orthogonal initialisation}
\label{ch5:orthog-init}
\begin{figure}[H]
    \vskip 0.2in
    \begin{center}
        \centerline{\includegraphics[width=\textwidth]{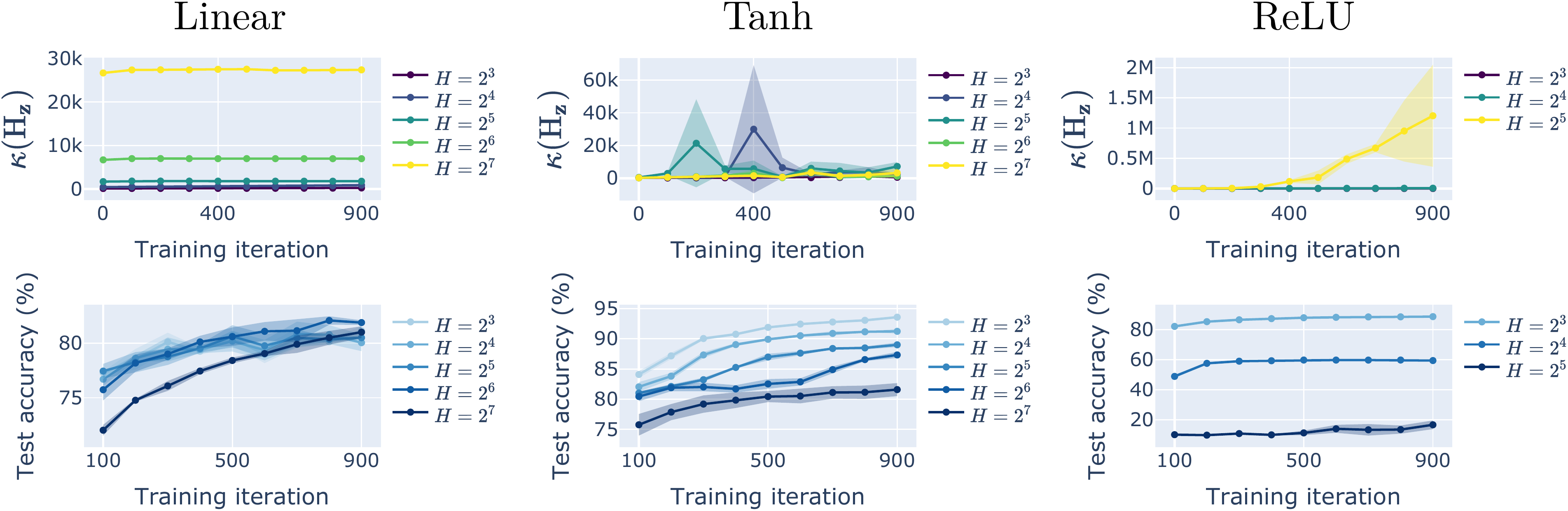}}
        \caption{\textbf{Test accuracies in Fig.~\ref{ch5:fig:mupc-vs-pc-mnist-accs} for orthogonal initialisation.} Note that performance is expected to drop for ReLU networks which cannot have stable forward passes with orthogonal weights (Fig.~\ref{ch5:fig:fwd-pass-stability-depth-params}). We also plot the condition number of the activity Hessian over training.}
        \label{ch5:fig:orthog-init-mnist}
    \end{center}
    \vskip -0.25in
\end{figure}

As mentioned in \S\ref{ch5:experiments}, in addition to $\mu$PC we also tested PCNs with orthogonal initialisation as a parameterisation ensuring stable forward passes at initialisation for some activation functions (\S \ref{ch5:desiderata}; Fig.~\ref{ch5:fig:fwd-pass-stability-depth-params}). In brief, we found that this initialisation was not as effective as $\mu$PC (Figs.~\ref{ch5:fig:orthog-init-mnist} \& \ref{ch5:fig:orthog-init-fashion}), likely due to loss of orthogonality of the weights during training. Adding an orthogonal regulariser could help, but at the cost of an extra hyperparameter to tune. We also find that, except for linear networks, the ill-conditioning of the inference landscape still grows and spikes during training, similar to other parameterisations (e.g. Fig.~\ref{ch5:fig:sp-train-cond-nums-GD-mnist}).

\subsection{$\mu$PC with one inference step}
\begin{figure}[H]
    \vskip 0.2in
    \begin{center}
        \centerline{\includegraphics[width=\textwidth]{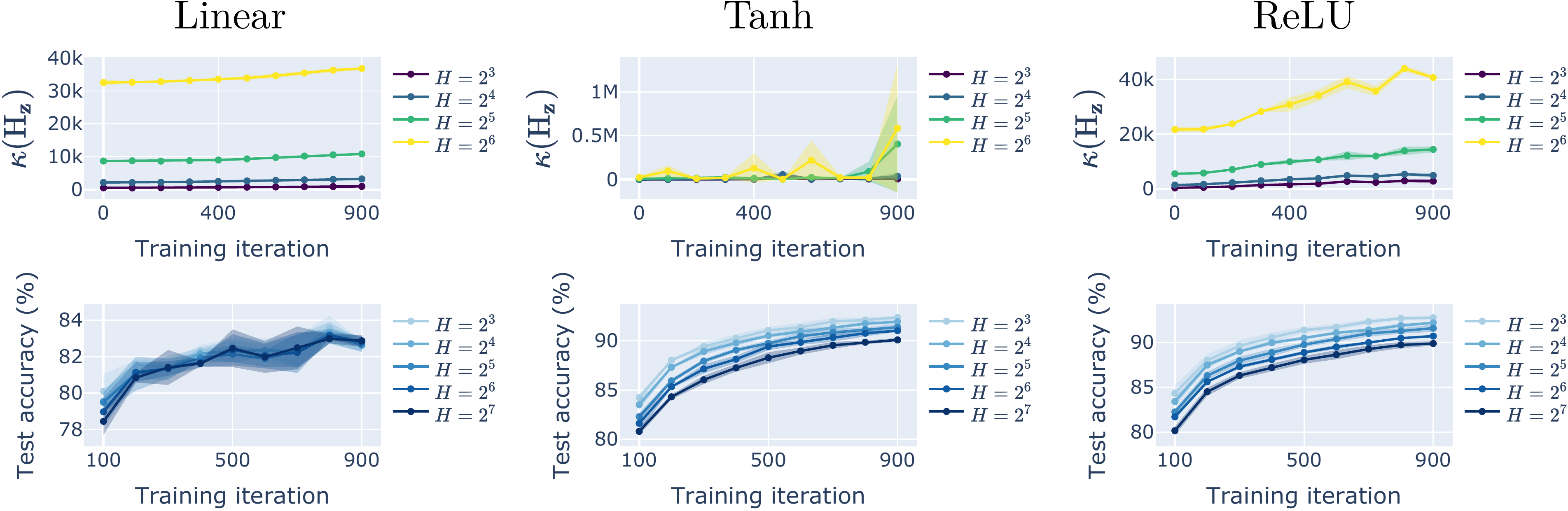}}
        \caption{\textbf{$\mu$PC test accuracies in Fig.~\ref{ch5:fig:mupc-vs-pc-mnist-accs} with one inference step.} We also plot the condition number of the activity Hessian during training.}
        \label{ch5:fig:mupc-one-step-mnist}
    \end{center}
    \vskip -0.25in
\end{figure}

All the experiments with $\mu$PC (e.g. Fig.~\ref{ch5:fig:mupc-vs-pc-mnist-accs}) used as many inference steps as hidden layers. Motivated by the results of \S \ref{ch5:activity-inits} showing that the forward pass of $\mu$PC seems to initialise the activities close to the analytical solution for DLNs (Eq.~\ref{ch5:eq:pc-infer-solution}), we also performed experiments with a single inference step. We found that this led a degradation in performance not only at initialisation but also as a function of depth (Figs.~\ref{ch5:fig:mupc-one-step-mnist} \& \ref{ch5:fig:mupc-one-step-fashion}), suggesting that some number of steps is still necessary despite $\mu$PC appearing to initialise the activities close to the inference solution (Fig.~\ref{ch5:fig:mupc-vs-pc-ffwd-lrs}). Similar to other parameterisations, we find that the ill-conditioning of the inference landscape grows and spikes during training.

\subsection{Is inference convergence sufficient for good generalisation?}
\label{ch5:infer-converge-sufficient}
Our analysis of the conditioning of the inference landscape (\S \ref{ch5:ill-cond-infer}) could be argued to rely on the assumption that converging to a solution of the inference dynamics is beneficial for learning and ultimately performance. This question has arguably not been fully resolved, with works like the one presented in the previous chapter showing both theoretical and empirical benefits for learning close to the inference equilibrium \cite{innocenti2025only}, while others argue to take only one step \cite{salvatori2022incremental}. As discussed in \S\ref{ch5:discussion}, our results suggest that convergence close to a solution is necessary for successful training (or monotonic decrease of the loss), which for brevity we will refer to as ``trainability''. In particular, $\mu$PC seems to the activities much closer to the solution than the SP (\S\ref{ch5:activity-inits}), and training $\mu$PC with one inference step leads to worse performance (e.g. Fig.~\ref{ch5:fig:mupc-one-step-mnist}) than with as many as hidden layers (e.g. Fig.~\ref{ch5:fig:mupc-vs-pc-mnist-accs}).

\begin{figure}[H]
    \vskip 0.2in
    \begin{center}
        \centerline{\includegraphics[width=\textwidth]{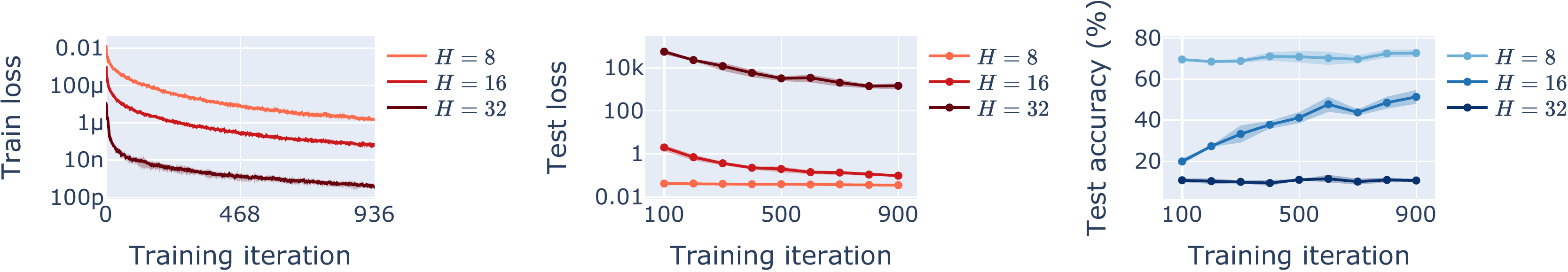}}
        \caption{\textbf{Train and test metrics of standard PCNs of varying depth trained with analytical inference (Eq.~\ref{ch5:eq:pc-infer-solution}).} We plot the training loss, test loss and test accuracy of ResNets ($N = 128$) trained with standard PC on MNIST by solving for inference analytically (using Eq.~\ref{ch5:eq:pc-infer-solution}). All experiments used Adam as optimiser with learning rate $\eta = 1e^{-3}$. Solid lines and shaded regions represent the mean and standard deviation across 3 random initialisations.}
        \label{ch5:fig:pc-analytic-infer}
    \end{center}
    \vskip -0.25in
\end{figure}

Here we report another experiment that speaks to this question and in particular suggests that \textit{while inference convergence is necessary for trainability, it is insufficient for good generalisation}, at least for standard PC. Training linear ResNets of varying depth on MNIST with ``perfect inference'' (using Eq.~\ref{ch5:eq:pc-infer-solution}), we observe that even the deepest ($H= 32$) networks now become trainable with standard PC in the sense that the training and test losses decrease monotonically (Fig.~\ref{ch5:fig:pc-analytic-infer}). However, the starting point of the test losses substantially increases with the depth, and the test accuracies of the deepest networks remain at chance level. These results do not contradict our analysis but highlight the important distinction between trainability and generalisation. Our analysis addresses the former, while the latter is beyond the scope of this work.

\section{Experimental details} 
\label{ch5:exp-details}
    Code to reproduce all the experiments is available at \sloppy{\url{https://github.com/thebuckleylab/jpc/experiments/mupc_paper}. We always used no biases, batch size $B = 64$, Adam as parameter optimiser, and GD as inference optimiser (with the exception of Figs.~\ref{ch5:fig:sp-train-cond-nums-adam-mnist} \& \ref{ch5:fig:sp-train-cond-nums-adam-fashion}). For the SP, all networks used Kaiming Uniform $(\matr{W}_\ell)_{ij} \sim \mathcal{U}(-1/N_{\ell-1}, 1/N_\ell)$ as the standard (PyTorch) initialisation used to train PCNs.

\paragraph{$\mu$PC experiments (e.g. Fig.~\ref{ch5:fig:mupc-vs-pc-mnist-accs}).} For the test accuracies in Figs.~\ref{ch5:fig:mupc-vs-pc-mnist-accs} \& \ref{ch5:fig:mupc-vs-pc-mnist-accs-all-act-fns}, we trained fully connected ResNets (Eq.~\ref{ch5:eq:resnet-energy}) to classify MNIST with standard PC, $\mu$PC and BP with Depth-$\mu$P. All networks had width $N = 512$ and always used as many GD inference iterations as the number of hidden layers $H \in \{2^i \}_{i=3}^7$. To save compute, we trained only for one epoch and evaluated the test accuracy every 300 iterations. For $\mu$PC, we selected runs based on the best results from the depth transfer (see \textbf{Hyperparameter transfer} below). For standard PC, we conducted the same grid search over the weight and activity learning rates as used for $\mu$PC. For BP, we performed a sweep over learning rates $\eta \in \{1e^0, 5e^{-1}, 1e^{-1}, 5e^{-2}, 1e^{-2}, 5e^{-3}, 1e^{-3}, 5e^{-4}, 1e{-4}\}$ at depth $H = 8$, and transferred the optimal value to the deepest ($H = 128$) networks presented.

Fig.~\ref{ch5:fig:wider_is_better_mnist} shows similar results for $\mu$PC based on the width transfer results. Fig.~\ref{ch5:fig:mupc-mnist-5-epochs} was obtained by extending the training of the 128 ReLU networks in Fig.~\ref{ch5:fig:mupc-vs-pc-mnist-accs} to 5 epochs. Figs.~\ref{ch5:fig:mupc-one-step-mnist} \& \ref{ch5:fig:mupc-one-step-fashion} were obtained with the same setup as Fig.~\ref{ch5:fig:mupc-vs-pc-mnist-accs} by running $\mu$PC for a single inference step. As noted in \S \ref{ch5:experiments}, the results on Fashion-MNIST (Fig.~\ref{ch5:fig:mupc-fashion-15-epochs}) were obtained with depth transfer by tuning 8-layer networks and transferring the optimal learning rates to 128 layers.

\paragraph{Hessian condition number at initialisation (e.g. Fig.~\ref{ch5:fig:sp-cond-nums-init}).} For different activation functions (Fig.~\ref{ch5:fig:sp-cond-nums-init}), architectures (Fig.~\ref{ch5:fig:resnets-cond-nums-init}) and parameterisations (Fig.~\ref{ch5:fig:des-trade-off}), we computed the condition number of the activity Hessian (Eq.~\ref{ch5:eq:activity-hessian}) at initialisation over widths and depths $N, H \in \{2^i\}^7_{i=1}$. This was the maximum range we could achieve to compute the full Hessian matrix given our memory resources. No biases were used since these do not affect the Hessian as explained in \S \ref{ch5:activity-grad-hess}. Results did not differ significantly across different seeds or input and output data dimensions, as predicted from the structure of the activity Hessian (Eq.~\ref{ch5:eq:activity-hessian}).

For the landscape insets of Fig.~\ref{ch5:fig:sp-cond-nums-init}, the energy landscape was sampled around the linear solution of the activities (Eq.~\ref{ch5:eq:pc-infer-solution}) along the maximum and minimum eigenvectors of the Hessian $\mathcal{F}(\mathbf{z}^* + \alpha \hat{\mathbf{v}}_{\text{min}} + \beta \hat{\mathbf{v}}_{\text{min}})$, with domain $\alpha, \beta \in [-2, 2]$ and $30\times 30$ resolution.

\paragraph{Hessian condition number over training (e.g. Fig.~\ref{ch5:fig:sp-train-cond-nums-GD-mnist}).} For different activations (e.g. Fig.~\ref{ch5:fig:sp-train-cond-nums-GD-mnist}), architectures (e.g. Fig.~\ref{ch5:fig:sp-train-cond-nums-skips-mnist}), algorithms (e.g. Fig.~\ref{ch5:fig:sp-train-cond-nums-adam-mnist}) and parameterisations (e.g. Fig.~\ref{ch5:fig:orthog-init-mnist}), we trained networks of width $N=128$ and hidden layers $H \in \{8, 16, 32\}$ to perform classification on MNIST and Fashion-MNIST. This set of widths and depths was chosen to allow for tractable computation of the full activity Hessian (Eq.~\ref{ch5:eq:activity-hessian}). Training was stopped after one epoch to illustrate the phenomenon of ill-conditioning. All experiments used weight learning rate $\eta = 1e^{-3}$ and performed a grid search over activity learning rates $\beta \in \{5e^{-1}, 1e^{-1}, 5e^{-2}\}$. A maximum number of $T=500$ steps was used, and inference was stopped when the norm of the activity gradients reached some tolerance.

\paragraph{Hyperparameter transfer (e.g. Fig.~\ref{ch5:fig:mupc-hyperparam-transfer-tanh}).} For the ResNets trained on MNIST with $\mu$PC (e.g. Fig.~\ref{ch5:fig:mupc-vs-pc-mnist-accs}), we performed a 2D grid search over the following learning rates: $\eta \in \{5e^{-1}, 1e^{-1}, 5e^{-2}, 1e^{-2}\}$ for the weights, and $\beta \in \{1e^{3}, 5e^{2}, 1e^{2}, 5e^{1}, 1e^{1}, 5e^{0}, 1e^{0}, 5e^{-1}, 1e^{-1}, 5e^{-2}, 1e^{-2} \}$ for the activities. We trained only for one epoch, in part to save compute and in part based on the results of \citep[][Fig.~B.3]{bordelon2023depthwise} showing that the optimal learning rate could be decided after just 3 epochs on CIFAR-10. The number of (GD) inference iterations was always the same as the number of hidden layers. For the width transfer results, we trained networks of 8 hidden layers and widths $N \in \{2^i \}_{i=6}^{10}$, while for the depth transfer we fixed the width to $N = 512$ and varied the depth $H \in \{2^i \}_{i=3}^{7}$. Note that this means that the plots with title $N = 512$ and $H = 8$ in Figs.~\ref{ch5:fig:mupc-hyperparam-transfer-tanh} \& \ref{ch5:fig:mupc-hyperparam-transfer-linear}-\ref{ch5:fig:mupc-hyperparam-transfer-relu} are the same. The landscape contours were averaged over 3 different random seeds, and the training loss is plotted on a log scale to aid interpretation.

\paragraph{Loss vs energy ratios (e.g. Fig.~\ref{ch5:fig:mupc-loss-energy-ratios-init-1e-1}).} We trained ResNets (Eq.~\ref{ch5:eq:resnet-energy}) to classify MNIST for one epoch with widths and depths $N, H \in \{2^i \}_{i=1}^6$. To replicate the successful setup of Fig.~\ref{ch5:fig:mupc-vs-pc-mnist-accs}, we used the same learning rate for the optimal linear networks trained on MNIST, $\eta = 1e^{-1}$. To verify Theorem~\ref{ch5:thm1}, at every training step we computed the ratio between the Depth-$\mu$P MSE loss $\mathcal{L}(\boldsymbol{\theta})$ and the equilibrated $\mu$PC energy $\mathcal{F}(\mathbf{z}^*, \boldsymbol{\theta})$ (Eq.~\ref{ch5:eq:resnet-equilib-energy}), where $\mathbf{z}^*$ was computed using Eq.~\ref{ch5:eq:pc-infer-solution}. All experiments used the weight learning rate $\eta = 1e^{-4}$. Fig.~\ref{ch5:fig:sp-loss-energy-ratios-init} shows the same results for the SP, which used a smaller learning rate $\eta = 1e^{-4}$ to avoid divergence at large depth. All the phase diagrams are plotted on a log scale for easier visualisation. Fig.~\ref{ch5:fig:example-sp-vs-mupc-loss-energy-ratios} shows an example of the ratio dynamics of $\mu$PC vs PC for a ResNet with 4 hidden layers and different widths. Results were similar across different random initialisations.

\section{Compute resources} 
\label{ch5:compute-res}
The experiments involving $\mu$PC, hyperparameter transfer, and the monitoring of the condition number of the Hessian during training were all run on an NVIDIA RTX A6000. The runtime varied by experiment, with the 128-layer networks trained for multiple epochs (Figs.~\ref{ch5:fig:mupc-mnist-5-epochs}-\ref{ch5:fig:mupc-fashion-15-epochs}) taking several days. All other experiments were run on a CPU and took between one hour and half a day, depending on the specific experiment.

\section{Supplementary figures} 
\label{ch5:supp-figures}
\begin{figure}[H]
    \begin{center}
        \centerline{\includegraphics[width=\textwidth]{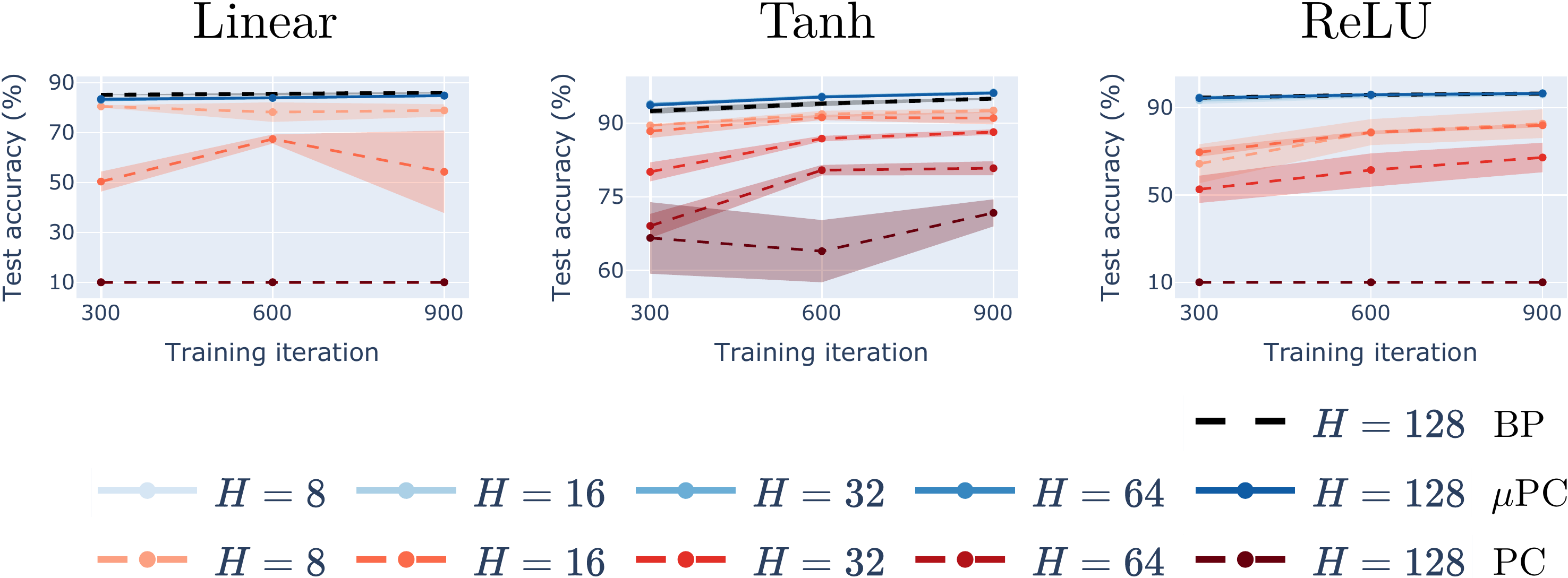}}
        \caption{\textbf{Test accuracies in Fig.~\ref{ch5:fig:mupc-vs-pc-mnist-accs} for different activation functions.} Solid lines and shaded regions indicate the mean and standard deviation across 3 random seeds, respectively. BP represents BP with Depth-$\mu$P.}
        \label{ch5:fig:mupc-vs-pc-mnist-accs-all-act-fns}
    \end{center}
\end{figure}
\begin{figure}[H]
    \begin{center}
        \centerline{\includegraphics[width=0.5\textwidth]{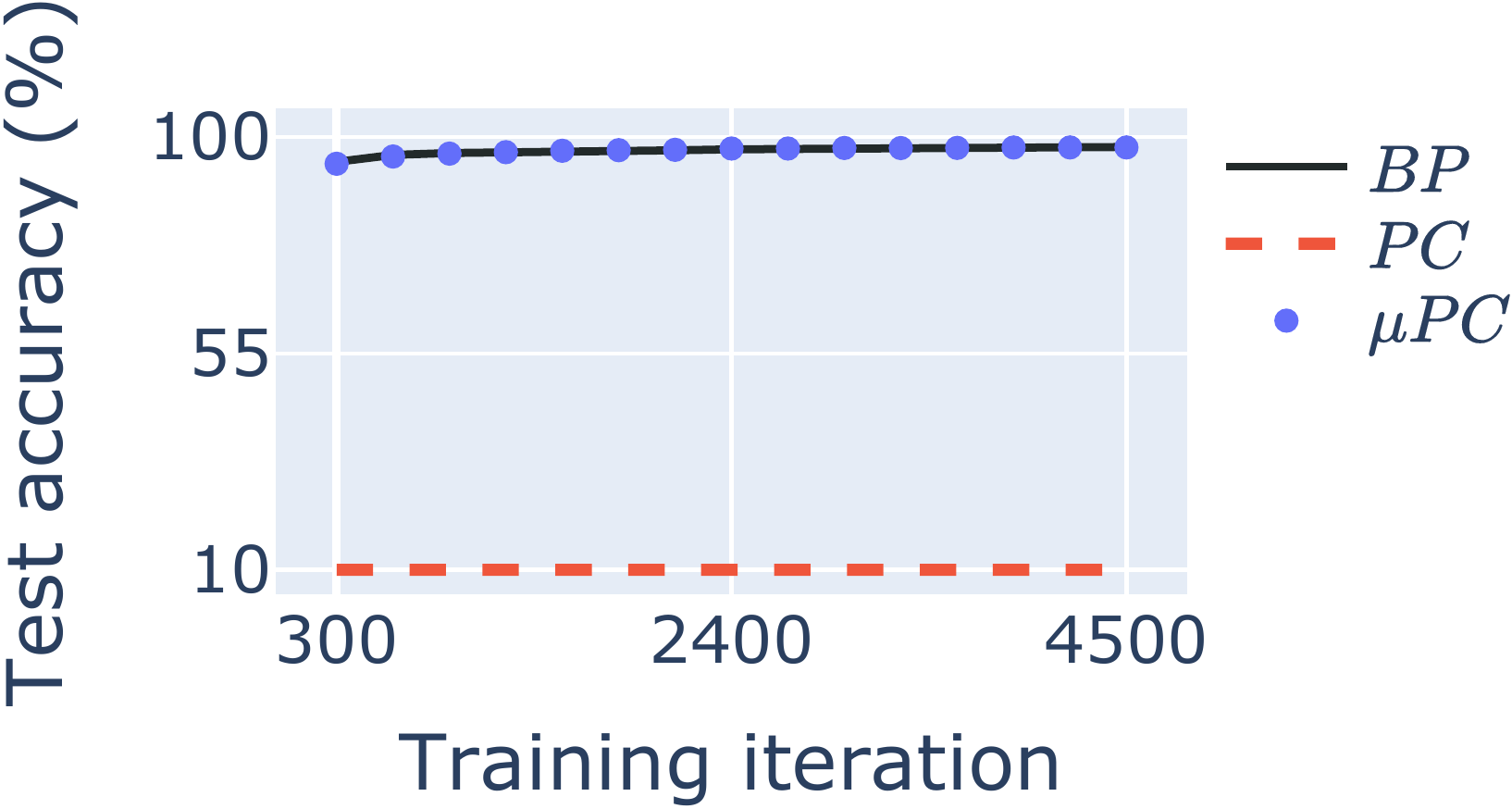}}
        \caption{\textbf{128-layer ReLU network trained with $\mu$PC on MNIST for 5 epochs.} Solid lines and (barely visible) shaded regions indicate the mean and standard deviation across 5 random seeds, respectively. BP represents BP with Depth-$\mu$P.}
        \label{ch5:fig:mupc-mnist-5-epochs}
    \end{center}
\end{figure}
\begin{figure}[H]
    \begin{center}
        \centerline{\includegraphics[width=0.5\textwidth]{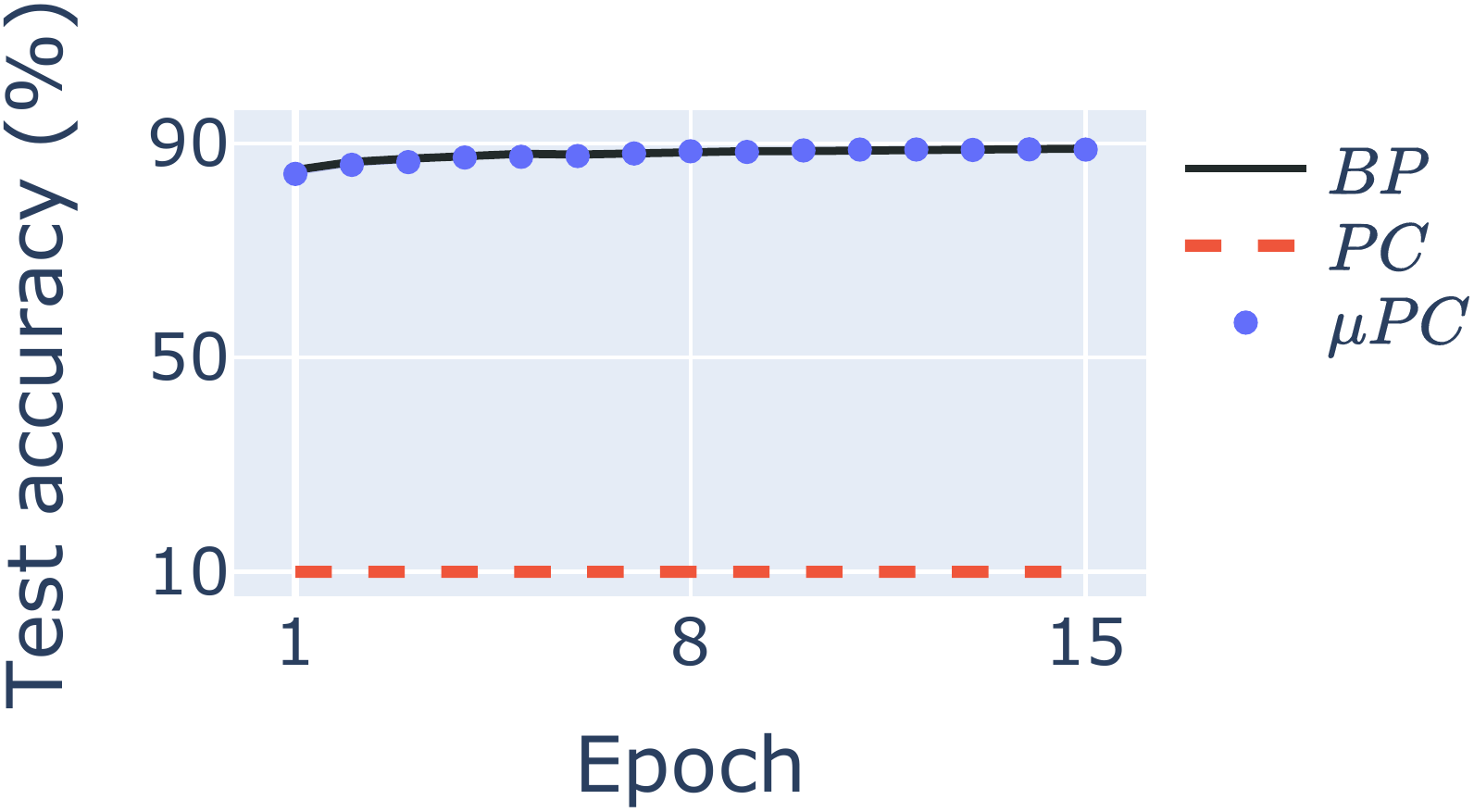}}
        \caption{\textbf{128-layer ReLU network trained with $\mu$PC on Fashion-MNIST.} Solid lines and (barely visible) shaded regions indicate the mean and standard deviation across 3 random seeds, respectively. BP represents BP with Depth-$\mu$P.}
        \label{ch5:fig:mupc-fashion-15-epochs} 
    \end{center}
\end{figure}
\begin{figure}[H]
    \begin{center}
        \centerline{\includegraphics[width=0.5\textwidth]{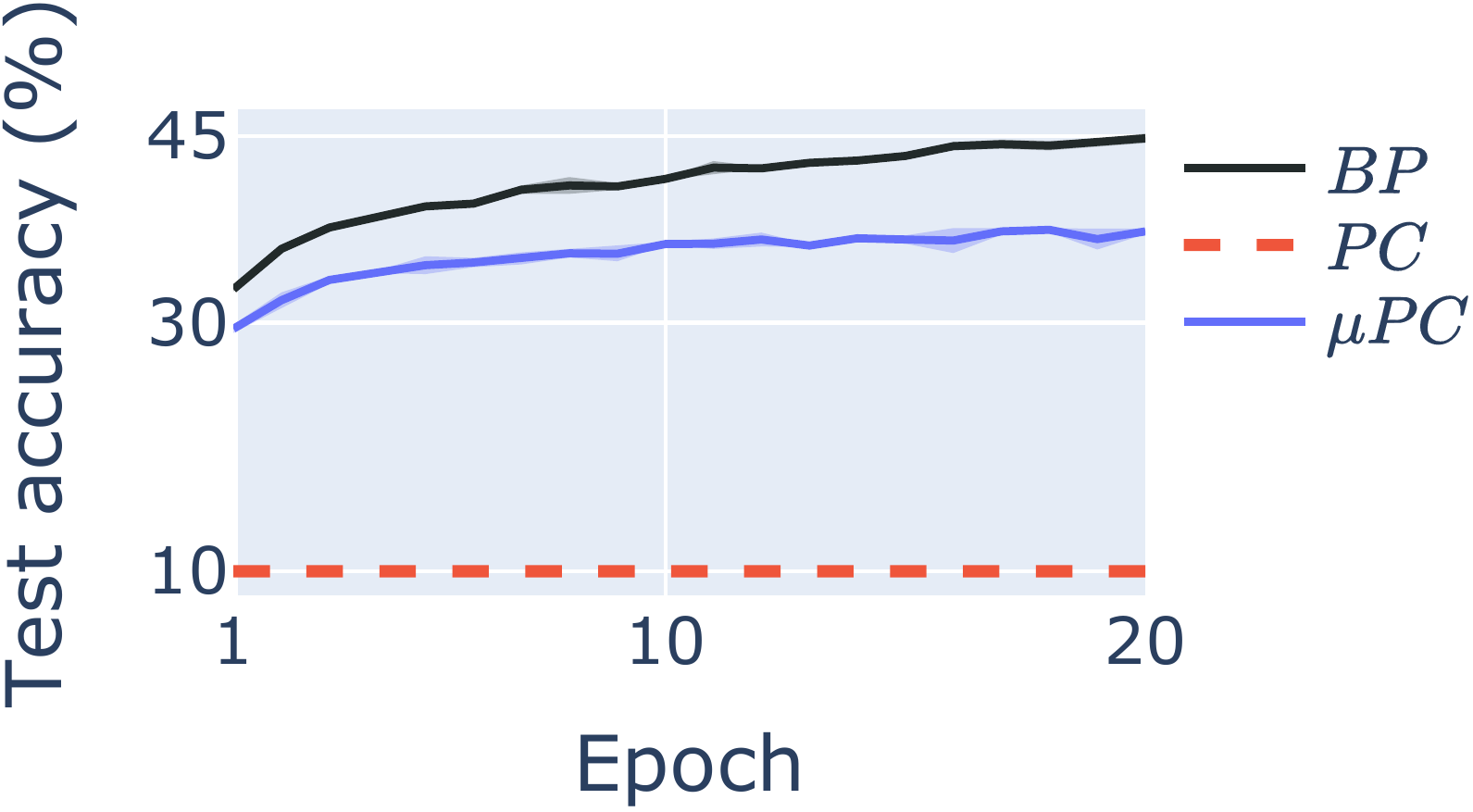}}
        \caption{\textbf{128-layer fully connected residual ReLU network trained with $\mu$PC on CIFAR10.} Solid lines and (barely visible) shaded regions indicate the mean and standard deviation across 3 random seeds, respectively. BP represents BP with Depth-$\mu$P. As for other datasets, we see that $\mu$PC remains capable of training such deep networks, although performance slightly lags behind BP. Note that accuracies for all algorithms are far from SOTA because of the fully connected (as opposed to convolutional) architecture used.}
        \label{ch5:fig:cifar-20-epochs}
    \end{center}
\end{figure}
\begin{figure}[H]
    \begin{center}
        \centerline{\includegraphics[width=\textwidth]{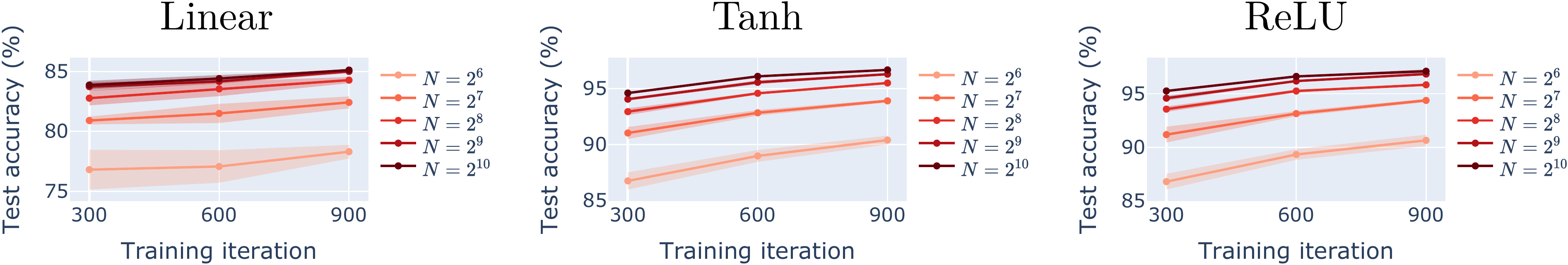}}
        \caption{\textbf{Same results as Fig.~\ref{ch5:fig:mupc-vs-pc-mnist-accs} varying the width $N$ and fixing the depth at $H = 8$, showing that ``wider is better'' \cite{yang2021tuning, ishikawa2024local}.}}
        \label{ch5:fig:wider_is_better_mnist}
    \end{center}
\end{figure}
\begin{figure}[H]
    \begin{center}
        \centerline{\includegraphics[width=0.5\textwidth]{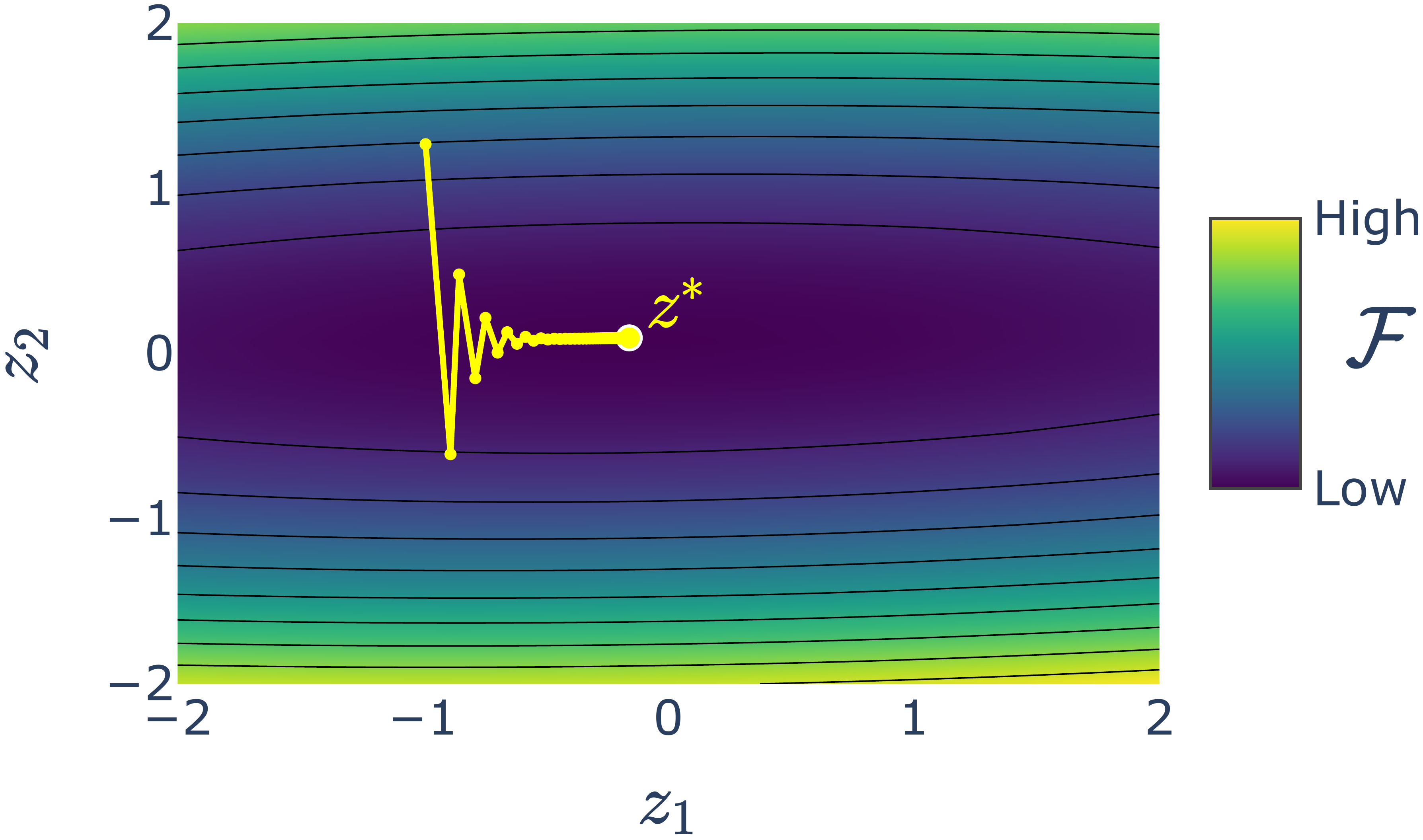}}
        \caption{\textbf{Toy illustration of the ill-conditioning of the inference landscape.} Plotted is the activity or inference landscape $\mathcal{F}(z_1, z_2)$ for a toy linear network with two hidden units $f(x) = w_3w_2w_1x$, along with the GD dynamics. One weight was artificially set to a much higher value than the others to induce ill-conditioning.}
        \label{ch5:toy-ill-cond}
    \end{center}
\end{figure}
\begin{figure}[H]
    \begin{center}
        \centerline{\includegraphics[width=\textwidth]{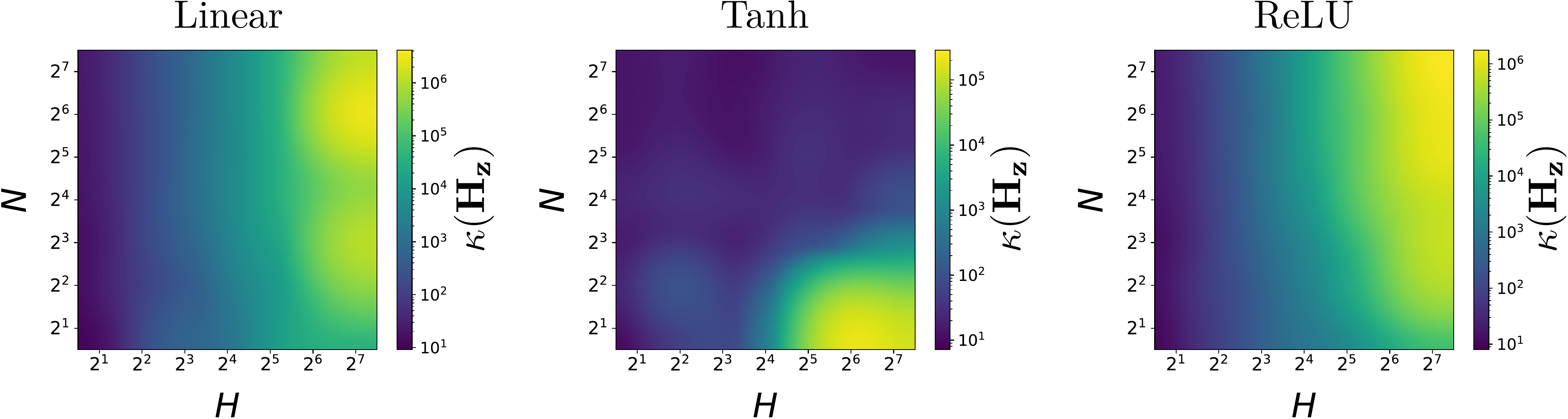}}
            \caption{\textbf{Same results as Fig.~\ref{ch5:fig:sp-cond-nums-init} for the activity Hessian of ResNets (Eq.~\ref{ch5:eq:resnets-activity-hessian}).}}
            \label{ch5:fig:resnets-cond-nums-init}
    \end{center}
\end{figure}
\begin{figure}[H]
    \begin{center}
        \centerline{\includegraphics[width=\textwidth]{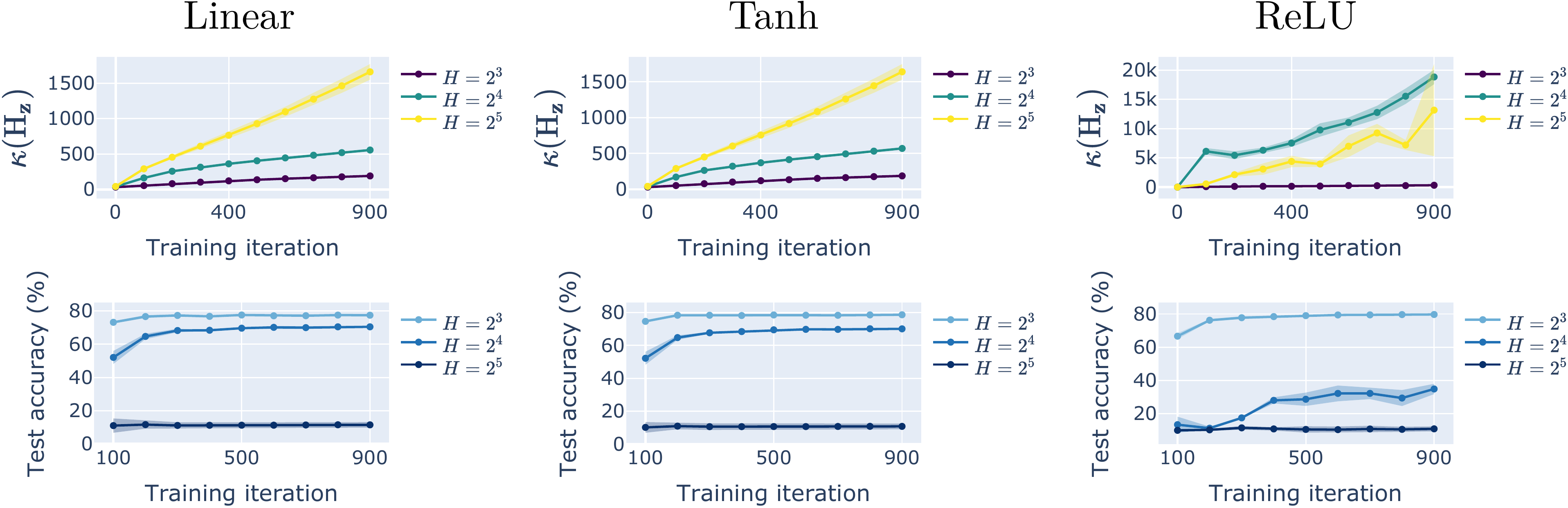}}
        \caption{\textbf{Same results as Fig.~\ref{ch5:fig:sp-train-cond-nums-GD-mnist} for Fashion-MNIST.}}
        \label{ch5:fig:sp-train-cond-nums-GD-fashion}
    \end{center}
\end{figure}
\begin{figure}[H]
    \begin{center}
        \centerline{\includegraphics[width=\textwidth]{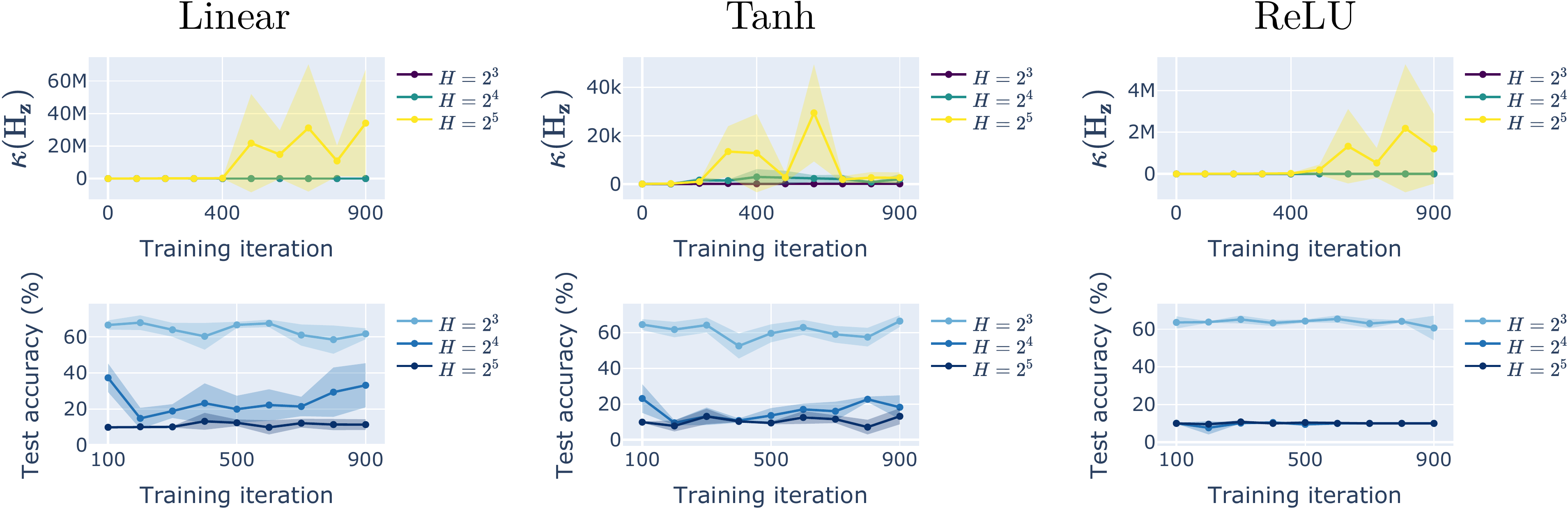}}
        \caption{\textbf{Same results as Fig.~\ref{ch5:fig:sp-train-cond-nums-adam-mnist} for Fashion-MNIST.}}
        \label{ch5:fig:sp-train-cond-nums-adam-fashion}
    \end{center}
\end{figure}
\begin{figure}[H]
    \begin{center}
        \centerline{\includegraphics[width=\textwidth]{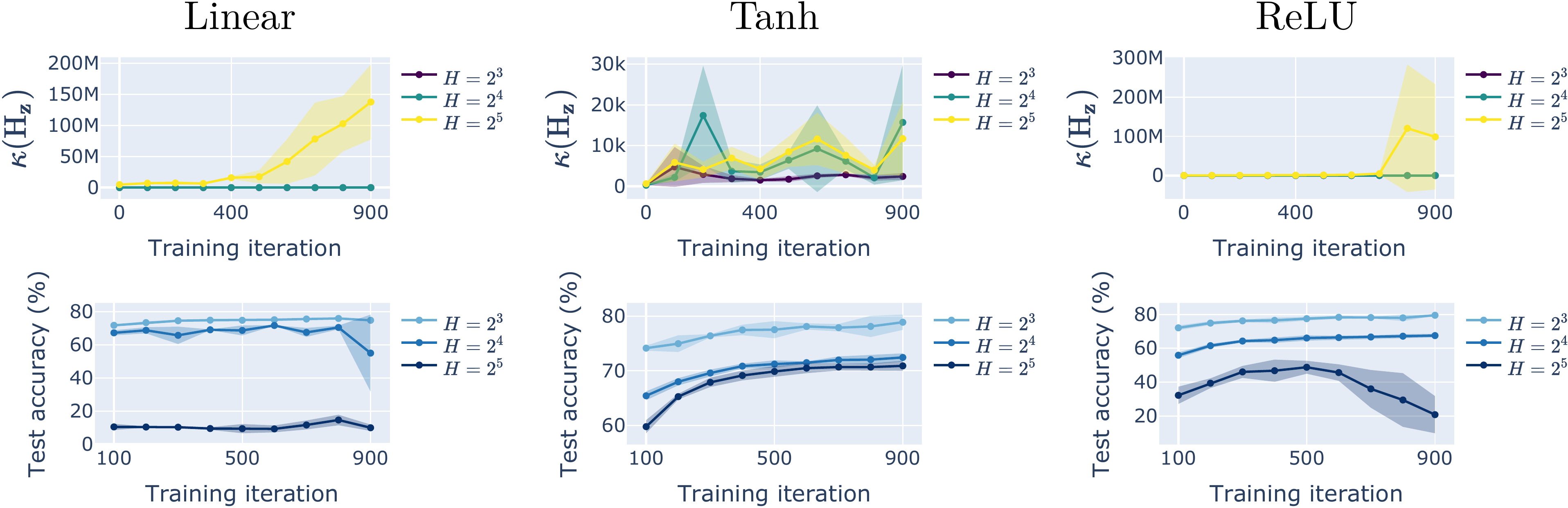}}
        \caption{\textbf{Same results as Fig.~\ref{ch5:fig:sp-train-cond-nums-skips-mnist} for Fashion-MNIST.}}
        \label{ch5:fig:sp-train-cond-nums-skips-fashion}
    \end{center}
\end{figure}
\begin{figure}[H]
    \begin{center}
        \centerline{\includegraphics[width=\textwidth]{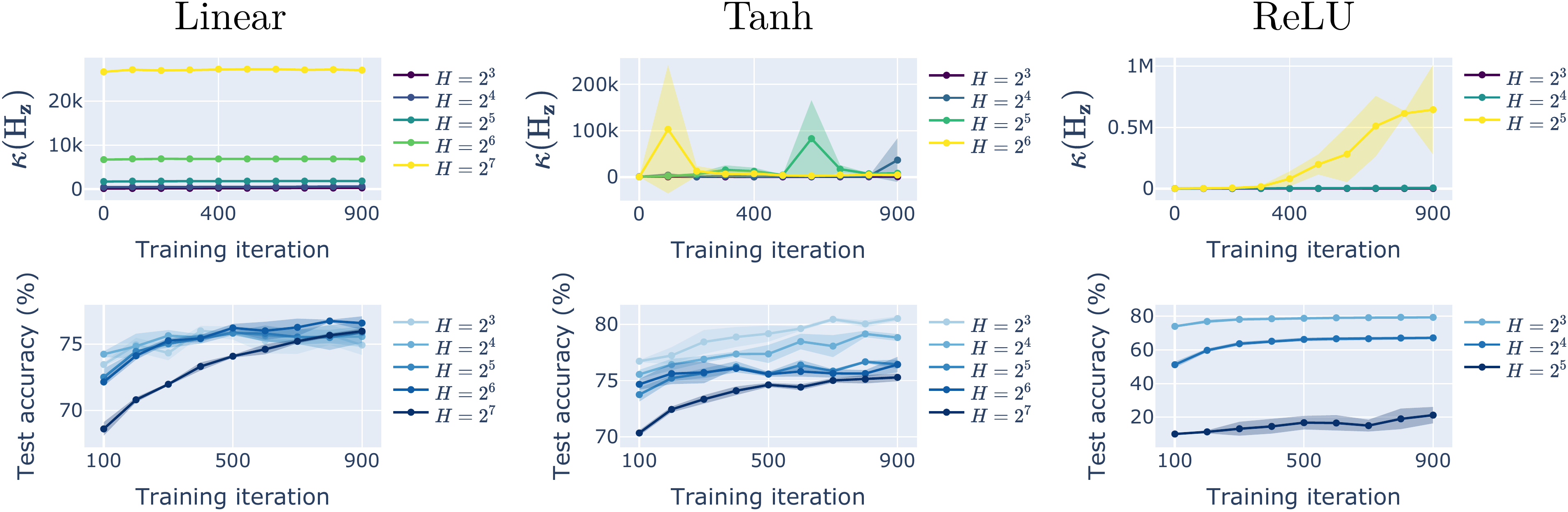}}
        \caption{\textbf{Same results as Fig.~\ref{ch5:fig:orthog-init-mnist} for Fashion-MNIST.}}
        \label{ch5:fig:orthog-init-fashion}
    \end{center}
\end{figure}
\begin{figure}[H]
    \begin{center}
        \centerline{\includegraphics[width=\textwidth]{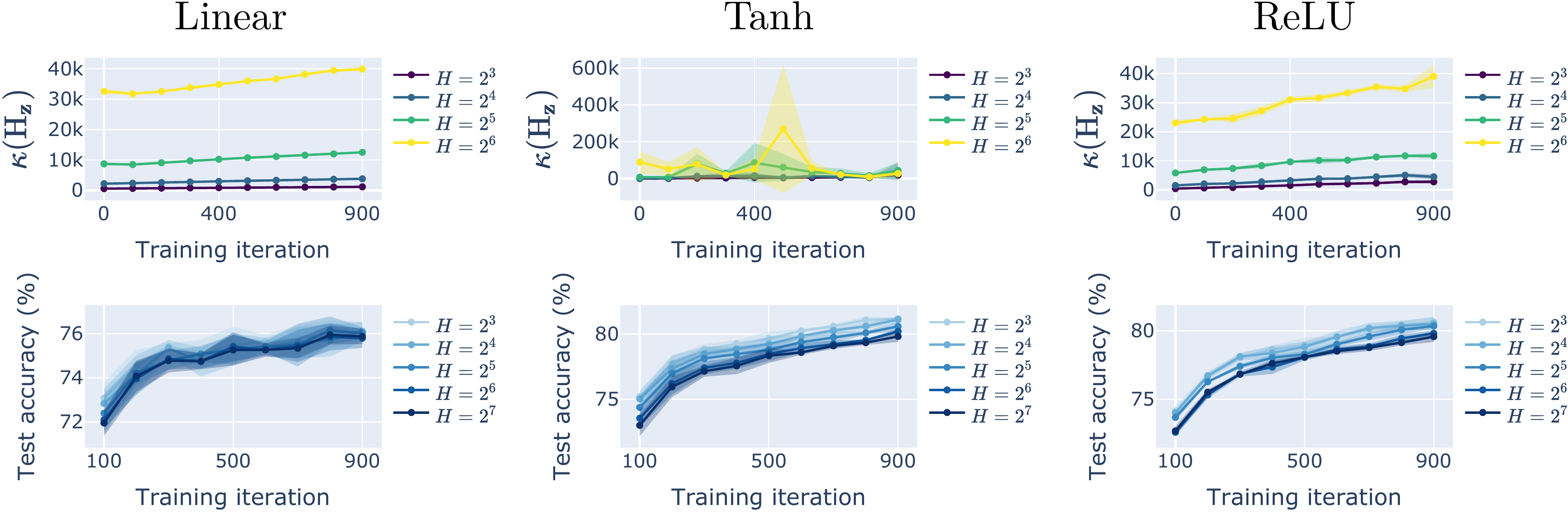}}
        \caption{\textbf{Same results as Fig.~\ref{ch5:fig:mupc-one-step-mnist} for Fashion-MNIST.}}
        \label{ch5:fig:mupc-one-step-fashion}
    \end{center}
\end{figure}
\begin{figure}[H]
    \begin{center}
        \centerline{\includegraphics[width=\textwidth]{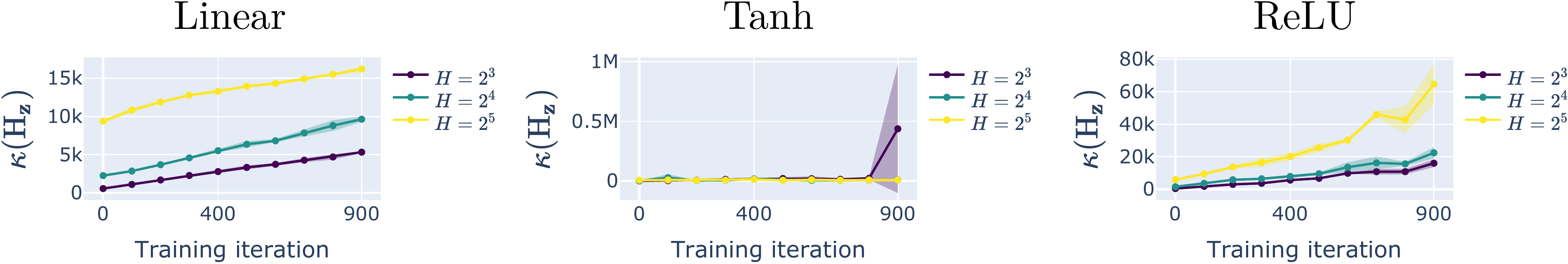}}
        \caption{\textbf{Inference conditioning during training for some $\mu$PC networks in Fig.~\ref{ch5:fig:mupc-vs-pc-mnist-accs}.}}
        \label{ch5:fig:mupc-train-cond-nums-mnist}
    \end{center}
\end{figure}
\begin{figure}[H]
    \begin{center}
        \centerline{\includegraphics[width=\textwidth]{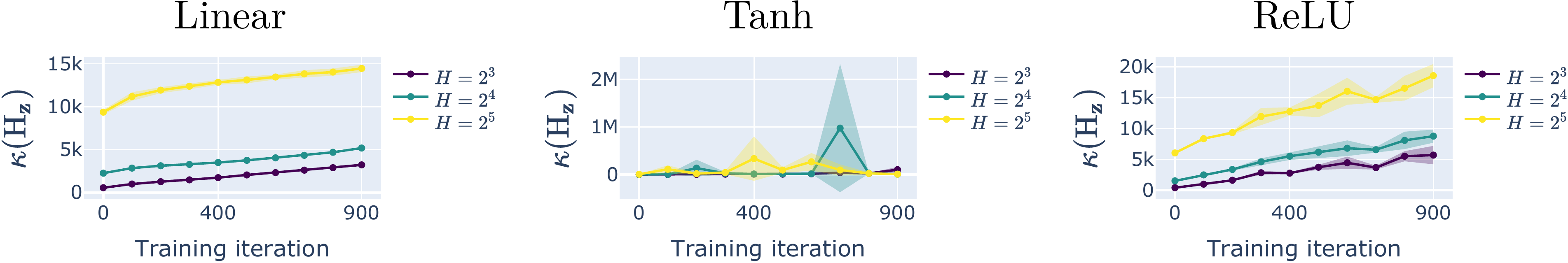}}
        \caption{\textbf{Same results as Fig.~\ref{ch5:fig:mupc-train-cond-nums-mnist} for Fashion-MNIST.}}
        \label{ch5:fig:mupc-train-cond-nums-fashion}
    \end{center}
\end{figure}
\begin{figure}[H]
    \vskip 0.2in
    \begin{center}
        \centerline{\includegraphics[width=\textwidth]{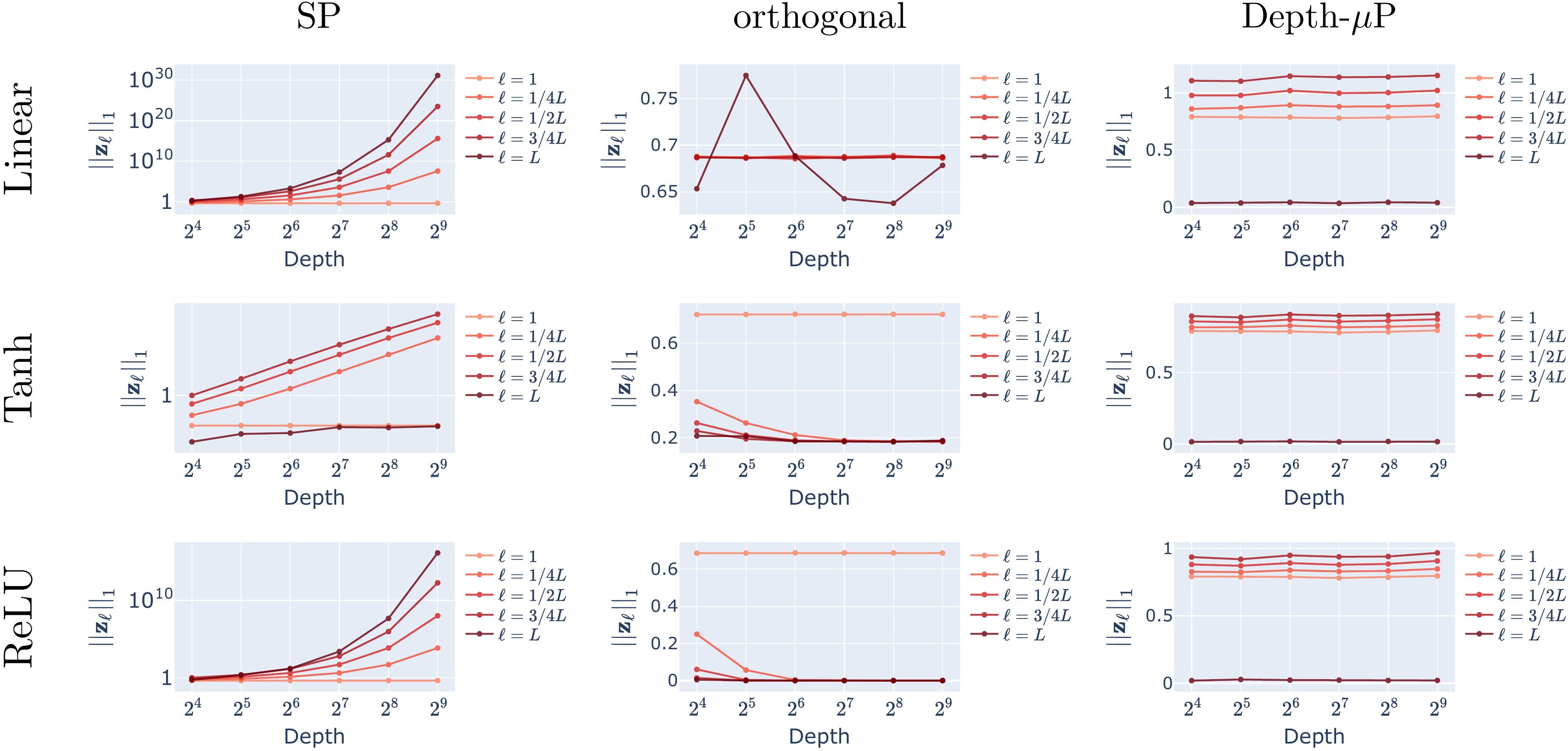}}
        \caption{\textbf{Forward pass (in)stability with network depth for different parameterisations.} For different activation functions and parameterisations, we plot the mean $\ell^1$ norm of the feedforward pass activities at initialisation as a function of the network depth $L$. Networks ($N=1024$) had skip connections for the standard parameterisation (SP) and Depth-$\mu$P but not orthogonal. Results were similar across different seeds.}
        \label{ch5:fig:fwd-pass-stability-depth-params}
    \end{center}
    \vskip -0.25in
\end{figure}
\begin{figure}[H]
    \begin{center}
        \centerline{\includegraphics[width=\textwidth]{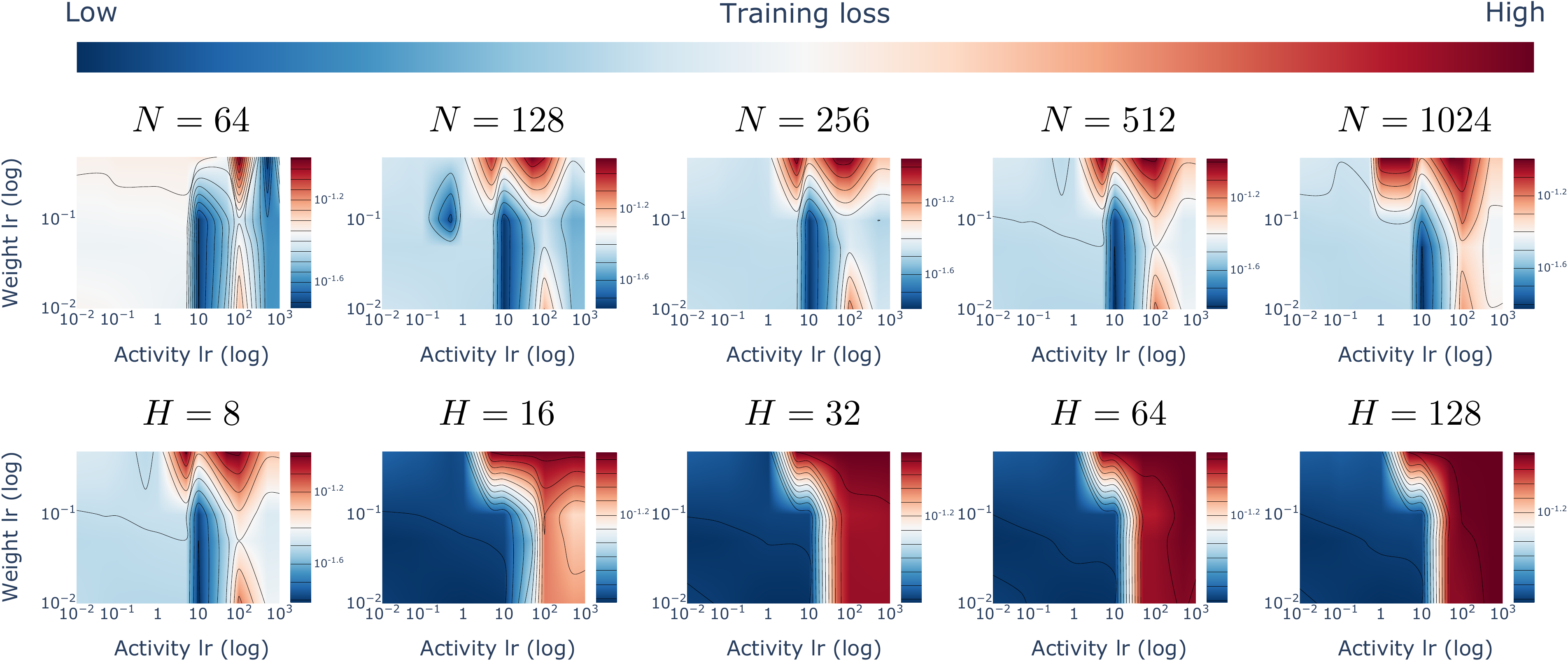}}
        \caption{\textbf{Same results as Fig.~\ref{ch5:fig:mupc-hyperparam-transfer-tanh} for Linear.}}
        \label{ch5:fig:mupc-hyperparam-transfer-linear}
    \end{center}
\end{figure}
\begin{figure}[H]
    \begin{center}
        \centerline{\includegraphics[width=\textwidth]{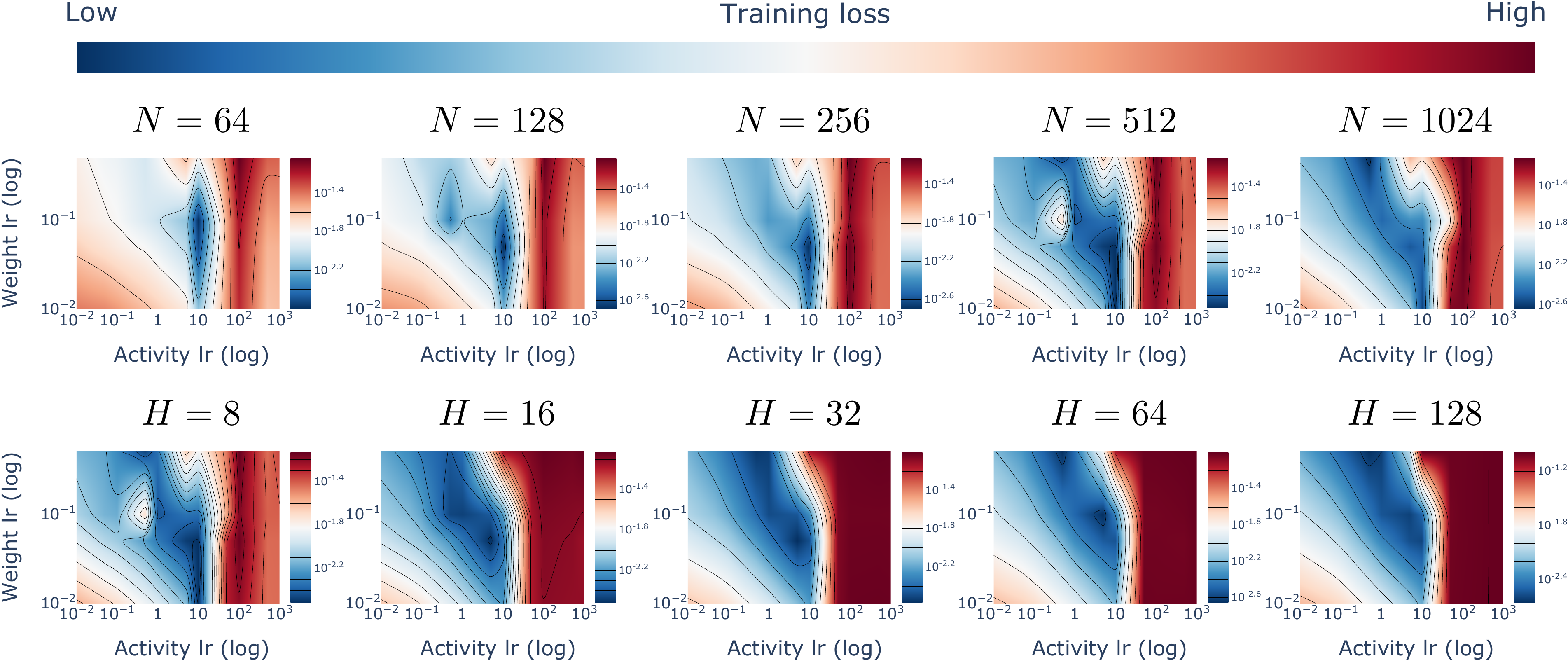}}
        \caption{\textbf{Same results as Fig.~\ref{ch5:fig:mupc-hyperparam-transfer-tanh} for ReLU.}}
        \label{ch5:fig:mupc-hyperparam-transfer-relu}
    \end{center}
\end{figure}
\begin{figure}[H]
    \begin{center}
        \centerline{\includegraphics[width=\textwidth]{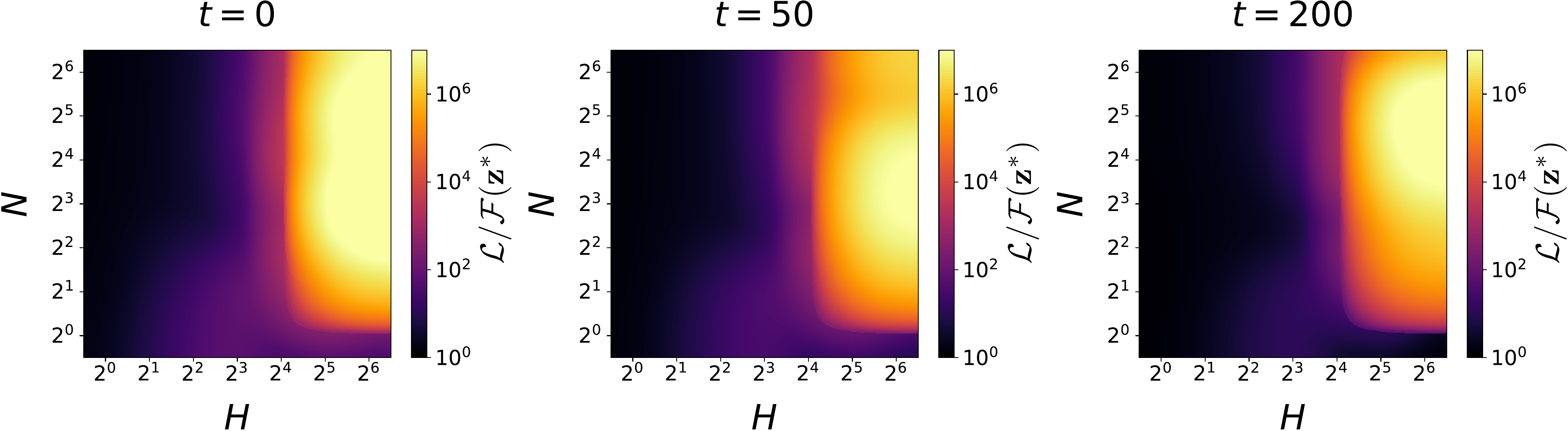}}
        \caption{\textbf{Same results as Fig.~\ref{ch5:fig:mupc-loss-energy-ratios-init-1e-1} for the standard parameterisation (SP).}}
        \label{ch5:fig:sp-loss-energy-ratios-init}
    \end{center}
\end{figure}
\begin{figure}[H]
    \begin{center}
        \centerline{\includegraphics[width=\textwidth]{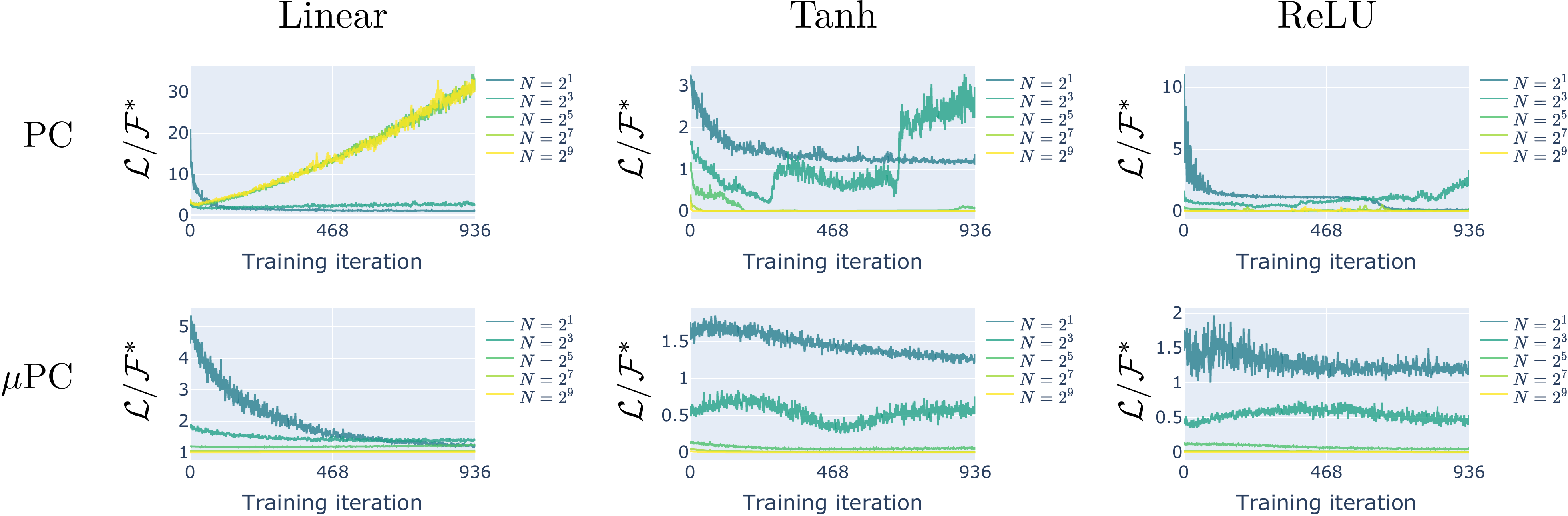}}
        \caption{\textbf{Example of the loss vs energy ratio dynamics of SP and $\mu$PC for $H = 4$.}}
        \label{ch5:fig:example-sp-vs-mupc-loss-energy-ratios}
    \end{center}
\end{figure}
\begin{figure}[H]
    \begin{center}
        \centerline{\includegraphics[width=\textwidth]{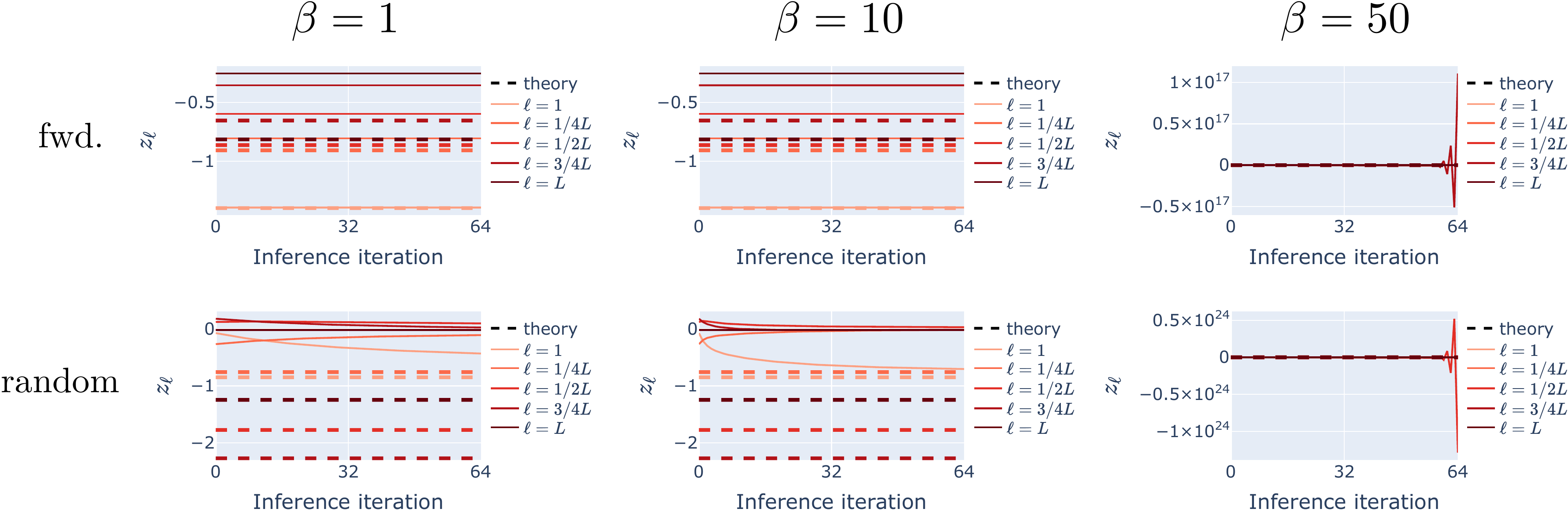}}
        \caption{\textbf{Same results as Fig.~\ref{ch5:fig:sp-activity-inits} for $\mu$PC.}}
        \label{ch5:fig:mupc-activity-inits}
    \end{center}
\end{figure}
\begin{figure}[H]
    \begin{center}
        \centerline{\includegraphics[width=\textwidth]{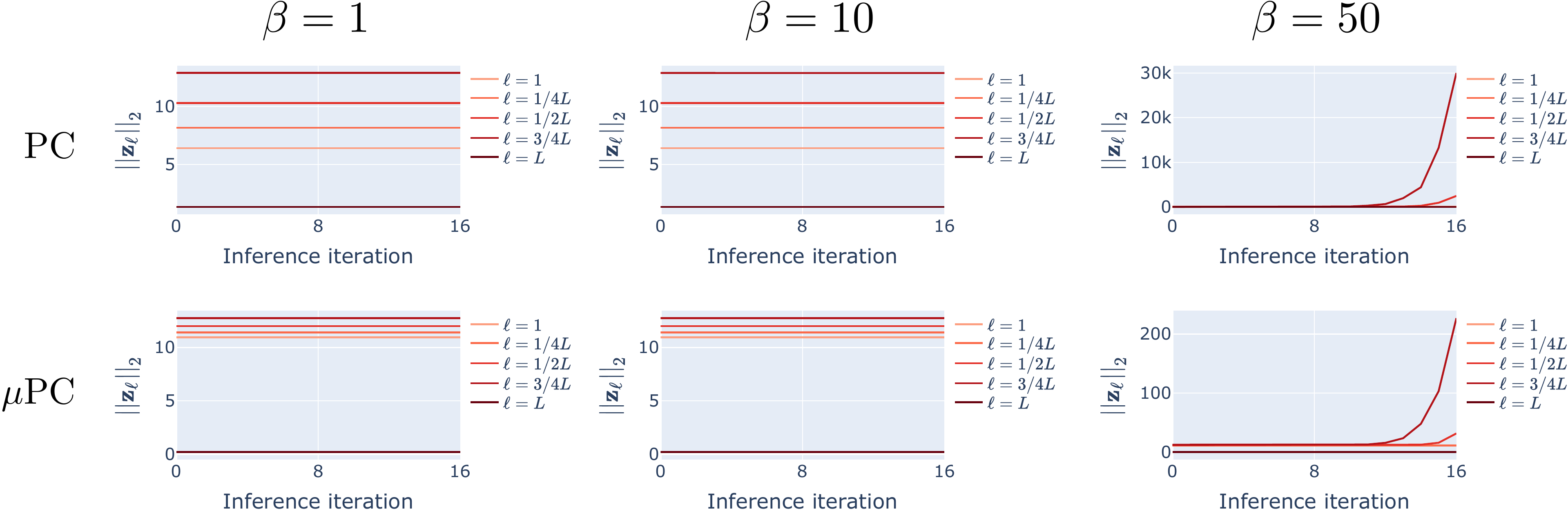}}
        \caption{\textbf{Same results as Fig.~\ref{ch5:fig:mupc-vs-pc-ffwd-lrs} for a ReLU network.}}
        \label{ch5:fig:mupc-vs-pc-ffwd-relu-lrs}
    \end{center}
\end{figure}

%% file: text/appendices/ch6.tex
\chapter{Appendix for Chapter 6}
\label{ch6:appendix}

\renewcommand{\thefigure}{D.\arabic{figure}}
\setcounter{figure}{0}

\section{Supplementary figures} 
\label{supp-figs}
\begin{figure}[H]
    \begin{center}
        \centerline{\includegraphics[width=\textwidth]{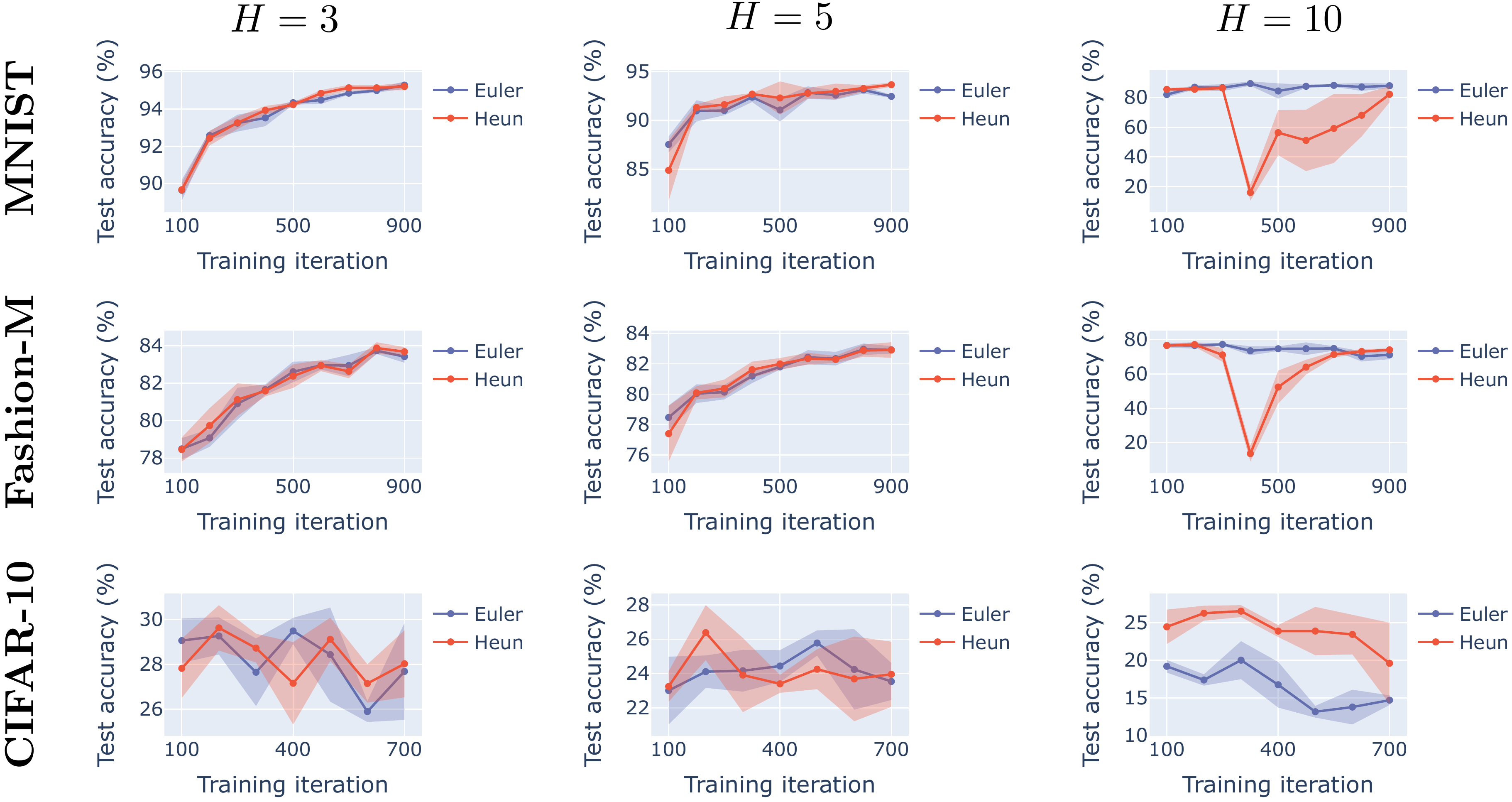}}
        \caption{\textbf{Test accuracies for Figure~\ref{ch6:fig:best-infer-runtimes}.} These accuracies were selected from Figures~\ref{ch6:fig:accs-per-ts-mnist}-\ref{ch6:fig:accs-per-ts-cifar} based on the lowest upper integration limit $T$ at which the maximum mean accuracy was achieved. Note that the experiments were not optimised for accuracy, since we were specifically interested in the runtime of different ODE solvers at comparable performance. We refer to \cite{pinchetti2024benchmarking} for a comprehensive performance benchmarking of PCNs.}
        \label{ch6:fig:best-test-accs}
    \end{center}
\end{figure}
\begin{figure}[H]
    \begin{center}
        \centerline{\includegraphics[width=\textwidth]{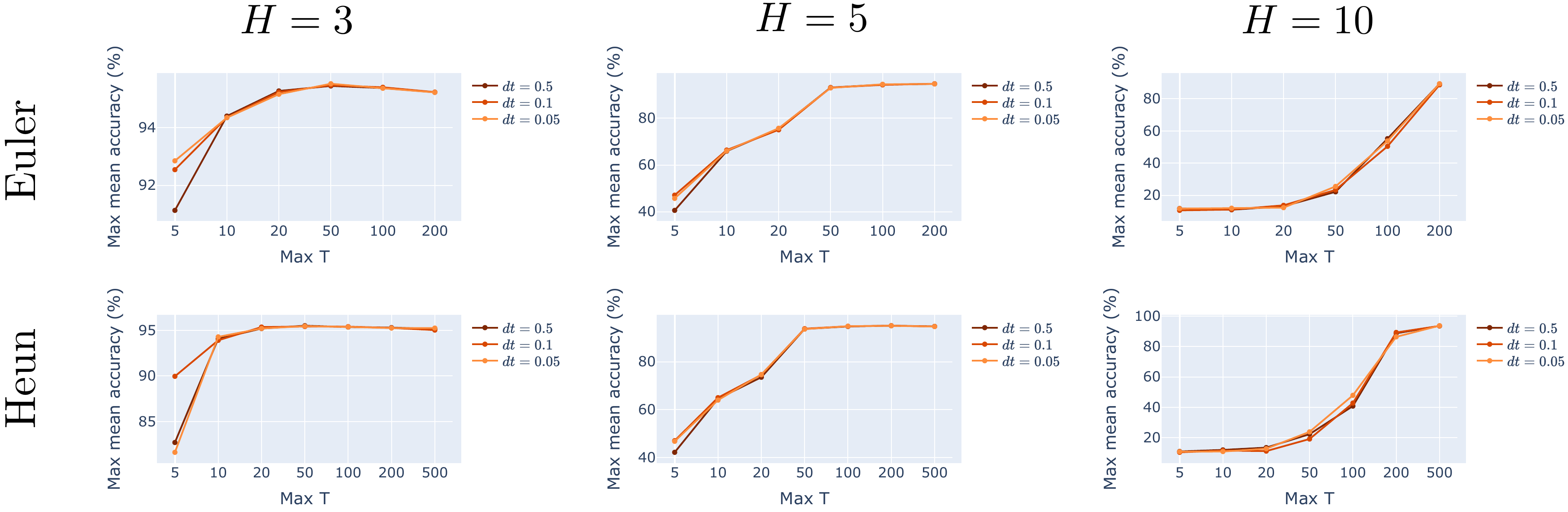}}
        \caption{\textbf{Maximum mean test accuracy on MNIST achieved with Euler and Heun as a function of different step sizes $dt$ and upper integration limits $T$.} For the results in Figure~\ref{ch6:fig:best-infer-runtimes} with $H=3$, we selected runs with $T=20$, and $dt=0.5$ for Euler and $dt=0.05$ for Heun. For $H=5$, we selected $T=50$, and $dt=0.5$ for Euler and $dt=0.05$ for Heun. Finally, for $H=10$, $T=200$ and $dt=0.05$ were chosen for both solvers.}
        \label{ch6:fig:accs-per-ts-mnist}
    \end{center}
\end{figure}
\begin{figure}[H]
    \begin{center}
        \centerline{\includegraphics[width=\textwidth]{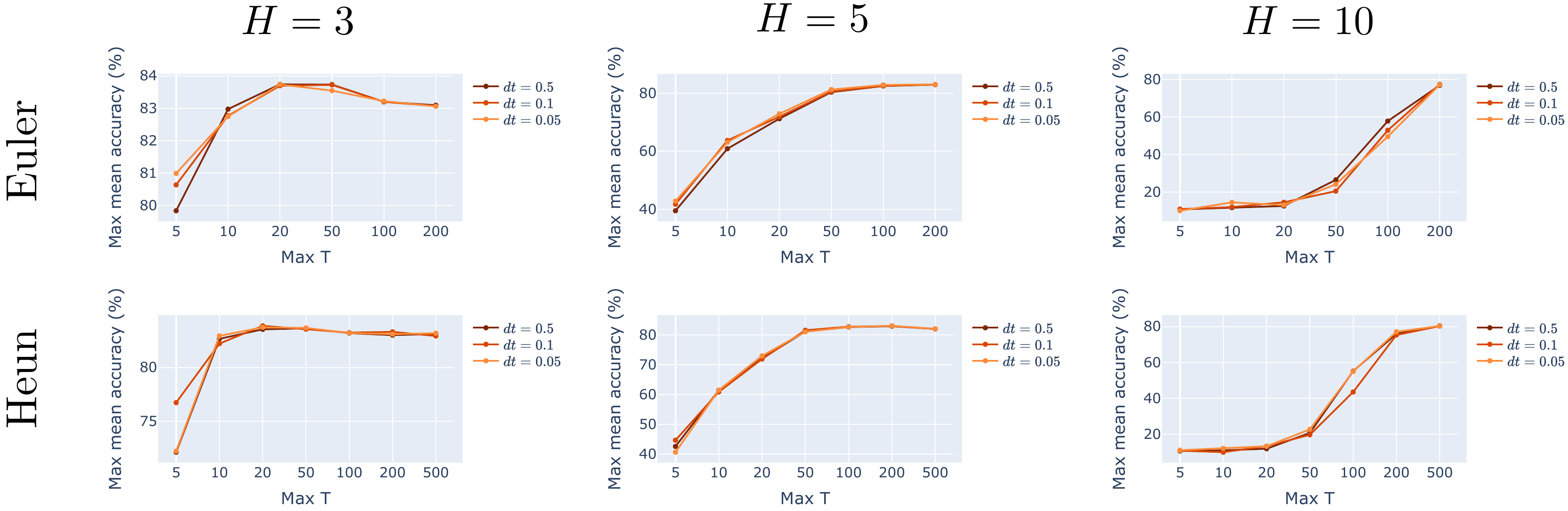}}
        \caption{\textbf{Same results as Figure~\ref{ch6:fig:accs-per-ts-mnist} for Fashion-MNIST.} For the results in Figure~\ref{ch6:fig:best-infer-runtimes} with $H=3$, we selected runs with $T=20$, and $dt=0.5$ for Euler and $dt=0.1$ for Heun. For the other network depths, the same hyperparameters were chosen for both solvers: $T=200$ and $dt=0.5$ for $H=5$, and $T=200$, and $dt=0.05$ for $H=10$.}
        \label{ch6:fig:accs-per-ts-fashion}
    \end{center}
\end{figure}
\begin{figure}[H]
    \begin{center}
        \centerline{\includegraphics[width=\textwidth]{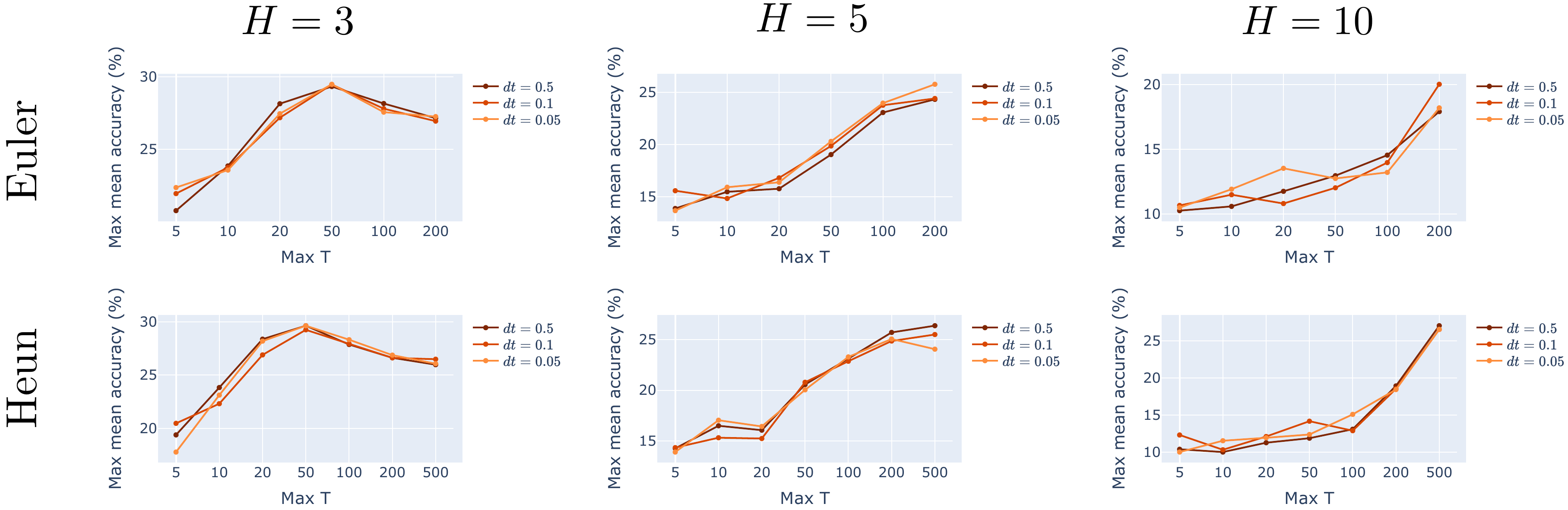}}
        \caption{\textbf{Same results as Figure~\ref{ch6:fig:accs-per-ts-mnist} for CIFAR-10.} For the results in Figure~\ref{ch6:fig:best-infer-runtimes} with $H=3$, we selected runs with $T=50$ and $dt=0.05$ for both solvers. For $H=5$, we selected $T=200$ and $dt = 0.05$ for Euler, and $T=500$ and $dt=0.5$ for Heun. Finally, for $H=10$, we selected $dt=0.1$, with $T=200$ for Euler and $T=500$ for Heun.}
        \label{ch6:fig:accs-per-ts-cifar}
    \end{center}
\end{figure}
\begin{figure}[H]
    \begin{center}
        \centerline{\includegraphics[width=\textwidth]{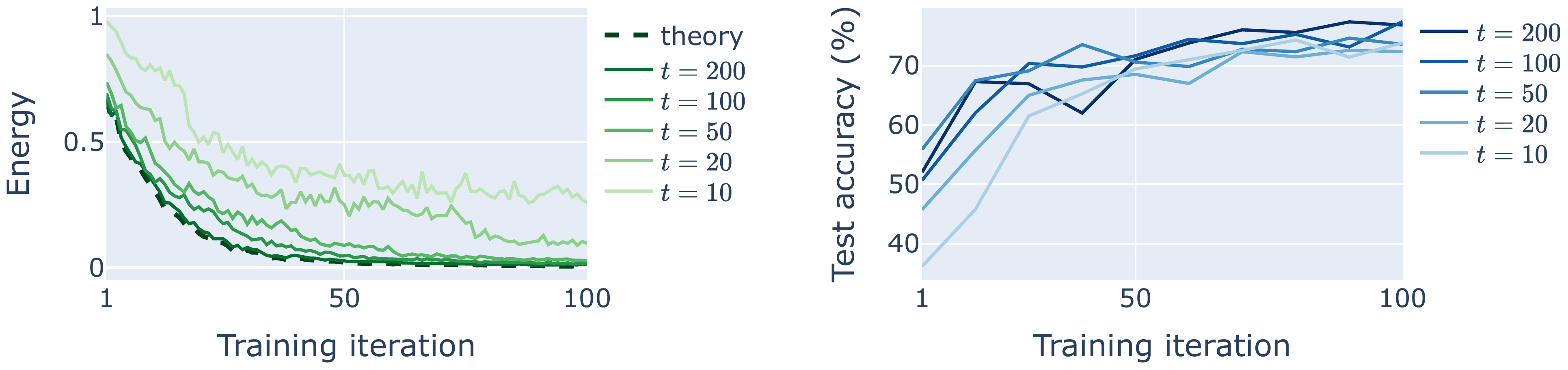}}
        \caption{\textbf{Same results as Figure~\ref{ch6:fig:energies-accs-mnist} for Fashion-MNIST.}}
        \label{ch6:fig:energies-accs-fashion}
    \end{center}
\end{figure}